%% file: LedHilBie_DATGAN.tex
\documentclass{article}

\usepackage{arxiv}

\usepackage[utf8]{inputenc} % allow utf-8 input
\usepackage[T1]{fontenc}    % use 8-bit T1 fonts
\usepackage[hyphens]{url}            % simple URL typesetting
\usepackage{booktabs}       % professional-quality tables
\usepackage{amsfonts}       % blackboard math symbols
\usepackage{nicefrac}       % compact symbols for 1/2, etc.
\usepackage{microtype}      % microtypography
\usepackage{lipsum}
\usepackage{amsmath}
\usepackage{amssymb}
\usepackage{bm}
\usepackage{bbold}
\usepackage{float}
\usepackage{graphicx}
\usepackage{subcaption}
\usepackage{multirow}
\usepackage{makecell}
\usepackage{tabularx}
\usepackage{blindtext}
\usepackage{xltabular}
\usepackage{pdflscape}
\usepackage{afterpage}
\usepackage{breakcites}
\usepackage[inline]{enumitem}

\newlist{inlinelist}{enumerate*}{1}
\setlist[inlinelist]{label=(\roman*)}

\usepackage{siunitx}
\usepackage{caption}
\usepackage[dvipsnames, table]{xcolor}
\definecolor{darkgreen}{HTML}{4F7942}

\usepackage[hyperindex,breaklinks]{hyperref}       % hyperlinks
\newcommand\myshade{85}
\colorlet{mylinkcolor}{violet}
\colorlet{mycitecolor}{Tan}
\colorlet{myurlcolor}{Aquamarine}

\hypersetup{
  linkcolor  = mylinkcolor!\myshade!black,
  citecolor  = mycitecolor!\myshade!black,
  urlcolor   = myurlcolor!\myshade!black,
  colorlinks = true,
}

\usepackage{tikz}
\usepackage{pgfplots}
\DeclareUnicodeCharacter{2212}{−}
\usepgfplotslibrary{groupplots,dateplot}
\pgfplotsset{compat=newest}

\usetikzlibrary{plotmarks}
\usetikzlibrary{positioning}
\usetikzlibrary{shapes}
\usetikzlibrary{arrows.meta}
\usetikzlibrary{patterns}

\usepackage[semicolon, authoryear]{natbib}

\usepackage{etoolbox}

\makeatletter
\pretocmd{\NAT@citex}{%
  \let\NAT@hyper@\NAT@hyper@citex
  \def\NAT@postnote{#2}%
  \setcounter{NAT@total@cites}{0}%
  \setcounter{NAT@count@cites}{0}%
  \forcsvlist{\stepcounter{NAT@total@cites}\@gobble}{#3}}{}{}
\newcounter{NAT@total@cites}
\newcounter{NAT@count@cites}
\def\NAT@postnote{}

\def\NAT@hyper@citex#1{%
  \stepcounter{NAT@count@cites}%
  \hyper@natlinkstart{\@citeb\@extra@b@citeb}#1%
  \ifnumequal{\value{NAT@count@cites}}{\value{NAT@total@cites}}
    {\ifNAT@swa\else\if*\NAT@postnote*\else%
     \NAT@cmt\NAT@postnote\global\def\NAT@postnote{}\fi\fi}{}%
  \ifNAT@swa\else\if\relax\NAT@date\relax
  \else\NAT@@close\global\let\NAT@nm\@empty\fi\fi% avoid compact citations
  \hyper@natlinkend}
\renewcommand\hyper@natlinkbreak[2]{#1}

\patchcmd{\NAT@citex}
  {\ifNAT@swa\else\if*#2*\else\NAT@cmt#2\fi
   \if\relax\NAT@date\relax\else\NAT@@close\fi\fi}{}{}{}
\patchcmd{\NAT@citex}
  {\if\relax\NAT@date\relax\NAT@def@citea\else\NAT@def@citea@close\fi}
  {\if\relax\NAT@date\relax\NAT@def@citea\else\NAT@def@citea@space\fi}{}{}

\makeatother

\newcommand{\tabitem}{~~\llap{\textbullet}~~}

\newcolumntype{C}{>{\centering\arraybackslash}X}
\newcolumntype{L}{>{\flushleft\arraybackslash}X}

\renewcommand\arraystretch{1.3}

\usepackage{algorithm}
\usepackage[noend]{algpseudocode}

\DeclareMathOperator*{\argmax}{arg\,max}
\DeclareSymbolFontAlphabet{\amsmathbb}{AMSb}%

\usepackage[show]{chato-notes} % set to ``hide for submission

\let\vec\bm

\newcommand{\nl}{\vskip \arraystretch pt}

\usepackage{pdfpages} % include pdfs
\usepackage{pgffor} % for loops
\usepackage{xr}
\def\supplementfilename{supp_mat/LedHilBie_DATGAN_Support_arxiv}

\pdfximage{\supplementfilename.pdf}
\def\numbersupplementpages{\the\pdflastximagepages}

\newif\ifarXiv
\arXivtrue 
\title{DATGAN: Integrating expert knowledge into deep learning for synthetic tabular data}

\author{
  Gael Lederrey \\
  Transport and Mobility Laboratory, \\ 
  \'Ecole Polytechnique F\'ed\'erale de Lausanne, \\
  Lausanne, Switzerland \\
  \texttt{gael.lederrey@epfl.ch}\\
   \And
  Tim Hillel \\
  Transport and Mobility Laboratory, \\ 
  \'Ecole Polytechnique F\'ed\'erale de Lausanne, \\
  Lausanne, Switzerland \\
  \texttt{tim.hillel@epfl.ch}\\
   \And
  Michel Bierlaire \\
  Transport and Mobility Laboratory, \\ 
  \'Ecole Polytechnique F\'ed\'erale de Lausanne, \\
  Lausanne, Switzerland \\
  \texttt{michel.bierlaire@epfl.ch}\\
}

\begin{document}
\maketitle

\input{text/00_abstract}

\newpage
\tableofcontents
\newpage

\input{text/01_introduction}
\input{text/02_lit_rev}
\input{text/03_methodology}
\input{text/04_case_study}
\input{text/05_results}

\input{text/06_conclusion}

\newpage
\bibliographystyle{plainnat-swapnames}
\bibliography{PhD}
\newpage

\input{text/07_appendix}

\ifarXiv
    

\section*{Supplementary material}

# Supplementary materials
# DATGAN: Integrating expert knowledge into deep learning for synthetic tabular data

## 1 Introduction

In this document, we present supplementary materials for the article names "*DATGAN: Integrating expert knowledge into deep learning for synthetic tabular data*". The original article provides the methodology to create multiple DATGAN versions. This document first summarizes these versions in Section 2. Then, in Section 3, we provide the comparison of the different DATGAN versions. This is used in the original article to choose the best-performing models depending on the case study. Next, Section 4 gives a notation table that can be helpful while following the methodology in the original article. Finally, Section 5 provide the complete results used for the rankings in the original article.

## 2 Summary of DATGAN versions

The methodology in the article provides a detailed overview of the different elements of the DATGAN model: the generator, the discriminator, the loss function, and the data encoding. The use of expert knowledge to build a DAG $\mathcal{G}$ and its usage to build the structure of the generator is the main contribution of this paper. Indeed, while this is respected, many things can change, such as the discriminator structure, keeping the DATGAN identity. Therefore, in our case, instead of sticking to a single model, we systematically test multiple aspects of the DATGAN:

- Three different loss functions:
  - standard loss (`SGAN`):
  $$\min_G \max_D \mathcal{L}^{\text{SGAN}}(D, G) = \mathbb{E}_{\{\widehat{\boldsymbol{v}}_{1:N_V}\} \sim \mathbb{P}(\widehat{\mathbf{T}})} \log D(\widehat{\boldsymbol{v}}_{1:N_V}) + \mathbb{E}_{\boldsymbol{z} \sim \mathcal{N}(0,1)} \log(1 - D(G(\boldsymbol{z}))) \qquad (1)$$
  - wasserstein loss (`WGAN`):
  $$\min_G \max_D \mathcal{L}^{\text{WGAN}}(D, G) = \mathbb{E}_{\{\widehat{\boldsymbol{v}}_{1:N_V}\} \sim \mathbb{P}(\widehat{\mathbf{T}})} D(\widehat{\boldsymbol{v}}_{1:N_V}) - \mathbb{E}_{\boldsymbol{z} \sim \mathcal{N}(0,1)} D(G(\boldsymbol{z})) \qquad (2)$$
  - wasserstein loss with gradient penalty (`WGGP`):
  $$\min_G \max_D \mathcal{L}^{\text{WGGP}}(D, G) = \mathbb{E}_{\{\widehat{\boldsymbol{v}}_{1:N_V}\} \sim \mathbb{P}(\widehat{\mathbf{T}})} D(\widehat{\boldsymbol{v}}_{1:N_V}) - \mathbb{E}_{\boldsymbol{z} \sim \mathcal{N}(0,1)} D(G(\boldsymbol{z})) + \lambda \mathbb{E}_{\widehat{\boldsymbol{v}} \sim \mathbb{P}(\widehat{\mathbf{T}}, G(\boldsymbol{z}))} (\|\nabla_{\widetilde{\boldsymbol{v}}} D(\widetilde{\boldsymbol{v}})\|_2 - 1)^2 \qquad (3)$$

- Three different label smoothing strategies:
  - none (`NO`)
  - one-sided on the original data (`OS`)
  - two-sided (`TS`)

- Four different sampling strategies:
  - $\arg\max$ on continuous and categorical columns (`AA`)
  - `simulation` on continuous columns and $\arg\max$ on categorical columns (`SA`)
  - $\arg\max$ on continuous columns and `simulation` on categorical columns (`AS`)
  - `simulation` on continuous and categorical columns (`SS`)

By combining all of these different strategies, we obtain 36 different modeling approaches for the DATGAN. They are enumerated in Table 1. The next section systematically compares these different approaches on multiple case studies to identify the best-performing combination.

Table 1: Summary of all the possible models using the DATGAN concept.

| Name | Loss function | Label smoothing | Sampling |
|---|---|---|---|
| SGAN_NO_AA | SGAN (Eq. 1) | none | $\arg\max$ for $C_t$ and $D_t$ |
| SGAN_NO_SA | | | simulation for $C_t$, $\arg\max$ for $D_t$ |
| SGAN_NO_AS | | | $\arg\max$ for $C_t$, simulation for $D_t$ |
| SGAN_NO_SS | | | simulation for $C_t$ and $D_t$ |
| SGAN_OS_AA | | one-sided | $\arg\max$ for $C_t$ and $D_t$ |
| SGAN_OS_SA | | | simulation for $C_t$, $\arg\max$ for $D_t$ |
| SGAN_OS_AS | | | $\arg\max$ for $C_t$, simulation for $D_t$ |
| SGAN_OS_SS | | | simulation for $C_t$ and $D_t$ |
| SGAN_TS_AA | | two-sided | $\arg\max$ for $C_t$ and $D_t$ |
| SGAN_TS_SA | | | simulation for $C_t$, $\arg\max$ for $D_t$ |
| SGAN_TS_AS | | | $\arg\max$ for $C_t$, simulation for $D_t$ |
| SGAN_TS_SS | | | simulation for $C_t$ and $D_t$ |
| WGAN_NO_AA | WGAN (Eq. 2) | none | $\arg\max$ for $C_t$ and $D_t$ |
| WGAN_NO_SA | | | simulation for $C_t$, $\arg\max$ for $D_t$ |
| WGAN_NO_AS | | | $\arg\max$ for $C_t$, simulation for $D_t$ |
| WGAN_NO_SS | | | simulation for $C_t$ and $D_t$ |
| WGAN_OS_AA | | one-sided | $\arg\max$ for $C_t$ and $D_t$ |
| WGAN_OS_SA | | | simulation for $C_t$, $\arg\max$ for $D_t$ |
| WGAN_OS_AS | | | $\arg\max$ for $C_t$, simulation for $D_t$ |
| WGAN_OS_SS | | | simulation for $C_t$ and $D_t$ |
| WGAN_TS_AA | | two-sided | $\arg\max$ for $C_t$ and $D_t$ |
| WGAN_TS_SA | | | simulation for $C_t$, $\arg\max$ for $D_t$ |
| WGAN_TS_AS | | | $\arg\max$ for $C_t$, simulation for $D_t$ |
| WGAN_TS_SS | | | simulation for $C_t$ and $D_t$ |
| WGGP_NO_AA | WGGP (Eq. 3) | none | $\arg\max$ for $C_t$ and $D_t$ |
| WGGP_NO_SA | | | simulation for $C_t$, $\arg\max$ for $D_t$ |
| WGGP_NO_AS | | | $\arg\max$ for $C_t$, simulation for $D_t$ |
| WGGP_NO_SS | | | simulation for $C_t$ and $D_t$ |
| WGGP_OS_AA | | one-sided | $\arg\max$ for $C_t$ and $D_t$ |
| WGGP_OS_SA | | | simulation for $C_t$, $\arg\max$ for $D_t$ |
| WGGP_OS_AS | | | $\arg\max$ for $C_t$, simulation for $D_t$ |
| WGGP_OS_SS | | | simulation for $C_t$ and $D_t$ |
| WGGP_TS_AA | | two-sided | $\arg\max$ for $C_t$ and $D_t$ |
| WGGP_TS_SA | | | simulation for $C_t$, $\arg\max$ for $D_t$ |
| WGGP_TS_AS | | | $\arg\max$ for $C_t$, simulation for $D_t$ |
| WGGP_TS_SS | | | simulation for $C_t$ and $D_t$ |

## 3 Comparison of DATGAN versions

This section compares the 36 DATGAN versions against each other using the result assessments methods presented in the original article on the first three case studies (CMAP, LPMC_half, and LPMC).

For the CMAP case study, we see that the DATGAN with the `WGGP` loss provides poorer results compared to the DATGAN with the `SGAN` or `WGAN` loss. It is the case with both assessment methods. This is quite unexpected since the `WGGP` loss is the most recent loss function available amongst the three. We do not clearly explain why the model fails with this loss on this particular dataset. Secondly, we see that all the models using the one-sided label smoothing (`OS`) tend to perform worse than their counterparts. This result was expected, especially when using simulation while sampling the synthetic variables. Indeed, if the model only uses one-sided label smoothing, the Generator will learn the noisy version of the categorical variables. If we use $\arg\max$ for selecting the categorical variable, it does not have a large effect since the noise is quite small. However, the final synthetic probability distribution does not correspond to the original one when using simulation. Therefore, the model will sample the values wrongly and thus lead to poor results. When we compare two-sided label smoothing (`TS`) against no label smoothing (`NO`), we see that the results are different for the `SGAN` and the `WGAN` loss functions. Indeed, it seems that it is better to use the two-sided label smoothing with the `WGAN` loss, and it is better to avoid label smoothing with the `SGAN` loss. Table 2 shows the ten best models in terms of average ranking on the two assessment methods. We see that both loss functions are viable. However, the `WGAN` loss seems to be slightly better on average. The sampling method seems to have less influence than the label smoothing and the loss function on the final results.

Table 2: Average rankings for the ten best DATGAN models on the CMAP dataset.

| Name | Avg. rank stats | Avg. rank ML | rank |
|---|---|---|---|
| `WGAN_TS_AS` | **2.88** | 8.0 | **5.44** |
| `WGAN_TS_AA` | 8.24 | 4.0 | 6.12 |
| `WGAN_TS_SS` | 4.04 | 9.5 | 6.77 |
| `SGAN_NO_SA` | 10.08 | 3.5 | 6.79 |
| `SGAN_NO_SS` | 8.56 | 6.0 | 7.28 |
| `WGAN_TS_SA` | 6.84 | 8.0 | 7.42 |
| `SGAN_NO_AS` | 12.68 | 4.0 | 8.34 |
| `SGAN_NO_AA` | 15.96 | **2.0** | 8.98 |
| `SGAN_TS_AS` | 8.68 | 9.5 | 9.09 |
| `SGAN_TS_SA` | 7.88 | 10.5 | 9.19 |

For the LPMC dataset, we have quite unexpected results compared to the CMAP case study. The conclusions on the label smoothing remain the same. However, the best loss function to use in this case is the `WGGP` loss. This is especially visible when using the Machine Learning efficacy method to assess the results. All the models with the `WGGP` loss consistently outperforms the models with the other losses. The `WGAN` loss performs slightly worse than the `SGAN` loss. Table 3 shows the ten best models in terms of average on both assessment methods. We see that both models using simulation for the categorical variables and two-sided label smoothing outperform the other models. It is interesting to note that, in this case, the sampling method for continuous variables does not affect much the final results. This result can be explained thanks to the encoding we use for the continuous variables. Indeed, all the encoded values $w_{t,j}$ are computed such that they all correspond to the continuous value $c_{t,j}$ once drawn from their respective Gaussian mixture. Therefore, if the algorithm fails to choose the right mixture, the sampling will return the same continuous synthetic value.

If we compare the two previous case studies, there are two hypotheses as to why the `WGAN` loss works better with the CMAP dataset and the `WGGP` loss with the LPMC dataset:(i) the number of data points in the dataset influences the optimization process with a given loss function; (ii) the ratio of continuous/categorical variables influences the choice of the loss function. The first hypothesis is easier to test since we can use the LPMC_half case study to compare with the LPMC case study. The results in the appendix show the same behavior with the smaller LPMC dataset than with the complete one. Indeed, the `WGGP` loss leads to significantly better results across both assessments methods compared to the other loss functions. Table 4 shows the ten best models in terms of average on both assessment methods. Once again, the version with the `WGGP` loss, two-sided label smoothing, and sampling using simulation for both continuous and categorical variables appears to be the best model. However, it is interesting to note that the versions with the `WGAN` loss function shows the best statistics when only looking at the first aggregation level on categorical variables. This hints at the fact that the `WGAN` loss is the most appropriate loss to use when a dataset contains a majority of categorical variables.

For further analyses, we settle on one particular version of the DATGAN. As seen across all the case studies, two-sided label smoothing is the best method for the discriminator. In addition, using simulation on both types of variables for the

Table 3: Average rankings for the ten best DATGAN models on the LPMC dataset.

| Name | Avg. rank stats | Avg. rank ML | rank |
|---|---|---|---|
| WGGP_TS_SS | **2.96** | 4.0 | **3.48** |
| WGGP_TS_AS | 5.16 | 3.0 | 4.08 |
| WGGP_NO_AS | 9.52 | 8.0 | 8.76 |
| WGGP_NO_SS | 9.76 | 8.5 | 9.13 |
| WGGP_OS_SA | 15.60 | 4.0 | 9.80 |
| WGGP_OS_AA | 18.36 | **2.5** | 10.43 |
| WGGP_TS_SA | 12.36 | 9.5 | 10.93 |
| WGGP_NO_AA | 12.72 | 9.5 | 11.11 |
| WGGP_TS_AA | 14.80 | 7.5 | 11.15 |
| WGGP_NO_SA | 13.04 | 10.0 | 11.52 |

Table 4: Average rankings for the ten best DATGAN models on the LPMC_half dataset.

| Name | Avg. rank stats | Avg. rank ML | rank |
|---|---|---|---|
| WGGP_TS_SS | **3.12** | 4.5 | **3.81** |
| WGGP_TS_AS | 4.84 | 4.0 | 4.42 |
| WGGP_NO_SS | 8.20 | 8.5 | 8.35 |
| WGGP_OS_SA | 14.36 | 4.0 | 9.18 |
| WGGP_NO_AS | 11.88 | 7.0 | 9.44 |
| WGGP_OS_AA | 16.00 | **3.5** | 9.75 |
| WGGP_OS_SS | 18.04 | 5.5 | 11.77 |
| WGGP_OS_AS | 20.24 | 5.0 | 12.62 |
| SGAN_TS_SS | 9.52 | 16.0 | 12.76 |
| SGAN_NO_AS | 10.96 | 16.0 | 13.48 |

sampling leads to better results for the two LPMC case studies. For the CMAP case study, it is amongst the best models. We, thus, choose to use this sampling method to stay consistent across both variable types. Finally, we cannot use the same one for all the datasets for the loss function. As stated earlier, the `WGAN` loss performs best when there are more categorical variables than continuous in the dataset. Therefore, the DATGANs presented in the original article are using the `WGAN` loss function for the CMAP case study and the `WGGP` loss function for the LPMC case studies.

## 4 Notations table

In this section, we present a notation table that can be used while following the methodology in the original article.

Table 5: Notations used in the methodology grouped by categories

| Notation | Name | Description |
|---|---|---|
| **Main elements** | | |
| $D$ | Discriminator | Neural network model used to discriminate/critic the original and synthetic data. (see Section 3.2) |
| $G$ | Generator | Neural network model used to generate synthetic data from random noise. (see Section 3.1) |

Table 5 – continued from previous page

| Notation | Name | Description |
|---|---|---|
| $\mathbf{T}$ | original dataset | Original dataset provided by the modeler. (size: $N_V \times N_{\text{rows}}$) |
| $\mathbb{P}(\mathbf{T})$ | distributions | Unknown joint distribution that the random variables $V_t$ follow. |
| $\mathbf{T_{synth}}$ | synthetic dataset | Final synthetic dataset after the sampling procedure; it corresponds to the final output of the generator (size: $N_V \times N_{\text{rows}}$). |
| $\widehat{\mathbf{T}}$ | encoded original dataset | Original dataset that has been encoded. (size: $\left(\sum_{t=1}^{N_C} 2N_{m,t} + \sum_{t=1}^{N_D} |D_t|\right) \times N_{\text{rows}}$) |
| $\widehat{\mathbf{T}}_{\mathbf{synth}}$ | encoded synthetic dataset | Synthetic dataset that is directly returned by the generator before the sampling step. (see Section 3.4.4; size: $\left(\sum_{t=1}^{N_C} 2N_{m,t} + \sum_{t=1}^{N_D} |D_t|\right) \times N_{\text{rows}}$) |
| **Main variables** | | |
| $V_t$ | random variable | $t$-th random variable. |
| $\boldsymbol{v}_t$ | column data in $\mathbf{T}$ | Column-vectors of data that define the table $\mathbf{T}$. (size: $N_{\text{rows}}$) |
| $v_{t,j}$ | single value in $\mathbf{T}$ | Scalar value in a column-vector. |
| $\boldsymbol{z}$ | noise | Random noise tensor used as an input for the Generator $G$. (size: $N_V \times N_{\text{rows}} \times N_{\text{sources}}$) |
| $\boldsymbol{v}_t^{\mathbf{synth}}$ | column data in $\mathbf{T_{synth}}$ | Column-vectors of data that define the table $\mathbf{T_{synth}}$. (size: $N_{\text{rows}}$) |
| $v_{t,j}^{\mathbf{synth}}$ | single value in $\mathbf{T_{synth}}$ | Scalar value in a column-vector. |
| $\widehat{\boldsymbol{v}}_t$ | encoded column data in $\mathbf{T}$ | Encoded version of $\boldsymbol{v}_t$. |
| $\widehat{v}_{t,j}$ | encoded single value in $\mathbf{T}$ | Encoded version of $v_{t,j}$. |
| $\widehat{\boldsymbol{v}}_t^{\mathbf{synth}}$ | encoded column data in $\mathbf{T_{synth}}$ | Encoded version of $\boldsymbol{v}_t^{\mathbf{synth}}$. |
| $\widehat{v}_{t,j}^{\mathbf{synth}}$ | encoded single value in $\mathbf{T_{synth}}$ | Encoded version of $v_{t,j}^{\mathbf{synth}}$. |
| **Directed Acyclic Graph** (Section 3.1.1) | | |
| $\mathcal{G}$ | DAG | Directed Acyclic Graph (DAG) used to represent the interdependencies between the variables. |
| $\mathcal{A}(V_t)$ | ancestors | Set of all the ancestors of the variable $V_t$ in the DAG $\mathcal{G}$. |
| $\mathcal{D}(V_t)$ | direct ancestors | Set of all the direct ancestors of the variable $V_t$ in the DAG $\mathcal{G}$. |
| $\mathcal{S}(V_t)$ | source nodes | Set of all the source nodes leading to variable $V_t$ in the DAG $\mathcal{G}$. |
| $\mathcal{E}(V_t)$ | in-edges | Set of all in-edges of the variable $V_t$ in the DAG $\mathcal{G}$. |
| **LSTM cells** (Section 3.1.2) | | |
| $\mathbf{LSTM_t}$ | LSTM cell | LSTM cell used to generate synthetic values for the variable $V_t$. |
| $C_t$ | cell state | Output cell state of the LSTM cell $\mathbf{LSTM_t}$. (size: $N_h \times N_b$) |
| $\boldsymbol{h}_t$ | output | Output tensor of the LSTM cell $\mathbf{LSTM_t}$. (size: $N_h \times N_b$) |
| $\boldsymbol{x}_t$ | generic input | Generic input tensor for a LSTM cell. (size: $N_x \times N_b$) |
| $\boldsymbol{i}_t$ | input | Input tensor of the LSTM cell $\mathbf{LSTM_t}$. (generic size: $(N_h + N_x) \times N_b$; DATGAN size: $(2N_h + N_z) \times N_b$) |
| $\boldsymbol{z}_t$ | noise | Random noise tensor used as an input for the LSTM cell $\mathbf{LSTM_t}$. (size: $N_z \times N_b$) |
| $\boldsymbol{f}_t$ | transformed output | Transformed output of the variable $\widehat{\boldsymbol{v}}_t^{\mathbf{synth}}$ to resize it according to the LSTM cell. (size: $N_h \times N_b$) |
| $\boldsymbol{a}_t$ | attention | Attention tensor used in the input of the LSTM cell $\mathbf{LSTM_t}$. (size: $N_h \times N_b$) |

Table 5 – continued from previous page

| Notation | Name | Description |
|---|---|---|
| $\boldsymbol{\alpha}_t$ | attention weights | Attention weight vector used to compute $a_t$. (size: $|\mathcal{A}(V(t))|$) |
| $\boldsymbol{h}'_t$ | reduced output | Reduced output tensor of the LSTM cell $\mathbf{LSTM_t}$ after using a convolutional layer. (size: $N_{\text{conv}} \times N_b$) |
| **Discriminator** (Section 3.2) | | |
| $\boldsymbol{l}_i$ | internal layer | Internal layer used in the discriminator. (undefined size) |
| $\widehat{\boldsymbol{l}}_i$ | standardized internal layer | Internal layer after passing it through a fully connected layer to resize it (size: $N_L \times N_b$) |
| $l_D$ | discriminator output | Final unbounded scalar result used as an output of the discriminator. |
| $\boldsymbol{l}_0$ | discriminator input | Input layer used for the discriminator. |
| **Loss function** (Section 3.3) | | |
| $\mathcal{L}(D, G)$ | loss function | Generic loss function. The generator is trying to minimize it while the discriminator is trying to maximize it at the same time. |
| $\mathcal{L}_G$ | generator loss function | Loss function applying only to the generator, which is trying to minimize it. |
| $\mathcal{L}_D$ | discriminator loss function | Loss function applying only to the discriminator, which is trying to minimize it. |
| **Data processing** (Section 3.4) | | |
| $\boldsymbol{c}_t$ | continuous column data in $\mathbf{T}$ | Column-vectors of continuous data that define the table $\mathbf{T}$. (size: $N_{\text{rows}}$) |
| $\boldsymbol{d}_t$ | categorical column data in $\mathbf{T}$ | Column-vectors of continuous data that define the table $\mathbf{T}$. (size: $N_{\text{rows}}$) |
| $\boldsymbol{c}_t^{\mathbf{synth}}$ | continuous column data in $\mathbf{T_{synth}}$ | Column-vectors of continuous data that define the table $\mathbf{T_{synth}}$. (size: $N_{\text{rows}}$) |
| $\boldsymbol{d}_t^{\mathbf{synth}}$ | categorical column data in $\mathbf{T_{synth}}$ | Column-vectors of continuous data that define the table $\mathbf{T_{synth}}$. (size: $N_{\text{rows}}$) |
| $\boldsymbol{\eta}_t$ | means of VGM | Vector containing the mean of each component in the Gaussian mixture for the continuous variable $C_t$. (size: $N_{m,t}$) |
| $\boldsymbol{\sigma}_t$ | variances of VGM | Vector containing the variance of each component in the Gaussian mixture for the continuous variable $C_t$. (size: $N_{m,t}$) |
| $\boldsymbol{w}_t$ | values of encoded continuous variable | Matrix containing the values of the encoded variable $\boldsymbol{c}_t$. (size: $N_{m,t} \times N_{\text{rows}}$) |
| $\boldsymbol{p}_t$ | probabilities of encoded continuous variable | Matrix containing the probabilities of the encoded variable $\boldsymbol{c}_t$. (size: $N_{m,t} \times N_{\text{rows}}$) |
| $\boldsymbol{o}_t$ | one-hot encoding of categorical variable | Matrix containing the one-hot encoding of the categorical variable $\boldsymbol{d}_t$. (size: $|D_t| \times N_{\text{rows}}$) |
| $\boldsymbol{w}_t^{\mathbf{synth}}$ | generated matrix similar to $\boldsymbol{w}_t$ | Output of the generator for the values of the encoded continuous variables. (size: $N_{m,t} \times N_{\text{rows}}$) |
| $\boldsymbol{p}_t^{\mathbf{synth}}$ | generated matrix similar to $\boldsymbol{p}_t$ | Output of the generator for the probabilities of the encoded continuous variables. (size: $N_{m,t} \times N_{\text{rows}}$) |
| $\boldsymbol{o}_t^{\mathbf{synth}}$ | generated matrix similar to $\boldsymbol{o}_t$ | Output of the generator for the one-hot encoded categorical variables. (size: $|D_t| \times N_{\text{rows}}$) |
| $\widetilde{\boldsymbol{o}}_t$ | Noisy version of $\boldsymbol{o}_t$ | Corresponds to $\boldsymbol{o}_t$ after adding uniform noise for the label smoothing (size: $|D_t| \times N_{\text{rows}}$) |

Table 5 – continued from previous page

| Notation | Name | Description |
|---|---|---|
| $\widetilde{\boldsymbol{o}}_t^{\mathbf{synth}}$ | Noisy version of $\boldsymbol{o}_t^{\mathbf{synth}}$ | Corresponds to $\boldsymbol{o}_t^{\mathbf{synth}}$ after adding uniform noise for the label smoothing (size: $\|D_t\| \times N_{\text{rows}}$) |
| **Results assessments** (Section 3.5) | | |
| $\boldsymbol{\pi}$ | Frequencies of original data | Frequency list computed on the original dataset $\mathbf{T}$. |
| $\boldsymbol{\pi}^{\mathbf{synth}}$ | Frequencies of synthetic data | Frequency list computed on the synthetic dataset $\mathbf{T}^{\mathbf{synth}}$. |
| MAE | Mean Absolute Error | Mean Absolute Error computed between frequencies lists on original and synthetic datasets. |
| RMSE | Root Mean Square Error | Root Mean Square Error computed between frequencies lists on original and synthetic datasets. |
| SRMSE | Standardized Root Mean Square Error | Standardized Root Mean Square Error computed between frequencies lists on original and synthetic datasets. |
| $R^2$ | Coefficient of determination | Coefficient of determination computed between frequencies lists on original and synthetic datasets. |
| $\rho_{\text{Pearson}}$ | Pearson's correlation | Pearson's correlation computed between frequencies lists on original and synthetic datasets. |
| $\mathcal{L}_{\text{MSE}}$ | Mean Square Error | Mean Square error loss used in the supervised learning validation on continuous variables. |
| $\mathcal{L}_{\text{log-loss}}$ | Log loss error | Log loss error used in the supervised learning validation on categorical variables. |
| $m_t$ | LightGBM model | LightGBM model trained on variable $\boldsymbol{v}_t$. It corresponds to a classification model ($m_{\text{class},t}$) for categorical variables, and to a regression model ($m_{\text{reg},t}$) for continuous variables. |
| $g_t^{\text{reg}}$ | Score for the continuous variables | Final score of the supervised learning-based validation for continuous variables based on a regression model. |
| $g_t^{\text{class}}$ | Score for the categorical variables | Final score of the supervised learning-based validation for categorical variables based on a classification model. |
| **DATGAN versions** | | |
| SGAN | SGAN loss function | Standard two-player minimax loss function is used while optimising the DATGAN, see Equation 9. |
| WGAN | WGAN loss function | Wasserstein loss function is used while optimising the DATGAN, see Equation 12. |
| WGGP | WGGP loss function | Wasserstein loss function with a gradient penalty is used while optimising the DATGAN, see Equation 15. |
| NO | no label smoothing | No label smoothing applied for the discriminator input, see Equation 28. |
| OS | one-sided label smoothing | One-sided smoothing applied on the original data for the discriminator input, see Equation 29. |
| TS | two-sided label smoothing | Two-sided smoothing applied on the original and synthetic data for the discriminator input, see Equation 30. |
| AA | $\arg\max$ sampling for continuous and categorical | Maxmimum probability assignment used to sample both final continuous and categorical synthetic column data. |

Table 5 – continued from previous page

| Notation | Name | Description |
|---|---|---|
| SA | simulation sampling for continuous and $\arg\max$ sampling for categorical | Probability assignment for continuous synthetic column data and maximum probability assignment for categorical synthetic column data. |
| AS | $\arg\max$ sampling for continuous and simulation sampling for categorical | Maximum probability assignment for continuous synthetic column data and probability assignment for categorical synthetic column data. |
| SS | simulation sampling for continuous and categorical | Probability assignement used to sample both continuous and categorical final synthetic column data. |
| **Sizes** | | |
| $N_V$ | #variables | Number of random variables in the original dataset $\mathbf{T}$. |
| $N_{\text{rows}}$ | #rows | Number of rows in the original dataset $\mathbf{T}$. |
| $N_{\text{sources}}$ | #sources | Number of source nodes in the DAG $\mathcal{G}$. |
| $N_h$ | size of hidden layer | Size of the hidden layers used in the LSTM cells. ($N_h = 100$) |
| $N_x$ | size of input | Size of the generic input of the LSTM cell. |
| $N_z$ | size of noise tensor | Number of elements in the noise vector. ($N_z = 200$) |
| $N_b$ | batch size | Batch size. ($N_b$ = 500) |
| $N_L$ | #layers in discriminator | Number of layers used in the discriminator. ($N_L = 1$) |
| $N_l$ | size of discriminator | Size of the hidden layers inside the discriminator. ($N_l = 100$) |
| $N_C$ | #continuous variables | Number of continuous variables in the original dataset $\mathbf{T}$. |
| $N_D$ | #categorical variables | Number of categorical variables in the original dataset $\mathbf{T}$. |
| $N_{m,t}$ | #modes | Number of modes used to encode the continuous variable $C_t$ with a VGM. |
| $|D_t|$ | #categories | Number of unique categories in the categorical variable $D_t$. |
| $N_{\text{conv}}$ | size of convolutional layer | Size of the layer used in the output transformer to act as a convolution. Ideally, we want to have $N_{\text{conv}} < N_h$. ($N_{\text{conv}} = 50$) |
| **Layers in Neural Networks** | | |
| FC | Fully Connected | Fully connected layer without any activation function. |
| LeakyReLU | LeakyReLU | Layer with a leaky reflect linear activation function. |
| BN | Batch Normalization | Batch Normalization. |
| LN | Layer Normalization | Layer Normalization as described by Ba et al. (2016). |
| div | mini-batch discrimination | Mini-batch discriminator vector as described by Salimans et al. (2016). |
| tanh | $\tanh$ FC | Fully connected layer FC with a $\tanh$ activation function. |
| softmax | softmax FC | Fully connected layer FC with a softmax activation function. |

# 5 Tables of results

In this section, we provide the complete tables of results used in the original article and in the previous section. Each section is named after the corresponding sections in the original article.

## 5.1 Comparison of DATGAN versions

### 5.1.1 CMAP case study

Table 6: Results of the statistics on the first aggregation level (all columns) for the CMAP dataset. Lighter grey tone corresponds to better results compared to darker ones.

| Name | | MAE | | RMSE | | $R^2$ | | SRMSE | | $\rho_{\text{Pearson}}$ | rank |
|---|---|---|---|---|---|---|---|---|---|---|---|
| SGAN_NO_AA | 12 | $7.52e-3$ | 12 | $9.31e-3$ | 17 | $9.80e-1$ | 15 | $6.63e-2$ | 24 | $9.92e-1$ | 16.0 |
| SGAN_NO_AS | 06 | $6.50e-3$ | 06 | $8.19e-3$ | 15 | $9.82e-1$ | 09 | $6.19e-2$ | 23 | $9.92e-1$ | 11.8 |
| SGAN_NO_SA | 07 | $6.70e-3$ | 07 | $8.55e-3$ | 09 | $9.85e-1$ | 07 | $6.15e-2$ | 17 | $9.94e-1$ | 9.4 |
| SGAN_NO_SS | 04 | $6.13e-3$ | 05 | $7.80e-3$ | 11 | $9.85e-1$ | 05 | $5.76e-2$ | 13 | $9.94e-1$ | 7.6 |
| SGAN_OS_AA | 24 | $9.80e-3$ | 24 | $1.22e-2$ | 20 | $9.73e-1$ | 24 | $8.31e-2$ | 21 | $9.93e-1$ | 22.6 |
| SGAN_OS_AS | 34 | $3.36e-2$ | 33 | $4.06e-2$ | 27 | $8.75e-1$ | 34 | $2.70e-1$ | 22 | $9.93e-1$ | 30.0 |
| SGAN_OS_SA | 23 | $9.70e-3$ | 23 | $1.21e-2$ | 19 | $9.74e-1$ | 23 | $8.28e-2$ | 16 | $9.94e-1$ | 20.8 |
| SGAN_OS_SS | 32 | $3.32e-2$ | 31 | $4.04e-2$ | 25 | $8.78e-1$ | 33 | $2.68e-1$ | 10 | $9.94e-1$ | 26.2 |
| SGAN_TS_AA | 11 | $7.19e-3$ | 11 | $8.92e-3$ | 06 | $9.86e-1$ | 12 | $6.32e-2$ | 08 | $9.95e-1$ | 9.6 |
| SGAN_TS_AS | 09 | $6.94e-3$ | 08 | $8.59e-3$ | 03 | $9.87e-1$ | 06 | $6.08e-2$ | 07 | $9.95e-1$ | 6.6 |
| SGAN_TS_SA | 08 | $6.94e-3$ | 09 | $8.72e-3$ | 04 | $9.87e-1$ | 10 | $6.22e-2$ | 06 | $9.95e-1$ | 7.4 |
| SGAN_TS_SS | 10 | $7.02e-3$ | 10 | $8.79e-3$ | 05 | $9.87e-1$ | 08 | $6.17e-2$ | 05 | $9.95e-1$ | 7.6 |
| WGAN_NO_AA | 22 | $9.56e-3$ | 22 | $1.18e-2$ | 18 | $9.80e-1$ | 18 | $7.90e-2$ | 20 | $9.93e-1$ | 20.0 |
| WGAN_NO_AS | 01 | $5.53e-3$ | 01 | $7.07e-3$ | 12 | $9.85e-1$ | 04 | $5.36e-2$ | 19 | $9.93e-1$ | 7.4 |
| WGAN_NO_SA | 21 | $9.55e-3$ | 21 | $1.17e-2$ | 16 | $9.81e-1$ | 17 | $7.87e-2$ | 15 | $9.94e-1$ | 18.0 |
| WGAN_NO_SS | 02 | $5.65e-3$ | 02 | $7.09e-3$ | 07 | $9.85e-1$ | 02 | $5.29e-2$ | 14 | $9.94e-1$ | 5.4 |
| WGAN_OS_AA | 13 | $7.58e-3$ | 14 | $9.48e-3$ | 14 | $9.83e-1$ | 14 | $6.58e-2$ | 12 | $9.94e-1$ | 13.4 |
| WGAN_OS_AS | 31 | $3.32e-2$ | 32 | $4.04e-2$ | 26 | $8.75e-1$ | 31 | $2.66e-1$ | 18 | $9.93e-1$ | 27.6 |
| WGAN_OS_SA | 15 | $7.64e-3$ | 15 | $9.51e-3$ | 13 | $9.83e-1$ | 16 | $6.67e-2$ | 09 | $9.94e-1$ | 13.6 |
| WGAN_OS_SS | 33 | $3.35e-2$ | 34 | $4.07e-2$ | 28 | $8.75e-1$ | 32 | $2.67e-1$ | 11 | $9.94e-1$ | 27.6 |
| WGAN_TS_AA | 16 | $7.75e-3$ | 16 | $9.58e-3$ | 10 | $9.85e-1$ | 13 | $6.41e-2$ | 04 | $9.96e-1$ | 11.8 |
| WGAN_TS_AS | 03 | $6.11e-3$ | 03 | $7.58e-3$ | 02 | $9.87e-1$ | 01 | $5.23e-2$ | 02 | $9.96e-1$ | 2.2 |
| WGAN_TS_SA | 14 | $7.63e-3$ | 13 | $9.34e-3$ | 08 | $9.85e-1$ | 11 | $6.23e-2$ | 01 | $9.96e-1$ | 9.4 |
| WGAN_TS_SS | 05 | $6.33e-3$ | 04 | $7.76e-3$ | 01 | $9.88e-1$ | 03 | $5.32e-2$ | 03 | $9.96e-1$ | 3.2 |
| WGGP_NO_AA | 20 | $8.76e-3$ | 20 | $1.10e-2$ | 23 | $9.65e-1$ | 22 | $8.26e-2$ | 28 | $9.86e-1$ | 22.6 |
| WGGP_NO_AS | 18 | $8.28e-3$ | 18 | $1.06e-2$ | 21 | $9.66e-1$ | 20 | $8.06e-2$ | 25 | $9.87e-1$ | 20.4 |
| WGGP_NO_SA | 19 | $8.65e-3$ | 19 | $1.08e-2$ | 22 | $9.66e-1$ | 21 | $8.14e-2$ | 27 | $9.86e-1$ | 21.6 |
| WGGP_NO_SS | 17 | $8.26e-3$ | 17 | $1.05e-2$ | 24 | $9.65e-1$ | 19 | $7.96e-2$ | 26 | $9.87e-1$ | 20.6 |
| WGGP_OS_AA | 30 | $2.65e-2$ | 30 | $3.45e-2$ | 34 | $6.03e-1$ | 30 | $2.52e-1$ | 36 | $9.61e-1$ | 32.0 |
| WGGP_OS_AS | 36 | $3.85e-2$ | 36 | $4.84e-2$ | 36 | $5.85e-1$ | 36 | $3.49e-1$ | 35 | $9.63e-1$ | 35.8 |
| WGGP_OS_SA | 29 | $2.49e-2$ | 29 | $3.17e-2$ | 29 | $7.79e-1$ | 29 | $2.25e-1$ | 34 | $9.68e-1$ | 30.0 |
| WGGP_OS_SS | 35 | $3.72e-2$ | 35 | $4.59e-2$ | 30 | $7.58e-1$ | 35 | $3.24e-1$ | 33 | $9.71e-1$ | 33.6 |
| WGGP_TS_AA | 28 | $1.71e-2$ | 28 | $2.05e-2$ | 35 | $5.97e-1$ | 28 | $1.35e-1$ | 31 | $9.79e-1$ | 30.0 |
| WGGP_TS_AS | 26 | $1.54e-2$ | 26 | $1.89e-2$ | 33 | $6.16e-1$ | 27 | $1.26e-1$ | 32 | $9.79e-1$ | 28.8 |
| WGGP_TS_SA | 27 | $1.62e-2$ | 27 | $1.95e-2$ | 32 | $6.20e-1$ | 26 | $1.25e-1$ | 29 | $9.81e-1$ | 28.2 |
| WGGP_TS_SS | 25 | $1.47e-2$ | 25 | $1.78e-2$ | 31 | $6.34e-1$ | 25 | $1.15e-1$ | 30 | $9.80e-1$ | 27.2 |

Table 7: Results of the statistics on the first aggregation level (continuous columns) for the CMAP dataset. Lighter grey tone corresponds to better results compared to darker ones.

| Name | | MAE | | RMSE | | $R^2$ | | SRMSE | | $\rho_{\text{Pearson}}$ | | rank |
|---|---|---|---|---|---|---|---|---|---|---|---|---|
| SGAN_NO_AA | 24 | $1.23e-2$ | 24 | $1.61e-2$ | 24 | $9.31e-1$ | 24 | $1.61e-1$ | 24 | $9.67e-1$ | | 24.0 |
| SGAN_NO_AS | 23 | $1.21e-2$ | 23 | $1.61e-2$ | 23 | $9.31e-1$ | 23 | $1.61e-1$ | 23 | $9.67e-1$ | | 23.0 |
| SGAN_NO_SA | 17 | $1.02e-2$ | 17 | $1.43e-2$ | 14 | $9.47e-1$ | 17 | $1.43e-1$ | 18 | $9.75e-1$ | | 16.6 |
| SGAN_NO_SS | 18 | $1.03e-2$ | 18 | $1.43e-2$ | 15 | $9.47e-1$ | 18 | $1.43e-1$ | 16 | $9.76e-1$ | | 17.0 |
| SGAN_OS_AA | 21 | $1.16e-2$ | 21 | $1.51e-2$ | 21 | $9.37e-1$ | 21 | $1.51e-1$ | 15 | $9.76e-1$ | | 19.8 |
| SGAN_OS_AS | 22 | $1.18e-2$ | 22 | $1.55e-2$ | 22 | $9.36e-1$ | 22 | $1.55e-1$ | 19 | $9.75e-1$ | | 21.4 |
| SGAN_OS_SA | 20 | $1.09e-2$ | 20 | $1.47e-2$ | 18 | $9.45e-1$ | 20 | $1.47e-1$ | 08 | $9.83e-1$ | | 17.2 |
| SGAN_OS_SS | 19 | $1.07e-2$ | 19 | $1.45e-2$ | 16 | $9.47e-1$ | 19 | $1.45e-1$ | 07 | $9.83e-1$ | | 16.0 |
| SGAN_TS_AA | 12 | $9.59e-3$ | 07 | $1.25e-2$ | 08 | $9.60e-1$ | 07 | $1.25e-1$ | 10 | $9.82e-1$ | | 8.8 |
| SGAN_TS_AS | 09 | $9.44e-3$ | 05 | $1.24e-2$ | 07 | $9.61e-1$ | 05 | $1.24e-1$ | 09 | $9.82e-1$ | | 7.0 |
| SGAN_TS_SA | 08 | $9.42e-3$ | 08 | $1.27e-2$ | 06 | $9.61e-1$ | 08 | $1.27e-1$ | 06 | $9.83e-1$ | | 7.2 |
| SGAN_TS_SS | 07 | $9.24e-3$ | 06 | $1.24e-2$ | 05 | $9.62e-1$ | 06 | $1.24e-1$ | 05 | $9.84e-1$ | | 5.8 |
| WGAN_NO_AA | 13 | $9.86e-3$ | 16 | $1.37e-2$ | 20 | $9.39e-1$ | 16 | $1.37e-1$ | 22 | $9.70e-1$ | | 17.4 |
| WGAN_NO_AS | 14 | $9.91e-3$ | 15 | $1.37e-2$ | 19 | $9.40e-1$ | 15 | $1.37e-1$ | 21 | $9.70e-1$ | | 16.8 |
| WGAN_NO_SA | 15 | $9.93e-3$ | 13 | $1.34e-2$ | 13 | $9.47e-1$ | 13 | $1.34e-1$ | 17 | $9.75e-1$ | | 14.2 |
| WGAN_NO_SS | 16 | $1.00e-2$ | 14 | $1.36e-2$ | 17 | $9.46e-1$ | 14 | $1.36e-1$ | 20 | $9.75e-1$ | | 16.2 |
| WGAN_OS_AA | 06 | $8.91e-3$ | 09 | $1.27e-2$ | 11 | $9.56e-1$ | 09 | $1.27e-1$ | 13 | $9.79e-1$ | | 9.6 |
| WGAN_OS_AS | 05 | $8.89e-3$ | 10 | $1.28e-2$ | 12 | $9.55e-1$ | 10 | $1.28e-1$ | 14 | $9.79e-1$ | | 10.2 |
| WGAN_OS_SA | 11 | $9.46e-3$ | 12 | $1.30e-2$ | 10 | $9.58e-1$ | 12 | $1.30e-1$ | 12 | $9.82e-1$ | | 11.4 |
| WGAN_OS_SS | 10 | $9.44e-3$ | 11 | $1.29e-2$ | 09 | $9.59e-1$ | 11 | $1.29e-1$ | 11 | $9.82e-1$ | | 10.4 |
| WGAN_TS_AA | 02 | $7.58e-3$ | 03 | $1.05e-2$ | 04 | $9.69e-1$ | 03 | $1.05e-1$ | 04 | $9.85e-1$ | | 3.2 |
| WGAN_TS_AS | 01 | $7.48e-3$ | 01 | $1.04e-2$ | 01 | $9.71e-1$ | 01 | $1.04e-1$ | 02 | $9.86e-1$ | | 1.2 |
| WGAN_TS_SA | 03 | $7.85e-3$ | 02 | $1.04e-2$ | 02 | $9.71e-1$ | 02 | $1.04e-1$ | 01 | $9.86e-1$ | | 2.0 |
| WGAN_TS_SS | 04 | $7.98e-3$ | 04 | $1.05e-2$ | 03 | $9.70e-1$ | 04 | $1.05e-1$ | 03 | $9.86e-1$ | | 3.6 |
| WGGP_NO_AA | 28 | $1.64e-2$ | 28 | $2.21e-2$ | 28 | $8.58e-1$ | 28 | $2.21e-1$ | 28 | $9.39e-1$ | | 28.0 |
| WGGP_NO_AS | 27 | $1.63e-2$ | 27 | $2.21e-2$ | 25 | $8.61e-1$ | 27 | $2.21e-1$ | 26 | $9.40e-1$ | | 26.4 |
| WGGP_NO_SA | 26 | $1.61e-2$ | 26 | $2.14e-2$ | 26 | $8.59e-1$ | 26 | $2.14e-1$ | 25 | $9.40e-1$ | | 25.8 |
| WGGP_NO_SS | 25 | $1.61e-2$ | 25 | $2.13e-2$ | 27 | $8.58e-1$ | 25 | $2.13e-1$ | 27 | $9.40e-1$ | | 25.8 |
| WGGP_OS_AA | 36 | $3.67e-2$ | 36 | $5.46e-2$ | 36 | $-5.05e-1$ | 36 | $5.46e-1$ | 36 | $8.31e-1$ | | 36.0 |
| WGGP_OS_AS | 35 | $3.67e-2$ | 35 | $5.45e-2$ | 35 | $-5.01e-1$ | 35 | $5.45e-1$ | 35 | $8.32e-1$ | | 35.0 |
| WGGP_OS_SA | 34 | $3.01e-2$ | 34 | $4.18e-2$ | 34 | $3.68e-1$ | 34 | $4.18e-1$ | 34 | $8.70e-1$ | | 34.0 |
| WGGP_OS_SS | 33 | $2.99e-2$ | 33 | $4.15e-2$ | 33 | $3.74e-1$ | 33 | $4.15e-1$ | 33 | $8.72e-1$ | | 33.0 |
| WGGP_TS_AA | 32 | $2.40e-2$ | 32 | $3.32e-2$ | 31 | $7.38e-1$ | 32 | $3.32e-1$ | 31 | $9.12e-1$ | | 31.6 |
| WGGP_TS_AS | 31 | $2.39e-2$ | 31 | $3.31e-2$ | 32 | $7.37e-1$ | 31 | $3.31e-1$ | 32 | $9.12e-1$ | | 31.4 |
| WGGP_TS_SA | 30 | $2.02e-2$ | 30 | $2.81e-2$ | 30 | $8.12e-1$ | 30 | $2.81e-1$ | 29 | $9.22e-1$ | | 29.8 |
| WGGP_TS_SS | 29 | $2.02e-2$ | 29 | $2.80e-2$ | 29 | $8.12e-1$ | 29 | $2.80e-1$ | 30 | $9.22e-1$ | | 29.2 |

Table 8: Results of the statistics on the first aggregation level (categorical columns) for the CMAP dataset. Lighter grey tone corresponds to better results compared to darker ones.

| Name | | MAE | | RMSE | | $R^2$ | | SRMSE | | $\rho_{\text{Pearson}}$ | | rank |
|---|---|---|---|---|---|---|---|---|---|---|---|---|
| SGAN_NO_AA | 12 | $6.33e-3$ | 08 | $7.60e-3$ | 11 | $9.92e-1$ | 08 | $4.25e-2$ | 11 | $9.98e-1$ | | 10.0 |
| SGAN_NO_AS | 03 | $5.09e-3$ | 04 | $6.23e-3$ | 04 | $9.95e-1$ | 04 | $3.73e-2$ | 05 | $9.99e-1$ | | 4.0 |

Table 8 – continued from previous page

| Name | | MAE | | RMSE | | $R^2$ | | SRMSE | | $\rho_{\text{Pearson}}$ | | rank |
|---|---|---|---|---|---|---|---|---|---|---|---|---|
| SGAN_NO_SA | 06 | $5.83\text{e}-3$ | 07 | $7.12\text{e}-3$ | 03 | $9.95\text{e}-1$ | 07 | $4.12\text{e}-2$ | 07 | $9.99\text{e}-1$ | | 6.0 |
| SGAN_NO_SS | 04 | $5.10\text{e}-3$ | 03 | $6.17\text{e}-3$ | 05 | $9.94\text{e}-1$ | 03 | $3.62\text{e}-2$ | 04 | $9.99\text{e}-1$ | | 3.8 |
| SGAN_OS_AA | 21 | $9.36\text{e}-3$ | 23 | $1.15\text{e}-2$ | 23 | $9.82\text{e}-1$ | 23 | $6.63\text{e}-2$ | 26 | $9.97\text{e}-1$ | | 23.2 |
| SGAN_OS_AS | 33 | $3.90\text{e}-2$ | 33 | $4.69\text{e}-2$ | 28 | $8.60\text{e}-1$ | 32 | $2.99\text{e}-1$ | 24 | $9.97\text{e}-1$ | | 30.0 |
| SGAN_OS_SA | 22 | $9.40\text{e}-3$ | 24 | $1.15\text{e}-2$ | 24 | $9.82\text{e}-1$ | 24 | $6.68\text{e}-2$ | 28 | $9.97\text{e}-1$ | | 24.4 |
| SGAN_OS_SS | 31 | $3.89\text{e}-2$ | 31 | $4.69\text{e}-2$ | 27 | $8.61\text{e}-1$ | 31 | $2.98\text{e}-1$ | 25 | $9.97\text{e}-1$ | | 29.0 |
| SGAN_TS_AA | 14 | $6.59\text{e}-3$ | 14 | $8.03\text{e}-3$ | 09 | $9.93\text{e}-1$ | 14 | $4.78\text{e}-2$ | 20 | $9.98\text{e}-1$ | | 14.2 |
| SGAN_TS_AS | 10 | $6.32\text{e}-3$ | 09 | $7.65\text{e}-3$ | 06 | $9.94\text{e}-1$ | 09 | $4.50\text{e}-2$ | 14 | $9.98\text{e}-1$ | | 9.6 |
| SGAN_TS_SA | 11 | $6.32\text{e}-3$ | 10 | $7.73\text{e}-3$ | 07 | $9.93\text{e}-1$ | 11 | $4.60\text{e}-2$ | 18 | $9.98\text{e}-1$ | | 11.4 |
| SGAN_TS_SS | 13 | $6.47\text{e}-3$ | 13 | $7.88\text{e}-3$ | 08 | $9.93\text{e}-1$ | 12 | $4.60\text{e}-2$ | 17 | $9.98\text{e}-1$ | | 12.6 |
| WGAN_NO_AA | 24 | $9.48\text{e}-3$ | 21 | $1.13\text{e}-2$ | 18 | $9.90\text{e}-1$ | 21 | $6.46\text{e}-2$ | 03 | $9.99\text{e}-1$ | | 17.4 |
| WGAN_NO_AS | 01 | $4.43\text{e}-3$ | 01 | $5.42\text{e}-3$ | 01 | $9.96\text{e}-1$ | 02 | $3.28\text{e}-2$ | 02 | $9.99\text{e}-1$ | | 1.4 |
| WGAN_NO_SA | 23 | $9.45\text{e}-3$ | 22 | $1.13\text{e}-2$ | 20 | $9.89\text{e}-1$ | 22 | $6.49\text{e}-2$ | 06 | $9.99\text{e}-1$ | | 18.6 |
| WGAN_NO_SS | 02 | $4.56\text{e}-3$ | 02 | $5.46\text{e}-3$ | 02 | $9.95\text{e}-1$ | 01 | $3.22\text{e}-2$ | 01 | $9.99\text{e}-1$ | | 1.6 |
| WGAN_OS_AA | 18 | $7.25\text{e}-3$ | 18 | $8.67\text{e}-3$ | 17 | $9.90\text{e}-1$ | 17 | $5.04\text{e}-2$ | 21 | $9.98\text{e}-1$ | | 18.2 |
| WGAN_OS_AS | 35 | $3.93\text{e}-2$ | 35 | $4.74\text{e}-2$ | 30 | $8.56\text{e}-1$ | 33 | $3.00\text{e}-1$ | 27 | $9.97\text{e}-1$ | | 32.0 |
| WGAN_OS_SA | 17 | $7.19\text{e}-3$ | 17 | $8.64\text{e}-3$ | 19 | $9.90\text{e}-1$ | 18 | $5.08\text{e}-2$ | 22 | $9.98\text{e}-1$ | | 18.6 |
| WGAN_OS_SS | 36 | $3.95\text{e}-2$ | 36 | $4.76\text{e}-2$ | 32 | $8.54\text{e}-1$ | 36 | $3.02\text{e}-1$ | 23 | $9.97\text{e}-1$ | | 32.6 |
| WGAN_TS_AA | 20 | $7.79\text{e}-3$ | 20 | $9.34\text{e}-3$ | 21 | $9.89\text{e}-1$ | 20 | $5.37\text{e}-2$ | 12 | $9.98\text{e}-1$ | | 18.6 |
| WGAN_TS_AS | 05 | $5.77\text{e}-3$ | 05 | $6.89\text{e}-3$ | 16 | $9.91\text{e}-1$ | 05 | $3.95\text{e}-2$ | 09 | $9.98\text{e}-1$ | | 8.0 |
| WGAN_TS_SA | 19 | $7.57\text{e}-3$ | 19 | $9.08\text{e}-3$ | 22 | $9.89\text{e}-1$ | 19 | $5.19\text{e}-2$ | 10 | $9.98\text{e}-1$ | | 17.8 |
| WGAN_TS_SS | 07 | $5.92\text{e}-3$ | 06 | $7.06\text{e}-3$ | 13 | $9.92\text{e}-1$ | 06 | $4.01\text{e}-2$ | 08 | $9.98\text{e}-1$ | | 8.0 |
| WGGP_NO_AA | 16 | $6.84\text{e}-3$ | 15 | $8.19\text{e}-3$ | 14 | $9.92\text{e}-1$ | 15 | $4.80\text{e}-2$ | 16 | $9.98\text{e}-1$ | | 15.2 |
| WGGP_NO_AS | 08 | $6.27\text{e}-3$ | 11 | $7.74\text{e}-3$ | 10 | $9.92\text{e}-1$ | 10 | $4.55\text{e}-2$ | 13 | $9.98\text{e}-1$ | | 10.4 |
| WGGP_NO_SA | 15 | $6.78\text{e}-3$ | 16 | $8.19\text{e}-3$ | 12 | $9.92\text{e}-1$ | 16 | $4.84\text{e}-2$ | 19 | $9.98\text{e}-1$ | | 15.6 |
| WGGP_NO_SS | 09 | $6.31\text{e}-3$ | 12 | $7.81\text{e}-3$ | 15 | $9.91\text{e}-1$ | 13 | $4.61\text{e}-2$ | 15 | $9.98\text{e}-1$ | | 12.8 |
| WGGP_OS_AA | 30 | $2.39\text{e}-2$ | 30 | $2.95\text{e}-2$ | 26 | $8.81\text{e}-1$ | 30 | $1.79\text{e}-1$ | 36 | $9.93\text{e}-1$ | | 30.4 |
| WGGP_OS_AS | 32 | $3.89\text{e}-2$ | 32 | $4.69\text{e}-2$ | 29 | $8.57\text{e}-1$ | 34 | $3.00\text{e}-1$ | 29 | $9.96\text{e}-1$ | | 31.2 |
| WGGP_OS_SA | 29 | $2.37\text{e}-2$ | 29 | $2.91\text{e}-2$ | 25 | $8.81\text{e}-1$ | 29 | $1.77\text{e}-1$ | 35 | $9.93\text{e}-1$ | | 29.4 |
| WGGP_OS_SS | 34 | $3.90\text{e}-2$ | 34 | $4.70\text{e}-2$ | 31 | $8.54\text{e}-1$ | 35 | $3.01\text{e}-1$ | 30 | $9.96\text{e}-1$ | | 32.8 |
| WGGP_TS_AA | 28 | $1.53\text{e}-2$ | 28 | $1.74\text{e}-2$ | 36 | $5.62\text{e}-1$ | 27 | $8.60\text{e}-2$ | 31 | $9.95\text{e}-1$ | | 30.0 |
| WGGP_TS_AS | 25 | $1.33\text{e}-2$ | 26 | $1.53\text{e}-2$ | 34 | $5.85\text{e}-1$ | 26 | $7.45\text{e}-2$ | 34 | $9.95\text{e}-1$ | | 29.0 |
| WGGP_TS_SA | 27 | $1.52\text{e}-2$ | 27 | $1.73\text{e}-2$ | 35 | $5.72\text{e}-1$ | 28 | $8.61\text{e}-2$ | 32 | $9.95\text{e}-1$ | | 29.8 |
| WGGP_TS_SS | 26 | $1.33\text{e}-2$ | 25 | $1.52\text{e}-2$ | 33 | $5.89\text{e}-1$ | 25 | $7.34\text{e}-2$ | 33 | $9.95\text{e}-1$ | | 28.4 |

Table 9: Results of the statistics on the second aggregation level for the CMAP dataset. Lighter grey tone corresponds to better results compared to darker ones.

| Name | | MAE | | RMSE | | $R^2$ | | SRMSE | | $\rho_{\text{Pearson}}$ | | rank |
|---|---|---|---|---|---|---|---|---|---|---|---|---|
| SGAN_NO_AA | 14 | $3.19\text{e}-3$ | 14 | $4.95\text{e}-3$ | 16 | $9.72\text{e}-1$ | 15 | $1.86\text{e}-1$ | 19 | $9.87\text{e}-1$ | | 15.6 |
| SGAN_NO_AS | 07 | $3.09\text{e}-3$ | 08 | $4.77\text{e}-3$ | 15 | $9.73\text{e}-1$ | 13 | $1.82\text{e}-1$ | 17 | $9.87\text{e}-1$ | | 12.0 |
| SGAN_NO_SA | 05 | $3.00\text{e}-3$ | 06 | $4.66\text{e}-3$ | 12 | $9.76\text{e}-1$ | 09 | $1.75\text{e}-1$ | 12 | $9.89\text{e}-1$ | | 8.8 |
| SGAN_NO_SS | 03 | $2.96\text{e}-3$ | 03 | $4.54\text{e}-3$ | 11 | $9.77\text{e}-1$ | 05 | $1.71\text{e}-1$ | 11 | $9.89\text{e}-1$ | | 6.6 |
| SGAN_OS_AA | 24 | $3.87\text{e}-3$ | 24 | $6.08\text{e}-3$ | 20 | $9.66\text{e}-1$ | 24 | $2.17\text{e}-1$ | 20 | $9.86\text{e}-1$ | | 22.4 |
| SGAN_OS_AS | 34 | $1.18\text{e}-2$ | 34 | $1.80\text{e}-2$ | 32 | $8.16\text{e}-1$ | 33 | $6.25\text{e}-1$ | 30 | $9.63\text{e}-1$ | | 32.6 |

Table 9 – continued from previous page

| Name | | MAE | | RMSE | | $R^2$ | | SRMSE | | $\rho_{\text{Pearson}}$ | | rank |
|---|---|---|---|---|---|---|---|---|---|---|---|---|
| SGAN_OS_SA | 23 | $3.83e-3$ | 23 | $6.01e-3$ | 19 | $9.68e-1$ | 23 | $2.13e-1$ | 14 | $9.88e-1$ | | 20.4 |
| SGAN_OS_SS | 33 | $1.18e-2$ | 33 | $1.79e-2$ | 31 | $8.19e-1$ | 32 | $6.21e-1$ | 28 | $9.65e-1$ | | 31.4 |
| SGAN_TS_AA | 12 | $3.17e-3$ | 13 | $4.83e-3$ | 10 | $9.77e-1$ | 11 | $1.77e-1$ | 10 | $9.89e-1$ | | 11.2 |
| SGAN_TS_AS | 10 | $3.15e-3$ | 11 | $4.81e-3$ | 07 | $9.78e-1$ | 12 | $1.78e-1$ | 08 | $9.90e-1$ | | 9.6 |
| SGAN_TS_SA | 08 | $3.09e-3$ | 07 | $4.73e-3$ | 05 | $9.78e-1$ | 06 | $1.74e-1$ | 05 | $9.90e-1$ | | 6.2 |
| SGAN_TS_SS | 11 | $3.16e-3$ | 12 | $4.82e-3$ | 06 | $9.78e-1$ | 10 | $1.77e-1$ | 06 | $9.90e-1$ | | 9.0 |
| WGAN_NO_AA | 22 | $3.68e-3$ | 22 | $5.69e-3$ | 18 | $9.70e-1$ | 19 | $2.03e-1$ | 16 | $9.87e-1$ | | 19.4 |
| WGAN_NO_AS | 15 | $3.28e-3$ | 16 | $5.02e-3$ | 14 | $9.73e-1$ | 16 | $1.86e-1$ | 18 | $9.87e-1$ | | 15.8 |
| WGAN_NO_SA | 21 | $3.66e-3$ | 21 | $5.65e-3$ | 17 | $9.72e-1$ | 18 | $2.02e-1$ | 13 | $9.88e-1$ | | 18.0 |
| WGAN_NO_SS | 16 | $3.31e-3$ | 15 | $5.02e-3$ | 13 | $9.74e-1$ | 14 | $1.85e-1$ | 15 | $9.88e-1$ | | 14.6 |
| WGAN_OS_AA | 09 | $3.12e-3$ | 09 | $4.77e-3$ | 09 | $9.78e-1$ | 07 | $1.74e-1$ | 09 | $9.89e-1$ | | 8.6 |
| WGAN_OS_AS | 31 | $1.16e-2$ | 31 | $1.77e-2$ | 29 | $8.23e-1$ | 30 | $6.15e-1$ | 25 | $9.69e-1$ | | 29.2 |
| WGAN_OS_SA | 13 | $3.18e-3$ | 10 | $4.80e-3$ | 08 | $9.78e-1$ | 08 | $1.75e-1$ | 07 | $9.90e-1$ | | 9.2 |
| WGAN_OS_SS | 32 | $1.16e-2$ | 32 | $1.78e-2$ | 30 | $8.22e-1$ | 31 | $6.17e-1$ | 26 | $9.69e-1$ | | 30.2 |
| WGAN_TS_AA | 06 | $3.01e-3$ | 05 | $4.63e-3$ | 04 | $9.81e-1$ | 04 | $1.64e-1$ | 03 | $9.92e-1$ | | 4.4 |
| WGAN_TS_AS | 01 | $2.78e-3$ | 01 | $4.21e-3$ | 01 | $9.83e-1$ | 01 | $1.53e-1$ | 02 | $9.92e-1$ | | 1.2 |
| WGAN_TS_SA | 04 | $2.98e-3$ | 04 | $4.54e-3$ | 03 | $9.82e-1$ | 03 | $1.61e-1$ | 01 | $9.92e-1$ | | 3.0 |
| WGAN_TS_SS | 02 | $2.83e-3$ | 02 | $4.27e-3$ | 02 | $9.83e-1$ | 02 | $1.56e-1$ | 04 | $9.92e-1$ | | 2.4 |
| WGGP_NO_AA | 20 | $3.52e-3$ | 20 | $5.33e-3$ | 24 | $9.57e-1$ | 21 | $2.03e-1$ | 24 | $9.81e-1$ | | 21.8 |
| WGGP_NO_AS | 17 | $3.42e-3$ | 18 | $5.30e-3$ | 21 | $9.58e-1$ | 20 | $2.03e-1$ | 21 | $9.81e-1$ | | 19.4 |
| WGGP_NO_SA | 19 | $3.48e-3$ | 17 | $5.27e-3$ | 22 | $9.57e-1$ | 17 | $2.02e-1$ | 22 | $9.81e-1$ | | 19.4 |
| WGGP_NO_SS | 18 | $3.42e-3$ | 19 | $5.31e-3$ | 23 | $9.57e-1$ | 22 | $2.04e-1$ | 23 | $9.81e-1$ | | 21.0 |
| WGGP_OS_AA | 30 | $9.46e-3$ | 30 | $1.69e-2$ | 36 | $4.37e-1$ | 34 | $6.40e-1$ | 34 | $9.31e-1$ | | 32.8 |
| WGGP_OS_AS | 36 | $1.35e-2$ | 36 | $2.13e-2$ | 35 | $5.23e-1$ | 36 | $7.78e-1$ | 36 | $9.18e-1$ | | 35.8 |
| WGGP_OS_SA | 29 | $8.87e-3$ | 29 | $1.53e-2$ | 33 | $7.06e-1$ | 29 | $5.61e-1$ | 33 | $9.44e-1$ | | 30.6 |
| WGGP_OS_SS | 35 | $1.29e-2$ | 35 | $2.00e-2$ | 34 | $7.05e-1$ | 35 | $7.20e-1$ | 35 | $9.31e-1$ | | 34.8 |
| WGGP_TS_AA | 28 | $6.30e-3$ | 28 | $9.80e-3$ | 28 | $8.91e-1$ | 28 | $3.31e-1$ | 31 | $9.62e-1$ | | 28.6 |
| WGGP_TS_AS | 27 | $6.08e-3$ | 27 | $9.39e-3$ | 27 | $8.97e-1$ | 27 | $3.19e-1$ | 32 | $9.61e-1$ | | 28.0 |
| WGGP_TS_SA | 26 | $6.04e-3$ | 26 | $9.29e-3$ | 26 | $9.07e-1$ | 26 | $3.10e-1$ | 27 | $9.65e-1$ | | 26.2 |
| WGGP_TS_SS | 25 | $5.80e-3$ | 25 | $8.88e-3$ | 25 | $9.12e-1$ | 25 | $2.99e-1$ | 29 | $9.64e-1$ | | 25.8 |

Table 10: Results of the statistics on the third aggregation level for the CMAP dataset. Lighter grey tone corresponds to better results compared to darker ones.

| Name | | MAE | | RMSE | | $R^2$ | | SRMSE | | $\rho_{\text{Pearson}}$ | | rank |
|---|---|---|---|---|---|---|---|---|---|---|---|---|
| SGAN_NO_AA | 13 | $1.19e-3$ | 14 | $2.19e-3$ | 14 | $9.51e-1$ | 14 | $3.75e-1$ | 16 | $9.77e-1$ | | 14.2 |
| SGAN_NO_AS | 10 | $1.19e-3$ | 12 | $2.14e-3$ | 13 | $9.53e-1$ | 13 | $3.70e-1$ | 15 | $9.77e-1$ | | 12.6 |
| SGAN_NO_SA | 06 | $1.15e-3$ | 08 | $2.09e-3$ | 12 | $9.57e-1$ | 10 | $3.58e-1$ | 12 | $9.80e-1$ | | 9.6 |
| SGAN_NO_SS | 05 | $1.15e-3$ | 05 | $2.06e-3$ | 10 | $9.58e-1$ | 08 | $3.53e-1$ | 11 | $9.80e-1$ | | 7.8 |
| SGAN_OS_AA | 24 | $1.37e-3$ | 24 | $2.58e-3$ | 20 | $9.42e-1$ | 24 | $4.16e-1$ | 18 | $9.76e-1$ | | 22.0 |
| SGAN_OS_AS | 34 | $3.30e-3$ | 33 | $6.42e-3$ | 32 | $7.42e-1$ | 33 | $1.14$ | 32 | $9.27e-1$ | | 32.8 |
| SGAN_OS_SA | 23 | $1.36e-3$ | 23 | $2.53e-3$ | 19 | $9.46e-1$ | 23 | $4.06e-1$ | 14 | $9.78e-1$ | | 20.4 |
| SGAN_OS_SS | 33 | $3.30e-3$ | 32 | $6.39e-3$ | 31 | $7.45e-1$ | 32 | $1.14$ | 31 | $9.28e-1$ | | 31.8 |
| SGAN_TS_AA | 14 | $1.20e-3$ | 13 | $2.15e-3$ | 11 | $9.57e-1$ | 09 | $3.58e-1$ | 10 | $9.80e-1$ | | 11.4 |
| SGAN_TS_AS | 12 | $1.19e-3$ | 11 | $2.13e-3$ | 09 | $9.58e-1$ | 12 | $3.66e-1$ | 09 | $9.80e-1$ | | 10.6 |
| SGAN_TS_SA | 08 | $1.18e-3$ | 09 | $2.10e-3$ | 07 | $9.59e-1$ | 07 | $3.50e-1$ | 05 | $9.81e-1$ | | 7.2 |

Table 10 – continued from previous page

| Name | MAE | | RMSE | | $R^2$ | | SRMSE | | $\rho_{\text{Pearson}}$ | | rank |
|---|---|---|---|---|---|---|---|---|---|---|---|
| SGAN_TS_SS | **09** | $1.18e-3$ | **10** | $2.11e-3$ | **08** | $9.59e-1$ | **11** | $3.62e-1$ | **07** | $9.81e-1$ | **9.0** |
| WGAN_NO_AA | 21 | $1.32e-3$ | 22 | $2.39e-3$ | 17 | $9.50e-1$ | 18 | $3.90e-1$ | 17 | $9.77e-1$ | 19.0 |
| WGAN_NO_AS | 20 | $1.31e-3$ | 20 | $2.36e-3$ | 18 | $9.49e-1$ | 22 | $4.03e-1$ | 20 | $9.75e-1$ | 20.0 |
| WGAN_NO_SA | 19 | $1.31e-3$ | 21 | $2.38e-3$ | 15 | $9.51e-1$ | 17 | $3.89e-1$ | 13 | $9.78e-1$ | 17.0 |
| WGAN_NO_SS | 22 | $1.32e-3$ | 19 | $2.35e-3$ | 16 | $9.50e-1$ | 21 | $4.01e-1$ | 19 | $9.76e-1$ | 19.4 |
| WGAN_OS_AA | **07** | $1.17e-3$ | **06** | $2.07e-3$ | **06** | $9.60e-1$ | **05** | $3.48e-1$ | **08** | $9.80e-1$ | **6.4** |
| WGAN_OS_AS | 31 | $3.21e-3$ | 30 | $6.26e-3$ | 29 | $7.52e-1$ | 30 | $1.13$ | 29 | $9.37e-1$ | 29.8 |
| WGAN_OS_SA | 11 | $1.19e-3$ | **07** | $2.08e-3$ | **05** | $9.60e-1$ | **06** | $3.48e-1$ | **06** | $9.81e-1$ | **7.0** |
| WGAN_OS_SS | 32 | $3.23e-3$ | 31 | $6.28e-3$ | 30 | $7.51e-1$ | 31 | $1.13$ | 30 | $9.36e-1$ | 30.8 |
| WGAN_TS_AA | **04** | $1.11e-3$ | **04** | $1.97e-3$ | **04** | $9.65e-1$ | **02** | $3.26e-1$ | **02** | $9.84e-1$ | **3.2** |
| WGAN_TS_AS | **01** | $1.08e-3$ | **01** | $1.90e-3$ | **01** | $9.66e-1$ | **03** | $3.27e-1$ | **03** | $9.83e-1$ | **1.8** |
| WGAN_TS_SA | **03** | $1.10e-3$ | **03** | $1.95e-3$ | **02** | $9.66e-1$ | **01** | $3.22e-1$ | **01** | $9.84e-1$ | **2.0** |
| WGAN_TS_SS | **02** | $1.09e-3$ | **02** | $1.92e-3$ | **03** | $9.65e-1$ | **04** | $3.31e-1$ | **04** | $9.83e-1$ | **3.0** |
| WGGP_NO_AA | 18 | $1.30e-3$ | 18 | $2.29e-3$ | 24 | $9.37e-1$ | 16 | $3.84e-1$ | 24 | $9.70e-1$ | 20.0 |
| WGGP_NO_AS | 16 | $1.26e-3$ | 17 | $2.27e-3$ | 21 | $9.39e-1$ | 19 | $3.92e-1$ | 21 | $9.71e-1$ | 18.8 |
| WGGP_NO_SA | 17 | $1.28e-3$ | 15 | $2.26e-3$ | 23 | $9.38e-1$ | 15 | $3.83e-1$ | 22 | $9.71e-1$ | 18.4 |
| WGGP_NO_SS | 15 | $1.25e-3$ | 16 | $2.27e-3$ | 22 | $9.38e-1$ | 20 | $3.96e-1$ | 23 | $9.70e-1$ | 19.2 |
| WGGP_OS_AA | 30 | $2.92e-3$ | 34 | $7.01e-3$ | 36 | $9.66e-2$ | 34 | $1.19$ | 34 | $8.97e-1$ | 33.6 |
| WGGP_OS_AS | 36 | $3.79e-3$ | 36 | $7.52e-3$ | 35 | $4.50e-1$ | 36 | $1.36$ | 36 | $8.67e-1$ | 35.8 |
| WGGP_OS_SA | 29 | $2.71e-3$ | 29 | $6.18e-3$ | 34 | $5.38e-1$ | 29 | $1.03$ | 33 | $9.15e-1$ | 30.8 |
| WGGP_OS_SS | 35 | $3.60e-3$ | 35 | $7.04e-3$ | 33 | $6.35e-1$ | 35 | $1.27$ | 35 | $8.84e-1$ | 34.6 |
| WGGP_TS_AA | 28 | $2.06e-3$ | 28 | $3.93e-3$ | 28 | $8.57e-1$ | 28 | $5.92e-1$ | 27 | $9.44e-1$ | 27.8 |
| WGGP_TS_AS | 27 | $1.97e-3$ | 27 | $3.71e-3$ | 27 | $8.70e-1$ | 27 | $5.85e-1$ | 28 | $9.43e-1$ | 27.2 |
| WGGP_TS_SA | 26 | $1.97e-3$ | 26 | $3.70e-3$ | 26 | $8.77e-1$ | 26 | $5.66e-1$ | 25 | $9.48e-1$ | 25.8 |
| WGGP_TS_SS | 25 | $1.88e-3$ | 25 | $3.51e-3$ | 25 | $8.85e-1$ | 25 | $5.65e-1$ | 26 | $9.46e-1$ | 25.2 |

Table 11: Results of the Machine Learning efficacy for the CMAP dataset. Lighter grey tone corresponds to better results compared to darker ones.

| Name | Continuous | | Categorical | | rank |
|---|---|---|---|---|---|
| SGAN_NO_AA | **02** | 3.04 | **02** | $6.23e-1$ | **2.0** |
| SGAN_NO_AS | **04** | 3.05 | **04** | $6.51e-1$ | 4.0 |
| SGAN_NO_SA | **06** | 3.06 | **01** | $6.15e-1$ | 3.5 |
| SGAN_NO_SS | **09** | 3.06 | **03** | $6.38e-1$ | 6.0 |
| SGAN_OS_AA | 13 | 3.07 | 27 | $1.61e2$ | 20.0 |
| SGAN_OS_AS | 27 | 3.38 | 23 | 2.95 | 25.0 |
| SGAN_OS_SA | 17 | 3.13 | 28 | $4.01e2$ | 22.5 |
| SGAN_OS_SS | 28 | 3.43 | 24 | 2.97 | 26.0 |
| SGAN_TS_AA | **10** | 3.06 | 11 | $7.25e-1$ | 10.5 |
| SGAN_TS_AS | **07** | 3.06 | 12 | $7.34e-1$ | **9.5** |
| SGAN_TS_SA | 12 | 3.07 | **09** | $7.02e-1$ | 10.5 |
| SGAN_TS_SS | 14 | 3.08 | **10** | $7.22e-1$ | 12.0 |
| WGAN_NO_AA | 21 | 3.16 | 32 | $7.04e3$ | 26.5 |
| WGAN_NO_AS | 19 | 3.14 | 17 | 1.18 | 18.0 |
| WGAN_NO_SA | 23 | 3.17 | 31 | $6.64e3$ | 27.0 |
| WGAN_NO_SS | 22 | 3.17 | 18 | 1.19 | 20.0 |

Table 11 – continued from previous page

| Name | Continuous | | Categorical | | rank |
|---|---|---|---|---|---|
| WGAN_OS_AA | 11 | 3.07 | 15 | $8.78e-1$ | 13.0 |
| WGAN_OS_AS | 25 | 3.32 | 21 | 2.80 | 23.0 |
| WGAN_OS_SA | 15 | 3.08 | 16 | $8.92e-1$ | 15.5 |
| WGAN_OS_SS | 26 | 3.34 | 22 | 2.82 | 24.0 |
| WGAN_TS_AA | **01** | 3.04 | **07** | $6.78e-1$ | **4.0** |
| WGAN_TS_AS | **03** | 3.04 | 13 | $7.47e-1$ | **8.0** |
| WGAN_TS_SA | **08** | 3.06 | **08** | $6.87e-1$ | **8.0** |
| WGAN_TS_SS | **05** | 3.05 | 14 | $7.56e-1$ | **9.5** |
| WGGP_NO_AA | 20 | 3.15 | 35 | 1.13e4 | 27.5 |
| WGGP_NO_AS | 16 | 3.12 | **06** | $6.72e-1$ | 11.0 |
| WGGP_NO_SA | 24 | 3.18 | 36 | 1.23e4 | 30.0 |
| WGGP_NO_SS | 18 | 3.13 | **05** | $6.71e-1$ | 11.5 |
| WGGP_OS_AA | 30 | 3.89 | 30 | 4.48e3 | 30.0 |
| WGGP_OS_AS | 36 | 4.50 | 25 | 3.10 | 30.5 |
| WGGP_OS_SA | 29 | 3.87 | 29 | 4.08e3 | 29.0 |
| WGGP_OS_SS | 35 | 4.25 | 26 | 3.30 | 30.5 |
| WGGP_TS_AA | 33 | 4.10 | 33 | 8.00e3 | 33.0 |
| WGGP_TS_AS | 34 | 4.15 | 20 | 1.28 | 27.0 |
| WGGP_TS_SA | 32 | 3.94 | 34 | 8.80e3 | 33.0 |
| WGGP_TS_SS | 31 | 3.91 | 19 | 1.20 | 25.0 |

### 5.1.2 LPMC case study

Table 12: Results of the statistics on the first aggregation level (all columns) for the LPMC dataset. Lighter grey tone corresponds to better results compared to darker ones.

| Name | MAE | | RMSE | | $R^2$ | | SRMSE | | $\rho_{\text{Pearson}}$ | | rank |
|---|---|---|---|---|---|---|---|---|---|---|---|
| SGAN_NO_AA | 22 | $1.26e-2$ | 22 | $1.74e-2$ | 24 | $8.72e-1$ | 22 | $1.36e-1$ | 24 | $9.69e-1$ | 22.8 |
| SGAN_NO_AS | 16 | $9.78e-3$ | 16 | $1.40e-2$ | 16 | $9.06e-1$ | 18 | $1.22e-1$ | 22 | $9.73e-1$ | 17.6 |
| SGAN_NO_SA | 20 | $1.22e-2$ | 20 | $1.65e-2$ | 21 | $8.86e-1$ | 19 | $1.27e-1$ | 20 | $9.74e-1$ | 20.0 |
| SGAN_NO_SS | 13 | $9.40e-3$ | 14 | $1.32e-2$ | 14 | $9.17e-1$ | 14 | $1.13e-1$ | 21 | $9.73e-1$ | 15.2 |
| SGAN_OS_AA | 24 | $1.41e-2$ | 24 | $1.84e-2$ | 30 | $3.91e-1$ | 23 | $1.43e-1$ | 36 | $9.36e-1$ | 27.4 |
| SGAN_OS_AS | 34 | $2.34e-2$ | 33 | $2.89e-2$ | 22 | $8.85e-1$ | 33 | $1.81e-1$ | 32 | $9.62e-1$ | 30.8 |
| SGAN_OS_SA | 23 | $1.31e-2$ | 21 | $1.68e-2$ | 29 | $4.07e-1$ | 20 | $1.27e-1$ | 35 | $9.40e-1$ | 25.6 |
| SGAN_OS_SS | 31 | $2.24e-2$ | 31 | $2.74e-2$ | 17 | $8.95e-1$ | 29 | $1.66e-1$ | 29 | $9.65e-1$ | 27.4 |
| SGAN_TS_AA | 15 | $9.71e-3$ | 13 | $1.28e-2$ | 15 | $9.13e-1$ | **09** | $1.00e-1$ | 18 | $9.74e-1$ | 14.0 |
| SGAN_TS_AS | **05** | $8.22e-3$ | **05** | $1.10e-2$ | **09** | $9.27e-1$ | **04** | $9.24e-2$ | 14 | $9.76e-1$ | **7.4** |
| SGAN_TS_SA | 14 | $9.41e-3$ | 11 | $1.25e-2$ | 13 | $9.17e-1$ | **06** | $9.75e-2$ | 13 | $9.76e-1$ | 11.4 |
| SGAN_TS_SS | **02** | $7.94e-3$ | **02** | $1.07e-2$ | **08** | $9.29e-1$ | **03** | $8.99e-2$ | 12 | $9.77e-1$ | **5.4** |
| WGAN_NO_AA | 36 | $2.42e-2$ | 36 | $3.25e-2$ | 26 | $8.01e-1$ | 36 | $2.30e-1$ | 17 | $9.75e-1$ | 30.2 |
| WGAN_NO_AS | 18 | $1.17e-2$ | 23 | $1.77e-2$ | 12 | $9.20e-1$ | 28 | $1.65e-1$ | 15 | $9.76e-1$ | 19.2 |
| WGAN_NO_SA | 32 | $2.25e-2$ | 34 | $2.93e-2$ | 25 | $8.22e-1$ | 35 | $1.98e-1$ | **06** | $9.80e-1$ | 26.4 |
| WGAN_NO_SS | 17 | $9.98e-3$ | 17 | $1.45e-2$ | **05** | $9.41e-1$ | 21 | $1.33e-1$ | **05** | $9.81e-1$ | 13.0 |
| WGAN_OS_AA | 26 | $1.61e-2$ | 26 | $2.12e-2$ | 32 | $2.09e-1$ | 27 | $1.63e-1$ | 26 | $9.67e-1$ | 27.4 |
| WGAN_OS_AS | 35 | $2.40e-2$ | 35 | $2.98e-2$ | 23 | $8.76e-1$ | 34 | $1.89e-1$ | 33 | $9.60e-1$ | 32.0 |
| WGAN_OS_SA | 25 | $1.52e-2$ | 25 | $1.97e-2$ | 31 | $2.21e-1$ | 24 | $1.48e-1$ | 31 | $9.63e-1$ | 27.2 |

| Name | | MAE | | RMSE | | $R^2$ | | SRMSE | | $\rho_{\text{Pearson}}$ | | rank |
|---|---|---|---|---|---|---|---|---|---|---|---|---|
| WGAN_OS_SS | 33 | $2.29\text{e}-2$ | 32 | $2.81\text{e}-2$ | 18 | $8.92\text{e}-1$ | 30 | $1.72\text{e}-1$ | 25 | $9.68\text{e}-1$ | | 27.6 |
| WGAN_TS_AA | 28 | $1.71\text{e}-2$ | 28 | $2.26\text{e}-2$ | 36 | $-7.22\text{e}-1$ | 32 | $1.79\text{e}-1$ | 16 | $9.75\text{e}-1$ | | 28.0 |
| WGAN_TS_AS | 11 | $9.03\text{e}-3$ | 15 | $1.32\text{e}-2$ | 04 | $9.49\text{e}-1$ | 16 | $1.18\text{e}-1$ | 03 | $9.85\text{e}-1$ | | 9.8 |
| WGAN_TS_SA | 27 | $1.70\text{e}-2$ | 27 | $2.21\text{e}-2$ | 35 | $-7.16\text{e}-1$ | 31 | $1.74\text{e}-1$ | 10 | $9.79\text{e}-1$ | | 26.0 |
| WGAN_TS_SS | 10 | $8.98\text{e}-3$ | 12 | $1.27\text{e}-2$ | 02 | $9.59\text{e}-1$ | 13 | $1.12\text{e}-1$ | 01 | $9.88\text{e}-1$ | | 7.6 |
| WGGP_NO_AA | 09 | $8.80\text{e}-3$ | 09 | $1.18\text{e}-2$ | 10 | $9.22\text{e}-1$ | 10 | $1.00\text{e}-1$ | 08 | $9.79\text{e}-1$ | | 9.2 |
| WGGP_NO_AS | 04 | $8.09\text{e}-3$ | 06 | $1.11\text{e}-2$ | 06 | $9.37\text{e}-1$ | 05 | $9.53\text{e}-2$ | 11 | $9.78\text{e}-1$ | | 6.4 |
| WGGP_NO_SA | 12 | $9.11\text{e}-3$ | 10 | $1.23\text{e}-2$ | 11 | $9.22\text{e}-1$ | 12 | $1.05\text{e}-1$ | 07 | $9.80\text{e}-1$ | | 10.4 |
| WGGP_NO_SS | 07 | $8.48\text{e}-3$ | 08 | $1.17\text{e}-2$ | 07 | $9.36\text{e}-1$ | 08 | $1.00\text{e}-1$ | 09 | $9.79\text{e}-1$ | | 7.8 |
| WGGP_OS_AA | 21 | $1.25\text{e}-2$ | 19 | $1.59\text{e}-2$ | 28 | $5.81\text{e}-1$ | 17 | $1.20\text{e}-1$ | 34 | $9.56\text{e}-1$ | | 23.8 |
| WGGP_OS_AS | 30 | $2.20\text{e}-2$ | 30 | $2.68\text{e}-2$ | 20 | $8.87\text{e}-1$ | 26 | $1.62\text{e}-1$ | 28 | $9.65\text{e}-1$ | | 26.8 |
| WGGP_OS_SA | 19 | $1.21\text{e}-2$ | 18 | $1.55\text{e}-2$ | 27 | $5.89\text{e}-1$ | 15 | $1.15\text{e}-1$ | 27 | $9.65\text{e}-1$ | | 21.2 |
| WGGP_OS_SS | 29 | $2.17\text{e}-2$ | 29 | $2.64\text{e}-2$ | 19 | $8.90\text{e}-1$ | 25 | $1.58\text{e}-1$ | 30 | $9.63\text{e}-1$ | | 26.4 |
| WGGP_TS_AA | 08 | $8.59\text{e}-3$ | 07 | $1.14\text{e}-2$ | 33 | $2.03\text{e}-1$ | 11 | $1.04\text{e}-1$ | 23 | $9.72\text{e}-1$ | | 16.4 |
| WGGP_TS_AS | 06 | $8.26\text{e}-3$ | 04 | $1.09\text{e}-2$ | 03 | $9.56\text{e}-1$ | 02 | $8.77\text{e}-2$ | 04 | $9.82\text{e}-1$ | | 3.8 |
| WGGP_TS_SA | 03 | $8.04\text{e}-3$ | 03 | $1.07\text{e}-2$ | 34 | $2.01\text{e}-1$ | 07 | $9.77\text{e}-2$ | 19 | $9.74\text{e}-1$ | | 13.2 |
| WGGP_TS_SS | 01 | $7.64\text{e}-3$ | 01 | $1.02\text{e}-2$ | 01 | $9.63\text{e}-1$ | 01 | $8.13\text{e}-2$ | 02 | $9.85\text{e}-1$ | | 1.2 |

Table 13: Results of the statistics on the first aggregation level (continuous columns) for the LPMC dataset. Lighter grey tone corresponds to better results compared to darker ones.

| Name | | MAE | | RMSE | | $R^2$ | | SRMSE | | $\rho_{\text{Pearson}}$ | | rank |
|---|---|---|---|---|---|---|---|---|---|---|---|---|
| SGAN_NO_AA | 30 | $1.39\text{e}-2$ | 30 | $2.16\text{e}-2$ | 33 | $8.87\text{e}-1$ | 30 | $2.16\text{e}-1$ | 33 | $9.67\text{e}-1$ | | 31.2 |
| SGAN_NO_AS | 31 | $1.39\text{e}-2$ | 29 | $2.16\text{e}-2$ | 34 | $8.86\text{e}-1$ | 29 | $2.16\text{e}-1$ | 34 | $9.67\text{e}-1$ | | 31.4 |
| SGAN_NO_SA | 23 | $1.28\text{e}-2$ | 23 | $1.96\text{e}-2$ | 31 | $9.18\text{e}-1$ | 23 | $1.96\text{e}-1$ | 27 | $9.75\text{e}-1$ | | 25.4 |
| SGAN_NO_SS | 24 | $1.29\text{e}-2$ | 24 | $1.97\text{e}-2$ | 32 | $9.17\text{e}-1$ | 24 | $1.97\text{e}-1$ | 28 | $9.75\text{e}-1$ | | 26.4 |
| SGAN_OS_AA | 22 | $1.28\text{e}-2$ | 22 | $1.91\text{e}-2$ | 21 | $9.36\text{e}-1$ | 22 | $1.91\text{e}-1$ | 23 | $9.76\text{e}-1$ | | 22.0 |
| SGAN_OS_AS | 21 | $1.28\text{e}-2$ | 21 | $1.90\text{e}-2$ | 23 | $9.36\text{e}-1$ | 21 | $1.90\text{e}-1$ | 24 | $9.76\text{e}-1$ | | 22.0 |
| SGAN_OS_SA | 14 | $1.07\text{e}-2$ | 14 | $1.58\text{e}-2$ | 11 | $9.56\text{e}-1$ | 14 | $1.58\text{e}-1$ | 08 | $9.84\text{e}-1$ | | 12.2 |
| SGAN_OS_SS | 13 | $1.07\text{e}-2$ | 13 | $1.58\text{e}-2$ | 12 | $9.56\text{e}-1$ | 13 | $1.58\text{e}-1$ | 07 | $9.84\text{e}-1$ | | 11.6 |
| SGAN_TS_AA | 12 | $1.05\text{e}-2$ | 11 | $1.51\text{e}-2$ | 18 | $9.47\text{e}-1$ | 11 | $1.51\text{e}-1$ | 16 | $9.80\text{e}-1$ | | 13.6 |
| SGAN_TS_AS | 11 | $1.05\text{e}-2$ | 12 | $1.52\text{e}-2$ | 17 | $9.47\text{e}-1$ | 12 | $1.52\text{e}-1$ | 15 | $9.80\text{e}-1$ | | 13.4 |
| SGAN_TS_SA | 07 | $9.97\text{e}-3$ | 07 | $1.46\text{e}-2$ | 13 | $9.54\text{e}-1$ | 07 | $1.46\text{e}-1$ | 10 | $9.83\text{e}-1$ | | 8.8 |
| SGAN_TS_SS | 08 | $9.97\text{e}-3$ | 08 | $1.47\text{e}-2$ | 14 | $9.54\text{e}-1$ | 08 | $1.47\text{e}-1$ | 09 | $9.83\text{e}-1$ | | 9.4 |
| WGAN_NO_AA | 36 | $1.99\text{e}-2$ | 36 | $3.18\text{e}-2$ | 35 | $8.75\text{e}-1$ | 36 | $3.18\text{e}-1$ | 35 | $9.64\text{e}-1$ | | 35.6 |
| WGAN_NO_AS | 35 | $1.99\text{e}-2$ | 35 | $3.17\text{e}-2$ | 36 | $8.75\text{e}-1$ | 35 | $3.17\text{e}-1$ | 36 | $9.64\text{e}-1$ | | 35.4 |
| WGAN_NO_SA | 34 | $1.65\text{e}-2$ | 34 | $2.51\text{e}-2$ | 30 | $9.25\text{e}-1$ | 34 | $2.51\text{e}-1$ | 25 | $9.76\text{e}-1$ | | 31.4 |
| WGAN_NO_SS | 33 | $1.64\text{e}-2$ | 33 | $2.50\text{e}-2$ | 29 | $9.25\text{e}-1$ | 33 | $2.50\text{e}-1$ | 26 | $9.76\text{e}-1$ | | 30.8 |
| WGAN_OS_AA | 26 | $1.35\text{e}-2$ | 26 | $2.04\text{e}-2$ | 24 | $9.36\text{e}-1$ | 26 | $2.04\text{e}-1$ | 20 | $9.79\text{e}-1$ | | 24.4 |
| WGAN_OS_AS | 25 | $1.35\text{e}-2$ | 25 | $2.03\text{e}-2$ | 22 | $9.36\text{e}-1$ | 25 | $2.03\text{e}-1$ | 19 | $9.79\text{e}-1$ | | 23.2 |
| WGAN_OS_SA | 18 | $1.15\text{e}-2$ | 18 | $1.72\text{e}-2$ | 04 | $9.63\text{e}-1$ | 18 | $1.72\text{e}-1$ | 04 | $9.87\text{e}-1$ | | 12.4 |
| WGAN_OS_SS | 15 | $1.13\text{e}-2$ | 17 | $1.69\text{e}-2$ | 03 | $9.63\text{e}-1$ | 17 | $1.69\text{e}-1$ | 03 | $9.88\text{e}-1$ | | 11.0 |
| WGAN_TS_AA | 29 | $1.38\text{e}-2$ | 31 | $2.17\text{e}-2$ | 19 | $9.39\text{e}-1$ | 31 | $2.17\text{e}-1$ | 11 | $9.82\text{e}-1$ | | 24.2 |
| WGAN_TS_AS | 32 | $1.40\text{e}-2$ | 32 | $2.18\text{e}-2$ | 20 | $9.38\text{e}-1$ | 32 | $2.18\text{e}-1$ | 14 | $9.82\text{e}-1$ | | 26.0 |
| WGAN_TS_SA | 28 | $1.37\text{e}-2$ | 28 | $2.06\text{e}-2$ | 08 | $9.59\text{e}-1$ | 28 | $2.06\text{e}-1$ | 01 | $9.89\text{e}-1$ | | 18.6 |

Table 13 – continued from previous page

| Name | | MAE | | RMSE | | $R^2$ | | SRMSE | | $\rho_{\text{Pearson}}$ | | rank |
|---|---|---|---|---|---|---|---|---|---|---|---|---|
| WGAN_TS_SS | 27 | $1.37e-2$ | 27 | $2.06e-2$ | 07 | $9.59e-1$ | 27 | $2.06e-1$ | 02 | $9.89e-1$ | | 18.0 |
| WGGP_NO_AA | 17 | $1.14e-2$ | 15 | $1.67e-2$ | 27 | $9.29e-1$ | 15 | $1.67e-1$ | 31 | $9.72e-1$ | | 21.0 |
| WGGP_NO_AS | 16 | $1.14e-2$ | 16 | $1.67e-2$ | 28 | $9.28e-1$ | 16 | $1.67e-1$ | 32 | $9.72e-1$ | | 21.6 |
| WGGP_NO_SA | 19 | $1.19e-2$ | 19 | $1.76e-2$ | 25 | $9.30e-1$ | 19 | $1.76e-1$ | 29 | $9.73e-1$ | | 22.2 |
| WGGP_NO_SS | 20 | $1.19e-2$ | 20 | $1.77e-2$ | 26 | $9.30e-1$ | 20 | $1.77e-1$ | 30 | $9.73e-1$ | | 23.2 |
| WGGP_OS_AA | 10 | $1.03e-2$ | 10 | $1.50e-2$ | 16 | $9.52e-1$ | 10 | $1.50e-1$ | 22 | $9.77e-1$ | | 13.6 |
| WGGP_OS_AS | 09 | $1.03e-2$ | 09 | $1.50e-2$ | 15 | $9.53e-1$ | 09 | $1.50e-1$ | 21 | $9.77e-1$ | | 12.6 |
| WGGP_OS_SA | 03 | $9.49e-3$ | 03 | $1.40e-2$ | 10 | $9.56e-1$ | 03 | $1.40e-1$ | 17 | $9.79e-1$ | | 7.2 |
| WGGP_OS_SS | 04 | $9.49e-3$ | 04 | $1.40e-2$ | 09 | $9.56e-1$ | 04 | $1.40e-1$ | 18 | $9.79e-1$ | | 7.8 |
| WGGP_TS_AA | 06 | $9.84e-3$ | 06 | $1.43e-2$ | 06 | $9.61e-1$ | 06 | $1.43e-1$ | 12 | $9.82e-1$ | | 7.2 |
| WGGP_TS_AS | 05 | $9.82e-3$ | 05 | $1.42e-2$ | 05 | $9.61e-1$ | 05 | $1.42e-1$ | 13 | $9.82e-1$ | | 6.6 |
| WGGP_TS_SA | 01 | $8.91e-3$ | 01 | $1.31e-2$ | 02 | $9.69e-1$ | 01 | $1.31e-1$ | 06 | $9.86e-1$ | | 2.2 |
| WGGP_TS_SS | 02 | $8.93e-3$ | 02 | $1.31e-2$ | 01 | $9.69e-1$ | 02 | $1.31e-1$ | 05 | $9.86e-1$ | | 2.4 |

Table 14: Results of the statistics on the first aggregation level (categorical columns) for the LPMC dataset. Lighter grey tone corresponds to better results compared to darker ones.

| Name | | MAE | | RMSE | | $R^2$ | | SRMSE | | $\rho_{\text{Pearson}}$ | | rank |
|---|---|---|---|---|---|---|---|---|---|---|---|---|
| SGAN_NO_AA | 19 | $1.13e-2$ | 19 | $1.34e-2$ | 17 | $8.58e-1$ | 17 | $6.17e-2$ | 19 | $9.70e-1$ | | 18.2 |
| SGAN_NO_AS | 07 | $5.93e-3$ | 07 | $6.93e-3$ | 09 | $9.23e-1$ | 07 | $3.43e-2$ | 13 | $9.79e-1$ | | 8.6 |
| SGAN_NO_SA | 20 | $1.15e-2$ | 20 | $1.37e-2$ | 18 | $8.56e-1$ | 18 | $6.24e-2$ | 16 | $9.72e-1$ | | 18.4 |
| SGAN_NO_SS | 10 | $6.18e-3$ | 10 | $7.17e-3$ | 11 | $9.16e-1$ | 09 | $3.54e-2$ | 17 | $9.72e-1$ | | 11.4 |
| SGAN_OS_AA | 24 | $1.53e-2$ | 24 | $1.78e-2$ | 30 | $-1.14e-1$ | 24 | $9.85e-2$ | 36 | $8.99e-1$ | | 27.6 |
| SGAN_OS_AS | 34 | $3.33e-2$ | 34 | $3.82e-2$ | 20 | $8.38e-1$ | 33 | $1.73e-1$ | 29 | $9.49e-1$ | | 30.0 |
| SGAN_OS_SA | 23 | $1.52e-2$ | 23 | $1.77e-2$ | 29 | $-1.03e-1$ | 23 | $9.79e-2$ | 35 | $9.00e-1$ | | 26.6 |
| SGAN_OS_SS | 33 | $3.33e-2$ | 33 | $3.81e-2$ | 19 | $8.39e-1$ | 34 | $1.74e-1$ | 31 | $9.47e-1$ | | 30.0 |
| SGAN_TS_AA | 18 | $8.98e-3$ | 18 | $1.06e-2$ | 16 | $8.81e-1$ | 16 | $5.26e-2$ | 22 | $9.69e-1$ | | 18.0 |
| SGAN_TS_AS | 09 | $6.13e-3$ | 09 | $7.08e-3$ | 13 | $9.08e-1$ | 12 | $3.75e-2$ | 14 | $9.72e-1$ | | 11.4 |
| SGAN_TS_SA | 17 | $8.90e-3$ | 17 | $1.05e-2$ | 15 | $8.83e-1$ | 15 | $5.20e-2$ | 18 | $9.70e-1$ | | 16.4 |
| SGAN_TS_SS | 08 | $6.06e-3$ | 08 | $6.99e-3$ | 14 | $9.05e-1$ | 11 | $3.71e-2$ | 15 | $9.72e-1$ | | 11.2 |
| WGAN_NO_AA | 29 | $2.81e-2$ | 29 | $3.32e-2$ | 25 | $7.32e-1$ | 29 | $1.49e-1$ | 07 | $9.85e-1$ | | 23.8 |
| WGAN_NO_AS | 02 | $4.04e-3$ | 02 | $4.75e-3$ | 01 | $9.62e-1$ | 02 | $2.45e-2$ | 03 | $9.87e-1$ | | 2.0 |
| WGAN_NO_SA | 30 | $2.81e-2$ | 30 | $3.32e-2$ | 26 | $7.27e-1$ | 30 | $1.50e-1$ | 10 | $9.84e-1$ | | 25.2 |
| WGAN_NO_SS | 01 | $3.98e-3$ | 01 | $4.67e-3$ | 05 | $9.56e-1$ | 01 | $2.42e-2$ | 04 | $9.86e-1$ | | 2.4 |
| WGAN_OS_AA | 25 | $1.86e-2$ | 25 | $2.20e-2$ | 31 | $-4.65e-1$ | 25 | $1.26e-1$ | 25 | $9.56e-1$ | | 26.2 |
| WGAN_OS_AS | 36 | $3.39e-2$ | 36 | $3.87e-2$ | 24 | $8.20e-1$ | 36 | $1.75e-1$ | 32 | $9.42e-1$ | | 32.8 |
| WGAN_OS_SA | 26 | $1.87e-2$ | 26 | $2.21e-2$ | 32 | $-4.68e-1$ | 26 | $1.26e-1$ | 33 | $9.40e-1$ | | 28.6 |
| WGAN_OS_SS | 35 | $3.37e-2$ | 35 | $3.85e-2$ | 23 | $8.25e-1$ | 35 | $1.74e-1$ | 28 | $9.50e-1$ | | 31.2 |
| WGAN_TS_AA | 28 | $2.00e-2$ | 28 | $2.35e-2$ | 35 | $-2.26$ | 27 | $1.44e-1$ | 20 | $9.69e-1$ | | 27.6 |
| WGAN_TS_AS | 03 | $4.45e-3$ | 03 | $5.21e-3$ | 02 | $9.59e-1$ | 03 | $2.48e-2$ | 02 | $9.87e-1$ | | 2.6 |
| WGAN_TS_SA | 27 | $2.00e-2$ | 27 | $2.35e-2$ | 36 | $-2.27$ | 28 | $1.44e-1$ | 21 | $9.69e-1$ | | 27.8 |
| WGAN_TS_SS | 04 | $4.61e-3$ | 04 | $5.38e-3$ | 03 | $9.59e-1$ | 04 | $2.50e-2$ | 01 | $9.88e-1$ | | 3.2 |
| WGGP_NO_AA | 11 | $6.36e-3$ | 11 | $7.25e-3$ | 10 | $9.16e-1$ | 13 | $3.77e-2$ | 05 | $9.86e-1$ | | 10.0 |
| WGGP_NO_AS | 05 | $5.01e-3$ | 05 | $5.79e-3$ | 07 | $9.45e-1$ | 05 | $2.83e-2$ | 11 | $9.83e-1$ | | 6.6 |
| WGGP_NO_SA | 13 | $6.48e-3$ | 12 | $7.39e-3$ | 12 | $9.14e-1$ | 14 | $3.77e-2$ | 06 | $9.85e-1$ | | 11.4 |

Table 14 – continued from previous page

| Name | | MAE | | RMSE | | $R^2$ | | SRMSE | | $\rho_{\text{Pearson}}$ | | rank |
|---|---|---|---|---|---|---|---|---|---|---|---|---|
| WGGP_NO_SS | 06 | $5.26e-3$ | 06 | $6.05e-3$ | 08 | $9.42e-1$ | 06 | $2.86e-2$ | 08 | $9.84e-1$ | | 6.8 |
| WGGP_OS_AA | 21 | $1.45e-2$ | 21 | $1.68e-2$ | 28 | $2.37e-1$ | 21 | $9.11e-2$ | 34 | $9.36e-1$ | | 25.0 |
| WGGP_OS_AS | 31 | $3.30e-2$ | 31 | $3.77e-2$ | 22 | $8.26e-1$ | 31 | $1.72e-1$ | 26 | $9.53e-1$ | | 28.2 |
| WGGP_OS_SA | 22 | $1.45e-2$ | 22 | $1.68e-2$ | 27 | $2.49e-1$ | 22 | $9.13e-2$ | 27 | $9.52e-1$ | | 24.0 |
| WGGP_OS_SS | 32 | $3.30e-2$ | 32 | $3.79e-2$ | 21 | $8.29e-1$ | 32 | $1.73e-1$ | 30 | $9.48e-1$ | | 29.4 |
| WGGP_TS_AA | 16 | $7.43e-3$ | 16 | $8.74e-3$ | 33 | $-5.01e-1$ | 20 | $6.80e-2$ | 23 | $9.63e-1$ | | 21.6 |
| WGGP_TS_AS | 14 | $6.82e-3$ | 14 | $7.88e-3$ | 06 | $9.51e-1$ | 10 | $3.70e-2$ | 12 | $9.82e-1$ | | 11.2 |
| WGGP_TS_SA | 15 | $7.23e-3$ | 15 | $8.49e-3$ | 34 | $-5.12e-1$ | 19 | $6.70e-2$ | 24 | $9.62e-1$ | | 21.4 |
| WGGP_TS_SS | 12 | $6.44e-3$ | 13 | $7.47e-3$ | 04 | $9.56e-1$ | 08 | $3.52e-2$ | 09 | $9.84e-1$ | | 9.2 |

Table 15: Results of the statistics on the second aggregation level for the LPMC dataset. Lighter grey tone corresponds to better results compared to darker ones.

| Name | | MAE | | RMSE | | $R^2$ | | SRMSE | | $\rho_{\text{Pearson}}$ | | rank |
|---|---|---|---|---|---|---|---|---|---|---|---|---|
| SGAN_NO_AA | 21 | $5.21e-3$ | 21 | $9.27e-3$ | 21 | $8.90e-1$ | 21 | $3.50e-1$ | 26 | $9.62e-1$ | | 22.0 |
| SGAN_NO_AS | 17 | $4.89e-3$ | 16 | $8.70e-3$ | 18 | $9.02e-1$ | 18 | $3.37e-1$ | 22 | $9.64e-1$ | | 18.2 |
| SGAN_NO_SA | 18 | $5.07e-3$ | 19 | $8.96e-3$ | 17 | $9.08e-1$ | 19 | $3.38e-1$ | 14 | $9.67e-1$ | | 17.4 |
| SGAN_NO_SS | 15 | $4.79e-3$ | 15 | $8.47e-3$ | 11 | $9.18e-1$ | 15 | $3.27e-1$ | 13 | $9.69e-1$ | | 13.8 |
| SGAN_OS_AA | 22 | $5.50e-3$ | 22 | $9.34e-3$ | 30 | $8.15e-1$ | 22 | $3.69e-1$ | 36 | $9.49e-1$ | | 26.4 |
| SGAN_OS_AS | 34 | $8.75e-3$ | 32 | $1.38e-2$ | 22 | $8.90e-1$ | 32 | $4.51e-1$ | 30 | $9.60e-1$ | | 30.0 |
| SGAN_OS_SA | 20 | $5.21e-3$ | 18 | $8.80e-3$ | 28 | $8.32e-1$ | 20 | $3.48e-1$ | 35 | $9.55e-1$ | | 24.2 |
| SGAN_OS_SS | 32 | $8.50e-3$ | 31 | $1.34e-2$ | 19 | $9.01e-1$ | 28 | $4.35e-1$ | 20 | $9.65e-1$ | | 26.0 |
| SGAN_TS_AA | 12 | $4.34e-3$ | 12 | $7.38e-3$ | 10 | $9.23e-1$ | 11 | $2.85e-1$ | 12 | $9.70e-1$ | | 11.4 |
| SGAN_TS_AS | 08 | $4.19e-3$ | 08 | $7.15e-3$ | 06 | $9.30e-1$ | 08 | $2.79e-1$ | 11 | $9.71e-1$ | | 8.2 |
| SGAN_TS_SA | 09 | $4.24e-3$ | 09 | $7.30e-3$ | 08 | $9.28e-1$ | 09 | $2.82e-1$ | 09 | $9.72e-1$ | | 8.8 |
| SGAN_TS_SS | 07 | $4.13e-3$ | 07 | $7.10e-3$ | 04 | $9.34e-1$ | 07 | $2.78e-1$ | 06 | $9.73e-1$ | | 6.2 |
| WGAN_NO_AA | 33 | $8.53e-3$ | 36 | $1.58e-2$ | 29 | $8.28e-1$ | 36 | $5.49e-1$ | 33 | $9.59e-1$ | | 33.4 |
| WGAN_NO_AS | 26 | $6.19e-3$ | 27 | $1.14e-2$ | 24 | $8.88e-1$ | 29 | $4.41e-1$ | 29 | $9.60e-1$ | | 27.0 |
| WGAN_NO_SA | 31 | $8.26e-3$ | 35 | $1.46e-2$ | 27 | $8.59e-1$ | 35 | $4.99e-1$ | 21 | $9.65e-1$ | | 29.8 |
| WGAN_NO_SS | 23 | $5.92e-3$ | 23 | $1.04e-2$ | 13 | $9.16e-1$ | 25 | $3.98e-1$ | 17 | $9.66e-1$ | | 20.2 |
| WGAN_OS_AA | 25 | $6.14e-3$ | 25 | $1.09e-2$ | 32 | $7.97e-1$ | 27 | $4.11e-1$ | 27 | $9.62e-1$ | | 27.2 |
| WGAN_OS_AS | 36 | $8.99e-3$ | 34 | $1.42e-2$ | 23 | $8.88e-1$ | 34 | $4.61e-1$ | 32 | $9.60e-1$ | | 31.8 |
| WGAN_OS_SA | 24 | $6.14e-3$ | 24 | $1.06e-2$ | 31 | $8.12e-1$ | 26 | $4.03e-1$ | 18 | $9.66e-1$ | | 24.6 |
| WGAN_OS_SS | 35 | $8.94e-3$ | 33 | $1.41e-2$ | 20 | $9.01e-1$ | 33 | $4.59e-1$ | 23 | $9.64e-1$ | | 28.8 |
| WGAN_TS_AA | 27 | $6.27e-3$ | 26 | $1.14e-2$ | 36 | $6.26e-1$ | 30 | $4.46e-1$ | 34 | $9.57e-1$ | | 30.6 |
| WGAN_TS_AS | 16 | $4.89e-3$ | 17 | $8.80e-3$ | 05 | $9.34e-1$ | 16 | $3.28e-1$ | 04 | $9.75e-1$ | | 11.6 |
| WGAN_TS_SA | 28 | $6.49e-3$ | 28 | $1.14e-2$ | 35 | $6.38e-1$ | 31 | $4.47e-1$ | 31 | $9.60e-1$ | | 30.6 |
| WGAN_TS_SS | 19 | $5.14e-3$ | 20 | $8.97e-3$ | 03 | $9.44e-1$ | 17 | $3.33e-1$ | 02 | $9.77e-1$ | | 12.2 |
| WGGP_NO_AA | 11 | $4.29e-3$ | 10 | $7.32e-3$ | 14 | $9.15e-1$ | 10 | $2.85e-1$ | 10 | $9.71e-1$ | | 11.0 |
| WGGP_NO_AS | 06 | $4.04e-3$ | 05 | $7.00e-3$ | 09 | $9.28e-1$ | 04 | $2.75e-1$ | 08 | $9.72e-1$ | | 6.4 |
| WGGP_NO_SA | 10 | $4.27e-3$ | 11 | $7.32e-3$ | 12 | $9.17e-1$ | 12 | $2.86e-1$ | 07 | $9.72e-1$ | | 10.4 |
| WGGP_NO_SS | 05 | $4.02e-3$ | 06 | $7.01e-3$ | 07 | $9.29e-1$ | 05 | $2.77e-1$ | 05 | $9.73e-1$ | | 5.6 |
| WGGP_OS_AA | 14 | $4.70e-3$ | 14 | $7.79e-3$ | 26 | $8.70e-1$ | 14 | $2.97e-1$ | 24 | $9.64e-1$ | | 18.4 |
| WGGP_OS_AS | 30 | $7.81e-3$ | 30 | $1.22e-2$ | 16 | $9.08e-1$ | 24 | $3.91e-1$ | 16 | $9.66e-1$ | | 23.2 |
| WGGP_OS_SA | 13 | $4.64e-3$ | 13 | $7.67e-3$ | 25 | $8.77e-1$ | 13 | $2.88e-1$ | 19 | $9.65e-1$ | | 16.6 |
| WGGP_OS_SS | 29 | $7.78e-3$ | 29 | $1.22e-2$ | 15 | $9.10e-1$ | 23 | $3.90e-1$ | 15 | $9.67e-1$ | | 22.2 |

Table 15 – continued from previous page

| Name | | MAE | | RMSE | | $R^2$ | | SRMSE | | $\rho_{\text{Pearson}}$ | | rank |
|---|---|---|---|---|---|---|---|---|---|---|---|---|
| WGGP_TS_AA | 04 | $3.85e-3$ | 04 | $6.39e-3$ | 34 | $7.89e-1$ | 06 | $2.77e-1$ | 28 | $9.62e-1$ | | 15.2 |
| WGGP_TS_AS | 02 | $3.70e-3$ | 02 | $6.16e-3$ | 02 | $9.47e-1$ | 02 | $2.35e-1$ | 03 | $9.75e-1$ | | **2.2** |
| WGGP_TS_SA | 03 | $3.71e-3$ | 03 | $6.23e-3$ | 33 | $7.91e-1$ | 03 | $2.71e-1$ | 25 | $9.64e-1$ | | 13.4 |
| WGGP_TS_SS | 01 | $3.52e-3$ | 01 | $5.91e-3$ | 01 | $9.51e-1$ | 01 | $2.26e-1$ | 01 | $9.78e-1$ | | **1.0** |

Table 16: Results of the statistics on the third aggregation level for the LPMC dataset. Lighter grey tone corresponds to better results compared to darker ones.

| Name | | MAE | | RMSE | | $R^2$ | | SRMSE | | $\rho_{\text{Pearson}}$ | | rank |
|---|---|---|---|---|---|---|---|---|---|---|---|---|
| SGAN_NO_AA | 20 | $1.87e-3$ | 21 | $4.21e-3$ | 23 | $8.64e-1$ | 20 | $6.48e-1$ | 24 | $9.49e-1$ | | 21.6 |
| SGAN_NO_AS | 18 | $1.84e-3$ | 18 | $4.09e-3$ | 20 | $8.75e-1$ | 18 | $6.37e-1$ | 20 | $9.51e-1$ | | 18.8 |
| SGAN_NO_SA | 16 | $1.81e-3$ | 17 | $4.02e-3$ | 18 | $8.86e-1$ | 17 | $6.33e-1$ | 18 | $9.55e-1$ | | 17.2 |
| SGAN_NO_SS | 15 | $1.80e-3$ | 16 | $3.98e-3$ | 15 | $8.94e-1$ | 16 | $6.32e-1$ | 15 | $9.56e-1$ | | 15.4 |
| SGAN_OS_AA | 21 | $1.97e-3$ | 20 | $4.20e-3$ | 33 | $8.25e-1$ | 22 | $6.96e-1$ | 36 | $9.35e-1$ | | 26.4 |
| SGAN_OS_AS | 34 | $2.79e-3$ | 32 | $5.75e-3$ | 24 | $8.57e-1$ | 32 | $8.38e-1$ | 29 | $9.44e-1$ | | 30.2 |
| SGAN_OS_SA | 19 | $1.86e-3$ | 15 | $3.94e-3$ | 27 | $8.54e-1$ | 21 | $6.64e-1$ | 30 | $9.44e-1$ | | 22.4 |
| SGAN_OS_SS | 32 | $2.71e-3$ | 31 | $5.60e-3$ | 21 | $8.72e-1$ | 31 | $8.34e-1$ | 23 | $9.50e-1$ | | 27.6 |
| SGAN_TS_AA | 13 | $1.62e-3$ | 13 | $3.49e-3$ | 10 | $8.99e-1$ | 14 | $5.62e-1$ | 12 | $9.58e-1$ | | 12.4 |
| SGAN_TS_AS | 10 | $1.59e-3$ | 11 | $3.44e-3$ | 09 | $9.06e-1$ | 12 | $5.60e-1$ | 10 | $9.59e-1$ | | 10.4 |
| SGAN_TS_SA | 09 | $1.58e-3$ | 10 | $3.40e-3$ | 06 | $9.08e-1$ | 11 | $5.55e-1$ | 07 | $9.61e-1$ | | **8.6** |
| SGAN_TS_SS | 07 | $1.56e-3$ | 09 | $3.39e-3$ | 04 | $9.13e-1$ | 13 | $5.60e-1$ | 05 | $9.62e-1$ | | **7.6** |
| WGAN_NO_AA | 33 | $2.73e-3$ | 36 | $6.71e-3$ | 34 | $7.81e-1$ | 36 | $9.66e-1$ | 32 | $9.43e-1$ | | 34.2 |
| WGAN_NO_AS | 27 | $2.27e-3$ | 30 | $5.22e-3$ | 30 | $8.48e-1$ | 29 | $8.08e-1$ | 31 | $9.44e-1$ | | 29.4 |
| WGAN_NO_SA | 31 | $2.71e-3$ | 35 | $6.20e-3$ | 32 | $8.25e-1$ | 34 | $8.89e-1$ | 27 | $9.47e-1$ | | 31.8 |
| WGAN_NO_SS | 28 | $2.28e-3$ | 29 | $5.00e-3$ | 19 | $8.79e-1$ | 27 | $7.73e-1$ | 26 | $9.48e-1$ | | 25.8 |
| WGAN_OS_AA | 23 | $2.05e-3$ | 24 | $4.75e-3$ | 31 | $8.34e-1$ | 26 | $7.47e-1$ | 22 | $9.50e-1$ | | 25.2 |
| WGAN_OS_AS | 35 | $2.86e-3$ | 33 | $5.86e-3$ | 25 | $8.57e-1$ | 33 | $8.59e-1$ | 34 | $9.43e-1$ | | 32.0 |
| WGAN_OS_SA | 25 | $2.11e-3$ | 23 | $4.65e-3$ | 26 | $8.55e-1$ | 25 | $7.42e-1$ | 19 | $9.53e-1$ | | 23.6 |
| WGAN_OS_SS | 36 | $2.92e-3$ | 34 | $6.00e-3$ | 22 | $8.65e-1$ | 35 | $9.06e-1$ | 28 | $9.46e-1$ | | 31.0 |
| WGAN_TS_AA | 24 | $2.08e-3$ | 28 | $4.98e-3$ | 36 | $7.47e-1$ | 30 | $8.09e-1$ | 35 | $9.41e-1$ | | 30.6 |
| WGAN_TS_AS | 17 | $1.83e-3$ | 19 | $4.14e-3$ | 08 | $9.06e-1$ | 15 | $6.22e-1$ | 04 | $9.62e-1$ | | 12.6 |
| WGAN_TS_SA | 26 | $2.21e-3$ | 26 | $4.95e-3$ | 35 | $7.66e-1$ | 28 | $8.02e-1$ | 33 | $9.43e-1$ | | 29.6 |
| WGAN_TS_SS | 22 | $1.99e-3$ | 22 | $4.30e-3$ | 03 | $9.14e-1$ | 19 | $6.47e-1$ | 03 | $9.63e-1$ | | 13.8 |
| WGGP_NO_AA | 14 | $1.64e-3$ | 14 | $3.51e-3$ | 14 | $8.96e-1$ | 09 | $5.49e-1$ | 11 | $9.58e-1$ | | 12.4 |
| WGGP_NO_AS | 06 | $1.54e-3$ | 08 | $3.35e-3$ | 07 | $9.08e-1$ | 04 | $5.36e-1$ | 08 | $9.60e-1$ | | **6.6** |
| WGGP_NO_SA | 12 | $1.61e-3$ | 12 | $3.48e-3$ | 11 | $8.99e-1$ | 10 | $5.50e-1$ | 09 | $9.59e-1$ | | 10.8 |
| WGGP_NO_SS | 05 | $1.52e-3$ | 06 | $3.33e-3$ | 05 | $9.10e-1$ | 05 | $5.37e-1$ | 06 | $9.61e-1$ | | **5.4** |
| WGGP_OS_AA | 11 | $1.60e-3$ | 07 | $3.33e-3$ | 13 | $8.97e-1$ | 08 | $5.44e-1$ | 16 | $9.56e-1$ | | 11.0 |
| WGGP_OS_AS | 30 | $2.43e-3$ | 27 | $4.95e-3$ | 17 | $8.89e-1$ | 24 | $7.31e-1$ | 17 | $9.56e-1$ | | 23.0 |
| WGGP_OS_SA | 08 | $1.57e-3$ | 05 | $3.29e-3$ | 12 | $8.98e-1$ | 07 | $5.42e-1$ | 13 | $9.57e-1$ | | **9.0** |
| WGGP_OS_SS | 29 | $2.40e-3$ | 25 | $4.91e-3$ | 16 | $8.90e-1$ | 23 | $7.31e-1$ | 14 | $9.56e-1$ | | 21.4 |
| WGGP_TS_AA | 04 | $1.45e-3$ | 04 | $3.03e-3$ | 29 | $8.48e-1$ | 06 | $5.39e-1$ | 25 | $9.48e-1$ | | 13.6 |
| WGGP_TS_AS | 02 | $1.39e-3$ | 02 | $2.90e-3$ | 02 | $9.31e-1$ | 02 | $4.61e-1$ | 02 | $9.66e-1$ | | **2.0** |
| WGGP_TS_SA | 03 | $1.40e-3$ | 03 | $2.96e-3$ | 28 | $8.51e-1$ | 03 | $5.33e-1$ | 21 | $9.50e-1$ | | 11.6 |
| WGGP_TS_SS | 01 | $1.32e-3$ | 01 | $2.79e-3$ | 01 | $9.35e-1$ | 01 | $4.50e-1$ | 01 | $9.69e-1$ | | **1.0** |

Table 17: Results of the Machine Learning efficacy for the LPMC dataset. Lighter grey tone corresponds to better results compared to darker ones.

| Name | Continuous | | Categorical | | rank |
|---|---|---|---|---|---|
| SGAN_NO_AA | 15 | 2.81e1 | 32 | 5.98 | 23.5 |
| SGAN_NO_AS | 16 | 2.94e1 | 19 | 3.89 | 17.5 |
| SGAN_NO_SA | 29 | 4.59e1 | 23 | 4.55 | 26.0 |
| SGAN_NO_SS | 30 | 4.74e1 | 18 | 3.77 | 24.0 |
| SGAN_OS_AA | 17 | 3.08e1 | 25 | 4.61 | 21.0 |
| SGAN_OS_AS | 20 | 3.37e1 | 27 | 4.73 | 23.5 |
| SGAN_OS_SA | 27 | 4.16e1 | 21 | 4.18 | 24.0 |
| SGAN_OS_SS | 28 | 4.44e1 | 24 | 4.55 | 26.0 |
| SGAN_TS_AA | 13 | 2.78e1 | 26 | 4.67 | 19.5 |
| SGAN_TS_AS | 14 | 2.80e1 | 16 | 3.71 | 15.0 |
| SGAN_TS_SA | 18 | 3.21e1 | 17 | 3.73 | 17.5 |
| SGAN_TS_SS | 19 | 3.31e1 | 14 | 3.48 | 16.5 |
| WGAN_NO_AA | 24 | 3.93e1 | 35 | 1.49e4 | 29.5 |
| WGAN_NO_AS | 26 | 4.14e1 | 29 | 5.00 | 27.5 |
| WGAN_NO_SA | 35 | 6.60e1 | 36 | 1.62e4 | 35.5 |
| WGAN_NO_SS | 36 | 6.87e1 | 28 | 4.94 | 32.0 |
| WGAN_OS_AA | 21 | 3.55e1 | 13 | 3.45 | 17.0 |
| WGAN_OS_AS | 22 | 3.90e1 | 30 | 5.30 | 26.0 |
| WGAN_OS_SA | 31 | 5.87e1 | 15 | 3.49 | 23.0 |
| WGAN_OS_SS | 33 | 6.15e1 | 31 | 5.40 | 32.0 |
| WGAN_TS_AA | 23 | 3.91e1 | 33 | 7.28e3 | 28.0 |
| WGAN_TS_AS | 25 | 4.05e1 | 20 | 4.14 | 22.5 |
| WGAN_TS_SA | 32 | 6.15e1 | 34 | 7.68e3 | 33.0 |
| WGAN_TS_SS | 34 | 6.34e1 | 22 | 4.31 | 28.0 |
| WGGP_NO_AA | **09** | 2.31e1 | 10 | 3.09 | **9.5** |
| WGGP_NO_AS | **10** | 2.32e1 | **06** | 2.65 | **8.0** |
| WGGP_NO_SA | 11 | 2.57e1 | **09** | 3.04 | **10.0** |
| WGGP_NO_SS | 12 | 2.60e1 | **05** | 2.61 | **8.5** |
| WGGP_OS_AA | **01** | 1.96e1 | **04** | 2.56 | **2.5** |
| WGGP_OS_AS | **02** | 2.03e1 | **07** | 2.94 | **4.5** |
| WGGP_OS_SA | **05** | 2.09e1 | **03** | 2.50 | **4.0** |
| WGGP_OS_SS | **06** | 2.16e1 | **08** | 2.94 | **7.0** |
| WGGP_TS_AA | **03** | 2.03e1 | 12 | 3.19 | **7.5** |
| WGGP_TS_AS | **04** | 2.03e1 | **02** | 2.21 | **3.0** |
| WGGP_TS_SA | **08** | 2.21e1 | 11 | 3.14 | **9.5** |
| WGGP_TS_SS | **07** | 2.20e1 | **01** | 2.19 | **4.0** |

### 5.1.3 LPMC_half case study

Table 18: Results of the statistics on the first aggregation level (all columns) for the LPMC_half dataset. Lighter grey tone corresponds to better results compared to darker ones.

| Name | MAE | | RMSE | | $R^2$ | | SRMSE | | $\rho_{\text{Pearson}}$ | | rank |
|---|---|---|---|---|---|---|---|---|---|---|---|
| SGAN_NO_AA | 22 | $1.59e-2$ | 23 | $2.18e-2$ | 19 | $8.15e-1$ | 21 | $1.62e-1$ | 10 | $9.63e-1$ | 19.0 |

Table 18 – continued from previous page

| Name | | MAE | | RMSE | | $R^2$ | | SRMSE | | $\rho_{\text{Pearson}}$ | | rank |
|---|---|---|---|---|---|---|---|---|---|---|---|---|
| SGAN_NO_AS | **10** | $1.12\mathrm{e}-2$ | 11 | $1.60\mathrm{e}-2$ | **05** | $9.04\mathrm{e}-1$ | 12 | $1.38\mathrm{e}-1$ | **04** | $9.71\mathrm{e}-1$ | | 8.4 |
| SGAN_NO_SA | 19 | $1.55\mathrm{e}-2$ | 21 | $2.09\mathrm{e}-2$ | 17 | $8.28\mathrm{e}-1$ | 17 | $1.55\mathrm{e}-1$ | **05** | $9.67\mathrm{e}-1$ | | 15.8 |
| SGAN_NO_SS | **07** | $1.09\mathrm{e}-2$ | **08** | $1.53\mathrm{e}-2$ | **02** | $9.17\mathrm{e}-1$ | **06** | $1.31\mathrm{e}-1$ | **02** | $9.74\mathrm{e}-1$ | | **5.0** |
| SGAN_OS_AA | 21 | $1.57\mathrm{e}-2$ | 22 | $2.12\mathrm{e}-2$ | 26 | $1.56\mathrm{e}-1$ | 22 | $1.72\mathrm{e}-1$ | 36 | $9.20\mathrm{e}-1$ | | 25.4 |
| SGAN_OS_AS | 32 | $2.56\mathrm{e}-2$ | 31 | $3.23\mathrm{e}-2$ | 20 | $8.08\mathrm{e}-1$ | 28 | $2.10\mathrm{e}-1$ | 32 | $9.37\mathrm{e}-1$ | | 28.6 |
| SGAN_OS_SA | 18 | $1.48\mathrm{e}-2$ | 18 | $1.97\mathrm{e}-2$ | 25 | $1.94\mathrm{e}-1$ | 19 | $1.57\mathrm{e}-1$ | 34 | $9.27\mathrm{e}-1$ | | 22.8 |
| SGAN_OS_SS | 30 | $2.45\mathrm{e}-2$ | 29 | $3.06\mathrm{e}-2$ | 14 | $8.43\mathrm{e}-1$ | 26 | $1.94\mathrm{e}-1$ | 26 | $9.50\mathrm{e}-1$ | | 25.0 |
| SGAN_TS_AA | 16 | $1.34\mathrm{e}-2$ | 16 | $1.85\mathrm{e}-2$ | 18 | $8.18\mathrm{e}-1$ | 16 | $1.50\mathrm{e}-1$ | 19 | $9.55\mathrm{e}-1$ | | 17.0 |
| SGAN_TS_AS | 11 | $1.12\mathrm{e}-2$ | **10** | $1.58\mathrm{e}-2$ | 15 | $8.39\mathrm{e}-1$ | 11 | $1.38\mathrm{e}-1$ | 18 | $9.55\mathrm{e}-1$ | | 13.0 |
| SGAN_TS_SA | 14 | $1.26\mathrm{e}-2$ | 14 | $1.70\mathrm{e}-2$ | 16 | $8.37\mathrm{e}-1$ | **09** | $1.34\mathrm{e}-1$ | 16 | $9.57\mathrm{e}-1$ | | 13.8 |
| SGAN_TS_SS | **03** | $1.01\mathrm{e}-2$ | **03** | $1.41\mathrm{e}-2$ | 11 | $8.67\mathrm{e}-1$ | **03** | $1.21\mathrm{e}-1$ | 11 | $9.62\mathrm{e}-1$ | | **6.2** |
| WGAN_NO_AA | 36 | $4.92\mathrm{e}-2$ | 36 | $6.44\mathrm{e}-2$ | 32 | $-3.48\mathrm{e}-1$ | 36 | $4.37\mathrm{e}-1$ | 35 | $9.21\mathrm{e}-1$ | | 35.0 |
| WGAN_NO_AS | 24 | $2.01\mathrm{e}-2$ | 30 | $3.09\mathrm{e}-2$ | 22 | $6.11\mathrm{e}-1$ | 34 | $2.94\mathrm{e}-1$ | 33 | $9.30\mathrm{e}-1$ | | 28.6 |
| WGAN_NO_SA | 35 | $4.28\mathrm{e}-2$ | 35 | $5.33\mathrm{e}-2$ | 29 | $-8.45\mathrm{e}-2$ | 35 | $3.26\mathrm{e}-1$ | 21 | $9.52\mathrm{e}-1$ | | 31.0 |
| WGAN_NO_SS | 17 | $1.37\mathrm{e}-2$ | 19 | $1.99\mathrm{e}-2$ | **08** | $8.78\mathrm{e}-1$ | 25 | $1.84\mathrm{e}-1$ | 12 | $9.62\mathrm{e}-1$ | | 16.2 |
| WGAN_OS_AA | 31 | $2.51\mathrm{e}-2$ | 33 | $3.37\mathrm{e}-2$ | 28 | $2.59\mathrm{e}-2$ | 32 | $2.53\mathrm{e}-1$ | 30 | $9.46\mathrm{e}-1$ | | 30.8 |
| WGAN_OS_AS | 34 | $2.79\mathrm{e}-2$ | 34 | $3.63\mathrm{e}-2$ | 21 | $7.77\mathrm{e}-1$ | 33 | $2.55\mathrm{e}-1$ | 31 | $9.39\mathrm{e}-1$ | | 30.6 |
| WGAN_OS_SA | 27 | $2.27\mathrm{e}-2$ | 27 | $2.94\mathrm{e}-2$ | 27 | $1.13\mathrm{e}-1$ | 29 | $2.11\mathrm{e}-1$ | 13 | $9.59\mathrm{e}-1$ | | 24.6 |
| WGAN_OS_SS | 33 | $2.57\mathrm{e}-2$ | 32 | $3.24\mathrm{e}-2$ | 13 | $8.60\mathrm{e}-1$ | 30 | $2.14\mathrm{e}-1$ | 22 | $9.52\mathrm{e}-1$ | | 26.0 |
| WGAN_TS_AA | 26 | $2.23\mathrm{e}-2$ | 26 | $2.93\mathrm{e}-2$ | 36 | $-1.09$ | 31 | $2.23\mathrm{e}-1$ | 15 | $9.57\mathrm{e}-1$ | | 26.8 |
| WGAN_TS_AS | 12 | $1.13\mathrm{e}-2$ | 12 | $1.65\mathrm{e}-2$ | **06** | $8.92\mathrm{e}-1$ | 14 | $1.47\mathrm{e}-1$ | **06** | $9.67\mathrm{e}-1$ | | **10.0** |
| WGAN_TS_SA | 25 | $2.14\mathrm{e}-2$ | 24 | $2.73\mathrm{e}-2$ | 35 | $-1.04$ | 27 | $2.05\mathrm{e}-1$ | **07** | $9.66\mathrm{e}-1$ | | 23.6 |
| WGAN_TS_SS | **04** | $1.07\mathrm{e}-2$ | **06** | $1.48\mathrm{e}-2$ | **01** | $9.26\mathrm{e}-1$ | **05** | $1.30\mathrm{e}-1$ | **01** | $9.75\mathrm{e}-1$ | | **3.4** |
| WGGP_NO_AA | 15 | $1.29\mathrm{e}-2$ | 15 | $1.74\mathrm{e}-2$ | 34 | $-5.08\mathrm{e}-1$ | 20 | $1.60\mathrm{e}-1$ | 28 | $9.47\mathrm{e}-1$ | | 22.4 |
| WGGP_NO_AS | **09** | $1.12\mathrm{e}-2$ | **09** | $1.54\mathrm{e}-2$ | 10 | $8.68\mathrm{e}-1$ | **08** | $1.33\mathrm{e}-1$ | 14 | $9.57\mathrm{e}-1$ | | **10.0** |
| WGGP_NO_SA | 13 | $1.25\mathrm{e}-2$ | 13 | $1.69\mathrm{e}-2$ | 33 | $-4.74\mathrm{e}-1$ | 18 | $1.55\mathrm{e}-1$ | 20 | $9.54\mathrm{e}-1$ | | 19.4 |
| WGGP_NO_SS | **05** | $1.08\mathrm{e}-2$ | **07** | $1.49\mathrm{e}-2$ | **07** | $8.86\mathrm{e}-1$ | **04** | $1.27\mathrm{e}-1$ | **09** | $9.63\mathrm{e}-1$ | | **6.4** |
| WGGP_OS_AA | 23 | $1.63\mathrm{e}-2$ | 20 | $2.04\mathrm{e}-2$ | 24 | $4.53\mathrm{e}-1$ | 15 | $1.49\mathrm{e}-1$ | 24 | $9.50\mathrm{e}-1$ | | 21.2 |
| WGGP_OS_AS | 29 | $2.45\mathrm{e}-2$ | 28 | $2.97\mathrm{e}-2$ | 12 | $8.65\mathrm{e}-1$ | 24 | $1.82\mathrm{e}-1$ | 27 | $9.48\mathrm{e}-1$ | | 24.0 |
| WGGP_OS_SA | 20 | $1.57\mathrm{e}-2$ | 17 | $1.96\mathrm{e}-2$ | 23 | $4.59\mathrm{e}-1$ | 13 | $1.42\mathrm{e}-1$ | 29 | $9.47\mathrm{e}-1$ | | 20.4 |
| WGGP_OS_SS | 28 | $2.39\mathrm{e}-2$ | 25 | $2.89\mathrm{e}-2$ | **09** | $8.72\mathrm{e}-1$ | 23 | $1.75\mathrm{e}-1$ | 23 | $9.50\mathrm{e}-1$ | | 21.6 |
| WGGP_TS_AA | **08** | $1.11\mathrm{e}-2$ | **05** | $1.44\mathrm{e}-2$ | 30 | $-2.19\mathrm{e}-1$ | **10** | $1.35\mathrm{e}-1$ | 25 | $9.50\mathrm{e}-1$ | | 15.6 |
| WGGP_TS_AS | **02** | $9.61\mathrm{e}-3$ | **02** | $1.28\mathrm{e}-2$ | **04** | $9.09\mathrm{e}-1$ | **02** | $1.05\mathrm{e}-1$ | **08** | $9.64\mathrm{e}-1$ | | **3.6** |
| WGGP_TS_SA | **06** | $1.08\mathrm{e}-2$ | **04** | $1.41\mathrm{e}-2$ | 31 | $-2.20\mathrm{e}-1$ | **07** | $1.33\mathrm{e}-1$ | 17 | $9.57\mathrm{e}-1$ | | 13.0 |
| WGGP_TS_SS | **01** | $9.32\mathrm{e}-3$ | **01** | $1.25\mathrm{e}-2$ | **03** | $9.11\mathrm{e}-1$ | **01** | $1.03\mathrm{e}-1$ | **03** | $9.73\mathrm{e}-1$ | | **1.8** |

Table 19: Results of the statistics on the first aggregation level (continuous columns) for the LPMC_half dataset. Lighter grey tone corresponds to better results compared to darker ones.

| Name | | MAE | | RMSE | | $R^2$ | | SRMSE | | $\rho_{\text{Pearson}}$ | | rank |
|---|---|---|---|---|---|---|---|---|---|---|---|---|
| SGAN_NO_AA | 20 | $1.57\mathrm{e}-2$ | 25 | $2.43\mathrm{e}-2$ | 15 | $8.97\mathrm{e}-1$ | 25 | $2.43\mathrm{e}-1$ | 15 | $9.69\mathrm{e}-1$ | | 20.0 |
| SGAN_NO_AS | 21 | $1.57\mathrm{e}-2$ | 26 | $2.45\mathrm{e}-2$ | 16 | $8.96\mathrm{e}-1$ | 26 | $2.45\mathrm{e}-1$ | 16 | $9.69\mathrm{e}-1$ | | 21.0 |
| SGAN_NO_SA | 15 | $1.51\mathrm{e}-2$ | 15 | $2.29\mathrm{e}-2$ | **09** | $9.31\mathrm{e}-1$ | 15 | $2.29\mathrm{e}-1$ | **01** | $9.78\mathrm{e}-1$ | | 11.0 |
| SGAN_NO_SS | 16 | $1.51\mathrm{e}-2$ | 16 | $2.30\mathrm{e}-2$ | 10 | $9.31\mathrm{e}-1$ | 16 | $2.30\mathrm{e}-1$ | **02** | $9.78\mathrm{e}-1$ | | 12.0 |
| SGAN_OS_AA | 22 | $1.58\mathrm{e}-2$ | 24 | $2.43\mathrm{e}-2$ | 30 | $8.30\mathrm{e}-1$ | 24 | $2.43\mathrm{e}-1$ | 20 | $9.66\mathrm{e}-1$ | | 24.0 |

Table 19 – continued from previous page

| Name | | MAE | | RMSE | | $R^2$ | | SRMSE | | $\rho_{\text{Pearson}}$ | | rank |
|---|---|---|---|---|---|---|---|---|---|---|---|---|
| SGAN_OS_AS | 19 | $1.56e-2$ | 23 | $2.41e-2$ | 29 | $8.33e-1$ | 23 | $2.41e-1$ | 19 | $9.67e-1$ | | 22.6 |
| SGAN_OS_SA | 12 | $1.38e-2$ | 12 | $2.11e-2$ | 17 | $8.95e-1$ | 12 | $2.11e-1$ | **08** | $9.73e-1$ | | 12.2 |
| SGAN_OS_SS | 11 | $1.37e-2$ | 11 | $2.09e-2$ | 18 | $8.94e-1$ | 11 | $2.09e-1$ | **09** | $9.73e-1$ | | 12.0 |
| SGAN_TS_AA | 17 | $1.55e-2$ | 21 | $2.40e-2$ | 31 | $8.27e-1$ | 21 | $2.40e-1$ | 21 | $9.65e-1$ | | 22.2 |
| SGAN_TS_AS | 18 | $1.56e-2$ | 22 | $2.40e-2$ | 32 | $8.26e-1$ | 22 | $2.40e-1$ | 22 | $9.65e-1$ | | 23.2 |
| SGAN_TS_SA | **10** | $1.37e-2$ | **10** | $2.06e-2$ | 22 | $8.69e-1$ | **10** | $2.06e-1$ | 14 | $9.71e-1$ | | 13.2 |
| SGAN_TS_SS | **09** | $1.36e-2$ | **09** | $2.05e-2$ | 21 | $8.70e-1$ | **09** | $2.05e-1$ | 12 | $9.72e-1$ | | 12.0 |
| WGAN_NO_AA | 36 | $3.61e-2$ | 36 | $5.76e-2$ | 35 | $2.86e-1$ | 36 | $5.76e-1$ | 36 | $8.80e-1$ | | 35.8 |
| WGAN_NO_AS | 35 | $3.60e-2$ | 35 | $5.75e-2$ | 36 | $2.85e-1$ | 35 | $5.75e-1$ | 35 | $8.80e-1$ | | 35.2 |
| WGAN_NO_SA | 34 | $2.27e-2$ | 34 | $3.47e-2$ | 28 | $8.34e-1$ | 34 | $3.47e-1$ | 30 | $9.47e-1$ | | 32.0 |
| WGAN_NO_SS | 33 | $2.27e-2$ | 33 | $3.47e-2$ | 27 | $8.35e-1$ | 33 | $3.47e-1$ | 29 | $9.47e-1$ | | 31.0 |
| WGAN_OS_AA | 32 | $2.17e-2$ | 31 | $3.41e-2$ | 33 | $7.35e-1$ | 31 | $3.41e-1$ | 32 | $9.46e-1$ | | 31.8 |
| WGAN_OS_AS | 31 | $2.17e-2$ | 32 | $3.41e-2$ | 34 | $7.33e-1$ | 32 | $3.41e-1$ | 31 | $9.46e-1$ | | 32.0 |
| WGAN_OS_SA | 27 | $1.68e-2$ | 27 | $2.53e-2$ | 13 | $9.16e-1$ | 27 | $2.53e-1$ | 11 | $9.72e-1$ | | 21.0 |
| WGAN_OS_SS | 28 | $1.69e-2$ | 28 | $2.54e-2$ | 14 | $9.15e-1$ | 28 | $2.54e-1$ | 13 | $9.71e-1$ | | 22.2 |
| WGAN_TS_AA | 30 | $1.74e-2$ | 30 | $2.71e-2$ | 26 | $8.47e-1$ | 30 | $2.71e-1$ | 26 | $9.54e-1$ | | 28.4 |
| WGAN_TS_AS | 29 | $1.73e-2$ | 29 | $2.71e-2$ | 25 | $8.49e-1$ | 29 | $2.71e-1$ | 25 | $9.54e-1$ | | 27.4 |
| WGAN_TS_SA | 23 | $1.58e-2$ | 17 | $2.33e-2$ | 11 | $9.30e-1$ | 17 | $2.33e-1$ | **05** | $9.74e-1$ | | 14.6 |
| WGAN_TS_SS | 26 | $1.59e-2$ | 18 | $2.35e-2$ | 12 | $9.30e-1$ | 18 | $2.35e-1$ | **07** | $9.73e-1$ | | 16.2 |
| WGGP_NO_AA | 25 | $1.59e-2$ | 19 | $2.36e-2$ | 24 | $8.53e-1$ | 19 | $2.36e-1$ | 34 | $9.40e-1$ | | 24.2 |
| WGGP_NO_AS | 24 | $1.59e-2$ | 20 | $2.36e-2$ | 23 | $8.53e-1$ | 20 | $2.36e-1$ | 33 | $9.40e-1$ | | 24.0 |
| WGGP_NO_SA | 14 | $1.48e-2$ | 14 | $2.23e-2$ | 20 | $8.83e-1$ | 14 | $2.23e-1$ | 28 | $9.51e-1$ | | 18.0 |
| WGGP_NO_SS | 13 | $1.47e-2$ | 13 | $2.21e-2$ | 19 | $8.84e-1$ | 13 | $2.21e-1$ | 27 | $9.51e-1$ | | 17.0 |
| WGGP_OS_AA | **08** | $1.30e-2$ | **08** | $1.85e-2$ | **08** | $9.31e-1$ | **08** | $1.85e-1$ | 24 | $9.64e-1$ | | 11.2 |
| WGGP_OS_AS | **07** | $1.29e-2$ | **07** | $1.83e-2$ | **07** | $9.32e-1$ | **07** | $1.83e-1$ | 23 | $9.64e-1$ | | 10.2 |
| WGGP_OS_SA | **06** | $1.20e-2$ | **06** | $1.71e-2$ | **05** | $9.37e-1$ | **06** | $1.71e-1$ | 17 | $9.67e-1$ | | **8.0** |
| WGGP_OS_SS | **05** | $1.20e-2$ | **03** | $1.69e-2$ | **06** | $9.36e-1$ | **03** | $1.69e-1$ | 18 | $9.67e-1$ | | **7.0** |
| WGGP_TS_AA | **04** | $1.18e-2$ | **05** | $1.71e-2$ | **04** | $9.43e-1$ | **05** | $1.71e-1$ | 10 | $9.73e-1$ | | **5.6** |
| WGGP_TS_AS | **03** | $1.17e-2$ | **04** | $1.70e-2$ | **03** | $9.45e-1$ | **04** | $1.70e-1$ | **06** | $9.74e-1$ | | **4.0** |
| WGGP_TS_SA | **02** | $1.13e-2$ | **02** | $1.66e-2$ | **01** | $9.48e-1$ | **02** | $1.66e-1$ | **03** | $9.75e-1$ | | **2.0** |
| WGGP_TS_SS | **01** | $1.12e-2$ | **01** | $1.65e-2$ | **02** | $9.47e-1$ | **01** | $1.65e-1$ | **04** | $9.75e-1$ | | **1.8** |

Table 20: Results of the statistics on the first aggregation level (categorical columns) for the LPMC_half dataset. Lighter grey tone corresponds to better results compared to darker ones.

| Name | | MAE | | RMSE | | $R^2$ | | SRMSE | | $\rho_{\text{Pearson}}$ | | rank |
|---|---|---|---|---|---|---|---|---|---|---|---|---|
| SGAN_NO_AA | 22 | $1.62e-2$ | 22 | $1.94e-2$ | 21 | $7.40e-1$ | 16 | $8.70e-2$ | 14 | $9.57e-1$ | | 19.0 |
| SGAN_NO_AS | **06** | $6.97e-3$ | **07** | $8.10e-3$ | **05** | $9.11e-1$ | **07** | $3.93e-2$ | **07** | $9.73e-1$ | | **6.4** |
| SGAN_NO_SA | 21 | $1.59e-2$ | 21 | $1.91e-2$ | 22 | $7.32e-1$ | 15 | $8.63e-2$ | 15 | $9.56e-1$ | | 18.8 |
| SGAN_NO_SS | **08** | $7.02e-3$ | **08** | $8.14e-3$ | **06** | $9.04e-1$ | **08** | $3.97e-2$ | **09** | $9.71e-1$ | | **7.8** |
| SGAN_OS_AA | 20 | $1.57e-2$ | 20 | $1.84e-2$ | 26 | $-4.69e-1$ | 22 | $1.07e-1$ | 36 | $8.77e-1$ | | 24.8 |
| SGAN_OS_AS | 32 | $3.48e-2$ | 32 | $3.99e-2$ | 20 | $7.86e-1$ | 32 | $1.80e-1$ | 34 | $9.10e-1$ | | 30.0 |
| SGAN_OS_SA | 19 | $1.57e-2$ | 19 | $1.83e-2$ | 25 | $-4.56e-1$ | 21 | $1.07e-1$ | 35 | $8.84e-1$ | | 23.8 |
| SGAN_OS_SS | 31 | $3.46e-2$ | 31 | $3.96e-2$ | 19 | $7.96e-1$ | 30 | $1.79e-1$ | 32 | $9.28e-1$ | | 28.6 |
| SGAN_TS_AA | 17 | $1.13e-2$ | 17 | $1.33e-2$ | 15 | $8.10e-1$ | 13 | $6.68e-2$ | 23 | $9.45e-1$ | | 17.0 |

Table 20 – continued from previous page

| Name | | MAE | | RMSE | | $R^2$ | | SRMSE | | $\rho_{\text{Pearson}}$ | | rank |
|---|---|---|---|---|---|---|---|---|---|---|---|---|
| SGAN_TS_AS | 10 | $7.15e-3$ | 10 | $8.20e-3$ | 12 | $8.51e-1$ | 10 | $4.29e-2$ | 21 | $9.47e-1$ | | 12.6 |
| SGAN_TS_SA | 18 | $1.17e-2$ | 18 | $1.37e-2$ | 17 | $8.08e-1$ | 14 | $6.77e-2$ | 24 | $9.44e-1$ | | 18.2 |
| SGAN_TS_SS | 07 | $6.97e-3$ | 06 | $8.05e-3$ | 11 | $8.63e-1$ | 09 | $4.23e-2$ | 18 | $9.54e-1$ | | 10.2 |
| WGAN_NO_AA | 36 | $6.15e-2$ | 36 | $7.07e-2$ | 29 | $-9.35e-1$ | 36 | $3.07e-1$ | 12 | $9.59e-1$ | | 29.8 |
| WGAN_NO_AS | 01 | $5.26e-3$ | 02 | $6.17e-3$ | 04 | $9.14e-1$ | 03 | $3.23e-2$ | 03 | $9.76e-1$ | | 2.6 |
| WGAN_NO_SA | 35 | $6.14e-2$ | 35 | $7.05e-2$ | 30 | $-9.37e-1$ | 35 | $3.07e-1$ | 13 | $9.58e-1$ | | 29.6 |
| WGAN_NO_SS | 02 | $5.29e-3$ | 01 | $6.15e-3$ | 03 | $9.19e-1$ | 01 | $3.20e-2$ | 04 | $9.76e-1$ | | 2.2 |
| WGAN_OS_AA | 28 | $2.82e-2$ | 28 | $3.33e-2$ | 27 | $-6.32e-1$ | 26 | $1.72e-1$ | 22 | $9.46e-1$ | | 26.2 |
| WGAN_OS_AS | 29 | $3.36e-2$ | 29 | $3.84e-2$ | 13 | $8.18e-1$ | 27 | $1.74e-1$ | 29 | $9.32e-1$ | | 25.4 |
| WGAN_OS_SA | 27 | $2.81e-2$ | 27 | $3.33e-2$ | 28 | $-6.34e-1$ | 25 | $1.72e-1$ | 20 | $9.47e-1$ | | 25.4 |
| WGAN_OS_SS | 30 | $3.39e-2$ | 30 | $3.88e-2$ | 16 | $8.09e-1$ | 28 | $1.76e-1$ | 27 | $9.35e-1$ | | 26.2 |
| WGAN_TS_AA | 26 | $2.69e-2$ | 26 | $3.13e-2$ | 36 | $-2.89$ | 31 | $1.79e-1$ | 10 | $9.61e-1$ | | 25.8 |
| WGAN_TS_AS | 03 | $5.71e-3$ | 03 | $6.60e-3$ | 01 | $9.31e-1$ | 02 | $3.20e-2$ | 01 | $9.79e-1$ | | 2.0 |
| WGAN_TS_SA | 25 | $2.67e-2$ | 25 | $3.10e-2$ | 35 | $-2.87$ | 29 | $1.78e-1$ | 11 | $9.60e-1$ | | 25.0 |
| WGAN_TS_SS | 04 | $5.79e-3$ | 04 | $6.72e-3$ | 02 | $9.23e-1$ | 04 | $3.23e-2$ | 02 | $9.77e-1$ | | 3.2 |
| WGGP_NO_AA | 13 | $1.02e-2$ | 13 | $1.16e-2$ | 34 | $-1.77$ | 17 | $9.02e-2$ | 19 | $9.53e-1$ | | 19.2 |
| WGGP_NO_AS | 05 | $6.78e-3$ | 05 | $7.74e-3$ | 08 | $8.82e-1$ | 05 | $3.80e-2$ | 06 | $9.73e-1$ | | 5.8 |
| WGGP_NO_SA | 15 | $1.04e-2$ | 14 | $1.19e-2$ | 33 | $-1.73$ | 18 | $9.15e-2$ | 16 | $9.56e-1$ | | 19.2 |
| WGGP_NO_SS | 09 | $7.12e-3$ | 09 | $8.14e-3$ | 07 | $8.88e-1$ | 06 | $3.90e-2$ | 05 | $9.75e-1$ | | 7.2 |
| WGGP_OS_AA | 24 | $1.95e-2$ | 24 | $2.23e-2$ | 24 | $9.64e-3$ | 24 | $1.16e-1$ | 26 | $9.37e-1$ | | 24.4 |
| WGGP_OS_AS | 34 | $3.53e-2$ | 34 | $4.03e-2$ | 18 | $8.03e-1$ | 34 | $1.81e-1$ | 30 | $9.32e-1$ | | 30.0 |
| WGGP_OS_SA | 23 | $1.92e-2$ | 23 | $2.19e-2$ | 23 | $1.54e-2$ | 23 | $1.15e-1$ | 33 | $9.28e-1$ | | 25.0 |
| WGGP_OS_SS | 33 | $3.49e-2$ | 33 | $4.00e-2$ | 14 | $8.12e-1$ | 33 | $1.81e-1$ | 28 | $9.34e-1$ | | 28.2 |
| WGGP_TS_AA | 16 | $1.04e-2$ | 16 | $1.19e-2$ | 31 | $-1.30$ | 19 | $1.02e-1$ | 31 | $9.29e-1$ | | 22.6 |
| WGGP_TS_AS | 12 | $7.64e-3$ | 12 | $8.90e-3$ | 10 | $8.75e-1$ | 12 | $4.59e-2$ | 17 | $9.55e-1$ | | 12.6 |
| WGGP_TS_SA | 14 | $1.03e-2$ | 15 | $1.19e-2$ | 32 | $-1.31$ | 20 | $1.02e-1$ | 25 | $9.40e-1$ | | 21.2 |
| WGGP_TS_SS | 11 | $7.56e-3$ | 11 | $8.80e-3$ | 09 | $8.78e-1$ | 11 | $4.57e-2$ | 08 | $9.71e-1$ | | 10.0 |

Table 21: Results of the statistics on the second aggregation level for the LPMC_half dataset. Lighter grey tone corresponds to better results compared to darker ones.

| Name | | MAE | | RMSE | | $R^2$ | | SRMSE | | $\rho_{\text{Pearson}}$ | | rank |
|---|---|---|---|---|---|---|---|---|---|---|---|---|
| SGAN_NO_AA | 21 | $6.14e-3$ | 21 | $1.09e-2$ | 13 | $8.69e-1$ | 17 | $4.03e-1$ | 09 | $9.58e-1$ | | 16.2 |
| SGAN_NO_AS | 09 | $5.40e-3$ | 11 | $9.56e-3$ | 06 | $8.94e-1$ | 11 | $3.72e-1$ | 06 | $9.62e-1$ | | 8.6 |
| SGAN_NO_SA | 19 | $5.99e-3$ | 18 | $1.05e-2$ | 07 | $8.92e-1$ | 14 | $3.85e-1$ | 05 | $9.64e-1$ | | 12.6 |
| SGAN_NO_SS | 08 | $5.33e-3$ | 09 | $9.32e-3$ | 04 | $9.15e-1$ | 05 | $3.59e-1$ | 03 | $9.68e-1$ | | 5.8 |
| SGAN_OS_AA | 22 | $6.37e-3$ | 22 | $1.13e-2$ | 26 | $7.08e-1$ | 24 | $4.58e-1$ | 30 | $9.35e-1$ | | 24.8 |
| SGAN_OS_AS | 31 | $9.39e-3$ | 30 | $1.51e-2$ | 20 | $8.30e-1$ | 26 | $5.14e-1$ | 25 | $9.45e-1$ | | 26.4 |
| SGAN_OS_SA | 20 | $6.03e-3$ | 20 | $1.05e-2$ | 24 | $7.50e-1$ | 20 | $4.21e-1$ | 26 | $9.42e-1$ | | 22.0 |
| SGAN_OS_SS | 30 | $9.09e-3$ | 29 | $1.45e-2$ | 15 | $8.62e-1$ | 25 | $4.83e-1$ | 15 | $9.52e-1$ | | 22.8 |
| SGAN_TS_AA | 17 | $5.82e-3$ | 19 | $1.05e-2$ | 22 | $8.27e-1$ | 19 | $4.18e-1$ | 20 | $9.50e-1$ | | 19.4 |
| SGAN_TS_AS | 11 | $5.45e-3$ | 15 | $9.75e-3$ | 18 | $8.44e-1$ | 16 | $4.00e-1$ | 16 | $9.52e-1$ | | 15.2 |
| SGAN_TS_SA | 12 | $5.51e-3$ | 13 | $9.67e-3$ | 16 | $8.54e-1$ | 13 | $3.81e-1$ | 11 | $9.55e-1$ | | 13.0 |
| SGAN_TS_SS | 06 | $5.11e-3$ | 07 | $8.94e-3$ | 12 | $8.72e-1$ | 07 | $3.63e-1$ | 07 | $9.58e-1$ | | 7.8 |
| WGAN_NO_AA | 36 | $1.63e-2$ | 36 | $3.13e-2$ | 36 | $1.27e-1$ | 36 | $1.09$ | 36 | $8.76e-1$ | | 36.0 |

Table 21 – continued from previous page

| Name | | MAE | | RMSE | | $R^2$ | | SRMSE | | $\rho_{\text{Pearson}}$ | rank |
|---|---|---|---|---|---|---|---|---|---|---|---|
| WGAN_NO_AS | 32 | $9.52e-3$ | 34 | $1.86e-2$ | 34 | $5.18e-1$ | 34 | $7.77e-1$ | 35 | $8.95e-1$ | 33.8 |
| WGAN_NO_SA | 35 | $1.43e-2$ | 35 | $2.52e-2$ | 33 | $5.22e-1$ | 35 | $8.02e-1$ | 34 | $9.20e-1$ | 34.4 |
| WGAN_NO_SS | 23 | $7.42e-3$ | 24 | $1.33e-2$ | 17 | $8.51e-1$ | 29 | $5.19e-1$ | 27 | $9.42e-1$ | 24.0 |
| WGAN_OS_AA | 29 | $8.92e-3$ | 32 | $1.65e-2$ | 29 | $6.29e-1$ | 33 | $6.13e-1$ | 32 | $9.33e-1$ | 31.0 |
| WGAN_OS_AS | 34 | $1.03e-2$ | 33 | $1.70e-2$ | 23 | $7.77e-1$ | 32 | $5.94e-1$ | 33 | $9.31e-1$ | 31.0 |
| WGAN_OS_SA | 26 | $8.29e-3$ | 28 | $1.44e-2$ | 25 | $7.41e-1$ | 28 | $5.18e-1$ | 19 | $9.50e-1$ | 25.2 |
| WGAN_OS_SS | 33 | $9.82e-3$ | 31 | $1.56e-2$ | 14 | $8.67e-1$ | 30 | $5.26e-1$ | 21 | $9.49e-1$ | 25.8 |
| WGAN_TS_AA | 25 | $7.99e-3$ | 27 | $1.42e-2$ | 35 | $4.90e-1$ | 31 | $5.49e-1$ | 29 | $9.37e-1$ | 29.4 |
| WGAN_TS_AS | 15 | $5.70e-3$ | 17 | $1.02e-2$ | 11 | $8.80e-1$ | 15 | $3.94e-1$ | 12 | $9.55e-1$ | 14.0 |
| WGAN_TS_SA | 24 | $7.92e-3$ | 26 | $1.35e-2$ | 32 | $5.43e-1$ | 27 | $5.17e-1$ | 23 | $9.47e-1$ | 26.4 |
| WGAN_TS_SS | 16 | $5.70e-3$ | 16 | $9.88e-3$ | 03 | $9.20e-1$ | 12 | $3.75e-1$ | 04 | $9.66e-1$ | 10.2 |
| WGGP_NO_AA | 13 | $5.65e-3$ | 14 | $9.71e-3$ | 31 | $5.90e-1$ | 22 | $4.29e-1$ | 31 | $9.35e-1$ | 22.2 |
| WGGP_NO_AS | 07 | $5.12e-3$ | 06 | $8.93e-3$ | 10 | $8.82e-1$ | 09 | $3.63e-1$ | 17 | $9.51e-1$ | 9.8 |
| WGGP_NO_SA | 10 | $5.42e-3$ | 10 | $9.39e-3$ | 30 | $6.07e-1$ | 18 | $4.15e-1$ | 28 | $9.41e-1$ | 19.2 |
| WGGP_NO_SS | 04 | $4.87e-3$ | 05 | $8.54e-3$ | 05 | $8.98e-1$ | 03 | $3.45e-1$ | 08 | $9.58e-1$ | 5.0 |
| WGGP_OS_AA | 18 | $5.86e-3$ | 12 | $9.58e-3$ | 21 | $8.30e-1$ | 06 | $3.61e-1$ | 18 | $9.51e-1$ | 15.0 |
| WGGP_OS_AS | 28 | $8.58e-3$ | 25 | $1.34e-2$ | 09 | $8.88e-1$ | 23 | $4.37e-1$ | 13 | $9.55e-1$ | 19.6 |
| WGGP_OS_SA | 14 | $5.69e-3$ | 08 | $9.30e-3$ | 19 | $8.34e-1$ | 04 | $3.52e-1$ | 14 | $9.53e-1$ | 11.8 |
| WGGP_OS_SS | 27 | $8.41e-3$ | 23 | $1.31e-2$ | 08 | $8.91e-1$ | 21 | $4.28e-1$ | 10 | $9.57e-1$ | 17.8 |
| WGGP_TS_AA | 05 | $4.91e-3$ | 04 | $8.18e-3$ | 28 | $6.55e-1$ | 10 | $3.67e-1$ | 24 | $9.46e-1$ | 14.2 |
| WGGP_TS_AS | 02 | $4.31e-3$ | 02 | $7.24e-3$ | 02 | $9.29e-1$ | 02 | $2.82e-1$ | 02 | $9.69e-1$ | 2.0 |
| WGGP_TS_SA | 03 | $4.76e-3$ | 03 | $8.02e-3$ | 27 | $6.55e-1$ | 08 | $3.63e-1$ | 22 | $9.48e-1$ | 12.6 |
| WGGP_TS_SS | 01 | $4.19e-3$ | 01 | $7.12e-3$ | 01 | $9.30e-1$ | 01 | $2.79e-1$ | 01 | $9.70e-1$ | 1.0 |

Table 22: Results of the statistics on the third aggregation level for the LPMC_half dataset. Lighter grey tone corresponds to better results compared to darker ones.

| Name | | MAE | | RMSE | | $R^2$ | | SRMSE | | $\rho_{\text{Pearson}}$ | rank |
|---|---|---|---|---|---|---|---|---|---|---|---|
| SGAN_NO_AA | 18 | $2.16e-3$ | 20 | $4.84e-3$ | 15 | $8.30e-1$ | 16 | $7.36e-1$ | 11 | $9.41e-1$ | 16.0 |
| SGAN_NO_AS | 13 | $2.06e-3$ | 12 | $4.46e-3$ | 11 | $8.55e-1$ | 08 | $6.92e-1$ | 08 | $9.45e-1$ | 10.4 |
| SGAN_NO_SA | 16 | $2.11e-3$ | 15 | $4.57e-3$ | 07 | $8.66e-1$ | 10 | $6.94e-1$ | 04 | $9.50e-1$ | 10.4 |
| SGAN_NO_SS | 11 | $2.04e-3$ | 11 | $4.36e-3$ | 04 | $8.84e-1$ | 06 | $6.72e-1$ | 03 | $9.53e-1$ | 7.0 |
| SGAN_OS_AA | 22 | $2.31e-3$ | 22 | $5.22e-3$ | 29 | $7.10e-1$ | 24 | $8.67e-1$ | 31 | $9.14e-1$ | 25.6 |
| SGAN_OS_AS | 31 | $3.03e-3$ | 30 | $6.29e-3$ | 21 | $7.93e-1$ | 28 | $9.40e-1$ | 26 | $9.24e-1$ | 27.2 |
| SGAN_OS_SA | 19 | $2.18e-3$ | 19 | $4.77e-3$ | 23 | $7.71e-1$ | 19 | $7.91e-1$ | 24 | $9.26e-1$ | 20.8 |
| SGAN_OS_SS | 30 | $2.93e-3$ | 26 | $6.00e-3$ | 16 | $8.29e-1$ | 25 | $9.00e-1$ | 17 | $9.33e-1$ | 22.8 |
| SGAN_TS_AA | 20 | $2.19e-3$ | 21 | $4.99e-3$ | 24 | $7.70e-1$ | 22 | $8.04e-1$ | 23 | $9.27e-1$ | 22.0 |
| SGAN_TS_AS | 14 | $2.09e-3$ | 16 | $4.67e-3$ | 20 | $7.95e-1$ | 17 | $7.75e-1$ | 19 | $9.30e-1$ | 17.2 |
| SGAN_TS_SA | 12 | $2.05e-3$ | 14 | $4.53e-3$ | 18 | $8.13e-1$ | 15 | $7.33e-1$ | 16 | $9.36e-1$ | 15.0 |
| SGAN_TS_SS | 09 | $1.96e-3$ | 09 | $4.28e-3$ | 14 | $8.35e-1$ | 12 | $7.10e-1$ | 13 | $9.40e-1$ | 11.4 |
| WGAN_NO_AA | 36 | $5.00e-3$ | 36 | $1.38e-2$ | 36 | $-1.01e-1$ | 36 | $1.98$ | 36 | $8.33e-1$ | 36.0 |
| WGAN_NO_AS | 34 | $3.41e-3$ | 34 | $8.55e-3$ | 35 | $3.75e-1$ | 34 | $1.41$ | 35 | $8.58e-1$ | 34.4 |
| WGAN_NO_SA | 35 | $4.42e-3$ | 35 | $1.06e-2$ | 34 | $5.02e-1$ | 35 | $1.42$ | 34 | $8.85e-1$ | 34.6 |
| WGAN_NO_SS | 28 | $2.82e-3$ | 29 | $6.29e-3$ | 19 | $7.97e-1$ | 30 | $9.81e-1$ | 30 | $9.14e-1$ | 27.2 |
| WGAN_OS_AA | 29 | $2.89e-3$ | 33 | $7.10e-3$ | 32 | $6.27e-1$ | 33 | $1.08$ | 32 | $9.10e-1$ | 31.8 |
| WGAN_OS_AS | 33 | $3.31e-3$ | 32 | $6.97e-3$ | 27 | $7.36e-1$ | 32 | $1.05$ | 33 | $9.04e-1$ | 31.4 |

Table 22 – continued from previous page

| Name | MAE | | RMSE | | $R^2$ | | SRMSE | | $\rho_{\text{Pearson}}$ | | rank |
|---|---|---|---|---|---|---|---|---|---|---|---|
| WGAN_OS_SA | 27 | $2.76e-3$ | 27 | $6.10e-3$ | 22 | $7.75e-1$ | 26 | $9.10e-1$ | 18 | $9.32e-1$ | 24.0 |
| WGAN_OS_SS | 32 | $3.22e-3$ | 31 | $6.58e-3$ | 17 | $8.25e-1$ | 31 | $9.88e-1$ | 22 | $9.27e-1$ | 26.6 |
| WGAN_TS_AA | 24 | $2.62e-3$ | 28 | $6.14e-3$ | 33 | $6.18e-1$ | 29 | $9.80e-1$ | 28 | $9.16e-1$ | 28.4 |
| WGAN_TS_AS | 17 | $2.13e-3$ | 18 | $4.73e-3$ | 13 | $8.45e-1$ | 14 | $7.29e-1$ | 14 | $9.40e-1$ | 15.2 |
| WGAN_TS_SA | 26 | $2.66e-3$ | 25 | $5.78e-3$ | 31 | $6.91e-1$ | 27 | $9.14e-1$ | 25 | $9.25e-1$ | 26.8 |
| WGAN_TS_SS | 21 | $2.21e-3$ | 17 | $4.72e-3$ | 03 | $8.86e-1$ | 13 | $7.19e-1$ | 05 | $9.49e-1$ | 11.8 |
| WGGP_NO_AA | 15 | $2.09e-3$ | 13 | $4.51e-3$ | 30 | $7.03e-1$ | 23 | $8.05e-1$ | 29 | $9.16e-1$ | 22.0 |
| WGGP_NO_AS | 07 | $1.91e-3$ | 08 | $4.15e-3$ | 12 | $8.55e-1$ | 07 | $6.89e-1$ | 15 | $9.38e-1$ | 9.8 |
| WGGP_NO_SA | 10 | $1.99e-3$ | 10 | $4.33e-3$ | 28 | $7.21e-1$ | 18 | $7.80e-1$ | 27 | $9.23e-1$ | 18.6 |
| WGGP_NO_SS | 04 | $1.81e-3$ | 06 | $3.94e-3$ | 05 | $8.72e-1$ | 05 | $6.56e-1$ | 07 | $9.46e-1$ | 5.4 |
| WGGP_OS_AA | 08 | $1.95e-3$ | 07 | $4.02e-3$ | 10 | $8.62e-1$ | 04 | $6.55e-1$ | 12 | $9.40e-1$ | 8.2 |
| WGGP_OS_AS | 25 | $2.65e-3$ | 24 | $5.38e-3$ | 08 | $8.66e-1$ | 21 | $8.03e-1$ | 09 | $9.44e-1$ | 17.4 |
| WGGP_OS_SA | 06 | $1.90e-3$ | 05 | $3.93e-3$ | 09 | $8.65e-1$ | 03 | $6.47e-1$ | 10 | $9.42e-1$ | 6.6 |
| WGGP_OS_SS | 23 | $2.60e-3$ | 23 | $5.27e-3$ | 06 | $8.69e-1$ | 20 | $7.92e-1$ | 06 | $9.46e-1$ | 15.6 |
| WGGP_TS_AA | 05 | $1.83e-3$ | 04 | $3.84e-3$ | 26 | $7.44e-1$ | 11 | $7.00e-1$ | 21 | $9.28e-1$ | 13.4 |
| WGGP_TS_AS | 02 | $1.62e-3$ | 02 | $3.37e-3$ | 02 | $9.09e-1$ | 02 | $5.37e-1$ | 02 | $9.59e-1$ | 2.0 |
| WGGP_TS_SA | 03 | $1.77e-3$ | 03 | $3.76e-3$ | 25 | $7.46e-1$ | 09 | $6.94e-1$ | 20 | $9.30e-1$ | 12.0 |
| WGGP_TS_SS | 01 | $1.57e-3$ | 01 | $3.32e-3$ | 01 | $9.11e-1$ | 01 | $5.34e-1$ | 01 | $9.60e-1$ | 1.0 |

Table 23: Results of the Machine Learning efficacy for the LPMC_half dataset. Lighter grey tone corresponds to better results compared to darker ones.

| Name | Continuous | | Categorical | | rank |
|---|---|---|---|---|---|
| SGAN_NO_AA | 18 | 4.04e1 | 24 | 5.66 | 21.0 |
| SGAN_NO_AS | 20 | 4.27e1 | 12 | 4.34 | 16.0 |
| SGAN_NO_SA | 29 | 6.41e1 | 17 | 4.94 | 23.0 |
| SGAN_NO_SS | 30 | 6.55e1 | 10 | 4.07 | 20.0 |
| SGAN_OS_AA | 11 | 3.41e1 | 19 | 5.10 | 15.0 |
| SGAN_OS_AS | 16 | 3.78e1 | 20 | 5.11 | 18.0 |
| SGAN_OS_SA | 24 | 4.66e1 | 21 | 5.12 | 22.5 |
| SGAN_OS_SS | 26 | 4.85e1 | 18 | 4.96 | 22.0 |
| SGAN_TS_AA | 12 | 3.60e1 | 16 | 4.70 | 14.0 |
| SGAN_TS_AS | 15 | 3.73e1 | 11 | 4.16 | 13.0 |
| SGAN_TS_SA | 22 | 4.50e1 | 15 | 4.64 | 18.5 |
| SGAN_TS_SS | 23 | 4.57e1 | 09 | 4.01 | 16.0 |
| WGAN_NO_AA | 27 | 5.88e1 | 36 | 4.79e4 | 31.5 |
| WGAN_NO_AS | 28 | 6.08e1 | 22 | 5.59 | 25.0 |
| WGAN_NO_SA | 35 | 1.01e2 | 35 | 4.74e4 | 35.0 |
| WGAN_NO_SS | 36 | 1.03e2 | 23 | 5.62 | 29.5 |
| WGAN_OS_AA | 21 | 4.42e1 | 32 | 6.48e3 | 26.5 |
| WGAN_OS_AS | 25 | 4.80e1 | 26 | 5.79 | 25.5 |
| WGAN_OS_SA | 33 | 8.27e1 | 31 | 6.40e3 | 32.0 |
| WGAN_OS_SS | 34 | 8.64e1 | 25 | 5.77 | 29.5 |
| WGAN_TS_AA | 17 | 3.84e1 | 33 | 1.93e4 | 25.0 |
| WGAN_TS_AS | 19 | 4.13e1 | 13 | 4.59 | 16.0 |
| WGAN_TS_SA | 31 | 6.74e1 | 34 | 1.95e4 | 32.5 |

Table 23 – continued from previous page

| Name | Continuous | | Categorical | | rank |
|---|---|---|---|---|---|
| WGAN_TS_SS | 32 | 6.76e1 | 14 | 4.59 | 23.0 |
| WGGP_NO_AA | 09 | 3.39e1 | 30 | 5.60e3 | 19.5 |
| WGGP_NO_AS | 10 | 3.39e1 | 04 | 2.60 | **7.0** |
| WGGP_NO_SA | 13 | 3.61e1 | 29 | 3.20e3 | 21.0 |
| WGGP_NO_SS | 14 | 3.63e1 | 03 | 2.57 | **8.5** |
| WGGP_OS_AA | 01 | 2.36e1 | 06 | 2.70 | **3.5** |
| WGGP_OS_AS | 02 | 2.43e1 | 08 | 3.10 | **5.0** |
| WGGP_OS_SA | 03 | 2.52e1 | 05 | 2.68 | **4.0** |
| WGGP_OS_SS | 04 | 2.59e1 | 07 | 3.10 | **5.5** |
| WGGP_TS_AA | 05 | 2.59e1 | 27 | 1.36e3 | 16.0 |
| WGGP_TS_AS | 06 | 2.62e1 | 02 | 2.54 | **4.0** |
| WGGP_TS_SA | 07 | 2.88e1 | 28 | 3.04e3 | 17.5 |
| WGGP_TS_SS | 08 | 2.90e1 | 01 | 2.54 | **4.5** |

## 5.2 Comparison with state-of-the-art models

### 5.2.1 CMAP case study

Table 24: Results of the statistical assessments between the best DATGAN version and the state-of-the-art models for the CMAP dataset. Lighter grey tone corresponds to better results compared to darker ones.

| Name | MAE | | RMSE | | $R^2$ | | SRMSE | | $\rho_{\text{Pearson}}$ | | rank |
|---|---|---|---|---|---|---|---|---|---|---|---|
| **First aggregation level - all columns** | | | | | | | | | | | |
| CTABGAN | 03 | $2.55e-2$ | 03 | $3.18e-2$ | 03 | $8.17e-1$ | 03 | $1.97e-1$ | 04 | $9.30e-1$ | 3.2 |
| CTGAN | 05 | $3.59e-2$ | 05 | $4.31e-2$ | 04 | $6.37e-1$ | 04 | $2.59e-1$ | 05 | $8.54e-1$ | 4.6 |
| DATGAN (WGAN) | 01 | $6.33e-3$ | 01 | $7.76e-3$ | 01 | $9.88e-1$ | 01 | $5.32e-2$ | 01 | $9.96e-1$ | **1.0** |
| TGAN | 02 | $1.16e-2$ | 02 | $1.50e-2$ | 02 | $9.49e-1$ | 02 | $1.14e-1$ | 02 | $9.81e-1$ | **2.0** |
| TVAE | 04 | $3.33e-2$ | 04 | $4.21e-2$ | 05 | $5.59e-1$ | 05 | $2.72e-1$ | 03 | $9.37e-1$ | 4.2 |
| **First aggregation level - continuous columns** | | | | | | | | | | | |
| CTABGAN | 02 | $1.23e-2$ | 02 | $1.71e-2$ | 02 | $9.37e-1$ | 02 | $1.71e-1$ | 03 | $9.73e-1$ | 2.2 |
| CTGAN | 04 | $2.10e-2$ | 04 | $2.87e-2$ | 04 | $8.27e-1$ | 04 | $2.87e-1$ | 04 | $9.24e-1$ | 4.0 |
| DATGAN (WGAN) | 01 | $7.98e-3$ | 01 | $1.05e-2$ | 01 | $9.70e-1$ | 01 | $1.05e-1$ | 01 | $9.86e-1$ | **1.0** |
| TGAN | 05 | $2.21e-2$ | 05 | $3.07e-2$ | 05 | $7.98e-1$ | 05 | $3.07e-1$ | 05 | $9.18e-1$ | 5.0 |
| TVAE | 03 | $1.32e-2$ | 03 | $1.72e-2$ | 03 | $9.32e-1$ | 03 | $1.72e-1$ | 02 | $9.80e-1$ | 2.8 |
| **First aggregation level - categorical columns** | | | | | | | | | | | |
| CTABGAN | 03 | $2.88e-2$ | 03 | $3.54e-2$ | 03 | $7.87e-1$ | 03 | $2.03e-1$ | 04 | $9.19e-1$ | 3.2 |
| CTGAN | 05 | $3.97e-2$ | 04 | $4.67e-2$ | 04 | $5.90e-1$ | 04 | $2.52e-1$ | 05 | $8.36e-1$ | 4.4 |
| DATGAN (WGAN) | 01 | $5.92e-3$ | 01 | $7.06e-3$ | 01 | $9.92e-1$ | 01 | $4.01e-2$ | 01 | $9.98e-1$ | **1.0** |
| TGAN | 02 | $9.01e-3$ | 02 | $1.10e-2$ | 02 | $9.87e-1$ | 02 | $6.63e-2$ | 02 | $9.96e-1$ | **2.0** |
| TVAE | 04 | $3.83e-2$ | 05 | $4.84e-2$ | 05 | $4.66e-1$ | 05 | $2.97e-1$ | 03 | $9.27e-1$ | 4.4 |
| **Second aggregation level** | | | | | | | | | | | |
| CTABGAN | 03 | $8.17e-3$ | 03 | $1.32e-2$ | 03 | $8.61e-1$ | 03 | $4.39e-1$ | 03 | $9.43e-1$ | 3.0 |
| CTGAN | 04 | $1.04e-2$ | 04 | $1.68e-2$ | 04 | $7.79e-1$ | 04 | $5.53e-1$ | 05 | $9.03e-1$ | 4.2 |
| DATGAN (WGAN) | 01 | $2.83e-3$ | 01 | $4.27e-3$ | 01 | $9.83e-1$ | 01 | $1.56e-1$ | 01 | $9.92e-1$ | **1.0** |
| TGAN | 02 | $4.73e-3$ | 02 | $7.66e-3$ | 02 | $9.29e-1$ | 02 | $2.92e-1$ | 02 | $9.71e-1$ | **2.0** |
| TVAE | 05 | $1.15e-2$ | 05 | $1.96e-2$ | 05 | $5.92e-1$ | 05 | $6.64e-1$ | 04 | $9.13e-1$ | 4.8 |

Table 24 – continued from previous page

| Name | | MAE | | RMSE | | $R^2$ | | SRMSE | | $\rho_{\text{Pearson}}$ | | rank |
|---|---|---|---|---|---|---|---|---|---|---|---|---|
| | | | | | **Third aggregation level** | | | | | | | |
| CTABGAN | 03 | $2.36e-3$ | 03 | $4.74e-3$ | 03 | $8.30e-1$ | 03 | $7.47e-1$ | 03 | $9.26e-1$ | | 3.0 |
| CTGAN | 04 | $2.86e-3$ | 04 | $5.81e-3$ | 04 | $7.51e-1$ | 04 | $9.15e-1$ | 04 | $8.85e-1$ | | 4.0 |
| DATGAN (WGAN) | 01 | $1.09e-3$ | 01 | $1.92e-3$ | 01 | $9.65e-1$ | 01 | $3.31e-1$ | 01 | $9.83e-1$ | | 1.0 |
| TGAN | 02 | $1.63e-3$ | 02 | $3.24e-3$ | 02 | $8.93e-1$ | 02 | $5.43e-1$ | 02 | $9.57e-1$ | | 2.0 |
| TVAE | 05 | $3.47e-3$ | 05 | $7.79e-3$ | 05 | $4.60e-1$ | 05 | $1.18$ | 05 | $8.80e-1$ | | 5.0 |

Table 25: Results of the Machine Learning efficacy between the best DATGAN version and the state-of-the-art models for the CMAP dataset. Lighter grey tone corresponds to better results compared to darker ones.

| Name | | Continuous | | | Categorical | | rank |
|---|---|---|---|---|---|---|---|
| CTABGAN | 03 | 3.33 | | 02 | 1.43 | | 2.5 |
| CTGAN | 05 | 3.45 | | 03 | 1.49 | | 4.0 |
| DATGAN (WGAN) | 01 | 3.05 | | 01 | $7.56e-1$ | | 1.0 |
| TGAN | 02 | 3.24 | | 04 | 8.10e1 | | 3.0 |
| TVAE | 04 | 3.36 | | 05 | 8.04e2 | | 4.5 |

### 5.2.2 LPMC case study

Table 26: Results of the statistical assessments between the best DATGAN version and the state-of-the-art models for the LPMC dataset. Lighter grey tone corresponds to better results compared to darker ones.

| Name | | MAE | | RMSE | | $R^2$ | | SRMSE | | $\rho_{\text{Pearson}}$ | | rank |
|---|---|---|---|---|---|---|---|---|---|---|---|---|
| | | | | **First aggregation level - all columns** | | | | | | | | |
| CTABGAN | 04 | $2.72e-2$ | 04 | $3.47e-2$ | 05 | $-3.97$ | 04 | $2.60e-1$ | 05 | $8.22e-1$ | | 4.4 |
| CTGAN | 03 | $2.62e-2$ | 03 | $3.28e-2$ | 04 | $-2.92$ | 03 | $2.27e-1$ | 04 | $8.71e-1$ | | 3.4 |
| DATGAN (WGGP) | 01 | $7.64e-3$ | 01 | $1.02e-2$ | 01 | $9.63e-1$ | 01 | $8.13e-2$ | 01 | $9.85e-1$ | | 1.0 |
| TGAN | 02 | $1.19e-2$ | 02 | $1.62e-2$ | 02 | $5.88e-1$ | 02 | $1.48e-1$ | 02 | $9.45e-1$ | | 2.0 |
| TVAE | 05 | $2.79e-2$ | 05 | $3.71e-2$ | 03 | $-2.24$ | 05 | $2.76e-1$ | 03 | $8.96e-1$ | | 4.2 |
| | | | | **First aggregation level - continuous columns** | | | | | | | | |
| CTABGAN | 04 | $1.68e-2$ | 04 | $2.60e-2$ | 02 | $9.44e-1$ | 04 | $2.60e-1$ | 02 | $9.80e-1$ | | 3.2 |
| CTGAN | 02 | $1.57e-2$ | 02 | $2.35e-2$ | 03 | $9.25e-1$ | 02 | $2.35e-1$ | 03 | $9.70e-1$ | | 2.4 |
| DATGAN (WGGP) | 01 | $8.93e-3$ | 01 | $1.31e-2$ | 01 | $9.69e-1$ | 01 | $1.31e-1$ | 01 | $9.86e-1$ | | 1.0 |
| TGAN | 03 | $1.68e-2$ | 03 | $2.46e-2$ | 05 | $8.72e-1$ | 03 | $2.46e-1$ | 05 | $9.50e-1$ | | 3.8 |
| TVAE | 05 | $1.72e-2$ | 05 | $2.79e-2$ | 04 | $8.98e-1$ | 05 | $2.79e-1$ | 04 | $9.58e-1$ | | 4.6 |
| | | | | **First aggregation level - categorical columns** | | | | | | | | |
| CTABGAN | 04 | $3.68e-2$ | 04 | $4.28e-2$ | 05 | $-8.54$ | 04 | $2.59e-1$ | 05 | $6.76e-1$ | | 4.4 |
| CTGAN | 03 | $3.60e-2$ | 03 | $4.14e-2$ | 04 | $-6.48$ | 03 | $2.18e-1$ | 04 | $7.79e-1$ | | 3.4 |
| DATGAN (WGGP) | 01 | $6.44e-3$ | 01 | $7.47e-3$ | 01 | $9.56e-1$ | 01 | $3.52e-2$ | 01 | $9.84e-1$ | | 1.0 |
| TGAN | 02 | $7.39e-3$ | 02 | $8.45e-3$ | 02 | $3.24e-1$ | 02 | $5.69e-2$ | 02 | $9.40e-1$ | | 2.0 |
| TVAE | 05 | $3.79e-2$ | 05 | $4.56e-2$ | 03 | $-5.16$ | 05 | $2.74e-1$ | 03 | $8.39e-1$ | | 4.2 |
| | | | | **Second aggregation level** | | | | | | | | |
| CTABGAN | 04 | $8.63e-3$ | 04 | $1.50e-2$ | 05 | $3.11e-2$ | 04 | $6.13e-1$ | 05 | $8.93e-1$ | | 4.4 |
| CTGAN | 03 | $8.30e-3$ | 03 | $1.42e-2$ | 03 | $4.28e-1$ | 03 | $5.39e-1$ | 03 | $9.07e-1$ | | 3.0 |
| DATGAN (WGGP) | 01 | $3.52e-3$ | 01 | $5.91e-3$ | 01 | $9.51e-1$ | 01 | $2.26e-1$ | 01 | $9.78e-1$ | | 1.0 |

Table 26 – continued from previous page

| Name | | MAE | | RMSE | | $R^2$ | | SRMSE | | $\rho_{\text{Pearson}}$ | | rank |
|---|---|---|---|---|---|---|---|---|---|---|---|---|
| TGAN | 02 | 4.83e−3 | 02 | 8.48e−3 | 02 | 8.19e−1 | 02 | 3.74e−1 | 02 | 9.47e−1 | | 2.0 |
| TVAE | 05 | 9.77e−3 | 05 | 1.77e−2 | 04 | 1.34e−1 | 05 | 6.82e−1 | 04 | 8.95e−1 | | 4.6 |
| **Third aggregation level** | | | | | | | | | | | | |
| CTABGAN | 04 | 2.52e−3 | 04 | 5.82e−3 | 04 | 5.09e−1 | 04 | 1.10 | 04 | 8.75e−1 | | 4.0 |
| CTGAN | 03 | 2.44e−3 | 03 | 5.54e−3 | 03 | 7.07e−1 | 03 | 9.67e−1 | 03 | 8.99e−1 | | 3.0 |
| DATGAN (WGGP) | 01 | 1.32e−3 | 01 | 2.79e−3 | 01 | 9.35e−1 | 01 | 4.50e−1 | 01 | 9.69e−1 | | 1.0 |
| TGAN | 02 | 1.71e−3 | 02 | 3.76e−3 | 02 | 8.30e−1 | 02 | 6.89e−1 | 02 | 9.34e−1 | | 2.0 |
| TVAE | 05 | 3.15e−3 | 05 | 7.54e−3 | 05 | 4.07e−1 | 05 | 1.21 | 05 | 8.70e−1 | | 5.0 |

Table 27: Results of the Machine Learning efficacy between the best DATGAN version and the state-of-the-art models for the LPMC dataset. Lighter grey tone corresponds to better results compared to darker ones.

| Name | | Continuous | | Categorical | | rank |
|---|---|---|---|---|---|---|
| CTABGAN | 03 | 3.49e1 | 04 | 5.48 | | 3.5 |
| CTGAN | 04 | 4.33e1 | 02 | 2.89 | | 3.0 |
| DATGAN (WGGP) | 01 | 2.20e1 | 01 | 2.19 | | 1.0 |
| TGAN | 02 | 2.82e1 | 03 | 2.91 | | 2.5 |
| TVAE | 05 | 4.91e1 | 05 | 4.88e2 | | 5.0 |

### 5.2.3 LPMC_half case study

Table 28: Results of the statistical assessments between the best DATGAN version and the state-of-the-art models for the LPMC_half dataset. Lighter grey tone corresponds to better results compared to darker ones.

| Name | | MAE | | RMSE | | $R^2$ | | SRMSE | | $\rho_{\text{Pearson}}$ | | rank |
|---|---|---|---|---|---|---|---|---|---|---|---|---|
| **First aggregation level - all columns** | | | | | | | | | | | | |
| CTABGAN | 03 | 2.21e−2 | 03 | 2.79e−2 | 04 | −2.46 | 03 | 2.06e−1 | 04 | 8.54e−1 | | 3.4 |
| CTGAN | 04 | 2.44e−2 | 04 | 3.00e−2 | 03 | −1.58 | 04 | 2.06e−1 | 05 | 8.24e−1 | | 4.0 |
| DATGAN (WGGP) | 01 | 9.32e−3 | 01 | 1.25e−2 | 01 | 9.11e−1 | 01 | 1.03e−1 | 01 | 9.73e−1 | | 1.0 |
| TGAN | 02 | 1.40e−2 | 02 | 1.88e−2 | 02 | 4.04e−1 | 02 | 1.65e−1 | 02 | 9.39e−1 | | 2.0 |
| TVAE | 05 | 2.97e−2 | 05 | 3.83e−2 | 05 | −2.69 | 05 | 2.68e−1 | 03 | 8.71e−1 | | 4.6 |
| **First aggregation level - continuous columns** | | | | | | | | | | | | |
| CTABGAN | 02 | 1.43e−2 | 03 | 2.21e−2 | 01 | 9.61e−1 | 03 | 2.21e−1 | 01 | 9.86e−1 | | 2.0 |
| CTGAN | 03 | 1.51e−2 | 02 | 2.19e−2 | 03 | 9.15e−1 | 02 | 2.19e−1 | 03 | 9.66e−1 | | 2.6 |
| DATGAN (WGGP) | 01 | 1.12e−2 | 01 | 1.65e−2 | 02 | 9.47e−1 | 01 | 1.65e−1 | 02 | 9.75e−1 | | 1.4 |
| TGAN | 05 | 1.81e−2 | 05 | 2.64e−2 | 05 | 8.46e−1 | 05 | 2.64e−1 | 05 | 9.36e−1 | | 5.0 |
| TVAE | 04 | 1.62e−2 | 04 | 2.57e−2 | 04 | 9.11e−1 | 04 | 2.57e−1 | 04 | 9.60e−1 | | 4.0 |
| **First aggregation level - categorical columns** | | | | | | | | | | | | |
| CTABGAN | 03 | 2.93e−2 | 03 | 3.33e−2 | 04 | −5.63 | 03 | 1.92e−1 | 04 | 7.31e−1 | | 3.4 |
| CTGAN | 04 | 3.30e−2 | 04 | 3.76e−2 | 03 | −3.89 | 04 | 1.94e−1 | 05 | 6.92e−1 | | 4.0 |
| DATGAN (WGGP) | 01 | 7.56e−3 | 01 | 8.80e−3 | 01 | 8.78e−1 | 01 | 4.57e−2 | 01 | 9.71e−1 | | 1.0 |
| TGAN | 02 | 1.03e−2 | 02 | 1.19e−2 | 02 | −5.40e−3 | 02 | 7.36e−2 | 02 | 9.42e−1 | | 2.0 |
| TVAE | 05 | 4.22e−2 | 05 | 5.01e−2 | 05 | −6.04 | 05 | 2.78e−1 | 03 | 7.89e−1 | | 4.6 |
| **Second aggregation level** | | | | | | | | | | | | |
| CTABGAN | 03 | 7.16e−3 | 03 | 1.24e−2 | 04 | 4.22e−1 | 03 | 5.06e−1 | 03 | 9.18e−1 | | 3.2 |

| Name | MAE | | RMSE | | $R^2$ | | SRMSE | | $\rho_{\text{Pearson}}$ | | rank |
|---|---|---|---|---|---|---|---|---|---|---|---|
| CTGAN | 04 | $8.05e-3$ | 04 | $1.36e-2$ | 03 | $5.59e-1$ | 04 | $5.08e-1$ | 04 | $9.09e-1$ | 3.8 |
| DATGAN (WGGP) | 01 | $4.19e-3$ | 01 | $7.12e-3$ | 01 | $9.30e-1$ | 01 | $2.79e-1$ | 01 | $9.70e-1$ | 1.0 |
| TGAN | 02 | $5.48e-3$ | 02 | $9.52e-3$ | 02 | $7.64e-1$ | 02 | $4.13e-1$ | 02 | $9.36e-1$ | 2.0 |
| TVAE | 05 | $1.01e-2$ | 05 | $1.81e-2$ | 05 | $1.54e-1$ | 05 | $6.54e-1$ | 05 | $8.94e-1$ | 5.0 |
| **Third aggregation level** | | | | | | | | | | | |
| CTABGAN | 03 | $2.21e-3$ | 03 | $4.98e-3$ | 04 | $6.85e-1$ | 03 | $9.24e-1$ | 03 | $9.04e-1$ | 3.2 |
| CTGAN | 04 | $2.49e-3$ | 04 | $5.52e-3$ | 03 | $7.34e-1$ | 04 | $9.24e-1$ | 04 | $8.99e-1$ | 3.8 |
| DATGAN (WGGP) | 01 | $1.57e-3$ | 01 | $3.32e-3$ | 01 | $9.11e-1$ | 01 | $5.34e-1$ | 01 | $9.60e-1$ | 1.0 |
| TGAN | 02 | $1.94e-3$ | 02 | $4.20e-3$ | 02 | $7.85e-1$ | 02 | $7.53e-1$ | 02 | $9.20e-1$ | 2.0 |
| TVAE | 05 | $3.24e-3$ | 05 | $7.66e-3$ | 05 | $4.22e-1$ | 05 | $1.16$ | 05 | $8.69e-1$ | 5.0 |

Table 29: Results of the Machine Learning efficacy between the best DATGAN version and the state-of-the-art models for the LPMC_half dataset. Lighter grey tone corresponds to better results compared to darker ones.

| Name | Continuous | | Categorical | | rank |
|---|---|---|---|---|---|
| CTABGAN | 04 | $4.25e1$ | 04 | $4.14$ | 4.0 |
| CTGAN | 05 | $5.13e1$ | 02 | $3.09$ | 3.5 |
| DATGAN (WGGP) | 01 | $2.90e1$ | 01 | $2.54$ | 1.0 |
| TGAN | 02 | $3.05e1$ | 03 | $3.34$ | 2.5 |
| TVAE | 03 | $4.24e1$ | 05 | $3.93e3$ | 4.0 |

### 5.2.4 ADULT case study

Table 30: Results of the statistical assessments between the best DATGAN version and the state-of-the-art models for the ADULT dataset. Lighter grey tone corresponds to better results compared to darker ones.

| Name | MAE | | RMSE | | $R^2$ | | SRMSE | | $\rho_{\text{Pearson}}$ | | rank |
|---|---|---|---|---|---|---|---|---|---|---|---|
| **First aggregation level - all columns** | | | | | | | | | | | |
| CTABGAN | 06 | $2.31e-2$ | 06 | $3.04e-2$ | 06 | $8.64e-1$ | 06 | $2.79e-1$ | 06 | $9.60e-1$ | 6.0 |
| CTGAN | 05 | $1.98e-2$ | 05 | $2.56e-2$ | 05 | $9.40e-1$ | 05 | $2.45e-1$ | 05 | $9.79e-1$ | 5.0 |
| DATGAN (WGAN) | 01 | $3.83e-3$ | 01 | $4.99e-3$ | 01 | $9.98e-1$ | 01 | $4.12e-2$ | 01 | $9.99e-1$ | 1.0 |
| DATGAN (WGGP) | 03 | $1.01e-2$ | 03 | $1.34e-2$ | 03 | $9.81e-1$ | 03 | $1.11e-1$ | 03 | $9.96e-1$ | 3.0 |
| TGAN | 02 | $5.01e-3$ | 02 | $6.89e-3$ | 02 | $9.94e-1$ | 02 | $7.48e-2$ | 02 | $9.98e-1$ | 2.0 |
| TVAE | 04 | $1.06e-2$ | 04 | $1.36e-2$ | 04 | $9.64e-1$ | 04 | $1.18e-1$ | 04 | $9.84e-1$ | 4.0 |
| **First aggregation level - continuous columns** | | | | | | | | | | | |
| CTABGAN | 06 | $1.56e-2$ | 06 | $2.45e-2$ | 06 | $9.26e-1$ | 06 | $2.31e-1$ | 06 | $9.77e-1$ | 6.0 |
| CTGAN | 04 | $1.14e-2$ | 04 | $1.70e-2$ | 05 | $9.63e-1$ | 04 | $1.66e-1$ | 05 | $9.84e-1$ | 4.4 |
| DATGAN (WGAN) | 01 | $5.88e-3$ | 01 | $8.10e-3$ | 01 | $9.96e-1$ | 01 | $7.13e-2$ | 01 | $9.98e-1$ | 1.0 |
| DATGAN (WGGP) | 05 | $1.29e-2$ | 05 | $2.03e-2$ | 04 | $9.66e-1$ | 05 | $1.85e-1$ | 04 | $9.92e-1$ | 4.6 |
| TGAN | 02 | $6.73e-3$ | 02 | $1.00e-2$ | 02 | $9.88e-1$ | 02 | $9.62e-2$ | 02 | $9.95e-1$ | 2.0 |
| TVAE | 03 | $7.61e-3$ | 03 | $1.09e-2$ | 03 | $9.87e-1$ | 03 | $1.04e-1$ | 03 | $9.94e-1$ | 3.0 |
| **First aggregation level - categorical columns** | | | | | | | | | | | |
| CTABGAN | 06 | $2.61e-2$ | 06 | $3.28e-2$ | 06 | $8.40e-1$ | 06 | $2.99e-1$ | 06 | $9.54e-1$ | 6.0 |
| CTGAN | 05 | $2.32e-2$ | 05 | $2.91e-2$ | 05 | $9.31e-1$ | 05 | $2.77e-1$ | 05 | $9.78e-1$ | 5.0 |
| DATGAN (WGAN) | 01 | $3.01e-3$ | 01 | $3.74e-3$ | 01 | $9.99e-1$ | 01 | $2.91e-2$ | 01 | $1.00$ | 1.0 |

Table 30 – continued from previous page

| Name | | MAE | | RMSE | | $R^2$ | | SRMSE | | $\rho_{\text{Pearson}}$ | | rank |
|---|---|---|---|---|---|---|---|---|---|---|---|---|
| DATGAN (WGGP) | 03 | $8.93e-3$ | 03 | $1.07e-2$ | 03 | $9.87e-1$ | 03 | $8.13e-2$ | 03 | $9.98e-1$ | | 3.0 |
| TGAN | 02 | $4.32e-3$ | 02 | $5.65e-3$ | 02 | $9.97e-1$ | 02 | $6.62e-2$ | 02 | $9.99e-1$ | | 2.0 |
| TVAE | 04 | $1.18e-2$ | 04 | $1.46e-2$ | 04 | $9.55e-1$ | 04 | $1.24e-1$ | 04 | $9.80e-1$ | | 4.0 |
| **Second aggregation level** | | | | | | | | | | | | |
| CTABGAN | 06 | $7.18e-3$ | 06 | $1.47e-2$ | 06 | $8.42e-1$ | 06 | $9.00e-1$ | 06 | $9.39e-1$ | | 6.0 |
| CTGAN | 05 | $5.94e-3$ | 05 | $1.21e-2$ | 05 | $9.25e-1$ | 05 | $7.44e-1$ | 05 | $9.70e-1$ | | 5.0 |
| DATGAN (WGAN) | 02 | $1.82e-3$ | 01 | $3.23e-3$ | 01 | $9.95e-1$ | 01 | $1.69e-1$ | 01 | $9.98e-1$ | | 1.2 |
| DATGAN (WGGP) | 03 | $3.22e-3$ | 03 | $6.45e-3$ | 03 | $9.78e-1$ | 03 | $3.45e-1$ | 03 | $9.93e-1$ | | 3.0 |
| TGAN | 01 | $1.76e-3$ | 02 | $3.54e-3$ | 02 | $9.91e-1$ | 02 | $2.31e-1$ | 02 | $9.96e-1$ | | 1.8 |
| TVAE | 04 | $3.46e-3$ | 04 | $6.78e-3$ | 04 | $9.60e-1$ | 04 | $3.78e-1$ | 04 | $9.81e-1$ | | 4.0 |
| **Third aggregation level** | | | | | | | | | | | | |
| CTABGAN | 06 | $2.06e-3$ | 06 | $6.43e-3$ | 06 | $7.82e-1$ | 06 | 1.82 | 06 | $9.12e-1$ | | 6.0 |
| CTGAN | 05 | $1.70e-3$ | 05 | $5.26e-3$ | 05 | $8.96e-1$ | 05 | 1.46 | 05 | $9.56e-1$ | | 5.0 |
| DATGAN (WGAN) | 02 | $6.16e-4$ | 01 | $1.57e-3$ | 01 | $9.90e-1$ | 01 | $4.18e-1$ | 01 | $9.96e-1$ | | 1.2 |
| DATGAN (WGGP) | 03 | $9.61e-4$ | 03 | $2.79e-3$ | 03 | $9.68e-1$ | 03 | $7.11e-1$ | 03 | $9.89e-1$ | | 3.0 |
| TGAN | 01 | $5.90e-4$ | 02 | $1.66e-3$ | 02 | $9.87e-1$ | 02 | $4.55e-1$ | 02 | $9.94e-1$ | | 1.8 |
| TVAE | 04 | $1.09e-3$ | 04 | $3.13e-3$ | 04 | $9.45e-1$ | 04 | $7.70e-1$ | 04 | $9.73e-1$ | | 4.0 |

Table 31: Results of the Machine Learning efficacy between the best DATGAN version and the state-of-the-art models for the ADULT dataset. Lighter grey tone corresponds to better results compared to darker ones.

| Name | | Continuous | | | Categorical | | rank |
|---|---|---|---|---|---|---|---|
| CTABGAN | 05 | 4.73 | | 04 | 8.56e3 | | 4.5 |
| CTGAN | 03 | 4.32 | | 03 | 1.04e3 | | 3.0 |
| DATGAN (WGAN) | 04 | 4.41 | | 02 | 1.91 | | 3.0 |
| DATGAN (WGGP) | 06 | 4.81 | | 01 | 1.09 | | 3.5 |
| TGAN | 01 | 4.13 | | 06 | 1.00e4 | | 3.5 |
| TVAE | 02 | 4.31 | | 05 | 8.64e3 | | 3.5 |

## 5.3 Sensitivity anaylsis of the DAG - Structure of the DAG

### 5.3.1 CMAP case study

Table 32: Results of the statistical assessments with the different DAGs for the CMAP dataset. Lighter grey tone corresponds to better results compared to darker ones.

| Name | | MAE | | RMSE | | $R^2$ | | SRMSE | | $\rho_{\text{Pearson}}$ | | rank |
|---|---|---|---|---|---|---|---|---|---|---|---|---|
| **First aggregation level - all columns** | | | | | | | | | | | | |
| full | 05 | $7.31e-3$ | 05 | $9.45e-3$ | 03 | $9.85e-1$ | 05 | $6.49e-2$ | 03 | $9.95e-1$ | | 4.2 |
| trans. red. | 03 | $6.70e-3$ | 04 | $8.68e-3$ | 01 | $9.88e-1$ | 03 | $6.19e-2$ | 01 | $9.95e-1$ | | 2.4 |
| linear | 04 | $6.71e-3$ | 03 | $8.48e-3$ | 02 | $9.86e-1$ | 02 | $6.00e-2$ | 02 | $9.95e-1$ | | 2.6 |
| prediction | 02 | $6.49e-3$ | 02 | $8.01e-3$ | 05 | $9.72e-1$ | 04 | $6.25e-2$ | 05 | $9.88e-1$ | | 3.6 |
| no links | 01 | $6.11e-3$ | 01 | $7.57e-3$ | 04 | $9.78e-1$ | 01 | $5.90e-2$ | 04 | $9.90e-1$ | | 2.2 |
| **First aggregation level - continuous columns** | | | | | | | | | | | | |
| full | 01 | $8.75e-3$ | 01 | $1.26e-2$ | 02 | $9.61e-1$ | 01 | $1.26e-1$ | 03 | $9.81e-1$ | | 1.6 |

Continues on next page...xxix

Table 32 – continued from previous page

| Name | MAE | | RMSE | | $R^2$ | | SRMSE | | $\rho_{\text{Pearson}}$ | | rank |
|---|---|---|---|---|---|---|---|---|---|---|---|
| trans. red. | 02 | 9.34e−3 | 03 | 1.35e−2 | 01 | 9.63e−1 | 03 | 1.35e−1 | 01 | 9.83e−1 | 2.0 |
| linear | 03 | 9.61e−3 | 02 | 1.32e−2 | 03 | 9.61e−1 | 02 | 1.32e−1 | 02 | 9.83e−1 | 2.4 |
| prediction | 05 | 1.56e−2 | 05 | 1.96e−2 | 05 | 8.82e−1 | 05 | 1.96e−1 | 05 | 9.42e−1 | 5.0 |
| no links | 04 | 1.39e−2 | 04 | 1.77e−2 | 04 | 9.06e−1 | 04 | 1.77e−1 | 04 | 9.54e−1 | 4.0 |
| **First aggregation level - categorical columns** | | | | | | | | | | | |
| full | 05 | 6.94e−3 | 05 | 8.65e−3 | 05 | 9.91e−1 | 05 | 4.95e−2 | 04 | 9.98e−1 | 4.8 |
| trans. red. | 04 | 6.04e−3 | 04 | 7.47e−3 | 02 | 9.94e−1 | 04 | 4.36e−2 | 03 | 9.98e−1 | 3.4 |
| linear | 03 | 5.99e−3 | 03 | 7.30e−3 | 04 | 9.93e−1 | 03 | 4.20e−2 | 05 | 9.98e−1 | 3.6 |
| prediction | 02 | 4.20e−3 | 02 | 5.10e−3 | 03 | 9.94e−1 | 01 | 2.91e−2 | 01 | 9.99e−1 | 1.8 |
| no links | 01 | 4.16e−3 | 01 | 5.03e−3 | 01 | 9.96e−1 | 02 | 2.94e−2 | 02 | 9.99e−1 | 1.4 |
| **Second aggregation level** | | | | | | | | | | | |
| full | 03 | 3.22e−3 | 03 | 5.08e−3 | 02 | 9.77e−1 | 03 | 1.85e−1 | 02 | 9.89e−1 | 2.6 |
| trans. red. | 01 | 3.13e−3 | 01 | 4.93e−3 | 01 | 9.78e−1 | 01 | 1.81e−1 | 01 | 9.90e−1 | 1.0 |
| linear | 02 | 3.21e−3 | 02 | 4.95e−3 | 03 | 9.76e−1 | 02 | 1.84e−1 | 03 | 9.89e−1 | 2.4 |
| prediction | 04 | 5.87e−3 | 04 | 9.27e−3 | 05 | 9.14e−1 | 05 | 3.48e−1 | 05 | 9.56e−1 | 4.6 |
| no links | 05 | 6.03e−3 | 05 | 9.49e−3 | 04 | 9.16e−1 | 04 | 3.48e−1 | 04 | 9.57e−1 | 4.4 |
| **Third aggregation level** | | | | | | | | | | | |
| full | 02 | 1.19e−3 | 02 | 2.20e−3 | 02 | 9.57e−1 | 02 | 3.80e−1 | 02 | 9.79e−1 | 2.0 |
| trans. red. | 01 | 1.19e−3 | 01 | 2.19e−3 | 01 | 9.58e−1 | 01 | 3.76e−1 | 01 | 9.80e−1 | 1.0 |
| linear | 03 | 1.23e−3 | 03 | 2.24e−3 | 03 | 9.54e−1 | 03 | 3.90e−1 | 03 | 9.78e−1 | 3.0 |
| prediction | 04 | 2.36e−3 | 04 | 4.55e−3 | 04 | 8.31e−1 | 04 | 7.85e−1 | 04 | 9.11e−1 | 4.0 |
| no links | 05 | 2.45e−3 | 05 | 4.70e−3 | 05 | 8.31e−1 | 05 | 7.91e−1 | 05 | 9.11e−1 | 5.0 |

Table 33: Results of the Machine Learning efficacy with the different DAGs for the CMAP dataset. Lighter grey tone corresponds to better results compared to darker ones.

| Name | Continuous | | Categorical | | rank |
|---|---|---|---|---|---|
| full | 01 | 3.13 | 01 | 8.69e−1 | 1.0 |
| trans. red. | 02 | 3.13 | 02 | 9.17e−1 | 2.0 |
| linear | 03 | 3.20 | 03 | 1.11 | 3.0 |
| prediction | 04 | 6.08 | 04 | 3.62 | 4.0 |
| no links | 05 | 6.22 | 05 | 3.95 | 5.0 |

### 5.3.2 LPMC case study

Table 34: Results of the statistical assessments with the different DAGs for the LPMC dataset. Lighter grey tone corresponds to better results compared to darker ones.

| Name | MAE | | RMSE | | $R^2$ | | SRMSE | | $\rho_{\text{Pearson}}$ | | rank |
|---|---|---|---|---|---|---|---|---|---|---|---|
| **First aggregation level - all columns** | | | | | | | | | | | |
| full | 02 | 7.74e−3 | 03 | 1.03e−2 | 02 | 9.54e−1 | 03 | 8.29e−2 | 02 | 9.83e−1 | 2.4 |
| trans. red. | 01 | 7.04e−3 | 01 | 9.50e−3 | 01 | 9.55e−1 | 02 | 7.73e−2 | 01 | 9.83e−1 | 1.2 |
| linear | 03 | 7.99e−3 | 02 | 1.03e−2 | 03 | 9.20e−1 | 01 | 7.60e−2 | 03 | 9.72e−1 | 2.4 |
| prediction | 04 | 1.59e−2 | 04 | 2.21e−2 | 05 | 8.71e−1 | 04 | 2.06e−1 | 05 | 9.37e−1 | 4.4 |
| no links | 05 | 1.60e−2 | 05 | 2.27e−2 | 04 | 8.79e−1 | 05 | 2.12e−1 | 04 | 9.49e−1 | 4.6 |

Table 34 – continued from previous page

| Name | MAE | | RMSE | | $R^2$ | | SRMSE | | $\rho_{\text{Pearson}}$ | | rank |
|---|---|---|---|---|---|---|---|---|---|---|---|
| | | | | | **First aggregation level - continuous columns** | | | | | | |
| full       | 03 | $9.34\text{e}-3$ | 03 | $1.35\text{e}-2$ | 02 | $9.55\text{e}-1$ | 03 | $1.35\text{e}-1$ | 03 | $9.80\text{e}-1$ | 2.8 |
| trans. red.| 02 | $8.46\text{e}-3$ | 02 | $1.26\text{e}-2$ | 03 | $9.53\text{e}-1$ | 02 | $1.26\text{e}-1$ | 02 | $9.81\text{e}-1$ | 2.2 |
| linear     | 01 | $7.92\text{e}-3$ | 01 | $1.15\text{e}-2$ | 01 | $9.75\text{e}-1$ | 01 | $1.15\text{e}-1$ | 01 | $9.88\text{e}-1$ | 1.0 |
| prediction | 04 | $2.79\text{e}-2$ | 04 | $4.01\text{e}-2$ | 05 | $7.84\text{e}-1$ | 04 | $4.01\text{e}-1$ | 05 | $8.93\text{e}-1$ | 4.4 |
| no links   | 05 | $2.80\text{e}-2$ | 05 | $4.14\text{e}-2$ | 04 | $7.92\text{e}-1$ | 05 | $4.14\text{e}-1$ | 04 | $9.06\text{e}-1$ | 4.6 |
| | | | | | **First aggregation level - categorical columns** | | | | | | |
| full       | 04 | $6.27\text{e}-3$ | 04 | $7.34\text{e}-3$ | 03 | $9.53\text{e}-1$ | 04 | $3.41\text{e}-2$ | 03 | $9.86\text{e}-1$ | 3.6 |
| trans. red.| 03 | $5.72\text{e}-3$ | 03 | $6.61\text{e}-3$ | 02 | $9.57\text{e}-1$ | 03 | $3.20\text{e}-2$ | 02 | $9.86\text{e}-1$ | 2.6 |
| linear     | 05 | $8.05\text{e}-3$ | 05 | $9.10\text{e}-3$ | 05 | $8.69\text{e}-1$ | 05 | $3.94\text{e}-2$ | 05 | $9.57\text{e}-1$ | 5.0 |
| prediction | 01 | $4.70\text{e}-3$ | 02 | $5.39\text{e}-3$ | 04 | $9.52\text{e}-1$ | 02 | $2.57\text{e}-2$ | 04 | $9.77\text{e}-1$ | 2.6 |
| no links   | 02 | $4.74\text{e}-3$ | 01 | $5.37\text{e}-3$ | 01 | $9.59\text{e}-1$ | 01 | $2.47\text{e}-2$ | 01 | $9.88\text{e}-1$ | 1.2 |
| | | | | | **Second aggregation level** | | | | | | |
| full       | 03 | $3.53\text{e}-3$ | 03 | $5.94\text{e}-3$ | 02 | $9.52\text{e}-1$ | 03 | $2.28\text{e}-1$ | 02 | $9.78\text{e}-1$ | 2.6 |
| trans. red.| 02 | $3.36\text{e}-3$ | 02 | $5.75\text{e}-3$ | 03 | $9.51\text{e}-1$ | 02 | $2.23\text{e}-1$ | 03 | $9.78\text{e}-1$ | 2.4 |
| linear     | 01 | $2.95\text{e}-3$ | 01 | $4.90\text{e}-3$ | 01 | $9.60\text{e}-1$ | 01 | $1.87\text{e}-1$ | 01 | $9.83\text{e}-1$ | 1.0 |
| prediction | 04 | $9.08\text{e}-3$ | 04 | $1.58\text{e}-2$ | 04 | $8.16\text{e}-1$ | 04 | $6.64\text{e}-1$ | 05 | $9.06\text{e}-1$ | 4.2 |
| no links   | 05 | $9.18\text{e}-3$ | 05 | $1.63\text{e}-2$ | 05 | $8.14\text{e}-1$ | 05 | $6.74\text{e}-1$ | 04 | $9.11\text{e}-1$ | 4.8 |
| | | | | | **Third aggregation level** | | | | | | |
| full       | 03 | $1.33\text{e}-3$ | 03 | $2.78\text{e}-3$ | 02 | $9.37\text{e}-1$ | 03 | $4.47\text{e}-1$ | 02 | $9.70\text{e}-1$ | 2.6 |
| trans. red.| 02 | $1.29\text{e}-3$ | 02 | $2.76\text{e}-3$ | 03 | $9.36\text{e}-1$ | 02 | $4.43\text{e}-1$ | 03 | $9.70\text{e}-1$ | 2.4 |
| linear     | 01 | $1.06\text{e}-3$ | 01 | $2.14\text{e}-3$ | 01 | $9.55\text{e}-1$ | 01 | $3.53\text{e}-1$ | 01 | $9.79\text{e}-1$ | 1.0 |
| prediction | 04 | $3.39\text{e}-3$ | 04 | $7.64\text{e}-3$ | 04 | $7.39\text{e}-1$ | 05 | $1.38$ | 05 | $8.64\text{e}-1$ | 4.4 |
| no links   | 05 | $3.45\text{e}-3$ | 05 | $7.92\text{e}-3$ | 05 | $7.29\text{e}-1$ | 04 | $1.37$ | 04 | $8.66\text{e}-1$ | 4.6 |

Table 35: Results of the Machine Learning efficacy with the different DAGs for the LPMC dataset. Lighter grey tone corresponds to better results compared to darker ones.

| Name | Continuous | | Categorical | | rank |
|---|---|---|---|---|---|
| full       | 01 | $2.14\text{e}1$ | 02 | $2.06$ | 1.5 |
| trans. red.| 02 | $2.16\text{e}1$ | 03 | $2.09$ | 2.5 |
| linear     | 03 | $2.18\text{e}1$ | 01 | $1.96$ | 2.0 |
| prediction | 04 | $4.94\text{e}2$ | 04 | $6.14$ | 4.0 |
| no links   | 05 | $5.06\text{e}2$ | 05 | $6.45$ | 5.0 |

### 5.3.3 LPMC_half case study

Table 36: Results of the statistical assessments with the different DAGs for the LPMC_half dataset. Lighter grey tone corresponds to better results compared to darker ones.

| Name | MAE | | RMSE | | $R^2$ | | SRMSE | | $\rho_{\text{Pearson}}$ | | rank |
|---|---|---|---|---|---|---|---|---|---|---|---|
| | | | | | **First aggregation level - all columns** | | | | | | |
| full       | 01 | $9.29\text{e}-3$ | 01 | $1.22\text{e}-2$ | 01 | $9.24\text{e}-1$ | 02 | $9.85\text{e}-2$ | 02 | $9.67\text{e}-1$ | 1.4 |
| trans. red.| 03 | $1.38\text{e}-2$ | 05 | $1.98\text{e}-2$ | 05 | $7.75\text{e}-1$ | 03 | $1.71\text{e}-1$ | 04 | $9.45\text{e}-1$ | 4.0 |
| linear     | 02 | $9.74\text{e}-3$ | 02 | $1.26\text{e}-2$ | 02 | $8.85\text{e}-1$ | 01 | $9.79\text{e}-2$ | 01 | $9.72\text{e}-1$ | 1.6 |

Table 36 – continued from previous page

| Name | MAE | | RMSE | | $R^2$ | | SRMSE | | $\rho_{\text{Pearson}}$ | | rank |
|---|---|---|---|---|---|---|---|---|---|---|---|
| prediction | 05 | $1.47\text{e}-2$ | 04 | $1.97\text{e}-2$ | 04 | $8.60\text{e}-1$ | 05 | $1.78\text{e}-1$ | 05 | $9.41\text{e}-1$ | 4.6 |
| no links | 04 | $1.42\text{e}-2$ | 03 | $1.90\text{e}-2$ | 03 | $8.62\text{e}-1$ | 04 | $1.72\text{e}-1$ | 03 | $9.45\text{e}-1$ | 3.4 |
| **First aggregation level - continuous columns** | | | | | | | | | | | |
| full | 02 | $1.10\text{e}-2$ | 02 | $1.60\text{e}-2$ | 02 | $9.45\text{e}-1$ | 02 | $1.60\text{e}-1$ | 02 | $9.73\text{e}-1$ | 2.0 |
| trans. red. | 03 | $1.90\text{e}-2$ | 03 | $2.97\text{e}-2$ | 05 | $7.12\text{e}-1$ | 03 | $2.97\text{e}-1$ | 03 | $9.29\text{e}-1$ | 3.4 |
| linear | 01 | $1.06\text{e}-2$ | 01 | $1.53\text{e}-2$ | 01 | $9.47\text{e}-1$ | 01 | $1.53\text{e}-1$ | 01 | $9.79\text{e}-1$ | 1.0 |
| prediction | 05 | $2.39\text{e}-2$ | 05 | $3.34\text{e}-2$ | 04 | $7.67\text{e}-1$ | 05 | $3.34\text{e}-1$ | 05 | $8.96\text{e}-1$ | 4.8 |
| no links | 04 | $2.34\text{e}-2$ | 04 | $3.25\text{e}-2$ | 03 | $7.69\text{e}-1$ | 04 | $3.25\text{e}-1$ | 04 | $9.02\text{e}-1$ | 3.8 |
| **First aggregation level - categorical columns** | | | | | | | | | | | |
| full | 03 | $7.66\text{e}-3$ | 03 | $8.69\text{e}-3$ | 03 | $9.06\text{e}-1$ | 03 | $4.11\text{e}-2$ | 04 | $9.61\text{e}-1$ | 3.2 |
| trans. red. | 04 | $8.91\text{e}-3$ | 05 | $1.06\text{e}-2$ | 04 | $8.34\text{e}-1$ | 05 | $5.43\text{e}-2$ | 05 | $9.61\text{e}-1$ | 4.6 |
| linear | 05 | $8.92\text{e}-3$ | 04 | $1.01\text{e}-2$ | 05 | $8.26\text{e}-1$ | 04 | $4.70\text{e}-2$ | 03 | $9.66\text{e}-1$ | 4.2 |
| prediction | 02 | $6.07\text{e}-3$ | 02 | $7.00\text{e}-3$ | 02 | $9.45\text{e}-1$ | 02 | $3.20\text{e}-2$ | 02 | $9.83\text{e}-1$ | 2.0 |
| no links | 01 | $5.65\text{e}-3$ | 01 | $6.46\text{e}-3$ | 01 | $9.48\text{e}-1$ | 01 | $2.93\text{e}-2$ | 01 | $9.86\text{e}-1$ | 1.0 |
| **Second aggregation level** | | | | | | | | | | | |
| full | 02 | $4.16\text{e}-3$ | 02 | $6.98\text{e}-3$ | 01 | $9.37\text{e}-1$ | 02 | $2.69\text{e}-1$ | 02 | $9.71\text{e}-1$ | 1.8 |
| trans. red. | 03 | $5.90\text{e}-3$ | 03 | $1.11\text{e}-2$ | 05 | $7.68\text{e}-1$ | 03 | $4.58\text{e}-1$ | 03 | $9.36\text{e}-1$ | 3.4 |
| linear | 01 | $3.88\text{e}-3$ | 01 | $6.51\text{e}-3$ | 02 | $9.29\text{e}-1$ | 01 | $2.54\text{e}-1$ | 01 | $9.73\text{e}-1$ | 1.2 |
| prediction | 05 | $8.40\text{e}-3$ | 05 | $1.46\text{e}-2$ | 03 | $8.17\text{e}-1$ | 05 | $6.10\text{e}-1$ | 05 | $9.10\text{e}-1$ | 4.6 |
| no links | 04 | $8.38\text{e}-3$ | 04 | $1.45\text{e}-2$ | 04 | $8.17\text{e}-1$ | 04 | $6.03\text{e}-1$ | 04 | $9.12\text{e}-1$ | 4.0 |
| **Third aggregation level** | | | | | | | | | | | |
| full | 02 | $1.52\text{e}-3$ | 02 | $3.21\text{e}-3$ | 01 | $9.22\text{e}-1$ | 02 | $5.19\text{e}-1$ | 02 | $9.63\text{e}-1$ | 1.8 |
| trans. red. | 03 | $2.05\text{e}-3$ | 03 | $5.03\text{e}-3$ | 05 | $6.95\text{e}-1$ | 03 | $8.69\text{e}-1$ | 03 | $9.19\text{e}-1$ | 3.4 |
| linear | 01 | $1.38\text{e}-3$ | 01 | $2.90\text{e}-3$ | 02 | $9.18\text{e}-1$ | 01 | $4.87\text{e}-1$ | 01 | $9.65\text{e}-1$ | 1.2 |
| prediction | 04 | $3.26\text{e}-3$ | 05 | $7.36\text{e}-3$ | 03 | $7.44\text{e}-1$ | 05 | $1.28$ | 05 | $8.69\text{e}-1$ | 4.4 |
| no links | 05 | $3.27\text{e}-3$ | 04 | $7.35\text{e}-3$ | 04 | $7.43\text{e}-1$ | 04 | $1.28$ | 04 | $8.70\text{e}-1$ | 4.2 |

Table 37: Results of the Machine Learning efficacy with the different DAGs for the LPMC_half dataset. Lighter grey tone corresponds to better results compared to darker ones.

| Name | Continuous | | Categorical | | rank |
|---|---|---|---|---|---|
| full | 02 | $2.86\text{e}1$ | 02 | $2.55$ | 2.0 |
| trans. red. | 03 | $3.67\text{e}1$ | 03 | $2.68$ | 3.0 |
| linear | 01 | $2.81\text{e}1$ | 01 | $2.44$ | 1.0 |
| prediction | 04 | $4.90\text{e}2$ | 04 | $6.07$ | 4.0 |
| no links | 05 | $4.99\text{e}2$ | 05 | $6.45$ | 5.0 |


\fi

\end{document}

%% file: text/00_abstract.tex
\begin{abstract}

Synthetic data can be used in various applications, such as correcting bias datasets or replacing scarce original data for simulation purposes. Generative Adversarial Networks (GANs) are considered state-of-the-art for developing generative models. However, these deep learning models are data-driven, and it is, thus, difficult to control the generation process. It can, therefore, lead to the following issues: lack of representativity in the generated data, the introduction of bias, and the possibility of overfitting the sample's noise. This article presents the Directed Acyclic Tabular GAN (DATGAN) to address these limitations by integrating expert knowledge in deep learning models for synthetic tabular data generation. This approach allows the interactions between variables to be specified explicitly using a Directed Acyclic Graph (DAG). The DAG is then converted to a network of modified Long Short-Term Memory (LSTM) cells to accept multiple inputs. Multiple DATGAN versions are systematically tested on multiple assessment metrics. We show that the best versions of the DATGAN outperform state-of-the-art generative models on multiple case studies. Finally, we show how the DAG can create hypothetical synthetic datasets.

\end{abstract}

\vfill

\keywords{Tabular Data Synthesis \and Generative Adversarial Networks \and Expert Knowledge}

%% file: text/01_introduction.tex
\section{Introduction}
\label{sec:introduction}

A massive increase in data availability has created tremendous opportunities for targeted modeling and a greater understanding of systems, particularly those involving human behavior. However, reliance on data creates a division based on data. For example, leading international cities in developed nations produce rich data about population movements and interactions with infrastructures. On the other hand, undeveloped nations have much lower data availability. The collection of such data, particularly socio-economic, can be prohibitively expensive. It can, thus, prevent non-data-rich areas from modeling. Furthermore, data can be controlled by certain groups (companies, government, or public agencies), who may be unwilling or unable to make full data publicly available. In addition, the sharing of detailed disaggregated socio-economic data has become increasingly complex with the current focus on data privacy (GDPR). Thus, synthetic data generation, \emph{i.e.} the creation of synthetic data samples which are consistent with the true population, has the opportunity to address many of these limitations.  

There are multiple use cases for synthetic tabular data: \begin{inlinelist} 
\item The most common use-case is dataset augmentation. It can allow researchers and modelers to approximate a large population from a smaller sample, thus reducing the cost of data collection. 
\item Secondly, synthetic data can be used for privacy preservation. It can, then, enable the sharing of detailed disaggregate populations without contravening GDPR and other data privacy laws. 
\item Another use case is bias correction. Synthetic data can correct bias in existing samples, allowing for reliable modeling of marginal and minority groups and behavior.
\item Finally, synthetic data generation models can be used as transfer learning methods. They can thus be used to transfer data from one city or context to a new context, allowing for detailed modeling where existing high-quality data is not available. 
\end{inlinelist}
In this paper, we focus mainly on synthetic population generation. Such populations are generally used for simulation in agent-based models, particularly for activity-based transport models. However, the techniques proposed and reviewed in this article can be used in any context where there is a need for detailed tabular datasets.

Many methods have been developed to generate such synthetic populations in existing studies. The two main approaches are statistical techniques such as Iterative Proportional Fitting (IPF)~\citep{deming_least_1940} or simulation using Gibbs sampling~\citep{geman_stochastic_1984}, and machine learning techniques. While the first approaches have been well studied within the transportation community, the latter comes from the Machine Learning community and generally focuses on general synthetic data. These deep learning methods, such as Generative Adversarial Networks (GANs)~\citep{goodfellow_generative_2014}, have already been tested against standard statistical techniques and outperform them while generating correlations in synthetic datasets. However, these methods are data-driven. They, therefore, lack control over the generation process. Without controlling the latter, it is impossible to know how well the deep learning models have understood the original sample, \emph{i.e.} which correlations between the variables in the original dataset the models have learned. It can, thus, lead to spurious correlations or propagation of existing bias in the sample. In addition, there is generally no focus on the representativity of the output, which is crucial for the accurate understanding of socio-economic characteristics. 

This article, thus, proposes a novel model that controls the generation process of such synthetic tabular data. We propose to let the researcher or modeler design a network to represent the interactions between the variables with a Directed Acyclic Graph (DAG). This DAG is then used to model the structure of the model that will generate such a population. Allowing researchers to control the process has three main advantages: they can tinker with the data generation process, create hypothetical datasets, and control the dependencies for forecasting. In this article, we thus present our new GAN model named Directed Acyclic Table GAN (DATGAN). We show that it outperforms state-of-the-art synthetic data generators on multiple metrics. These metrics have been created to allow for systematic testing both using formal statistical analysis and supervised learning-based approaches. We also provide a sensitivity analysis on the DAG to show its effect on the data generation process. Finally, we show how the DAG can create hypothetical situations and generate a synthetic dataset based on the new rules.

The rest of this article is laid as follows. In the next section, we present the Literature Review. We first introduce the existing approaches for population synthesis and then discuss the different research axes. Finally, we conclude the literature review with the opportunities and limitations of existing research. In Section~\ref{sec:methodology}, we present the whole methodology for the DATGAN. We discuss how to preprocess the data, what models are used for the generator and the discriminator, and how to use the DAG to create the generator's structure using LSTM cells. Section~\ref{sec:case_study} presents the case studies and Section~\ref{sec:results} shows the results. We conclude this article in Section~\ref{sec:conclusion} and give ideas for future work. In addition to this article, we provide supplementary materials containing a notation table for the methodology, a comparison of multiple DATGAN versions, and all the detailed results used in Section~\ref{sec:results}.

%% file: text/02_lit_rev.tex
\section{Literature Review}
\label{sec:lit_rev}

There are five main research axes for synthetic tabular data generation: simulation/activity-based modeling, Machine Learning efficacy, bias correction, privacy preservation, and transfer learning. These research axes are discussed in detail in Section~\ref{sec:research_axes}. The literature review first focuses on population synthesis with older methods such as Iterative Proportional Fitting (IPF) and Gibbs sampling. Then, we look at more general Machine Learning techniques for synthetic tabular data generation. These techniques are discussed in detail in Section~\ref{sec:existing_approaches}. Then, in Section~\ref{sec:sota_models}, we discuss in more detail some state-of-the-art models that we selected to compare to the model presented in this article. Next, Section~\ref{sec:model_eval} is dedicated to model evaluation and shows how the transportation and Machine Learning communities are evaluating generated synthetic datasets. Finally, in Section~\ref{sec:opportunities}, we discuss the opportunities and limitations of these techniques linked to the five research axes. 

\afterpage{
\begin{landscape}
\mbox{}\vfill
\begin{table}[!h]
    \centering
    \caption{Main methods for synthetic tabular data generation found in the transportation literature and in the Machine Learning community.}
    \label{tab:existing_techniques} 
    \begin{tabularx}{\hsize}{>{\hsize=.5\hsize}C|>{\hsize=.5\hsize}C||>{\hsize=1.45\hsize}X|>{\hsize=.85\hsize}C|>{\hsize=1.35\hsize}L|>{\hsize=1.35\hsize}L}
        \multicolumn{1}{c|}{\textbf{Categories}} & \multicolumn{1}{c||}{\textbf{Methods}} & \multicolumn{1}{c|}{\textbf{Description}} & \multicolumn{1}{c|}{\textbf{References}} & \multicolumn{1}{c|}{\textbf{Advantages}} & \multicolumn{1}{c}{\textbf{Disadvantages}}  \\ \midrule[1.5pt] 
        
        \multirow{6}{*}{\makecell{Statistical\\techniques}} & 
        IPF & 
        Iterative method using marginals to create a synthetic table. & 
        \cite{auld_population_2009};
        \cite{barthelemy_synthetic_2013};
        \cite{rich_large-scale_2018}
        &
        \tabitem Efficient in its basic form \nl
        \hspace*{0pt}\tabitem Simple to implement 
        & 
        \tabitem No interactions between variables \nl
        \hspace*{0pt}\tabitem Computationally expensive if more complexity added \nl
        \hspace*{0pt}\tabitem Prone to sampling zero issue \nl
        \hspace*{0pt}\tabitem No differences between data types
        \\ \cline{2-6}
        
        & 
        Gibbs sampling & 
        Gibbs sampler trained until reaching stationary state on prepared conditionals. & 
        \cite{farooq_simulation_2013}; 
        \cite{casati_synthetic_2015}; 
        \cite{kim_simulated_2016}
        & 
        \tabitem Learn from marginals  \nl
        \hspace*{0pt}\tabitem Outperforms IPF  \nl
        \hspace*{0pt}\tabitem Can be linked to other methods
        & 
        \tabitem No interactions between variables \nl
        \hspace*{0pt}\tabitem Computationally expensive \nl
        \hspace*{0pt}\tabitem Probability distributions are an assumption \\ \midrule
        
        & 
        Bayesian networks & 
        Probabilistic graphical model used to determine probabilistic inferences between the variables. & 
        \cite{sun_bayesian_2015};
        \cite{zhang_connected_2019}
        & 
        \tabitem Dependencies of variables defined prior to training \nl
        \hspace*{0pt}\tabitem Probabilistic model
        &
        \tabitem Requires prior information on the dataset \nl
        \hspace*{0pt}\tabitem Computationally expensive when dealing with large and sparse datasets
        \\ \cline{2-6}
        Machine Learning techniques & 
        VAE & 
        Pair of neural networks composed of an encoder and a decoder. Transforms data in a latent space to reduce its dimensionality. Encoding-decoding scheme has to be learned. &
        \cite{garrido_prediction_2019};
        \cite{xu_modeling_2019}
        & 
        \tabitem Aims at learning a latent representation of the variables \nl
        \hspace*{0pt}\tabitem Latent space is suitable for inference and completion of data
        & 
        \tabitem Might not be able to learn the true posterior distribution \nl
        \hspace*{0pt}\tabitem Usually outperformed by GANs
        \\ \cline{2-6}
        & 
        GAN & 
        Pair of neural networks composed of a generator and a discriminator. The generator is trained to fool the discriminator. Learning process is a two players minimax game. &
        \cite{goodfellow_generative_2014};
        \cite{xu_synthesizing_2018};
        \cite{xu_modeling_2019};
        \cite{zhao_ctab-gan_2021}
        & 
        \tabitem Generator never sees true data (privacy ensured) \nl
        \hspace*{0pt}\tabitem Architectures of both neural networks are flexible \nl
        \hspace*{0pt}\tabitem Current state-of-the-art generative model
        & 
        \tabitem Equilibrium between both neural networks difficult to achieve \nl
        \tabitem Dependencies of variables cannot be controlled
    \end{tabularx}
\end{table}
\vfill
\end{landscape}
}

\subsection{Existing approaches for synthetic tabular data generation}
\label{sec:existing_approaches}

One of the primary uses of synthetic tabular data has been for the creation of synthetic populations, in particular for transportation research. As a result, many research contributions focus specifically on this topic, using different techniques. Therefore, we focus this review first on synthetic population generation and then introduce more general-purpose Machine Learning algorithms that have been proposed in the literature, which could also be applied for this purpose. 

We can identify two main categories of techniques for synthetic tabular data generation: Statistical techniques such as resampling methods and simulation-based methods (see Section~\ref{sec:stat_techniques}), and Machine Learning methods (see Section~\ref{sec:deep_learning}). In the following sections, we discuss these different methods in detail. Table~\ref{tab:existing_techniques} gives a summary of the main methods in each category with a list of references, advantages, and disadvantages.

\subsubsection{Statistical techniques}
\label{sec:stat_techniques}

There are two main methods to generate synthetic populations within the transportation community: resampling and simulation-based approaches. The first one is based on Iterative Proportional Fitting methods (IPF)~\citep{deming_least_1940}. It consists of proportionally adjusting a matrix to produce a new table such that the specified marginals are individually conserved. \cite{beckman_creating_1996}, first, use this methodology to create a synthetic population based on the SF3 (San Francisco area) census data. \cite{auld_population_2009} and \cite{barthelemy_synthetic_2013} both propose to improve the IPF methodology using a multi-step procedures. While IPF methods are simple to implement, this technique has multiple significant limitations in generating highly detailed and realistic synthetic tabular data. Firstly, there is no interaction between the variables with the basic algorithm. It is possible to add these interactions by adding more dimensions to the table. However, for each level of interaction, one more dimension has to be added to the table. It, thus, quickly becomes a computationally expensive algorithm. In addition, IPF cannot differentiate between structural and sampling zeros. Multiple methods have been suggested to avoid sampling zero issues in the literature, such as \cite{auld_population_2009}. Finally, IPF cannot differentiate between the different data types (categorical and continuous). Thus, researchers have developed new techniques to generate synthetic populations, such as Markov Chain Monte Carlo (MCMC) simulation.

\cite{farooq_simulation_2013} proposes to use a Markov Chain Monte Carlo (MCMC) simulation using Gibbs sampling to generate synthetic populations. The idea is to draw from an (unknown) multi-dimensional random variable characterizing the distribution of individuals in the population using a Gibbs sampler. The marginal distributions used by the Gibbs sampler are generated from real data. Since the full-conditionals are rarely available for all the attributes in the original data, the authors use a parametric model to construct the missing conditional distributions. The authors show that this simulation technique outperforms IPF methods using multiple statistical metrics such as $R^2$ and Standardized Root Mean Squared Error (SRMSE)~\citep{muller_population_2010}. Multiple improvements have been made on the original method~\citep{casati_synthetic_2015, kim_simulated_2016, philips_fine_2017}. However, while simulation-based techniques outperform IPF techniques, these methods still have limitations in the context of synthetic population generation. The main issue is that the models are working with conditionals only. This can be an advantage if only this information is available. However, since MCMC methods must assume the type of probability distributions that the variables follow, wrong assumptions can lead to fundamentally incorrect distributions. 

While these statistical methods have been widely used in the transportation community, they are outdated compared to Machine Learning techniques. For example, \cite{borysov_how_2019} show that their Machine Learning-based approaches outperform MCMC-based approaches on multiple criteria. Therefore, in this article, we concentrate on testing our methodology against state-of-the-art Machine Learning methods.

\subsubsection{Machine Learning techniques}
\label{sec:deep_learning}

Recent advances in Machine Learning and data generation techniques have enabled new approaches for generating synthetic populations. We identify three primary Machine Learning-based approaches that have been used for this purpose: Bayesian networks, Variational AutoEncoder (VAE), and Generative Adversarial Networks (GANs).

\cite{sun_bayesian_2015} use Bayesian networks to generate such populations. Bayesian networks are graphical models used to encode probability distributions for a set of variables. They use a Directed Acyclic Graph (DAG) to represent the dependencies between the variables and a set of local probability distributions for each variable in the original table and given conditional probabilities. The authors show that their model outperforms both IPF and Gibbs sampling. \cite{zhang_connected_2019} extended this concept further by using a three-step procedure to generate a population and its social network. They use a Bayesian network to create a synthetic population of households, an integer problem with Langrangian relaxation for the assignment problem, and an Exponential Random Graph Model (ERGM) for the social network simulation. 

Variational AutoEncoders (VAEs)~\citep{kingma_auto-encoding_2014} aim to reduce the dimensionality of the data into an encoded vector in the latent space. Data can then be generated more easily in this latent space since it is smaller in dimensionality. For example, \cite{borysov_how_2019} have used a VAE to generate a synthetic population. They demonstrated that their VAE model outperforms both IPF and Gibbs sampling for generating complex data. However, this type of method has quickly been outperformed by the current state-of-the-art method for generating synthetic data: Generative Adversarial Networks (GANs).

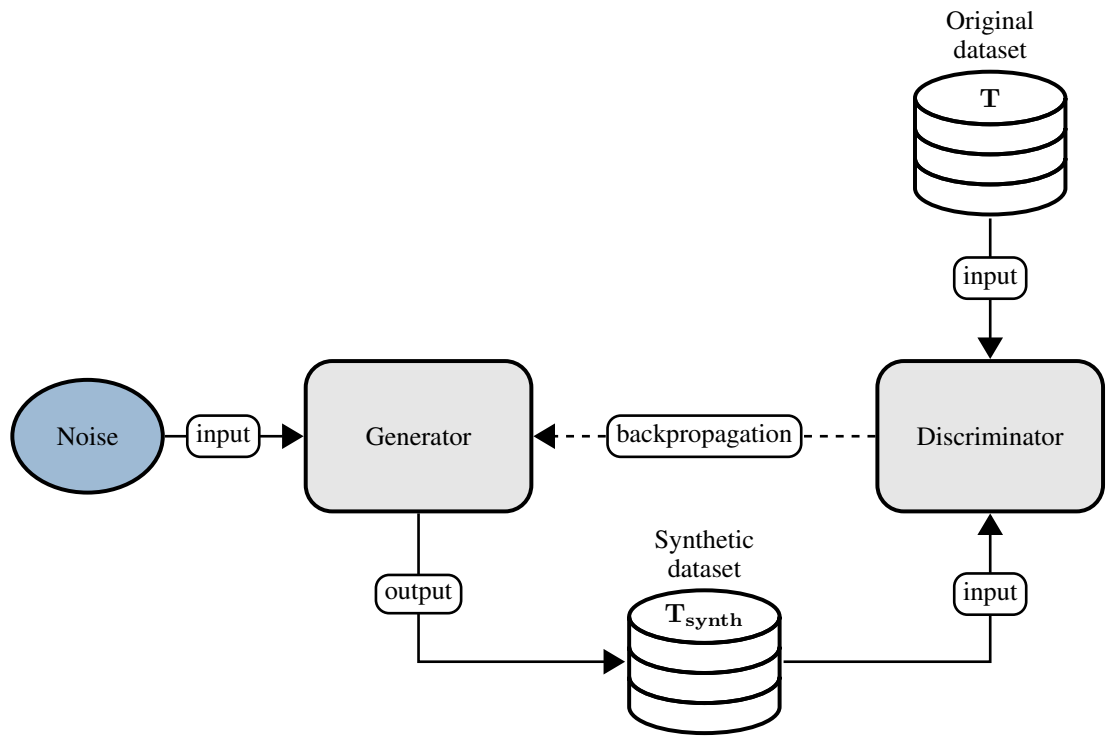
\begin{figure}[h]
    \centering
    \input{tikz/GAN}
    \caption{Schematic representation of the standard GAN structure.}
    \label{fig:GAN}
\end{figure}

Generative Adversarial Networks (GANs)~\citep{goodfellow_generative_2014} make use of two neural networks, the \emph{generator} and the \emph{discriminator}, which compete against each other on independent, unsupervised tasks. The generator processes random noise to produce synthetic data. The discriminator (or critic) then evaluates the synthetic data against original data to provide a classification or continuous score on each data point on whether the data is original or synthetic. The generator is then trained through backpropagation. Figure~\ref{fig:GAN} shows the schematic representation of a GAN. 

GANs have quickly evolved to become more specialized. For example, \cite{arjovsky_wasserstein_2017} demonstrate that the use of a discrete loss function results in issues such as vanishing gradients. They thus propose an alternative continuous loss function based on the Wasserstein distance. This GAN is therefore named Wasserstein GAN (WGAN). Further key developments in GAN research include the introduction of a penalty on the gradient during model training~\citep{gulrajani_improved_2017} or the addition of conditionality~\citep{mirza_conditional_2014}. While the primary application of GANs has been the generation of image data, with a particular focus on human faces~\citep{alqahtani_applications_2021}, researchers have also developed specific architectures for tabular data. It, thus, allowed researchers to switch their focus to more general synthetic tabular data generation rather than synthetic population generation.

TableGAN~\citep{park_data_2018} and TGAN~\citep{xu_synthesizing_2018} are two specific GANs models for tabular data. TableGAN has been developed with privacy-preservation techniques in mind. This model is based on Deep Convolutional GAN (DCGAN)~\citep{radford_unsupervised_2016}. On the other hand, TGAN has been developed to reproduce tabular data as realistically as possible using Long Short Term Memory (LSTM) cells for the generator~\citep{hochreiter_long_1997}. The authors demonstrated that TGAN outperforms tableGAN. Researchers have also developed their own GAN structures to generate synthetic populations in the transportation community. For example, \cite{garrido_prediction_2019} develop their own GAN structure based on the WGAN to use tabular data. They show that this new model was statistically better than IPF techniques, Gibbs sampling, and the VAE of \cite{borysov_how_2019}. Finally, \cite{badu-marfo_composite_2020} created a new GAN named Composite Travel GAN (CTGAN). Their GAN is based on the Coupled GAN (CoGAN)~\citep{liu_coupled_2016} and is used to generate the table of attributes for the population and the sequence of Origin-Destination segments. They show that the CTGAN outperforms the VAE statistically. While these models are showing outstanding performances compared to previous methods, the switch to data-driven methods has hindered the control of the researchers or modelers on the generation process. The lack of control during this process can hinder the final results depending on the research axis. Thus, in the next section, we discuss multiple axes found in the literature, some requiring high control on the generated synthetic data.

\subsection{Research axes}
\label{sec:research_axes}

The primary focus of existing population synthesis in the transportation domain has been for direct use in simulation models. On the other hand, the deep learning community motivates their research by stating that using more data improves the efficacy of Machine Learning models. For example, \cite{jha_impact_2019} show that a larger and more complete dataset leads to better validation and fewer uncertainties. Other examples discussing the dataset size can be found in the literature~\citep{barbedo_impact_2018,linjordet_impact_2019}. However, this does not represent the only research axis for synthetic tabular data generation. In the remainder of this section, we present and discuss five different research axes for synthetic generation in the literature.

\subsubsection{Simulation and agent-based modeling}
\label{sec:simulation}

Agent-based models~\citep{bonabeau_agent-based_2002} are used to simulate the actions and interactions of autonomous agents in order to understand the behavior of a system. These models are extensively used in the transportation community~\citep{kagho_agent-based_2020} and require a large amount of data to be adequately trained. Synthetic data are, thus, often used to replace scarce and expensive original data. 

In the transportation community, the predominant focus for the population synthesis papers introduced in Section~\ref{sec:existing_approaches} is for direct use in simulation. For example, \cite{beckman_creating_1996}, \cite{barthelemy_synthetic_2013}, \cite{farooq_simulation_2013}, \cite{borysov_how_2019}, and \cite{garrido_prediction_2019} all motivate their work by discussing the generation of synthetic population for (agent-based) simulation models. 

\subsubsection{Privacy preservation}
\label{sec:privacy}

Privacy preservation techniques ensure that any private information is not disclosed using data or Machine Learning models. Synthetic data can, thus, replace highly sensitive original data when privacy is of concern. Methods using only the conditionals, such as IPF or Gibbs sampling, are especially effective since the methods never use the original data to generate the synthetic data. 

For example, \cite{barthelemy_synthetic_2013} motivate their model (a three-step procedure based on simulation techniques) to improve the privacy preservation of the standard IPF methods. They state that the standard method tends to repeat observations, and thus it is possible to retrieve information from the original dataset. More recently, the tableGAN~\citep{park_data_2018} has been specifically designed to preserve the original datasets' privacy. Multiple GAN models have been created in computer vision, with privacy preservation as the core motivation. For example, \cite{liu_ppgan_2019} created the Privacy-Preserving GAN (PPGAN). This GAN uses differential privacy by adding specifically designed noise to the gradient during the learning procedure. \cite{yin_gans_2018}, on the other hand, directly generated protected data within the generator of their GAN by removing some sensible information and encoding them in the generated data. They tested their synthetic data against attack models to show that their GAN could generate more complex data to be deciphered. More recently, \cite{zhao_ctab-gan_2021} motivate and test their model on privacy preservation. 

\subsubsection{Machine Learning efficacy}
\label{sec:ml_efficacy}

The generation of synthetic populations also enables the augmentation of existing real-world datasets with synthetic individuals, therefore increasing the size and variability of the dataset. Several studies have investigated the estimation of Machine Learning techniques on augmented or fully synthetic data. This concept is already widely used on images~\citep{shorten_survey_2019}. While simple techniques such as rotating or scaling images can be used in Computer Vision, applying such simple tricks on tabular data is impossible. Thus, researchers have been developing models aiming at augmenting tabular data. 

For example, \cite{xu_synthesizing_2018} motivate the development of the TGAN because organizations are using Machine Learning on relational tabular data to augment process workflows carried out by humans. Furthermore, they state that these synthetic datasets can either be used as an augmentation for the existing datasets or as a means to preserve privacy, which is discussed in Section~\ref{sec:privacy}. On the other hand, \cite{xu_modeling_2019} do not give a clear motivation on the usage of synthetic datasets. However, they test their models on Machine Learning efficacy by replacing the training data with the generated synthetic data. More recently, both \cite{wen_causal-tgan_2021} and \cite{zhao_ctab-gan_2021} motivate their models using Machine Learning efficacy as the core method to assess their generated synthetic datasets.  

\subsubsection{Bias correction}
\label{sec:bias_correction}

Synthetic data can also be applied to correct bias in existing datasets by controlling the data generation process. It builds on standard resampling methods~\citep{rubin_use_1973} to rebalance the dataset, which reduces the signal-to-noise ratio of the existing data (by removing oversampled data or resampling undersampled data). Indeed, data generation techniques can also be used to augment an existing dataset and rebalance it. 

For example, Conditional GANs~\citep{mirza_conditional_2014} have been created to tackle such imbalance issues. The idea of such GANs is to generate synthetic data using prior information. They, thus, increase the probability of generating synthetic data with the given information. \cite{xu_modeling_2019} have adapted this methodology to tabular data with the Conditional Table GAN (CTGAN). They show that the conditionality is truly efficient for Machine Learning models when the data is highly imbalanced. They create synthetic datasets addressing the imbalance and trained Machine Learning models on this synthetic and the original dataset. The models trained on the synthetic datasets perform better than those trained on the original dataset. Previously, \cite{farooq_simulation_2013} motivate their research on population synthesis with Gibbs sampling using the fact that it can complete datasets. However, the authors have not formally evaluated this use of synthetic data.

\subsubsection{Transfer learning}
\label{sec:transfer_learning}

The use of Conditional GANs, and other conditional data generation approaches, enables the possibility for transfer learning, where knowledge from one context with large data availability can be transferred to another context with lower data availability. 

For example, \cite{noguchi_image_2019} propose a new method using the BigGAN to transfer the knowledge learned on large datasets and apply this knowledge to a dataset with only 25 images. They show that they can add a new class to a pre-trained generator without disturbing the performance of the original domain. \cite{wang_minegan_2020} propose to use a miner network that identifies which distribution of multiple pre-trained GANs is the most beneficial for a specific target. This mining pushed the sampling towards more suitable regions in the latent space. Therefore, the MineGAN can transfer the knowledge of multiple GANs such as the BigGAN and the Progressive GAN to a domain with fewer images. Other relevant methods for transfer knowledge can be found in the articles of \cite{jeon_t-gd_2020} and \cite{fregier_mind2mind_2020}. 

While this research axis has already been explored in the computer vision community, it has not been explored in population synthesis. Indeed, while the transfer of knowledge between two tabular datasets might not make sense, it could be used for tabular data of populations. For example, census data are often collected regularly, \emph{e.g.} every two of five years. We could, thus, imagine using GANs trained on data from previous years to transfer their knowledge to the most recent years.

\subsection{State-of-the-art models}
\label{sec:sota_models}

We present a detailed overview of four different approaches demonstrated to achieve the best performance when generating synthetic tabular data. As such, these approaches represent the state-of-the-art in this field. The four approaches, which all make use of deep learning algorithms, are introduced across three key articles: \cite{xu_synthesizing_2018}, \cite{xu_modeling_2019}, and \cite{zhao_ctab-gan_2021}. A summary of the models is given in Table~\ref{tab:summary_sota}.

\begin{table}[H]
    \centering
    \caption{Summary of the state-of-the-art models selected for comparison with the DATGAN.}
    \label{tab:summary_sota} 
    \begin{tabularx}{0.9\textwidth}{c||>{\hsize=.35\hsize}C|>{\hsize=1.65\hsize}L}
        \multicolumn{1}{c||}{\textbf{Model}} & \multicolumn{1}{c|}{\textbf{Article}} & \multicolumn{1}{c}{\textbf{Information}} \\ \midrule[1.5pt] 
        TGAN & \cite{xu_synthesizing_2018} & \tabitem \emph{Generator:} LSTM cells in linear arrangement \nl
        \hspace*{0pt}\tabitem \emph{Discriminator:} Fully-connected neural network \nl
        \hspace*{0pt}\tabitem \emph{Data preprocessing:} Continuous vs categorical \nl
        \hspace*{0pt}\tabitem \emph{Loss function:} Cross-entropy loss \nl
        \hspace*{0pt}\tabitem \emph{Conditionality:} None \\ \midrule
        
        CTGAN & \cite{xu_modeling_2019} & \tabitem \emph{Generator:} Fully-connected neural network \nl
        \hspace*{0pt}\tabitem \emph{Discriminator:} Fully connected neural network \nl
        \hspace*{0pt}\tabitem \emph{Data preprocessing:} Continuous vs categorical \nl
        \hspace*{0pt}\tabitem \emph{Loss function:} Wasserstein loss with gradient-penalty \nl
        \hspace*{0pt}\tabitem \emph{Conditionality:} On categorical variables \\ \midrule
        
        TVAE & \cite{xu_modeling_2019} & \tabitem \emph{Encoder:} Updated structure for preprocessed data \nl
        \hspace*{0pt}\tabitem \emph{Decoder:} Similar to conventional VAE \nl
        \hspace*{0pt}\tabitem \emph{Data preprocessing:} Continuous vs categorical \nl
        \hspace*{0pt}\tabitem \emph{Loss function:} Evidence lower-bound (ELBO) loss \nl
        \hspace*{0pt}\tabitem \emph{Conditionality:} None \\ \midrule
        
        CTAB-GAN & \cite{zhao_ctab-gan_2021} & \tabitem \emph{Generator:} Convolutional neural network \nl
        \hspace*{0pt}\tabitem \emph{Discriminator:} Convolutional neural network \nl
        \hspace*{0pt}\tabitem \emph{Classifier:} Multi-layer perceptron \nl
        \hspace*{0pt}\tabitem \emph{Data preprocessing:} Continuous vs categorical vs mixed \nl
        \hspace*{0pt}\tabitem \emph{Loss function:} Cross-entropy, information and classification losses \nl
        \hspace*{0pt}\tabitem \emph{Conditionality:} On categorical variables 
    \end{tabularx}
\end{table}

While some GANs have been developed for privacy preservation, TGAN~\citep{xu_synthesizing_2018} focuses on learning the marginal distributions using recurrent neural networks. Since our focus is on creating representative synthetic data, we thus selected TGAN as the first model to be compared. It uses Long Short-Term Memory (LSTM) cells~\citep{hochreiter_long_1997} to generate each variable in the table. The LSTM cells are arranged linearly following the order of the variables in the dataset. The authors make the difference between categorical and continuous variables. Both variable types are encoded differently: \begin{inlinelist}
    \item continuous variables are encoded using Gaussian mixtures; 
    \item categorical variables are one-hot encoded. 
\end{inlinelist}
Finally, the TGAN is trained using the standard minimax loss function~\citep{goodfellow_generative_2014}, and it is compared to other data synthesizers such as Gaussian Copula and Bayesian Networks. The authors show that TGAN outperforms all these methods.  

CTGAN~\citep{xu_modeling_2019} uses a fully-connected neural network for both the generator and the critic. Like TGAN, the variables are differentiated between categorical and continuous variables. CTGAN uses the same encoding procedure for both variable types with a slight difference for continuous variables: a Variational Gaussian Mixture (VGM) is used instead of standard Gaussian mixtures. In addition, this model uses conditionality by adding a conditional vector on categorical variables. Finally, CTGAN is trained using the Wasserstein loss with gradient-penalty~\citep{gulrajani_improved_2017}. 

The Tabular VAE (TVAE)~\citep{xu_modeling_2019} is an adaptation of a standard Variational AutoEncoder (VAE) by modifying the loss function and preprocessing the data. The variables are encoded using the same procedure as in CTGAN. TVAE uses the evidence lower-bounder (ELBO) loss~\citep{kingma_auto-encoding_2014}. While the encoder is slightly updated compared to conventional VAEs, the decoder keeps a usual structure. The authors have compared TVAE, CTGAN, and other methods for synthesizing tabular data such as tableGAN. They show that both TVAE and CTGAN outperform other methods. On multiple metrics, TVAE performs better than CTGAN. However, as stated by the authors, CTGAN achieves differential privacy~\citep{jordon_pate-gan_2018} easier than TVAE since the generator never sees the original data.

Finally, the particularity of CTAB-GAN~\citep{zhao_ctab-gan_2021} compared to the previous models is that it aims at fixing issues with skewed continuous distributions. Indeed, continuous distributions can take many forms, such as long-tailed, exponential, or mixed distributions. Therefore, this model implements multiple data preprocessing methods for different distributions. CTAB-GAN comprises three neural networks: a generator, a discriminator, and an additional classifier. The latter is used to learn the semantic integrity of the original data and predict the synthetic data classes. This helps produce more accurate labeled synthetic data. The generator and discriminator are convolutional neural networks, while the classifier is a multi-layer perceptron. In addition, CTAB-GAN uses conditionality to counter the imbalance in the training dataset to improve the learning process. Finally, CTAB-GAN is trained using the standard cross-entropy loss function with the addition of an information loss and a classification loss. The authors have tested their model against other state-of-the-art models such as tableGAN and CTGAN. They have shown that their model outperforms all the other models using Machine Learning efficacy and statistical similarity metrics.

\subsection{Model evaluation}
\label{sec:model_eval}

Model evaluation is intrinsically linked to the research axis for which a model was developed. Indeed, a generator developed to correct bias should not be tested on the same characteristics as a model developed for privacy preservation. Thus, researchers have come up with different methods for assessing generated synthetic datasets. In this section, we discuss two types of methods that are primarily used to assess the representativity of a synthetic dataset compared to its original counterpart: statistical methods (in Section~\ref{sec:stat_assess}) and Machine Learning methods (in Section~\ref{sec:ml_techniques}).

\subsubsection{Statistical assessments}
\label{sec:stat_assess}

Multiple statistical tests can be used to compare two distributions, such as the $\Xi^2$ test or the Student's t-test. While these tests can provide good information when comparing the distributions of each variable separately, it does not consider the correlation between the variables. Since this aspect is essential for creating representative synthetic populations, researchers in the transportation community have been developing new statistical tests to address this issue. The Standardized Root Mean Squared Error (SRMSE)~\citep{muller_population_2010} is used in most transportation articles working on population synthesis to assess the generated datasets. The test consists in selecting one or multiple variables in a dataset and creating a frequency list based on the appearance of each unique value. We can, then, apply the SRMSE (see Equation~\ref{eq:SRMSE}) formulation on this frequency list. While this technique has been shown to work well compared to other statistical methods, it has two main flaws:
\begin{inlinelist}
\item the frequency lists are computed by counting the unique values (or combinations of unique values). Therefore, it is preferably used on discrete values.
\item The choice of variables (or combination of variables) is up to the researcher or modeler. Thus, articles using this methodology tend only to test a couple of combinations.
\end{inlinelist}
In order to address the first flaws, we can transform the continuous values into discrete values by assigning them to specific buckets. This is a relatively simple fix, but if the discretization is done correctly, SRMSE should still provide valid results. However, the second flaw is more problematic. Indeed, when generating a synthetic dataset, we want to make sure that all the correlations are correctly generated. Therefore, it is required to update the methodology of the SRMSE to do systematic testing on all the variables and their possible combinations.

\subsubsection{Machine Learning techniques}
\label{sec:ml_techniques}

In Machine Learning, many datasets are considered classification datasets. Thus, they are used with a Machine Learning model to predict future instances of a unique variable. Therefore, researchers developing generators to improve the efficacy of such models usually only test the synthetic datasets using the predictive power of Machine Learning models on a single variable. While this technique works well in this specific case, it does not provide enough information if one is trying to assess the representativity of a synthetic dataset compared to an original one. Indeed, there might be missing correlations between the other variables in the dataset that the Machine Learning models will overlook. It is, thus, possible to update this technique such that a Machine Learning model is used to predict each variable in the dataset instead of a single one. If there are issues with correlations between the variables, the efficacy of the Machine Learning models will drop while predicting the other variables, thus providing more information.

\subsection{Opportunities and limitations}
\label{sec:opportunities}

The role of the generator in a GAN is to produce batches of synthetic data, taking only noise as an input. As such, the structure of the generator network should be closely matched to the underlying structure of the data being replicated. For instance, in images, each variable represents a pixel whose meaning is image-specific and only defined relative to other pixels in the image. In other words, the meaning of a pixel in an image is dependent on its relative position and value, not its absolute position and value, and the meaning of a single pixel in one image is (largely) independent from the meaning of the corresponding pixel in another image in the same dataset. It is therefore typical for generators used in image generation to make use of convolutional neural networks~\citep{radford_unsupervised_2016, isola_image--image_2017, zhu_unpaired_2018}, which model the relative definitions of the pixel values learned over thousands or millions of images in a dataset.

Unlike images, the variables in tabular data typically have a specific meaning and can be understood by their absolute positions and values (within a single dataset). For example, a column representing an individual's age in socio-economic data defines the age of every instance (row) in the table. Furthermore, the age value of each row can be understood without needing to know the values of the other variables. While the variables in a dataset have a fixed position-specific meaning, their values are dependent on the other variables in the dataset. As such, the generator must capture the interdependencies between these variables. 

There are several different approaches to this in the literature. The first approach mimics what is done with images. Indeed, several models are built using fully connected networks~\citep{xu_modeling_2019} or convolutional neural networks~\citep{park_data_2018, zhao_ctab-gan_2021}. The generator has to learn the structure from the data during the learning process using backpropagation. This can be rather cumbersome for the generator since it never has access to the original data. Therefore, another approach is to fix the structure of the data. For example, the TGAN~\citep{xu_synthesizing_2018} uses a sequential order based on the columns' order. The structure is implicitly learned using attention vectors used with each variable in the dataset.

In both approaches, the generator learns the relationships between the variables from the available data via backpropagation of the discriminator loss. However, there are two primary limitations of this approach. Firstly, the generator, which needs to be highly flexible, can overfit the noise in the data and generalize to relationships between the columns which do not actually exist in unseen data. Secondly, the generator has to use the limited signal in the data to learn the core structure of the data, which is often already known to some degree by the modeler. Both can cause issues when the signal-to-noise ratio is high, as is often the case with socio-economic datasets of limited size.

Therefore, there is an opportunity to develop techniques that address these flaws and use the learning power of GANs. Indeed, by defining the relationships between the variables beforehand using expert knowledge, we can force the generator only to learn specific correlations. In addition, if the researcher or modeler provides these relationships, the model starts its learning process with more information than a fully connected network that has to learn all these connections. Therefore, we can overcome the issues with GANs while keeping their strengths.

%% file: tikz/GAN.tex
\begin{tikzpicture}[every text node part/.style={align=center}]

\makeatletter
\tikzset{
    database top segment style/.style={draw},
    database middle segment style/.style={draw},
    database bottom segment style/.style={draw},
    database/.style={
        path picture={
            \path [database bottom segment style]
                (-\db@r,-0.5*\db@sh) 
                -- ++(0,-1*\db@sh) 
                arc [start angle=180, end angle=360,
                    x radius=\db@r, y radius=\db@ar*\db@r]
                -- ++(0,1*\db@sh)
                arc [start angle=360, end angle=180,
                    x radius=\db@r, y radius=\db@ar*\db@r];
            \path [database middle segment style]
                (-\db@r,0.5*\db@sh) 
                -- ++(0,-1*\db@sh) 
                arc [start angle=180, end angle=360,
                    x radius=\db@r, y radius=\db@ar*\db@r]
                -- ++(0,1*\db@sh)
                arc [start angle=360, end angle=180,
                    x radius=\db@r, y radius=\db@ar*\db@r];
            \path [database top segment style]
                (-\db@r,1.5*\db@sh) 
                -- ++(0,-1*\db@sh) 
                arc [start angle=180, end angle=360,
                    x radius=\db@r, y radius=\db@ar*\db@r]
                -- ++(0,1*\db@sh)
                arc [start angle=360, end angle=180,
                    x radius=\db@r, y radius=\db@ar*\db@r];
            \path [database top segment style]
                (0, 1.5*\db@sh) circle [x radius=\db@r, y radius=\db@ar*\db@r];
        },
        minimum width=2*\db@r + \pgflinewidth,
        minimum height=3*\db@sh + 2*\db@ar*\db@r + \pgflinewidth,
    },
    database segment height/.store in=\db@sh,
    database radius/.store in=\db@r,
    database aspect ratio/.store in=\db@ar,
    database segment height=0.1cm,
    database radius=0.25cm,
    database aspect ratio=0.35,
    database top segment/.style={
        database top segment style/.append style={#1}},
    database middle segment/.style={
        database middle segment style/.append style={#1}},
    database bottom segment/.style={
        database bottom segment style/.append style={#1}}
}
\makeatother

\makeatletter
\tikzset{ loop/.style={ % requires library shapes.misc
        draw,
        chamfered rectangle,
        chamfered rectangle xsep=2cm
    }
}
\makeatother

\definecolor{lightgray}{rgb}{.9,.9,.9}
\definecolor{lightblue}{rgb}{.62,.73,.83}
\definecolor{lightred}{rgb}{.83, .73, .62}
\definecolor{lightorange}{rgb}{.99,.82,.60}

% Helpers
\def\n{20}
%\draw[help lines,xstep=1,ystep=1, color=black!20] (-\n,-\n) grid (\n,\n);
%\draw[help lines,xstep=.5,ystep=.5, color=black!10, dotted] (-\n,-\n) grid (\n,\n);
%\foreach \x in {-\n,...,\n} { \node [anchor=north, color=black!30] at (\x,0) {\x}; }
%\foreach \y in {-\n,...,\n} { \node [anchor=east, color=black!30] at (0,\y) {\y}; }

\def\dy{3}
\def\dx{4}

\node[ellipse, fill=lightblue, line width=1.5, draw=black, minimum width=2cm, minimum height=1.5cm] (B) at (0,0) {Noise};

\node[rounded corners=10pt, fill=lightgray, line width=1.5, draw=black, minimum width=3cm, minimum height=2cm] (C) at (1.1*\dx,0) {Generator};

\node[rounded corners=10pt, fill=lightgray, line width=1.5, draw=black, minimum width=3cm, minimum height=2cm] (D) at (3*\dx,0) {Discriminator};

\node[database,label=above:Synthetic\\dataset,database radius=1cm,database segment height=0.4cm, line width=1.5] (I) at (2.05*\dx,-\dy)  {};
\node at (2.05*\dx,-\dy+0.6) {$\bf T_{synth}$};

\node[database,label=above:Original\\dataset,database radius=1cm,database segment height=0.4cm, line width=1.5] (J) at (3*\dx,1.3*\dy)  {};
\node at (3*\dx,1.3*\dy+0.6) {$\bf T$};

\draw[-{Triangle[scale=1.5]}, line width=1] (C) |- (I);
\draw[-{Triangle[scale=1.5]}, line width=1] (I) -| (D);

\node[fill=white, draw=black, rounded corners=5pt, line width=1] at (3*\dx,-0.7*\dy) {input};
\node[fill=white, draw=black, rounded corners=5pt, line width=1] at (1.1*\dx,-0.7*\dy) {output};

\draw[-{Triangle[scale=1.5]}, line width=1] (J) -- node[fill=white, above=-0.15cm, draw=black, rounded corners=5pt] {input} (D);

\draw[-{Triangle[scale=1.5]}, line width=1] (B)  -- node[left=-0.35cm, fill=white, draw=black, rounded corners=5pt] {input}(C);

\draw[-{Triangle[scale=1.5]}, line width=1, dashed] (D) -- node[fill=white, draw=black, rounded corners=5pt, solid] {backpropagation}  (C);

\end{tikzpicture}

%% file: text/03_methodology.tex
\section{Methodology}
\label{sec:methodology}

Following several previous works in the literature, our approach for synthetic tabular data generation makes use of a GAN~\citep{goodfellow_generative_2014} to generate synthetic data. Our primary contribution is to closely match the generator structure to the underlying structure of the data through a Directed Acyclic Graph (DAG) specified by the modeler. According to their prior expert knowledge of the data structure, the DAG allows the modeler to define the structure of the correlations between variables in the dataset as a series of directed links between nodes in a graph. Each link in the DAG represents a causal link that the generator can capture, and if no links (either direct or indirect) exist between two variables, then the generator treats those variables independently. This has two primary advantages over the existing approaches within the context of the limitations identified in the literature review. Firstly, by restricting the set of permissible links between the variables in the datasets, the DAG represents an expert regularisation of the model and restricts the ability of the GAN to overfit noise in the training sample. Secondly, giving the generator a headstart in knowing the underlying structure of the data allows the GAN to make more efficient use of the training sample when learning to generate data. These benefits could enable the DATGAN to use limited available original data more efficiently when learning to produce realistic and representative synthetic data samples.

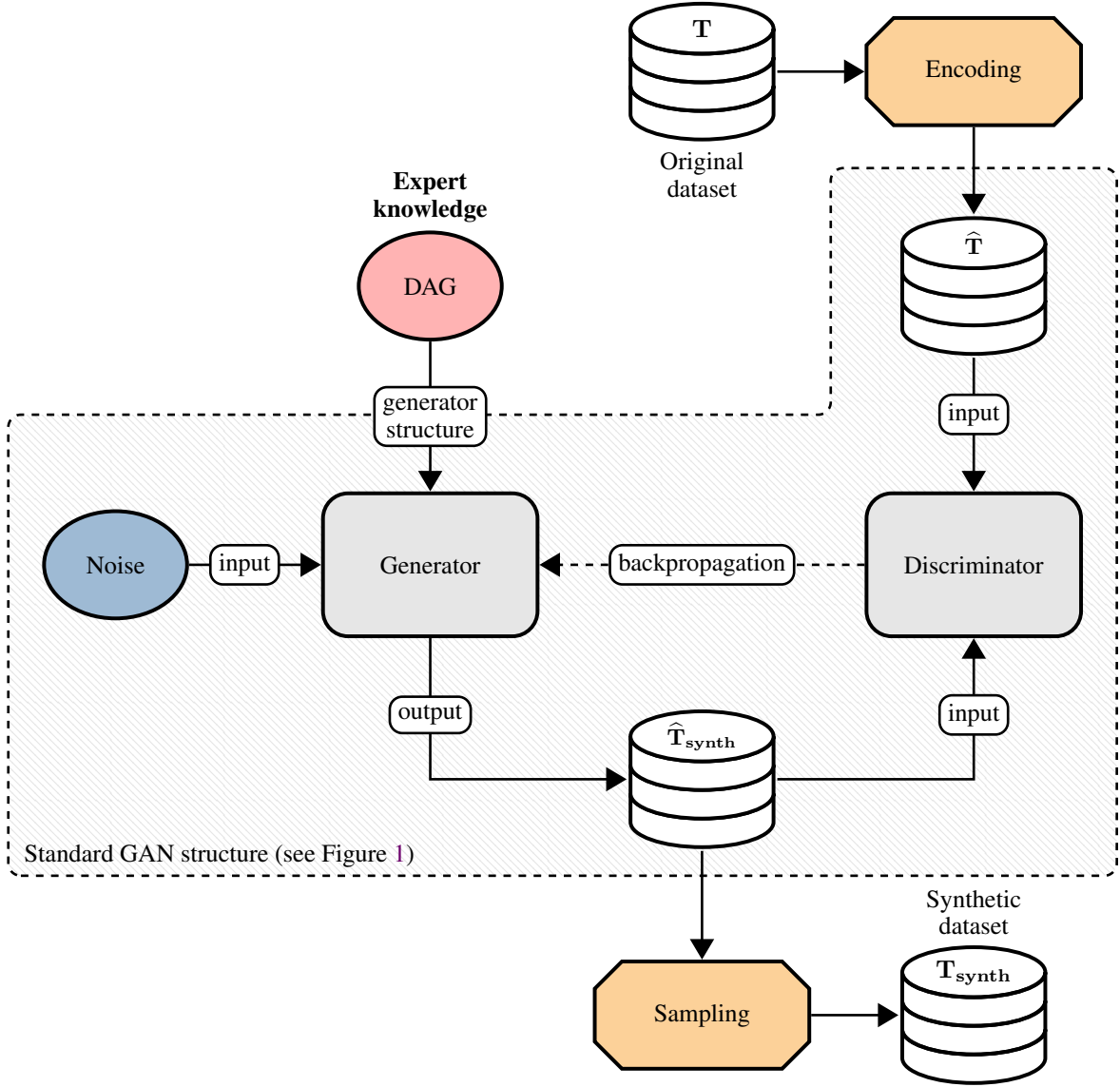
\begin{figure}[h!]
    \centering
    \input{tikz/DATGAN}
    \caption{Global schematic representation of the DATGAN. The different element in this figure are presented in the following sections: the Generator is presented in Section~\ref{sec:generator}, the DAG in Section~\ref{sec:DAG}, the Discriminator in Section~\ref{sec:discriminator}, the Encoding in Section~\ref{sec:encoding}, and the Sampling in Section~\ref{sec:sampling}.}
    \label{fig:DATGAN}
\end{figure}

Figure~\ref{fig:DATGAN} provides a high-level overview of the DATGAN data generation process. As is typical with GANs, the generator in the DATGAN (described in Section~\ref{sec:generator}) never sees the original data. It generates data purely from a random noise input. The generator's structure is determined according to a DAG which specifies the structural relationships between the variables in the data (\emph{i.e.} the expert knowledge), as defined by the modeler. We present the DAG in Section~\ref{sec:DAG} and the process to automatically create the generator structure from the DAG in Section~\ref{sec:LSTM}. At the same time, the discriminator (described in Section~\ref{sec:discriminator}) is trained to classify/critique the original and synthetic data. Therefore, it can be considered a competitive game between two adversaries (the generator and the discriminator). The loss functions used to optimize both models are presented in Section~\ref{sec:loss_function}. Since tabular data can contain attributes of different types (\emph{e.g.} continuous, nominal, and ordinal), original and synthetic data have to be processed before being used. We, thus, introduce several new data processing steps specific to the DATAGAN model in Section~\ref{sec:data_processing}. At the end of this section, we present the result assessments methods used to compare the synthetic datasets in Section~\ref{sec:result_assessments}. We conclude the methodology by providing some implementation notes in Section~\ref{sec:implementation}. The supplementary materials include a notation table summarizing all the notations used in the methodology and a summary of the possible DATGAN versions using the different components presented here.

Formally, we consider a table $\bf T$ containing $N_{V}$ columns. Each column in the table $\bf T$ is represented by $\vec{v}_t$ for $t=1,\ldots,N_{V}$. We, thus, have ${\bf T} = \left\{\vec{v}_{1:N_{V}}\right\}$. These columns have been drawn from an unknown joint distribution of random variables $V_t$, \emph{i.e.} the values in $\bf T$ are drawn from $\mathbb{P}\left(V_{1:N_{V}}\right)$. We, thus, usally refer to the variables in $\bf T$ using $V_t$. We represent the rows of $\bf T$ by $\left\{v_{1:N_{V},i}\right\}$ for $i=1,\ldots,N_{\text{rows}}$ where $N_{\text{rows}}$ corresponds to the number of rows in $\bf T$. We assume that each row of $\bf T$ is sampled independently, \emph{i.e.} it is cross-sectional and does not contain panel or sequential data. Our goal is to learn a generative model $G(\vec{z})$, where $\vec{z}$ is a tensor of random noise, such that the samples generated from $G$ create a synthetic table $\bf T_{synth}$. For neural networks, we work in standardized space consisting of values between -1 and 1. We, thus, denote processed datasets with the character $\widehat{\cdot}$, as shown in Figure~\ref{fig:DATGAN}. In the meantime, the discriminator $D$ is trained to differentiate between original data $\bf \widehat{T}$ and synthetic data $\bf \widehat{T}_{synth}$.

\subsection{Generator}
\label{sec:generator}

The role of the generator is to produce batches of synthetic data, taking only noise as an input. Within DATGAN, the generator structure is defined using a Directed Acyclic Graph (DAG), which specifies the interdependencies between each variable $V_t$ in the original dataset $\bf T$. We first present the DAG, including how the modeler should construct it, in Section~\ref{sec:DAG}. We then demonstrate how the DAG is used to automatically create the generator network in Section~\ref{sec:LSTM}, through the use of Long Short Term Memory (LSTM) cells~\citep{gers_learning_2000}. This includes defining a new multi-input LSTM cell required to capture complex correlations specified in the DAG. 

\subsubsection{Directed Acyclic Graph (DAG)}
\label{sec:DAG}

The DAG $\mathcal{G}$ is specified by the modeler to define the correlations between the variables in the data. However, similarly to Bayesian networks, a DAG must represent causal links between variables. Indeed, correlations do not have a direction, while causal links do. The main reason to use a directed graph instead of an undirected one is due to the nature of the representation of the variables in the Generator. As explained in Section~\ref{sec:LSTM}, each variable $\vec{v}_t$ in $\bf T$ is represented by a single LSTM cell. These cells communicate with each other in a directed manner, \emph{i.e.} the previous cell sends information to the next one. Therefore, to better reflect this behavior, a directed graph is required. 

The mathematical definition of a directed acyclic graph is given by:
\begin{itemize}
    \item The graph $\mathcal{G}$ must be directed, \emph{i.e.} each edge in the graph has only direction. 
    \item The graph $\mathcal{G}$ must not contain any cycle, \emph{i.e.} the starting vertex of any given path cannot be the same as the ending vertex.
\end{itemize}
These two properties ensure that the DAG is a topological sorting. It means that we can extract a linear ordering of the vertices such that for every directed edge $\vec{v}_{t_1}\rightarrow\vec{v}_{t_2}$ from vertex $\vec{v}_{t_1}$ to vertex $\vec{v}_{t_2}$, $\vec{v}_{t_1}$ comes before $\vec{v}_{t_2}$ in the ordering. 

With these rules in mind, the modeler can define the DAG for the DATGAN. It is possible to get inspiration from Bayesian networks and how their respective DAG is created manually~\citep{lucas_bayesian_2004}. However, there are some slight differences between the two DAGs. Therefore, we provide some insights about the components of our DAG:
\begin{itemize}
    \item Each variable $\vec{v}_t$ in the table $\bf T$ must be associated to a node in the graph $\mathcal{G}$. 
    \item A directed edge between two vertices, \emph{i.e.} $\vec{v}_{t_1}\rightarrow\vec{v}_{t_2}$, means that the generation of the first variable $\vec{v}_{t_1}$ will influence the generation of the second variable $\vec{v}_{t_2}$. Usually, the direction of the edge is a matter of judgement and should not influence the final result. 
    \item The absence of a link between two variables means that their correlation is not \emph{directly} learned by the Generator. However, it is possible to obtain some correlation in the final synthetic dataset if these two variables have a common ancestor in the graph $\mathcal{G}$. Therefore, two variables will not show any correlations in the synthetic dataset if they do not have any common ancestors or links between them in the DAG.
    \item The graph $\mathcal{G}$ can be composed of multiple DAGs as long as the first rule is respected. By creating multiple DAGs, the modeler ensures that the different parts of the dataset are not correlated. While this approach is not common, it could be used in a dataset containing variables about multiple unrelated topics. 
\end{itemize}

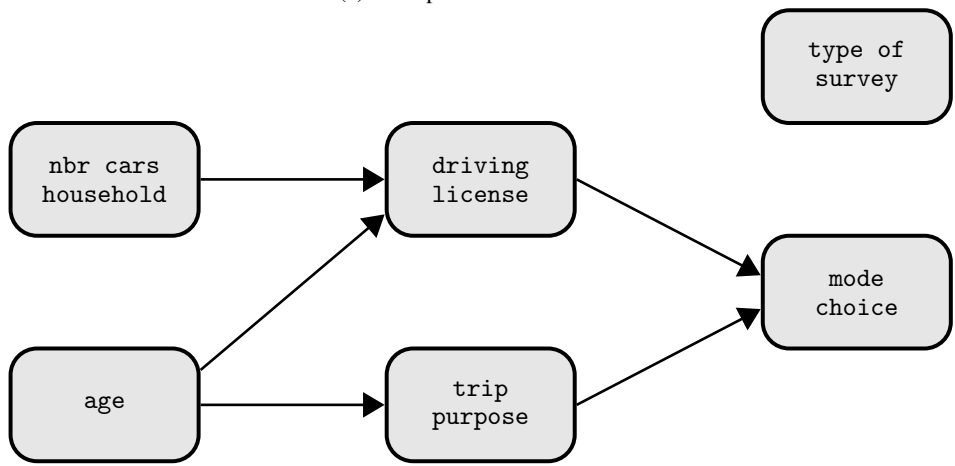
\begin{figure}[h!]
\begin{subfigure}{\textwidth}
    \centering
    \renewcommand\arraystretch{1.5}
    \begin{tabularx}{0.9\textwidth}{|*{6}{C|}}
    \hline 
    \thead{age} & \thead{driving \\license} & \thead{trip \\ purpose} & \thead{type of \\ survey} & \thead{nbr cars \\household} & \thead{mode \\choice} \\ \hline
    \bf continuous & \bf boolean & \bf nominal & \bf nominal & \bf ordinal & \bf nominal \\[0.5cm]
    \makecell{0-100} & \makecell{True \\ False} & \makecell{Work \\ Leisure \\ ...} & \makecell{Internet \\ Phone \\ ...} & \makecell{0 \\ 1 \\ 2 \\ ...} & \makecell{Driving \\ Soft Modes \\ ...} \\\hline
    \end{tabularx}
    \caption{Example of a mock dataset}
    \label{fig:data}
\end{subfigure}\\
\begin{subfigure}{\textwidth}
    \centering
    \input{tikz/DAG}
    \caption{Example of a DAG used to represent the structure of the variables}
    \label{fig:DAG}
\end{subfigure}
\caption{Example of the structure of the data. Figure~\ref{fig:data} shows the structure of a table with six variables. Figure~\ref{fig:DAG} shows one possible DAG used to represent the variables in Figure~\ref{fig:data}.}
\label{fig:mock}
\end{figure}

As for Bayesian networks, there is no unique way to create a DAG for a given dataset. Using the example shown in Figure~\ref{fig:mock}, we present different possibilities. The first one consists of following the variables' order in the datasets. This creates a simple ordered list of the variables. For example, the TGAN~\citep{xu_synthesizing_2018} uses this specific list to link each of its LSTM cells. The advantage of such a DAG is its simplicity since no prior knowledge of the data is required. However, this DAG defines causal links based on an arbitrary order. It, thus, does not make use of expert knowledge and results in poorer results. Another possibility is to create a DAG centered around predicting a given variable. For example, in the case of Table~\ref{fig:data}, one could want to predict the variable \texttt{mode choice}. Therefore, a possible structure for the DAG is to link all the other nodes in the table to a single sink node representing the variable that we want to predict. However, while this DAG would capture all possible correlations between each variable and the one that needs to be predicted, it will not capture other correlations. Therefore, a population generated using this DAG would fail basic correlation tests. 

While the two possibilities presented above allow creating a DAG without prior knowledge of the data, they will fail to deliver a synthetic dataset that correctly models the correlations between the variables in a table $\bf T$. We recommend building a DAG containing as many links as possible. It is always possible to perform a transitive reduction of $\mathcal{G}$, \emph{i.e.} removing paths such that for all vertices $\vec{v}_{t_1}$ and $\vec{v}_{t_2}$ there exists only a unique path that goes from $\vec{v}_{t_1}$ to $\vec{v}_{t_2}$, after its definition. There are no strict rules to decide whether one should add a causal link between two variables. It is a matter of judgment, and multiple trials and errors will be needed. However, we provide a set of instructions that can help the modeler define such a DAG:

\begin{itemize}
    \item If the dataset is used to predict one variable, define this variable as a sink node in $\mathcal{G}$. For example, in Figure~\ref{fig:mock}, the dataset can be used to predict the variable \texttt{mode choice}. It is, thus, the sink node of the graph. It is also possible to define multiple sink nodes.
    \item Usually, datasets contain different categories of variables. For example, a travel survey dataset might contain trips, individuals, and households variables. It is, thus, generally easier to define the causal links between variables belonging to similar semantic groups. 
    \item The next step consists in defining the source nodes. However, there are no specific rules for this. It is entirely up to the modeler.
    \item Finally, the modeler has to choose the direction of the causal links. Again, there are no rules for this. In the example of Figure~\ref{fig:DAG}, one could decide that the variables \texttt{driving license} and \texttt{age} have an inverted causal link. This would slightly change the DAG but should not fundamentally change the results. 
\end{itemize}

As shown in Figure~\ref{fig:mock}, we present a mock dataset (see Figure~\ref{fig:data}) and one possible DAG (see Figure~\ref{fig:DAG}) representing the causal links between the variables. This dataset is a travel survey dataset. We thus define the mode choice as the sink node. We can define the following category of variables: \begin{inlinelist}
    \item trip-related variables: \texttt{mode choice} and \texttt{trip purpose};
    \item individual-related variables: \texttt{age} and \texttt{driving license};
    \item household-related variables: \texttt{nbr cars household};
    \item survey-related variables: \texttt{type of survey}.
\end{inlinelist} For each of these categories, we want to make sure that the variables are linked together. Since \texttt{mode choice} is the sink node, we can create an edge from \texttt{trip purpose} to \texttt{mode choice}. For the individual-related variables, the direction of the causal link can be either direction. For the source nodes, we decide to set the variables \texttt{age} and \texttt{nbr cars household} as the source nodes. Finally, we can add some more links to complete the DAG. We decided, on purpose, to let the variable \texttt{type of survey} out of the DAG not to influence the data generation. One could argue that it could be linked to the age since older individuals are less familiar with internet technologies. However, as stated earlier, the modeler has to make choices while constructing the DAG, requiring trials and errors. 

Once the DAG $\mathcal{G}$ has been created, we can define several valuable sets. These sets are used later when representing the structure of the DAG in the generator. 

\begin{itemize}
    \item $\mathcal{A}(V_t)$: the set of ancestors of the variable $V_t$.
    \item $\mathcal{D}(V_t)$: the set of direct ancestors of the variable $V_t$.
    \item $\mathcal{S}(V_t)$: the set of sources nodes leading to the variable $V_t$.
    \item $\mathcal{E}(V_t)$: the set of in-edges of the variable $V_t$.
\end{itemize}

If we use the variable \texttt{mode choice} from Figure~\ref{fig:DAG} as an example, we can define these different sets:
\begin{itemize}
    \item $\mathcal{A}(\texttt{mode choice}) = \left\{ \texttt{nbr cars households};\ \texttt{age};\ \texttt{driving license};\ \texttt{trip purpose}\right\}$
    \item $\mathcal{D}(\texttt{mode choice}) = \left\{  \texttt{driving license};\ \texttt{trip purpose} \right\}$
    \item $\mathcal{S}(\texttt{mode choice}) = \left\{\texttt{nbr cars households};\ \texttt{age}\right\}$
    \item $\mathcal{E}(\texttt{mode choice}) = \left\{ \texttt{driving license} \rightarrow \texttt{mode choice};\ \texttt{trip purpose} \rightarrow \texttt{mode choice} \right\}$
\end{itemize}

\subsubsection{Representation of the DAG}
\label{sec:LSTM}

As explained in the introduction of this section, the DAG is used to represent the causal links between the variables. Thus, we want to develop an architecture for the generator similar to the specified DAG. Tabular data cannot be considered sequential data since the order of the variables in a dataset is usually random. However, the DAG allows us to have a sequence of variables with a specific order. We can, thus, use Neural Networks models that have been to work well with this type of data. More specifically, we use Long Short Term Memory (LSTM) cells~\citep{hochreiter_long_1997}, a type of recurrent neural network, to generate synthetic values for each variable $V_t$. We denote the LSTM cell associated to the variable $V_t$ by $\mathbf{LSTM_t}$. The advantage of using recurrent neural networks is that the previous output affects the current state of the neural network. Using the sequence defined by the DAG $\mathcal{G}$, we can, thus, use previous outputs, \emph{i.e.} synthetic values of previous variables, to influence the generation process of a given variable $V_t$.

\begin{figure}[H]
    \centering
    \input{tikz/generic_LSTM}
    \caption{Main components of a LSTM cell following \cite{gers_learning_2000}. The blue hexagons represent variables, the red rectangles neural network layers, and the orange circles mathematical operations. The different gates used to transform the input and the cell state are shown in dark gray.}
    \label{fig:comp_LSTM}
\end{figure}
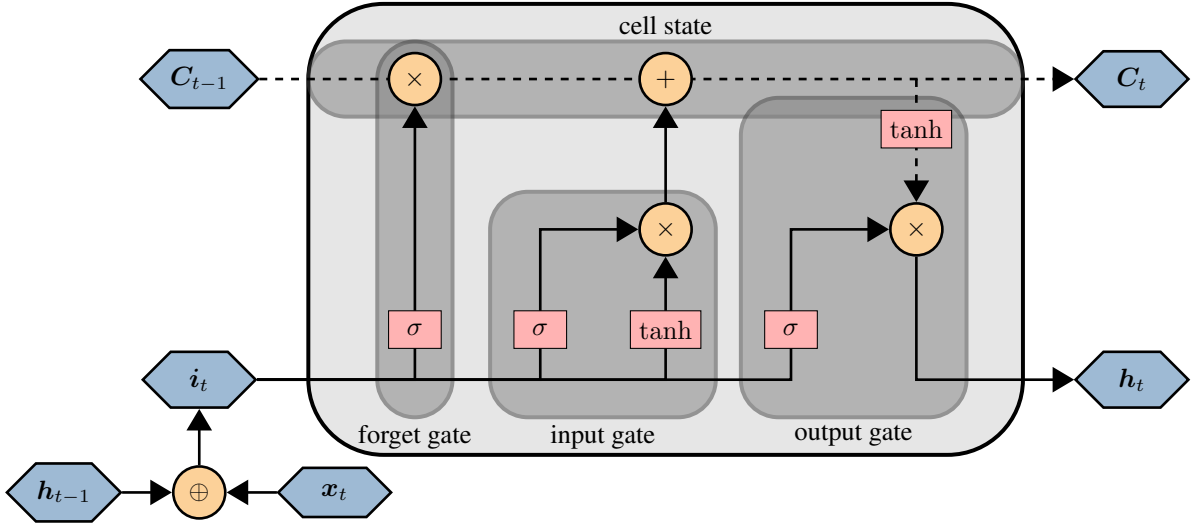

The key elements to an LSTM cell are the cell state and the multiple gates used to protect and control the cell state, as shown in Figure~\ref{fig:comp_LSTM}. For conciseness, we do not present a detailed overview of the mathematical operations of an LSTM cell. For a full description, we direct the reader to \cite{gers_learning_2000}. In Figure~\ref{fig:comp_LSTM}, the input cell state is characterized by $\vec{C}_{t-1}$ and the output cell state by $\vec{C}_t$. The cell will receive an input that is the concatenation between the output of the previous cell ($\vec{h}_{t-1}$) and an input vector ($\vec{x}_t$). It is thus given by:
\begin{equation}
    \vec{i}_t = \vec{h}_{t-1} \oplus \vec{x}_t
\end{equation}
This input vector will pass through three different gates to transform the cell state as it is necessary:
\begin{enumerate}
    \item {\bf the forget gate} is used to decide which old information is forgotten in the cell state.
    \item {\bf the input gate} is used to decide which new information is stored kept in the new cell state
    \item {\bf the output gate} is used to decide the output of the cell using information from both the input $\vec{i}_t$ and the new cell state $C_t$. 
\end{enumerate}
The modeler has to define the size of the hidden layers $N_h$ in the LSTM cell and the batch size $N_b$. The first defines the size of the output vector $\vec{h}_t$ as well as the cell state $\vec{C}_{t}$. The latter corresponds to the number of data points fed in the network. Therefore, we usually define the output $\vec{h}_t$ and cell state $\vec{C}_t$ as tensors of size $N_h \times N_b$. The input $x_t$ usually takes a different size and is thus characterized by a tensor of size $N_x \times N_b$ (the batch size has to remain the same between all the tensors).

In the case of the DATGAN, we have to modify the inputs and outputs of the LSTM cell according to the principles of GANs. Figure~\ref{fig:lstm_datgan} provides a schema of the LSTM structure in the DATGAN. The insides of the LSTM cell $\bf LSTM_t$ are the same as the one shown in Figure~\ref{fig:comp_LSTM}. The first main modification concerns the inputs. Indeed, the generator in a GAN takes random noise as an input instead of an input vector such as $\vec{x}_t$. In addition, we add an attention vector to the input tensor $\vec{i}_t$. The idea behind the attention vector is to keep information of intermediate encoders and pass it to a new encoder. This mimics cognitive attention and, thus, helps with the long-term memory of the LSTM cells. Therefore, the input tensor $\vec{i}_t$ corresponds to the concatenation of three tensors:

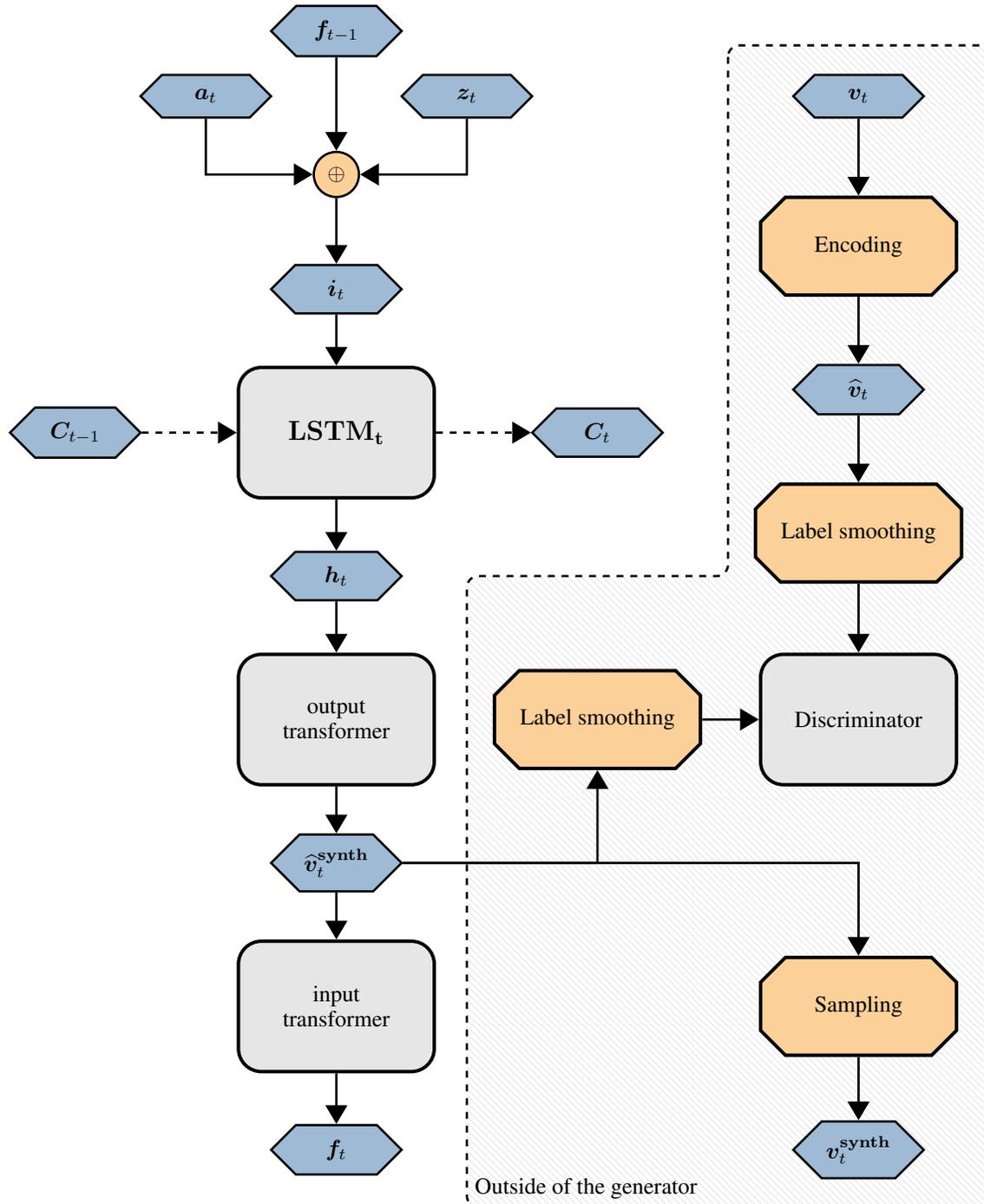
\begin{figure}[hb!]
    \centering
    \input{tikz/DATGAN_LSTM}
    \caption{Schematic representation of how the LSTM cells are used withing the DATGAN to generate the synthetic variable $\widehat{\vec{v}}_t^{\bf synth}$. In Section~\ref{sec:LSTM}, we mainly discuss the left part of the diagram. The right part shows how the new generated variable is passed to the discriminator and sampled. These different elements are presented in the following sections: the Discriminator is presented in Section~\ref{sec:discriminator}, the Encoding in Section~\ref{sec:encoding}, the Label smoothing in Section~\ref{sec:discr_input}, and the Sampling in Section~\ref{sec:sampling}.}
    \label{fig:lstm_datgan}
\end{figure}

\begin{itemize}
    \item $\vec{z}_{t}$ is a tensor of Gaussian noise with dimension $N_z \times N_b$. For each source node in the DAG $\mathcal{G}$, we randomly draw values from $\mathcal{N}(0,1)$. For all the other variables $V_t$, the noise vector is a concatenation of the noise from the source nodes passed through a fully connected layer without any activation function, \emph{i.e.}
    \begin{equation}
        \vec{z}_t = \texttt{FC}\left(\bigoplus_{k\in\mathcal{S}(V_t)} \vec{z}_k, N_z\right)
    \end{equation}
    If two different variables $V_{t_1}$ and $V_{t_2}$ have the same source nodes, \emph{i.e.} $\mathcal{S}(V_{t_1}) = \mathcal{S}(V_{t_2})$, the noise tensor is the same for both variables, \emph{i.e.} $\vec{z}_{t_1} = \vec{z}_{t_2}$. There are two reasons to apply such a rule: 
    \begin{inlinelist} 
        \item it removes pointless computation by creating new variables;
        \item if two variables have a unique source node, they will receive the same noise as an input. We must, therefore, follow this rule if there is more than one source node.
    \end{inlinelist}
    \item $\vec{f}_{t-1}$ can be compared to the previous output tensor $\vec{h}_{t-1}$ in Figure~\ref{fig:comp_LSTM}.
    However, one of the differences between the DATGAN and a usual LSTM network is that we do not directly use the output tensor of the LSTM $\vec{h}_t$. Indeed, as shown in Figure~\ref{fig:lstm_datgan}, we transform it into the encoded synthetic variable $\widehat{\vec{v}}_t^{\bf synth}$. Since $\widehat{\vec{v}}_t^{\bf synth}$ do not have a standard size, we thus need to transform it in order to obtain the usable tensor $\vec{f}_t$ of dimension $N_h \times N_b$. We can thus say that $\vec{f}_t$ corresponds to the transformed output of the LSTM cell $\bf LSTM_t$. If the variable $V_t$ is a source node, this tensor is randomly initialized and learned during the optimization process. 
    \item $\vec{a}_{t}$ is an attention tensor that allows the cell to learn which previous cell outputs are relevant to the input. It is defined as a weighted average over all the LSTM outputs. If the current variable $V_t$ has at least one ancestor, we learn an attention weight vector $\alpha_t\in\mathbb{R}^{|\mathcal{A}(t)|}$. The context vector is thus computed as:
    \begin{equation}
        \vec{a}_t = \sum_{k\in\mathcal{A}(t)} \frac{\exp\alpha_{t}^{(k)}}{\sum_{j}\exp\alpha_{t}^{(j)}}\vec{h}_k 
    \end{equation}
    This context tensor has a dimension $N_h \times N_b$. If variable $V_t$ is a source node, we define $\vec{a}_t$ as a zero-vector of dimension $N_h \times N_b$.
\end{itemize}

LSTMs cells are initially designed to work in sequence, \emph{i.e.} each cell is linked to a unique following cell. However, as shown in the DAG in Figure~\ref{fig:DAG}, some variables can have multiple direct ancestors. For example, the variable \texttt{mode choice} has two ancestors: \texttt{driving license} and \texttt{trip purpose}. We, thus, need a way to connect multiple LSTM cells. If one cell has multiple outputs, we send the output of the cell to the next cells, \emph{e.g.} the inputs of the cells for the variables \texttt{driving license} and \texttt{trip purpose} coming from the variable \texttt{age} are the same. The main issue lies in having multiple cell inputs. The attention tensor $\vec{a}_t$ and the noise tensor $\vec{z}_t$ are defined based on all the ancestors of the current variable. Therefore, we do not have to change these definitions. On the other, the previous cell state $\vec{C}_{t-1}$ and the transformed output $\vec{f}_{t-1}$ are defined to work in sequence. Therefore, we define the multi-input cell state and multi-input transformed output by concatenating the cell states and transformed outputs from the direct ancestors and passing them through a fully connected layer to resize them:
\begin{align}
    \vec{C}_{t-1} &= \texttt{FC}\left(\bigoplus_{k\in\mathcal{D}(V_t)} \vec{C}_k, N_h\right) \\
    \vec{f}_{t-1} &= \texttt{FC}\left(\bigoplus_{k\in\mathcal{D}(V_t)} \vec{f}_k, N_h\right)
\end{align}
During the training process, the weights of the two layers have to be learned. We can, thus, feed the LSTM cell with homogeneous inputs.

\begin{algorithm*}[h!]
\caption{Ordering of the variables using a DAG}\label{algo:ordering}
\begin{algorithmic}[1]
\Require{DAG: $\mathcal{G}$}
\Ensure{ordered list of variables: \texttt{treated}}
\State Compute a dictionary \verb+in_edges+ with $V_t$ as the key and $\mathcal{E}(V_t)$ as the value for all $t=1,\ldots,N_{V}$.
\State Initialize \verb+untreated+ as a set with all the variables names and \verb+treated+ an empty list
\State Initialize \verb+to_treat+ as a list containing all the variables with 0 in-edges
\While {$|\texttt{untreated}|>0$}
    \ForAll {$n \in \texttt{to\_treat}$}
        \State Remove $n$ from \verb+untreated+ and add it to \verb+treated+
    \EndFor
    \State Set \verb+to_treat+ as an empty list
    \ForAll {$e\in \mathcal{G}.E$}\Comment{$e$ is an edge and it is a tuple with 2 values: the out-vertex and the in-vertex}
        \State Initialize boolean \verb+all_ancestors_treated+ to True
        \ForAll {$\ell \in \texttt{in\_edges}[e[1]]$}
            \If {$\ell \notin \texttt{treated}$}
                \State Set \verb+all_ancestors_treated+ to False
            \EndIf
        \EndFor
        \If {$e[0]\in\texttt{treated}$ \textbf{and} $\texttt{all\_ancestors\_treated}$\text{ is True} \textbf{and} $e[1]\notin\texttt{treated}$ \textbf{and} $e[1]\notin\texttt{to\_treat}$}
            \State Add $e[1]$ to the list \verb+to_treat+
        \EndIf
    \EndFor
\EndWhile
\State \textbf{return} \verb+treated+
\end{algorithmic}
\end{algorithm*}

One final issue remains in constructing the structure of the generator. Indeed, we know how to generate each variable separately and connect them using the DAG and the multi-input LSTM cells. However, while building the generator's structure, we cannot start with any random variable in the DAG. Indeed, as per the definition of the inputs of the LSTM cell, the ancestors $\mathcal{A}(V_t)$ must have already been built first. For example, the attention tensor uses the outputs of all the ancestors' cells. Therefore, we need an algorithm that creates an ordered list based on the provided DAG $\mathcal{G}$. This algorithm is given in Algorithm~\ref{algo:ordering}. The goal of this algorithm is to take the DAG $\mathcal{G}$ and return an ordered list of the variables $V_t$ such that all the ancestors of a given variable have a smaller index in the list, \emph{i.e.} they appear first in the list. The algorithm is built recursively. The idea is to define two lists, one with untreated nodes \texttt{untreated}, containing all the nodes at the beginning of the algorithm, and one with treated nodes \texttt{treated}, empty at the beginning. We, then, start by selecting all source nodes and adding to a list named \texttt{to\_treat}. Then, while the list of untreated nodes is not empty, \emph{i.e.} while there are still nodes to be added to the final list, we start by assigning all nodes in the list \texttt{to\_treat} in the \texttt{treated} list. Then, we check each edge in the DAG $\mathcal{G}$ and check if all the ancestors of the out-vertex have been treated. If it is the case, we can add this node to the \texttt{to\_treat} list and assign it later to the final list. The algorithm stops when all the nodes have been added to the \texttt{treated} list.

Now that every component has been defined for the generator, we can build it following the ordered list provided by Algorithm~\ref{algo:ordering}. Each time that we create a LSTM cell for the variable $V_t$, as in Figure~\ref{fig:lstm_datgan}, we check the direct ancestors $\mathcal{D}(V_t)$. If more than one direct ancestor, we apply the multi-input technique to the LSTM cell. The generator is finished once one LSTM cell has been created for each variable in the ordered list.

\subsection{Discriminator}
\label{sec:discriminator}

As seen in Figure~\ref{fig:DATGAN}, the generator is used to create the synthetic dataset $\bf \widehat{T}_{synth}$. The role of the discriminator is to compare this dataset with the encoded original dataset $\bf \widehat{T}$. We, thus, want to train the discriminator to be able to identify original and synthetic data. That way, the generator will have to produce better synthetic data if it wants to fool the discriminator. 

Following \cite{xu_synthesizing_2018}, we use a fully connected neural network with $N_{L}$-layers for the discriminator, where the internal layers, for $i=1,\ldots,N_L$, are given by:
\begin{align}
    \widehat{\vec{l}}_{i} &= \texttt{FC}\left(\vec{l}_{i-1}, N_l\right)\\
    \vec{l}_i &= \texttt{LeakyReLU}\left(\texttt{BN}\left(\widehat{\vec{l}}_{i}\oplus\texttt{div}\left(\widehat{\vec{l}}_{i}\right)\right)\right)
\end{align}
where \begin{inlinelist}
    \item $\texttt{div}(\cdot)$ represents the mini-batch discrimination vector presented by \cite{salimans_improved_2016};
    \item $\texttt{BN}(\cdot)$ corresponds to the batch normalization;
    \item $\texttt{LeakyReLU}(\cdot)$ is the leaky reflect linear activation function.
\end{inlinelist}
The output of the discriminator is computed using a fully connected layer with a size of 1 to return an unbounded scalar:
\begin{equation}
    l_{D} = \texttt{FC}\left(\vec{l}_{N_L}, 1\right)
\end{equation}
The input vector $\vec{l}_0$ of the discriminator is different depending on the data it is using:
\begin{itemize}
    \item For the original dataset, $\vec{l}_0$corresponds to the concatenation of all the column vectors $\left\{ \widehat{\vec{v}}_{1:N_{V}}\right\}$ after an encoding step, as shown in Figure~\ref{fig:DATGAN}.
    \item For the synthetic dataset, $\vec{l}_0$ corresponds to the concatenation of all the usable outputs $\left\{ \widehat{\vec{v}}^{\bf synth}_{1:N_{V}}\right\}$ given by the generator, as shown in Figure~\ref{fig:lstm_datgan}.
\end{itemize}

However, as seen in Figure~\ref{fig:lstm_datgan}, a label smoothing step has to be performed before feeding these matrices to the discriminator. This step is discussed in Section~\ref{sec:discr_input}.

\subsection{Loss function}
\label{sec:loss_function}

The loss function sets up the game between the discriminator and the generator. The discriminator $D$ aims to maximize the loss function when comparing the synthetic data produced by the generator against the original data. Meanwhile, the generator $G$ aims to minimize the same loss function by generating synthetic data, fooling the discriminator. The generator thus learns from the discriminator in this process by backpropagating the discriminator loss. 

Since the loss function drives the optimization process to obtain the best possible model, we argue that our model does not have to be characterized by a single loss function. Therefore, we systematically test three different loss functions. The first one is the standard cross-entropy loss defined by~\cite{goodfellow_generative_2014}, the second one is the Wasserstein or Earth-Mover distance defined by~\cite{arjovsky_wasserstein_2017}, and the third one is the Wasserstein distance with Gradient-Penalty defined by~\cite{gulrajani_improved_2017}.

\paragraph{Standard loss} The first loss function is the standard loss function used in the original GAN by~\cite{goodfellow_generative_2014}. We name it $\mathcal{L}^{\texttt{SGAN}}$ with SGAN standing for Standard GAN. It is given by: 
\begin{equation}
    \label{eq:std_loss}
    \min_{G} \max_{D} \mathcal{L}^{\texttt{SGAN}}(D,G) = \mathop{\mathbb{E}}_{\left\{\widehat{\vec{v}}_{1:N_{V}}\right\}\sim\mathbb{P}({\bf\widehat{T}})} \log D\left(\widehat{\vec{v}}_{1:N_{V}}\right) + \mathop{\mathbb{E}}_{\vec{z}\sim\mathcal{N}(0,1)} \log\left(1-D(G(\vec{z}))\right)
\end{equation}

This loss function requires the discriminator to produce a probability for each data point to be either original or synthetic. However, as it has been defined in Section~\ref{sec:discriminator}, the discriminator outputs an unbounded scalar and not a probability. In order to use this discriminator with this loss function, we thus pass the output through an additional sigmoid layer to produce bounded $[0,1]$ probabilities. 

As explained by~\cite{goodfellow_generative_2014}, $\log\left(1-D(G(\vec{z}))\right)$ tends to saturate during the training process. Therefore, instead of training $G$ to minimize the full loss function, we can instead train it to maximize $\log D(G(\vec{z}))$. We can thus define the loss function for both networks separately. The goal is to minimize both losses simultaneously during the training process. They are given by:
\begin{align}
    \label{eq:SGAN_gen}
    \mathcal{L}^{\texttt{SGAN}}_{G} &= -\mathop{\mathbb{E}}_{\vec{z}\sim\mathcal{N}(0,1)} \log D(G(\vec{z})) \\
    \mathcal{L}^{\texttt{SGAN}}_{D} &= -\mathop{\mathbb{E}}_{\left\{\widehat{\vec{v}}_{1:N_{V}}\right\}\sim\mathbb{P}({\bf \widehat{T}})} \log D\left(\widehat{\vec{v}}_{1:N_{V}}\right) + \mathop{\mathbb{E}}_{\vec{z}\sim\mathcal{N}(0,1)} \log D(G(\vec{z}))
\end{align}
As suggested by the authors, we train our models using the Adam optimizer~\citep{kingma_adam_2014}.

\paragraph{Wasserstein loss} The second loss function has been implemented in the Wasserstein GAN (WGAN) by~\cite{arjovsky_wasserstein_2017}. It is defined using the Earth-Mover distance:
\begin{equation}
    \label{eq:wgan_loss}
    \min_{G} \max_{D} \mathcal{L}^{\texttt{WGAN}}(D,G) = \mathop{\mathbb{E}}_{\left\{\widehat{\vec{v}}_{1:N_{V}}\right\}\sim\mathbb{P}({\bf \widehat{T}})} D\left(\widehat{\vec{v}}_{1:N_{V}}\right) - \mathop{\mathbb{E}}_{\vec{z}\sim\mathcal{N}(0,1)} D(G(\vec{z}))
\end{equation}
There are multiple advantages to use this loss function instead of the standard loss: \begin{inlinelist}
    \item the main advantage is the fact that the discriminator becomes a critic since it does not need to produce a 0-1 output anymore. Indeed, we can use the output of the discriminator $l_D$ as it is defined. It thus results in less vanishing gradients and an easier learning process for the generator $G$;
    \item the loss function correlates with the quality of the sample, contrary to the SGAN loss. It is, thus, possible to determine when the GAN has converged.
\end{inlinelist}

We can, now, define the separate loss functions for both networks as:
\begin{align}
    \mathcal{L}^{\texttt{WGAN}}_{G} &= -\mathop{\mathbb{E}}_{\vec{z}\sim\mathcal{N}(0,1)} D(G(\vec{z})) \\
    \mathcal{L}^{\texttt{WGAN}}_{D} &= -\mathop{\mathbb{E}}_{\left\{\widehat{\vec{v}}_{1:N_{V}}\right\}\sim\mathbb{P}({\bf \widehat{T}})} D\left(\widehat{\vec{v}}_{1:N_{V}}\right) + \mathop{\mathbb{E}}_{\vec{z}\sim\mathcal{N}(0,1)} D(G(\vec{z}))
\end{align}
As suggested by the authors, we train our models using the RMSProp optimizer~\citep{tieleman_lecture_2012}.

\paragraph{Wasserstein loss with gradient penalty} The loss for the WGAN-GP is the same as the Wasserstein loss with the addition of a gradient penalty~\citep{gulrajani_improved_2017}. It is given by:
\begin{equation}
    \label{eq:wggp_loss}
    \min_{G} \max_{D} \mathcal{L}^{\texttt{WGGP}}(D,G) = \mathop{\mathbb{E}}_{\left\{\widehat{\vec{v}}_{1:N_{V}}\right\}\sim\mathbb{P}({\bf \widehat{T}})} D\left(\widehat{\vec{v}}_{1:N_{V}}\right) - \mathop{\mathbb{E}}_{\vec{z}\sim\mathcal{N}(0,1)} D(G(\vec{z})) + \lambda \mathop{\mathbb{E}}_{\vec{\widehat{v}}\sim\mathbb{P}({\bf \widehat{T}}, G(\vec{z}))}\left(\left\lVert \nabla_{\vec{\widetilde{v}}} D(\vec{\widetilde{v}})\right\rVert_2 -1 \right)^2
\end{equation}
where $\lambda$ is a parameter defined by the modeler. The main issue with the WGAN is that it needs to enforce the Lipschitz contsraint on the critic. It does that by clipping the weights of the critic. \cite{gulrajani_improved_2017} show that it leads undesired behaviour in the generator samples. We can, thus, fix this issue by adding a gradient penalty on the critic. In Equation~\ref{eq:wggp_loss}, the mid-value $\widetilde{\vec{v}}$ is sampled uniformely along straight lines between pair of points sampled from the original dataset ${\bf T}$ ($\widehat{v}$) and generated data $G(\vec{z})$ ($\widehat{v}^{\bf synth}$).

The separate loss functions for each networks are, therefore, defined as:
\begin{align}
    \mathcal{L}^{\texttt{WGGP}}_{G} &= -\mathop{\mathbb{E}}_{\vec{z}\sim\mathcal{N}(0,1)} D(G(\vec{z})) \\
    \mathcal{L}^{\texttt{WGGP}}_{D} &= -\mathop{\mathbb{E}}_{\left\{\widehat{\vec{v}}_{1:N_{V}}\right\}\sim\mathbb{P}({\bf \widehat{T}})} D\left(\widehat{\vec{v}}_{1:N_{V}}\right) + \mathop{\mathbb{E}}_{\vec{z}\sim\mathcal{N}(0,1)} D(G(\vec{z})) + \lambda \mathop{\mathbb{E}}_{\vec{\widehat{v}}\sim\mathbb{P}({\bf \widehat{T}}, G(\vec{z}))}\left(\left\lVert \nabla_{\vec{\widehat{v}}} D(\vec{\widehat{v}})\right\rVert_2 -1 \right)^2
\end{align}
Following \cite{gulrajani_improved_2017}, we replace the batch normalization in the discriminator with a layer normalization~\citep{ba_layer_2016}, we set $\lambda = 10$, and we train both models using the Adam optimizer~\citep{kingma_adam_2014}.

Following \cite{xu_synthesizing_2018}, we include a Kullback-Leibler (KL) divergence term to all the generator losses. For two discrete probability distributions $P$ and $Q$ defined on the same probability space $\mathcal{X}$, the KL divergence is given by:

\begin{equation}
    \text{KL}\left(P, Q\right) = \sum_{x\in\mathcal{X}} P(x) \log\left(\frac{P(x)}{Q(x)}\right)
\end{equation}

Therefore, we can use this divergence for any discrete probability distributions in the original and synthetic datasets. The use of this divergence has two main consequences: \begin{inlinelist}
    \item it gives a boost when starting the training of the generator since it is trying to make discrete probability distributions as close as possible;
    \item it makes the model more stable under training.
\end{inlinelist}
We discuss which variables are concerned by this divergence in Section~\ref{sec:encoding} for the original dataset $\bf \widehat{T}$ and in Section~\ref{sec:gen_output} for the synthetic dataset $\bf \widehat{T}_{synth}$.

\subsection{Data processing}
\label{sec:data_processing}

Tabular data are generally composed of multiple data types, as seen in Figure~\ref{fig:data}. In the context of this article, we consider two different variable types:
\begin{description}
    \item[Continuous data] corresponds to a random variable following a continuous distribution (\emph{e.g.} the distance to travel to a destination). The variable can then be rounded to obtain discrete values, as is usually the case with an individual's age. 
    \item[Categorical data] corresponds to all other types of data such as: \\
    $\bullet$\: binary random variables (\emph{e.g.} whether someone is retired or not). \\
    $\bullet$\: nominal random variables, \emph{i.e.} discrete random variable with three or more possible values, where there is no order nor notion of distance between the different values (\emph{e.g.} a color). \\
    $\bullet$\: ordinal random variables, \emph{i.e.} discrete random variables with three or more possible values with a defined order (and possibly also distance) between each possible value (\emph{e.g.} education level). Contrary to nominal data, we can define an order (and possibly a distance) between the different categories.
\end{description}

Typically, in Machine Learning, neural networks work with data ranging from -1 to 1 or 0 to 1. However, these four types of data are not designed in such a way. We thus need to encode the original dataset $\bf T$ in a dataset $\bf \widehat{T}$, as shown in Figure~\ref{fig:DATGAN}, that transforms the different data types into more homogeneous types. 

The table $\bf T$ contains $N_C$ continuous random variables $\left\{C_1,\ldots,C_{N_C}\right\}$ and $N_D$ categorical random variables $\left\{D_1,\ldots,D_{N_D}\right\}$ such that $N_C + N_D = N_V$. We can thus define the table $\bf T$ using vectors of continuous and categorical variables, \emph{i.e.} ${\bf T} = \left\{\vec{c}_{1:N_C}, \vec{d}_{1:N_D}\right\}$. Similarly, the synthetic dataset is defined as ${\bf T_{synth}} = \left\{\vec{c}^{\bf synth}_{1:N_C}, \vec{d}^{\bf synth}_{1:N_D}\right\}$.

Since we are considering two different data types in the table $\bf T$, we cannot process them the same way. Thus, Section~\ref{sec:encoding} explains how the encoding is done for each type. Section~\ref{sec:gen_output} explains how these types of data are generated. Section~\ref{sec:discr_input} discusses how the synthetic data and the original data are passed to the discriminator. Finally, Section~\ref{sec:sampling} shows how the data is sampled from the generator's output to create the final synthetic dataset. 

\subsubsection{Encoding}
\label{sec:encoding}

The DATGAN model takes as input only $[-1,1]$ or $[0,1]$ bounded vectors. Therefore, we need to encode unbounded continuous variables and categorical variables to be processed by the GAN. Continuous data tend to follow multimodal distributions. We thus build on the previous methodology of \cite{xu_modeling_2019} who apply a Variational Gaussian Mixture (VGM) model~\citep{bishop_pattern_2006} to cluster continuous values into a discrete number of Gaussian mixtures. In this work, we develop this further by automatically determining the number of components from the data. 

\begin{algorithm*}[h!]
\caption{Continuous variables preprocessing}\label{algo:cont_var}
\begin{algorithmic}[1]
\Require{List of float values ($\vec{c}_t$)}
\Ensure{Matrix of probabilities ($\vec{p}_t$) and values ($\vec{w}_t$)}
\State Set the initial number of modes $N_{m,t} = 10$
\While{True}
\State Sample $\vec{s}_t$ from $\vec{c}_t$ and fit the \texttt{BayesianGaussianMixture} to $\vec{s}_t$ with $N_{m,t}$ modes
\State Predict the class on $\vec{s}_t$ and compute $N_{\text{pred}}$ the number of unique classes predicted by the VGM
\State Define $N_{\text{weights}}$, the number of weights of the model above a threshold $\varepsilon_w=0.01$
\If{$N_{\text{pred}} < N_{m,t}$ {\bf or} $N_{\text{weights}} < N_{m,t}$}
\State $N_{m,t} = \min\left(N_{\text{pred}}, N_{\text{weights}}\right)$
\Else
\State break
\EndIf
\EndWhile
\State Fit the \texttt{BayesianGaussianMixture} to $\vec{c}_t$ with $N_{m,t}$ modes. 
\State Means and standard deviations of the $N_{m,t}$ Gaussian mixtures are given by
\[
\vec{\eta}_t = \left(\eta_t^{(1)},\ldots,\eta_t^{(N_{m,t})}\right) 
\ \text{and}\ 
\vec{\sigma}_t = \left(\sigma_t^{(1)},\ldots,\sigma_t^{(N_{m,i})}\right)
\]
\State Compute the posterior probability of $c_{t,j}$ coming from each of the $N_{m,t}$ mixtures as a vector $\vec{p}_{t,j} = \left(p_{t,j}^{(1)},\ldots,p_{t,j}^{(N_{m,t})}\right)$. It corresponds to a normalized probability distributions over the $n_{m,i}$ Gaussian distributions.
\State Normalize $c_{t,j}$ for each Gaussian mixture using Equation~\ref{eq:normalize}.
\State Clip each value $w_{t,j}^{(k)}$ between -0.99 and 0.99 and set $\vec{w}_{t,j} = \left(w_{t,j}^{(1)}, \ldots, w_{t,j}^{(N_{m,t})}\right)$.
\State \textbf{return} $\vec{p}_t$ and $\vec{w}_t$
\end{algorithmic}
\end{algorithm*}

For each continuous variable $C_t$ in the dataset, we first train a VGM on a random subset of the data with a high number of components ($N_{m,t}=10$). We then determine how many components are needed to capture the distribution by comparing the component weights against a threshold and the number of predicted components $N_{\text{pred}}$ with the original number of components $N_{m,t}$. We repeat this process until convergence. Finally, we retrain the model on the entire column vector $\vec{c}_t$ using only the number of significant components. From this trained model, we extract the means $\vec{\eta}_t$ and standard deviations $\vec{\sigma}_t$ from the VGM. We can then normalize the values $c_{t,j}$ using the following formula:
\begin{equation}
    \label{eq:normalize}
    w_{t,j}^{(k)} = \frac{c_{t,j}-\eta_t^{(k)}}{\delta \sigma_t^{(k)}}\quad \text{for } k = 1, \ldots, N_{m,t},
\end{equation}
where $\delta$ is a parameter specified by the modeller. Following \cite{xu_synthesizing_2018}, we use a value of $\delta=2$ and clip the values of $w_{t,j}^{(k)}$ between -0.99 and 0.99. At the same time, we compute the posterior probability vectors $\vec{p}_{t,j}$ that the value $c_{t,j}$ belongs to each of the $N_{m,t}$ mixtures. Thus, each value $c_{t,j}$ in $\vec{c}_t$ are represented by the vector of probabilities $\vec{p}_{t,j} \in [0,1]^{N_{m,t}}$ and the vector of values $\vec{w}_{t,j} \in [-0.99,0.99]^{N_{m,t}}$. Algorithm~\ref{algo:cont_var} shows a summary of the procedure used to preprocess continuous variables.

For categorical variables\footnote{The same treatment is applied to categorical and boolean variables due to the label smoothing explained in Section~\ref{sec:discr_input}.}, we transform them using one-hot encoding. We consider $\vec{d}_t$ the realizations of the random variable $D_t$. $\vec{d}_t$ is transformed using $|D_t|$-dimensional one-hot vector $\vec{o}_t$ where $|D_t|$ corresponds to the number of unique categories in $D_t$.

We can thus convert the initial table ${\bf T}$ into an intermediate table ${\bf \widehat{T}} = \left\{\vec{w}_1, \vec{p}_1, \ldots, \vec{w}_{N_C}, \vec{p}_{N_C},\vec{o}_1, \ldots, \vec{o}_{N_D}\right\}$. As a simplification, we write ${\bf \widehat{T}} = \left\{\vec{w}_{1:N_C}, \vec{p}_{1:N_C}, \vec{o}_{1:N_D}\right\}$. The dimension of this new table is given by $\sum_{t=1}^{N_C}2N_{m,t} + \sum_{t=1}^{N_D}|D_t|$. While this new encoded table is larger than the original table, we can now use define the KL divergence on multiple variables. Indeed, the vector $\vec{p}_{t,j}$ corresponds to a discrete probability distribution for a given row. We can, thus, apply the KL divergence on this term. In addition, the one-hot encoded vector $\vec{o}_{t,j}$ also corresponds to a discrete probability distributions. The only difference is that this distribution is composed of only 1s and 0s. Nevertheless, the KL divergence is also applicable on this variable.

\subsubsection{Generator output}
\label{sec:gen_output}

In Figure~\ref{fig:lstm_datgan}, we show how the LSTM cell $\bf LSTM_t$ produces an output $\vec{h}_t$ before it is transformed into the encoded synthetic variable $\widehat{\vec{v}}_t^{\bf synth}$. However, in Section~\ref{sec:encoding}, we show that we distinguish between two variable types. Thus, the output transformer in Figure~\ref{fig:lstm_datgan} differs depending on if we are working with a continuous or a categorical variable. Nevertheless, the first step for the output transformer is similar in both cases. Indeed, the goal is to use some semblance of convolution on the LSTM output $\vec{h}_t$ to improve the results. We, thus, transform this output through a hidden layer:
\begin{equation}
    \vec{h}'_t = \texttt{tanh}\left(\vec{h}_{t}, N_{\text{conv}} \right)
\end{equation}
The final transformation into the synthetic encoded variable depends on the variable type. For continuous variables, we thus pass the reduced output $\vec{h}'_t$ through two different fully connected layers to extract both the vector of probabilities $\vec{p}^{\bf synth}_t$ and the vector of corresponding values $\vec{w}^{\bf synth}_t$:
\begin{align}
    \vec{w}^{\bf synth}_t &= \texttt{tanh}(\vec{h}'_{t}, N_{m,t}) \\
    \vec{p}^{\bf synth}_t &= \texttt{softmax}(\vec{h}'_{t}, N_{m,t})
\end{align}

For the categorical variables, we pass the reduced output $\vec{h}'_t$ through a single fully connected layer to extract the output probabilities $\vec{o}^{\bf synth}_t$ belonging to each class:
\begin{equation}
    \vec{o}^{\bf synth}_t = \texttt{softmax}(\vec{h}'_{t}, |D_t|)
\end{equation}

Both matrices of discrete probabilities $\vec{p}^{\bf synth}_t$ and $\vec{o}^{\bf synth}_t$ are using a softmax activation function in order to ensure that the sum along the rows is equal to one. This ensures that the rows of these matrices correspond to discrete probability vectors. The matrix $\vec{w}^{\bf synth}_t$ uses a $\tanh$ activation function since we are allowing this matrix to take values between -1 and 1.

Since these encoded synthetic variables do not have homogeneous sizes, we cannot use them directly as the input of the next LSTM cell. This, thus, explains why we are passing the encoded synthetic values $\widehat{\vec{v}}_t^{\bf synth}$ through an input transformer. The goal of this transformer is to take $\widehat{\vec{v}}_t^{\bf synth}$ and transform it back to the same tensor for all the different variables $V_t$. Therefore, we have to distinguish between continuous and categorical variables again. For the continuous variables, we concatenate the transformed synthetic variables together and pass them through a fully connected layer to obtain $\vec{f}_t$:
\begin{equation}
    \vec{f}_t = \texttt{FC}(\vec{w}^{\bf synth}_t\oplus\vec{p}^{\bf synth}_t, N_{h})
\end{equation}
For the categorical variables, we just pass $\vec{o}^{\bf synth}_t$ through a fully connected layer:
\begin{equation}
    \vec{f}_t = \texttt{FC}(\vec{o}^{\bf synth}_t, N_{h})
\end{equation}

Finally, the tensors $\vec{w}^{\bf synth}_t$, $\vec{p}^{\bf synth}_t$, and $\vec{o}^{\bf synth}_t$ are combined to form the encoded synthetic table ${\bf \widehat{T}_{synth}} = \left\{\vec{w}^{\bf synth}_{1:n_C}, \vec{p}^{\bf synth}_{1:n_C}, \vec{o}^{\bf synth}_{1:n_D}\right\}$. This synthetic table is passed to the discriminator as an input for the optimization process. We thus directly compare it to the encoded table ${\bf \widehat{T}}$ defined in Section~\ref{sec:encoding}. 

\subsubsection{Discriminator input}
\label{sec:discr_input}

As explained in Section~\ref{sec:discriminator}, the input tensor $\vec{l}_0$ corresponds to the concatenation of all the variables in ${\bf \widehat{T}}$ for the original data or $\bf \widehat{T}_{synth}$ for the synthetic data. 

For categorical variables, where the original data are one-hot encoded, it would be trivial for a discriminator to differentiate between the original and synthetic values (as the generator produces probabilities over each class, which will not be $\{0,1\}$ vectors). In addition, as explained by \cite{goodfellow_nips_2017}, deep networks tend to produce overconfident results when adversarially constructed. The author thus suggests using one-sided label smoothing for the standard loss function, as defined in Section~\ref{sec:loss_function}. It means that we perturb the $\{0,1\}$ vectors with additive uniform noise and rescale them to produce $[0,1]$ bound vectors. Formally, label smoothing is defined as follows:
\begin{align}
    \widetilde{o}^{(k)}_{t, j} &= o^{(k)}_{t, j} + \mathcal{U}_{[0,\gamma]}, \quad k=0,\ldots, |D_t| \nonumber \\
    \widetilde{\vec{o}}_{t} &= \widetilde{\vec{o}}_{t}/||\widetilde{\vec{o}}_{t}||
\end{align}
where $\gamma$ is a parameter defined by the modeler. $\widetilde{\vec{o}}_{t}$ now corresponds to a noisy version of the original one-hot encoded vector $\vec{o}_{t}$. 

An issue with applying label smoothing is that the generator output tries to match the distorted representation of the data, and so the generator probability outputs will be biased towards low proability values. To address this, we propose here to apply equivalent smoothing to the generator output before passing it to the discriminator:
\begin{align}
    \widetilde{o}^{{\bf synth},(k)}_{t, j} &= o^{{\bf synth},(k)}_{t, j} + \mathcal{U}_{[0,\gamma]}, \quad k=0,\ldots, |D_t| \nonumber \\
    \widetilde{\vec{o}}^{\bf synth}_{t} &= \widetilde{\vec{o}}^{\bf synth}_{t}/||\widetilde{\vec{o}}^{\bf synth}_{t}||
\end{align}
where $\gamma$ should match the parameter used for the input smoothing. It removes the bias in the generator output and, thus, it is effectively trying to learn the original $[0,1]$ representations. Therefore, it produces unbiased probabilities. We refer to this as two-sided label smoothing. 

To investigate the benefits of label smoothing, we systematically investigate three possible strategies for categorical variables:

\begin{align}
    \label{eq:NO}
    \text{no label smoothing:}&& \vec{l}_0^{\texttt{NO}} &= \begin{cases}
    \vec{w}_{1:N_C}\oplus\vec{p}_{1:N_C}\oplus\vec{o}_{1:N_D} & \text{for original data} \\
    \vec{w}^{\bf synth}_{1:N_C}\oplus\vec{p}^{\bf synth}_{1:N_C}\oplus\vec{o}^{\bf synth}_{1:N_D} & \text{for synthetic data}
    \end{cases} \\
    \label{eq:OS}
    \text{one-sided label smoothing:}&& \vec{l}_0^{\texttt{OS}} &= \begin{cases}
    \vec{w}_{1:N_C}\oplus\vec{p}_{1:N_C}\oplus\widetilde{\vec{o}}_{1:N_D} & \text{for original data} \\
    \vec{w}^{\bf synth}_{1:N_C}\oplus\vec{p}^{\bf synth}_{1:N_C}\oplus\vec{o}^{\bf synth}_{1:N_D} & \text{for synthetic data}
    \end{cases} \\
    \label{eq:TS}
    \text{two-sided label smoothing:}&& \vec{l}_0^{\texttt{TS}} &= \begin{cases}
    \vec{w}_{1:N_C}\oplus\vec{p}_{1:N_C}\oplus\widetilde{\vec{o}}_{1:N_D} & \text{for original data} \\
    \vec{w}^{\bf synth}_{1:N_C}\oplus\vec{p}^{\bf synth}_{1:N_C}\oplus\widetilde{\vec{o}}^{\bf synth}_{1:N_D} & \text{for synthetic data}
    \end{cases}
\end{align}

\subsubsection{Sampling}
\label{sec:sampling}

Once the generator has been trained against the discriminator, we need to be able to generate the final synthetic dataset $\bf T_{synth}$. However, the dataset created by the generator $\bf \widehat{T}_{synth}$ corresponds to the encoded dataset $\bf \widehat{T}$. Therefore, we need to decode $\bf \widehat{T}_{synth}$ to obtain the final synthetic dataset. 

In previous works \citep{xu_synthesizing_2018,xu_modeling_2019}, the synthetic value is sampled from the probability distribution by simply assigning the value to the highest probability class (\emph{i.e.} $\argmax$ assignment). However, this approach does not result in representative mode shares. Instead, as is typical in choice modeling scenarios~\citep{ben-akiva_discrete_1985}, we propose to sample the synthetic value through simulation, \emph{i.e.} drawing according to the output probability values (without any label smoothing applied). Thus, for categorical variables, we have two different ways to obtain the final value:
\begin{align}
    d^{{\bf synth}, \argmax}_{t,j} &= \argmax \vec{o}^{\bf synth}_{t,j} \\
    d^{{\bf synth}, \texttt{simul}}_{t,j} &= \texttt{simulation}\left[\vec{o}^{\bf synth}_{t,j}\right]
\end{align}
Similary, by inverting Equation~\ref{eq:normalize}, we have for continuous variables:
\begin{align}
    c^{{\bf synth}, \argmax}_{t, j} &= \delta w_{t,j}^{{\bf synth}, (k)}\sigma_t^{(k)} + \eta_t^{(k)} && \text{where} \ k = \argmax \vec{p}^{\bf synth}_{t,j} \\
    c^{{\bf synth}, \texttt{simul}}_{t, j} &= \delta w_{t,j}^{{\bf synth}, (k)}\sigma_t^{(k)} + \eta_t^{(k)} && \text{where} \ k = \texttt{simulation}\left[\vec{p}^{\bf synth}_{t,j}\right]
\end{align}
where $\delta$ corresponds to the same values used to encode the continuous variables $\vec{c}_t$ in Section~\ref{sec:encoding}.

In order to test the impacts of simulation versus maximum probability assignment for the categorical and continuous variables, we systematically test four different sampling strategies:

\begin{align}
    \label{eq:AA}
    \text{Only $\argmax$:}&& {\bf T}_{\bf synth}^{\texttt{AA}} &= \left\{\vec{c}^{{\bf synth}, \argmax}_{1:N_C}, \vec{d}^{{\bf synth}, \argmax}_{1:N_D}\right\} \\
    \label{eq:SA}
    \text{\texttt{simulation} for continuous and $\argmax$ for categorical:}&& {\bf T}_{\bf synth}^{\texttt{SA}} &=  \left\{\vec{c}^{{\bf synth}, \texttt{simul}}_{1:N_C}, \vec{d}^{{\bf synth}, \argmax}_{1:N_D}\right\}\\
    \label{eq:AS}
    \text{$\argmax$ for continuous and \texttt{simulation} for categorical:}&& {\bf T}_{\bf synth}^{\texttt{AS}} &=  \left\{\vec{c}^{{\bf synth}, \argmax}_{1:N_C}, \vec{d}^{{\bf synth}, \texttt{simul}}_{1:N_D}\right\}\\
    \label{eq:SS}
    \text{Only \texttt{simulation}:}&& {\bf T}_{\bf synth}^{\texttt{SS}} &= \left\{\vec{c}^{{\bf synth}, \texttt{simul}}_{1:N_C}, \vec{d}^{{\bf synth}, \texttt{simul}}_{1:N_D}\right\}
\end{align}

As a side note, we would like to add that the sampling process is entirely independent of the optimization process of both the generator and the discriminator. It is, therefore, possible to train a single model and test the different sampling methods afterward. 

The supplementary materials summarize the possible DATGAN versions using the proposed loss functions, label smoothing strategies, and sampling strategies. In addition, we compare these versions against each other in order to select the best-performing one. 

\subsection{Result assessments}
\label{sec:result_assessments}

For assessing the quality of synthetic datasets, compared to the original datasets, we use two main methods: \begin{inlinelist}
\item a statistical method;
\item a machine learning-based method.
\end{inlinelist} The goal of the first method (see Section~\ref{sec:res_stats}) is to verify that the synthetic datasets display the same statistical properties compared to the original dataset. In order to do this, we compare the distributions of each column individually between the synthetic datasets and the original datasets. We then test combinations of multiple columns to study if the models can grasp more complex correlations between the variables. The second method (see Section~\ref{sec:res_ml}) is closer to a real-world problem one can face. Indeed, the goal is to study if the synthetic datasets can be used in classification/regression context in Machine Learning. 

\subsubsection{Statistical tests}
\label{sec:res_stats}

For the statistical tests, we build on existing approaches in the transportation literature~\citep{garrido_prediction_2019, borysov_how_2019, badu-marfo_composite_2020}. The idea is to compute frequency lists (i.e. frequency count of each unique value) for each column on both the original dataset $\vec{\pi}$ and the synthetic dataset $\vec{\pi}^{\bf synth}$. In the literature, authors typically only calculate the frequency lists for single columns (i.e. marginal distributions) for a few relevant variables and test them against each other. In this paper, we build on this in two ways:
(i)~calculating joint frequency lists for $n$~columns simultaneously (therefore assessing joint distributions of order~$n$) and
(ii)~systematically testing all possible combinations of columns at each aggregation level. 

If we only compute only the frequency lists for single variables, we will only assess the marginal distribution of each variable independently. Whilst this verifies whether each column of the synthetic data matches distribution of the corresponding column in the original data, it does not provide any information of the correlations between the columns in either the synthetic or original data (i.e. it assesses each column independently of all other columns). To assess whether relationships between variables in the the synthetic data matches that of the original data, it is necessary to investigate the joint distributions of multiple columns simultaneously. To address this, we calculate the joint frequency lists for multiple columns simultaneously. Furthermore, at each aggregation level (i.e. number of columns), we calculate the frequency lists for all possible combinations of columns.

Since the number of possible combinations is at each aggregation level increases factorially, we limit the level of aggregation to one, two, or three columns as follows: 
\begin{description}
    \item[First order:] Columns are compared to each other separately. It returns $N_{V}$ different aggregated lists.
    \item[Second order:] Columns are aggregated two-by-two, giving $\binom{N_{V}}{2}$ different aggregated lists.
    \item[Third order:] Columns are aggregated three-by-three, giving $\binom{N_{V}}{3}$ different aggregated lists.
\end{description}
Note that continuous variables must be binned to calculate frequency lists.
We arbitrarily set the number of bins for each continuous column to 10, such that the second order and third order frequency lists of continuous columns will have 100 and 1000 unique values respectively. 

Once these frequency lists have been computed, we can compare them using standard statistic metrics defined in the literature. We select five different metrics:
\begin{itemize}
    \item Mean Absolute Error:
    \begin{equation}
        \text{MAE}\left(\vec{{\pi}}^{\bf synth}, \vec{\pi}\right) = \dfrac{\sum_{i=1}^{N_{\text{cnt}}} |\pi^{\bf synth}_i - \pi_i|}{N_{\vec{pi}}}
    \end{equation}
    where $N_{\vec{\pi}}$ corresponds to the size of the frequency list $\vec{\pi}$.
    \item Root Mean Square Error:
    \begin{equation}
        \text{RMSE}\left(\vec{\pi}^{\bf synth}, \vec{\pi}\right) = \left(\dfrac{\sum_{i=1}^{N_{\text{cnt}}} \left(\pi^{\bf synth}_i - \pi_i\right)^2}{N_{\text{cnt}}}\right)^{1/2}
    \end{equation}
    \item Standardized Root Mean Square Error~\citep{muller_hierarchical_2011}:
    \begin{equation}
        \label{eq:SRMSE}
        \text{SRMSE}\left(\vec{\pi}^{\bf synth}, \vec{\pi}\right) = \dfrac{\text{RMSE}\left(\vec{\pi^{\bf synth}}, \vec{\pi}\right)}{\vec{\overline{\pi}}}
    \end{equation}
    where $\vec{\overline{\pi}}$ corresponds to the average value of $\vec{\pi}$.
    \item Coefficient of determination:
    \begin{equation}
        R^2\left(\vec{\pi^{\bf synth}}, \vec{\pi}\right) = 1 - \dfrac{\sum_{i=1}^{N_{\text{cnt}}} \left(\pi^{\bf synth}_i - \pi_i\right)^2}{\sum_{i=1}^{N_{\text{cnt}}} \left(\overline{\pi}_i - \pi_i\right)^2}
    \end{equation}
    \item Pearson's correlation:
    \begin{equation}
        \rho_{\text{Pearson}}\left(\vec{\pi}^{\bf synth}, \vec{\pi}\right) = \dfrac{\text{cov}\left(\vec{\pi}^{\bf synth}, \vec{\pi}\right)}{\sigma_{\vec{\pi}}\sigma_{\vec{\pi^{\bf synth}}}}
    \end{equation}
    where $\text{cov}$ corresponds to the covariance matrix and $\sigma_{\vec{X}}$ the standard deviation of a given vector $\vec{X}$.
\end{itemize}

Finally, the results can be averaged over all combinations to obtain a single number per synthetic dataset, statistic, and aggregation level.

\subsubsection{Supervised learning-based validation}
\label{sec:res_ml}

We propose in this paper a new supervised learning-based validation method for synthetic data which makes use of supervised classification and regression models to approximate the full conditional distributions of each variable, given all other variables in the dataset. The general approach is to estimate two regression or classification models for each variable $\vec{v}_t$. The first model ($m_{t}$) is estimated on a training portion of the original data, and the second ($m^{\bf synth}_t$) is estimated on a corresponding training portion of the synthetic data. In each case, the model tries to predict the values in the corresponding column conditional on all other columns in the dataset.

Each model is then validated on the same test portion of the original data, providing two loss scores. The expected value of the loss of the model estimated on the synthetic data (which approximates the conditionals in the synthetic dataset) should be greater than or equal to the expected loss of the model estimated on the original data (which approximates the true conditionals in the original data). The closer the loss scores of the models estimated on the synthetic and original data, the more closely the synthetic data has captured the conditional distributions of the original data. The approach is detailed in Algorithm~\ref{algo:ml_efficiency}. 

\begin{algorithm*}[h!]
\caption{Supervised learning-based validation}\label{algo:ml_efficiency}
\begin{algorithmic}[1]
\Require{Original data $\bf T$, synthetic data~$\bf T_{synth}$}
\Ensure{Similarity score for each variable $\vec{v}_t \in {\bf T}$}
\ForAll {$\vec{v}_t \in \bf{T}$}
    \State $\vec{y}_t = \vec{v}_t$
    \State $\vec{X}_t = {\bf T} \setminus \vec{v}_t $
    \State Divide $\vec{y}_t$ and $\vec{X}_t$ into training set $(\vec{y}_{t,\text{train}}, \vec{X}_{t,\text{train}})$ and test set $(\vec{y}_{t,\text{test}}, \vec{X}_{t,\text{test}})$
    \State $\vec{y}^{\bf synth}_t = \vec{v}^{\bf synth}_t$
    \State $\vec{X}^{\bf synth}_t = {\bf T_{synth}} \setminus \vec{v}^{\bf synth}_t $
    \State Sample training set $(\vec{y}^{\bf synth}_{t,\text{train}}$, $\vec{X}^{\bf synth}_{t,\text{train}})$ from $\vec{y}^{\bf synth}_t$ and $\vec{X}^{\bf synth}_t$, with the same dimensions as $(\vec{y}_{t,\text{train}}, \vec{X}_{t,\text{train}})$. 
    \If {$\vec{v}_t \in \vec{c}_{1:N_C}$}
        \State  Estimate regression model $m_{t,\text{reg}}$ on $(\vec{y}_{t,\text{train}},\vec{X}_{t,\text{train}})$
        \State Estimate regression model $m^{\bf synth}_{t,\text{reg}}$ on $(\vec{y}^{\bf synth}_{t,\text{train}},\vec{X}^{\bf synth}_{t,\text{train}})$ 
        %\Comment{$f^{\text{reg}}_{t}$ is an estimate of the full conditional of $\vec{v}^\text{synth}_t|T^\text{synth} - \vec{v}^\text{synth}_t$}
        \State $g^\text{reg}_t = \mathcal{L}_\text{MSE}(\vec{y}_{t,\text{test}}, m^{\bf synth}_{t,\text{reg}}(\vec{X}_{t,\text{test}}))/\mathcal{L}_\text{MSE}(\vec{y}^{\bf synth}_{t,\text{test}}, m_{t,\text{reg}}(\vec{X}^{\bf synth}_{t,\text{test}}))$ 
        %\Comment{$\mathcal{L}_\text{MSE}$ is Mean-squared error}
        \State \textbf{return} $g^\text{reg}_t$
    \Else
        \State Estimate probabilistic classification model $m_{t,\text{class}}$ on $(\vec{y}_{t,\text{train}},\vec{X}_{t,\text{train}})$
        \State Estimate probabilistic classification model $m^{\bf synth}_{t,\text{class}}$ on $(\vec{y}^{\bf synth}_{t,\text{train}},\vec{X}^{\bf synth}_{t,\text{train}})$ 
        %\Comment{$f^{\text{reg}}_{t}$ is an estimate of the full conditional of $\vec{v}^\text{synth}_t|T^\text{synth} - \vec{v}^\text{synth}_t$}
        \State $g^\text{class}_t = \mathcal{L}_\text{log-loss}(\vec{y}_{t,\text{test}}, m^{\bf synth}_{t,\text{class}}(\vec{X}_{t,\text{test}}))-\mathcal{L}_\text{log-loss}(\vec{y}^{\bf synth}_{t,\text{test}}, m_{t,\text{class}}(\vec{X}^{\bf synth}_{t,\text{test}}))$ 
        %\Comment{$\mathcal{L}_\text{MSE}$ is Mean-squared error}
        \State \textbf{return} $g^\text{class}_t$
    \EndIf
\EndFor
\end{algorithmic}
\end{algorithm*}

We make use of gradient boosting ensembles of decision trees for both $m_{\text{reg}, t}$ and $m_{\text{class}, t}$ as \begin{inlinelist}
    \item they can be easily applied to both regression and probabilistic classification problems;
    \item are computationally efficient to estimate;
    \item have been shown to have high predictive performance on a wide variety of supervised learning tasks;
    \item can determine appropriate regularisation automatically using early stopping
\end{inlinelist}. We specifically make use of the LightGBM library~\citep{ke_lightgbm_2017} which inherently handles categorical input features, thus avoiding the need for one-hot encoding of categorical variables. 

For continuous variables, the score $g^\text{reg}_t$ is the ratio of the mean-squared error of the model estimated on the synthetic dataset to the mean-squared error of the model estimated on the original dataset, with a score of 1 indicating a perfect match, and a higher score representing a worse fit. For categorical variables, the score $g^\text{class}_t$ is the absolute difference between the normalized log-loss of the model estimated on the synthetic dataset and the normalized log-loss of the model estimated on the original dataset, with a score of 0 indicating a perfect match, and a higher score representing a worse fit. The scores can be summed over all columns to give aggregate scores for all continuous and categorical variables, respectively. 

While Algorithm~\ref{algo:ml_efficiency} describes a single train-test split, the same algorithm can be used with $k$-fold cross-validation to obtain more accurate estimates of the model losses. We use 5-fold cross-validation and stratified sampling to select training folds for the categorical data. 

\subsection{Implementation notes}
\label{sec:implementation}

The code for the DATGAN has been implemented using Python 3.7.9. We use the libraries \texttt{tensorflow} (v1.15.5) \citep{abadi_tensorflow_2016} and \texttt{tensorpack} (v0.9.4) \citep{yuxin_tensorpack_2016} for the main components of the neural networks. In addition, we use the library \texttt{networkx} (v2.5) \citep{hagberg_exploring_2008} for specifying the DAG $\mathcal{G}$ discussed in Section~\ref{sec:DAG}. This library already has built-in functions to verify that a user-specified graph is a directed acyclic graph. 

For the optimization process using the different loss functions presented in Section~\ref{sec:loss_function}, we follow the instructions of the authors of the different articles for the hyperparameters. During initial tests, we investigated different values of the learning rate and decided on the following learning rates for each loss, which appeared to work best in these initial tests:
\begin{description}
    \item[Standard loss] learning rate of $1\cdot 10^{-3}$
    \item[Wasserstein loss] learning rate of $2\cdot 10^{-4}$
    \item[Wasserstein loss with gradient-penalty] learning rate of $1\cdot 10^{-4}$
\end{description}
We do not provide any specific results for this hyperparameter since it is not the main focus of our article. Finally, the sizes of the different components presented in the methodology are directly given in the notation table in the supplementary materials.

The complete code for this project, including the different versions of the DATGAN, the case studies, and the results, can be found on Github: \href{https://github.com/glederrey/SynthPop}{https://github.com/glederrey/SynthPop}. A Python library has been created with the original DATGAN model as well as an updated version written with \texttt{tensorflow} (v2.8.0). The code can be found here: \href{https://github.com/glederrey/DATGAN}{https://github.com/glederrey/DATGAN}.

%% file: tikz/DATGAN.tex
\begin{tikzpicture}[every text node part/.style={align=center}]

\makeatletter
\tikzset{
    database top segment style/.style={draw},
    database middle segment style/.style={draw},
    database bottom segment style/.style={draw},
    database/.style={
        path picture={
            \path [database bottom segment style]
                (-\db@r,-0.5*\db@sh) 
                -- ++(0,-1*\db@sh) 
                arc [start angle=180, end angle=360,
                    x radius=\db@r, y radius=\db@ar*\db@r]
                -- ++(0,1*\db@sh)
                arc [start angle=360, end angle=180,
                    x radius=\db@r, y radius=\db@ar*\db@r];
            \path [database middle segment style]
                (-\db@r,0.5*\db@sh) 
                -- ++(0,-1*\db@sh) 
                arc [start angle=180, end angle=360,
                    x radius=\db@r, y radius=\db@ar*\db@r]
                -- ++(0,1*\db@sh)
                arc [start angle=360, end angle=180,
                    x radius=\db@r, y radius=\db@ar*\db@r];
            \path [database top segment style]
                (-\db@r,1.5*\db@sh) 
                -- ++(0,-1*\db@sh) 
                arc [start angle=180, end angle=360,
                    x radius=\db@r, y radius=\db@ar*\db@r]
                -- ++(0,1*\db@sh)
                arc [start angle=360, end angle=180,
                    x radius=\db@r, y radius=\db@ar*\db@r];
            \path [database top segment style]
                (0, 1.5*\db@sh) circle [x radius=\db@r, y radius=\db@ar*\db@r];
        },
        minimum width=2*\db@r + \pgflinewidth,
        minimum height=3*\db@sh + 2*\db@ar*\db@r + \pgflinewidth,
    },
    database segment height/.store in=\db@sh,
    database radius/.store in=\db@r,
    database aspect ratio/.store in=\db@ar,
    database segment height=0.1cm,
    database radius=0.25cm,
    database aspect ratio=0.35,
    database top segment/.style={
        database top segment style/.append style={#1}},
    database middle segment/.style={
        database middle segment style/.append style={#1}},
    database bottom segment/.style={
        database bottom segment style/.append style={#1}}
}
\makeatother

\makeatletter
\tikzset{ loop/.style={ % requires library shapes.misc
        draw,
        chamfered rectangle,
        chamfered rectangle xsep=2cm
    }
}
\makeatother

\definecolor{lightgray}{rgb}{.9,.9,.9}
\definecolor{lightblue}{rgb}{.62,.73,.83}
\definecolor{lightred}{rgb}{.83, .73, .62}
\definecolor{lightorange}{rgb}{.99,.82,.60}

% Helpers
\def\n{20}
%\draw[help lines,xstep=1,ystep=1, color=black!20] (-\n,-\n) grid (\n,\n);
%\draw[help lines,xstep=.5,ystep=.5, color=black!10, dotted] (-\n,-\n) grid (\n,\n);
%\foreach \x in {-\n,...,\n} { \node [anchor=north, color=black!30] at (\x,0) {\x}; }
%\foreach \y in {-\n,...,\n} { \node [anchor=east, color=black!30] at (0,\y) {\y}; }

\def\dy{3}
\def\dx{4}

\draw[dashed, line width=1, pattern=north west lines, pattern color=lightgray, rounded corners=5pt] (-\dx/2+0.5,-1.45*\dy) -- (3.5*\dx, -1.45*\dy) -- (3.5*\dx, 1.85*\dy) -- (2.5*\dx, 1.85*\dy) -- (2.5*\dx, 0.7*\dy) -- (-\dx/2+0.5, 0.7*\dy) -- cycle;
\node[anchor=south west] at (-\dx/2+0.6,-1.45*\dy) {Standard GAN structure (see Figure~\ref{fig:GAN})};

\node[ellipse, fill=red!30!white, line width=1.5, draw=black, minimum width=2cm, minimum height=1.5cm] (A) at (1.1*\dx,1.3*\dy) {DAG};
\node[above=0cm of A] {\bf Expert\\\bf knowledge};

\node[ellipse, fill=lightblue, line width=1.5, draw=black, minimum width=2cm, minimum height=1.5cm] (B) at (0,0) {Noise};

\node[rounded corners=10pt, fill=lightgray, line width=1.5, draw=black, minimum width=3cm, minimum height=2cm] (C) at (1.1*\dx,0) {Generator};

\node[rounded corners=10pt, fill=lightgray, line width=1.5, draw=black, minimum width=3cm, minimum height=2cm] (D) at (3*\dx,0) {Discriminator};

\node[loop, fill=lightorange, line width=1.5, draw=black, minimum width=3cm, minimum height=1.5cm] (E) at (3*\dx,2.3*\dy) {Encoding};

\node[database,label=below:Original\\dataset,database radius=1cm,database segment height=0.4cm, line width=1.5] (F) at (2.05*\dx,2.3*\dy) {};
\node at (2.05*\dx,2.3*\dy+0.6) {$\bf T$};

\node[loop, fill=lightorange, line width=1.5, draw=black, minimum width=3cm, minimum height=1.5cm] (G) at (2.05*\dx,-2.1*\dy) {Sampling};

\node[database,label=above:Synthetic\\dataset,database radius=1cm,database segment height=0.4cm, line width=1.5]  (H) at (3*\dx,-2.1*\dy) {};
\node at (3*\dx,-2.1*\dy+0.6) {$\bf T_{synth}$};

\node[database,database radius=1cm,database segment height=0.4cm, line width=1.5, database bottom segment={fill=white}, database middle segment={fill=white}, database top segment={fill=white}]  (I) at (2.05*\dx,-\dy) {};
\node at (2.05*\dx,-\dy+0.6) {\small $\bf\widehat{T}_{synth}$};

\node[database,database radius=1cm,database segment height=0.4cm, line width=1.5, database bottom segment={fill=white}, database middle segment={fill=white}, database top segment={fill=white}]  (J) at (3*\dx,1.3*\dy) {};
\node at (3*\dx,1.3*\dy+0.6) {\small $\bf\widehat{T}$};

\draw[-{Triangle[scale=1.5]}, line width=1] (C) |- (I);
\draw[-{Triangle[scale=1.5]}, line width=1] (I) -| (D);
\draw[-{Triangle[scale=1.5]}, line width=1] (I) -- (G);

\node[fill=white, draw=black, rounded corners=5pt, line width=1] at (1.1*\dx,-0.7*\dy) {output};
\node[fill=white, draw=black, rounded corners=5pt, line width=1] at (3*\dx,-0.7*\dy) {input};

\draw[-{Triangle[scale=1.5]}, line width=1] (E) --(J);
\draw[-{Triangle[scale=1.5]}, line width=1] (J) -- node[fill=white, above=-0.15cm, draw=black, rounded corners=5pt] {input} (D);

\draw[-{Triangle[scale=1.5]}, line width=1] (G) -- (H);
\draw[-{Triangle[scale=1.5]}, line width=1] (F) --(E);

\draw[-{Triangle[scale=1.5]}, line width=1] (A)  --  node[fill=white, draw=black, rounded corners=5pt] {generator\\structure} (C);
\draw[-{Triangle[scale=1.5]}, line width=1] (B)  -- node[left=-0.35cm, fill=white, draw=black, rounded corners=5pt] {input}(C);

\draw[-{Triangle[scale=1.5]}, line width=1, dashed] (D) -- node[fill=white, draw=black, rounded corners=5pt, solid] {backpropagation}  (C);

\end{tikzpicture}

%% file: tikz/DAG.tex
\begin{tikzpicture}[every text node part/.style={align=center}]

\definecolor{lightgray}{rgb}{.9,.9,.9}

% Helpers
\def\n{20}
%\draw[help lines,xstep=1,ystep=1, color=black!20] (-\n,-\n) grid (\n,\n);
%\draw[help lines,xstep=.5,ystep=.5, color=black!10, dotted] (-\n,-\n) grid (\n,\n);
%\foreach \x in {-\n,...,\n} { \node [anchor=north, color=black!30] at (\x,0) {\x}; }
%\foreach \y in {-\n,...,\n} { \node [anchor=east, color=black!30] at (0,\y) {\y}; }

\def\dy{1.5}
\def\dx{5}
\def\minwidth{2.5cm}
\def\minheight{1.5cm}

\node[rounded corners=10pt, fill=lightgray, line width=1.5, draw=black, minimum width=\minwidth, minimum height=\minheight] (A) at (0,0) {\ttfamily nbr cars\\\ttfamily household};

\node[rounded corners=10pt, fill=lightgray, line width=1.5, draw=black, minimum width=\minwidth, minimum height=\minheight] (B) at (0,-2*\dy) {\ttfamily age};

\node[rounded corners=10pt, fill=lightgray, line width=1.5, draw=black, minimum width=\minwidth, minimum height=\minheight] (C) at (\dx,0) {\ttfamily driving\\\ttfamily license};

\node[rounded corners=10pt, fill=lightgray, line width=1.5, draw=black, minimum width=\minwidth, minimum height=\minheight] (D) at (\dx,-2*\dy) {\ttfamily trip\\\ttfamily purpose};

\node[rounded corners=10pt, fill=lightgray, line width=1.5, draw=black, minimum width=\minwidth, minimum height=\minheight] (E) at (2*\dx,-\dy) {\ttfamily mode\\\ttfamily choice};

\node[rounded corners=10pt, fill=lightgray, line width=1.5, draw=black, minimum width=\minwidth, minimum height=\minheight] (F) at (2*\dx,\dy) {\ttfamily type of\\\ttfamily survey};

\draw[-{Triangle[scale=1.5]}, line width=1] (A) -- (C);

\draw[-{Triangle[scale=1.5]}, line width=1] (B.20) -- (C.200);
\draw[-{Triangle[scale=1.5]}, line width=1] (B) -- (D);

\draw[-{Triangle[scale=1.5]}, line width=1] (C.0) -- (E.170);
\draw[-{Triangle[scale=1.5]}, line width=1] (D.0) -- (E.190);

\end{tikzpicture}

%% file: tikz/generic_LSTM.tex
\begin{tikzpicture}[every text node part/.style={align=center}]

\makeatletter
\tikzset{ loop/.style={ % requires library shapes.misc
        draw,
        chamfered rectangle,
        chamfered rectangle xsep=2cm
    }
}
\makeatother

\definecolor{lightgray}{rgb}{.9,.9,.9}
\definecolor{othergray}{rgb}{.7,.7,.7}
\definecolor{lightblue}{rgb}{.62,.73,.83}
\definecolor{lightorange}{rgb}{.99,.82,.60}

% Helpers
%\def\n{20}
%\draw[help lines,xstep=1,ystep=1, color=black!20] (-\n,-\n) grid (\n,\n);
%\draw[help lines,xstep=.5,ystep=.5, color=black!10, dotted] (-\n,-\n) grid (\n,\n);
%\foreach \x in {-\n,...,\n} { \node [anchor=north, color=black!30] at (\x,0) {\x}; }
%\foreach \y in {-\n,...,\n} { \node [anchor=east, color=black!30] at (0,\y) {\y}; }

\def\dy{2}
\def\dx{1.666666666666}

\node[rectangle, rounded corners=1cm, fill=lightgray, line width=1.5, draw=black, minimum width=9.5cm, minimum height=6cm] (A) at (0,0) {};

\node[rectangle, rounded corners=0.5cm, fill=gray, line width=1.5, draw=darkgray, minimum width=9.5cm, minimum height=1cm, opacity=0.5] (M) at (0,\dy) {};

\node[rectangle, rounded corners=0.5cm, fill=gray, line width=1.5, draw=darkgray, minimum width=1cm, minimum height=5cm, opacity=0.5] (N) at (-5+\dx,0) {};

\node[rectangle, rounded corners=0.5cm, fill=gray, line width=1.5, draw=darkgray, minimum width=3cm, minimum height=3cm, opacity=0.5] (O) at (-5+2.5*\dx,- \dy/2) {};

\node[rectangle, rounded corners=0.5cm, fill=gray, line width=1.5, draw=darkgray, minimum width=3cm, minimum height=4.25cm, opacity=0.5] (P) at (-5+4.5*\dx,- \dy/2+0.75-0.125) {};

\node[loop, fill=lightblue, draw=black, line width=1, minimum width=1.5cm] (B) at (-6.2, \dy) {$\vec{C}_{t-1}$};

\node[loop, fill=lightblue, draw=black, line width=1, minimum width=1.5cm] (C) at (6.2, \dy) {$\vec{C}_{t}$};
\draw[-{Triangle[scale=1.5]}, line width=1, dashed] (B) -- (C);

\node[loop, fill=lightblue, draw=black, line width=1, minimum width=1.5cm] (D) at (-6.2, -\dy) {$\vec{i}_{t}$};

\node[circle, fill=lightorange, draw=black, line width=1] (E) at (-6.2,-1.75*\dy) {$\oplus$};
\draw[-{Triangle[scale=1.5]}, line width=1] (E) -- (D);

\node[loop, fill=lightblue, draw=black, line width=1, minimum width=1.5cm] (F) at (-8, -1.75*\dy) {$\vec{h}_{t-1}$};
\draw[-{Triangle[scale=1.5]}, line width=1] (F) -- (E);

\node[loop, fill=lightblue, draw=black, line width=1, minimum width=1.5cm] (G) at (-4.4, -1.75*\dy) {$\vec{x}_{t}$};
\draw[-{Triangle[scale=1.5]}, line width=1] (G) -- (E);

\node[loop, fill=lightblue, draw=black, line width=1, minimum width=1.5cm] (H) at (6.2,- \dy) {$\vec{h}_{t}$};

\node[circle, fill=lightorange, draw=black, line width=1] (I) at (-5+\dx,\dy) {$\times$};
\draw[-{Triangle[scale=1.5]}, line width=1] (D) -| (I);

\node[circle, fill=lightorange, draw=black, line width=1] (J) at (5-3*\dx,0) {$\times$};
\draw[-{Triangle[scale=1.5]}, line width=1] (D) -| (J);

\node[circle, fill=lightorange, draw=black, line width=1] (K) at (5-3*\dx,\dy) {$+$};
\draw[-{Triangle[scale=1.5]}, line width=1] (J) -| (K);

\draw[-{Triangle[scale=1.5]}, line width=1] (D) -- ++(right:4.533333333cm) |- (J);

\node[circle, fill=lightorange, draw=black, line width=1] (L) at (5-\dx,0) {$\times$};
\draw[-{Triangle[scale=1.5]}, line width=1] (L) |- (H);

\draw[-{Triangle[scale=1.5]}, line width=1] (D) -- ++(right:7.866666666cm) |- (L);

\draw[-{Triangle[scale=1.5]}, line width=1, dashed] (K)++(right:2*\dx) -- (L);

\node[rectangle, fill=red!30!white,minimum width=0.7cm, minimum height=0.5cm, draw=black] at (-5+\dx, -1.33333333333) {$\sigma$};
\node[rectangle, fill=red!30!white,minimum width=0.7cm, minimum height=0.5cm, draw=black] at (-5+2*\dx, -1.33333333333) {$\sigma$};
\node[rectangle, fill=red!30!white,minimum width=0.7cm, minimum height=0.5cm, draw=black] at (-5+3*\dx, -1.33333333333) {$\tanh$};
\node[rectangle, fill=red!30!white,minimum width=0.7cm, minimum height=0.5cm, draw=black] at (-5+4*\dx, -1.33333333333) {$\sigma$};

\node[rectangle, fill=red!30!white,minimum width=0.7cm, minimum height=0.5cm, draw=black] at (-5+5*\dx, 1.33333333333) {$\tanh$};

\node[above=-0.05cm of M] {cell state};
\node[below=-0.05cm of N] {forget gate};
\node[below=-0.05cm of O] {input gate};
\node[below=-0.05cm of P] {output gate};

\end{tikzpicture}

%% file: tikz/DATGAN_LSTM.tex
\begin{tikzpicture}[every text node part/.style={align=center}]

\makeatletter
\tikzset{ loop/.style={ % requires library shapes.misc
        draw,
        chamfered rectangle,
        chamfered rectangle xsep=2cm
    }
}
\makeatother

\definecolor{lightgray}{rgb}{.9,.9,.9}
\definecolor{lightblue}{rgb}{.62,.73,.83}
\definecolor{lightorange}{rgb}{.99,.82,.60}

% Helpers
\def\n{20}
%\draw[help lines,xstep=1,ystep=1, color=black!20] (-\n,-\n) grid (\n,\n);
%\draw[help lines,xstep=.5,ystep=.5, color=black!10, dotted] (-\n,-\n) grid (\n,\n);
%\foreach \x in {-\n,...,\n} { \node [anchor=north, color=black!30] at (\x,0) {\x}; }
%\foreach \y in {-\n,...,\n} { \node [anchor=east, color=black!30] at (0,\y) {\y}; }

\def\dy{2.2}
\def\dx{4}

\node[rounded corners=10pt, fill=lightgray, line width=1.5, draw=black, minimum width=3cm, minimum height=2cm] (A) at (0,0) {\large $\mathbf{LSTM_t}$};

\node[loop, fill=lightblue, draw=black, line width=1, minimum width=2cm] (B) at (-\dx, 0) {$\vec{C}_{t-1}$};
\draw[-{Triangle[scale=1.5]}, line width=1, dashed] (B) -- (A);

\node[loop, fill=lightblue, draw=black, line width=1, minimum width=2cm] (P) at (0,\dy) {$\vec{i}_{t}$};
\draw[-{Triangle[scale=1.5]}, line width=1] (P) -- (A);

\node[circle, fill=lightorange, draw=black, line width=1] (C) at (0, 1.8*\dy) {$\oplus$};
\draw[-{Triangle[scale=1.5]}, line width=1] (C) -- (P);

\node[loop, fill=lightblue, draw=black, line width=1, minimum width=2cm] (D) at (\dx,0) {$\vec{C}_{t}$};
\draw[-{Triangle[scale=1.5]}, line width=1, dashed] (A) -- (D);

\node[loop, fill=lightblue, draw=black, line width=1, minimum width=2cm] (E) at (0,-\dy) {$\vec{h}_{t}$};
\draw[-{Triangle[scale=1.5]}, line width=1] (A) -- (E);

\node[loop, fill=lightblue, draw=black, line width=1, minimum width=2cm] (F) at (-\dx/2,2.8*\dy-1) {$\vec{a}_{t}$};
\draw[-{Triangle[scale=1.5]}, line width=1] (F) |- (C);
\node[loop, fill=lightblue, draw=black, line width=1, minimum width=2cm] (G) at (0,2.8*\dy) {$\vec{f}_{t-1}$};
\draw[-{Triangle[scale=1.5]}, line width=1] (G) -- (C);
\node[loop, fill=lightblue, draw=black, line width=1, minimum width=2cm] (H) at (\dx/2,2.8*\dy-1) {$\vec{z}_{t}$};
\draw[-{Triangle[scale=1.5]}, line width=1] (H) |- (C);

\node[rounded corners=10pt, fill=lightgray, line width=1.5, draw=black, minimum width=3cm, minimum height=2cm] (I) at (0,-2*\dy) {output\\ transformer};
\draw[-{Triangle[scale=1.5]}, line width=1] (E) -- (I);

\node[rounded corners=10pt, fill=lightgray, line width=1.5, draw=black, minimum width=3cm, minimum height=2cm] (J) at (0,-4*\dy) {input\\ transformer};

\node[loop, fill=lightblue, draw=black, line width=1, minimum width=2cm] (K) at (0,-5*\dy) {$\vec{f}_{t}$};
\draw[-{Triangle[scale=1.5]}, line width=1] (J) -- (K);

\node[loop, fill=lightblue, draw=black, line width=1, minimum width=2cm] (L) at (0,-3*\dy) {$\widehat{\vec{v}}_t^{\bf synth}$};
\draw[-{Triangle[scale=1.5]}, line width=1] (I) -- (L);
\draw[-{Triangle[scale=1.5]}, line width=1] (L) -- (J);

\draw[dashed, line width=1, pattern=north west lines, pattern color=lightgray, rounded corners=5pt] (1.5*\dx, 2.7*\dy) -- (1.5*\dx, -\dy) -- (0.5*\dx, -\dy)  -- (0.5*\dx, -5.4*\dy) -- (2.5*\dx, -5.4*\dy) -- (2.5*\dx, 2.7*\dy) -- cycle;

\node[loop, fill=lightorange, line width=1.5, draw=black, minimum width=3cm, minimum height=1.5cm] (M) at (2*\dx,-4*\dy) {Sampling};

\node[loop, fill=lightblue, draw=black, line width=1, minimum width=2cm]  (N) at (2*\dx, -5*\dy) {$\vec{v}_t^{\bf synth}$};

\draw[-{Triangle[scale=1.5]}, line width=1] (M) -- (N);

\node[rounded corners=10pt, fill=lightgray, line width=1.5, draw=black, minimum width=3cm, minimum height=2cm] (O) at (2*\dx,-2*\dy) {Discriminator};

\node[loop, fill=lightorange, line width=1.5, draw=black, minimum width=3cm, minimum height=1.5cm] (P) at (2*\dx,-0.7*\dy) {Label smoothing};

\node[loop, fill=lightblue, draw=black, line width=1, minimum width=2cm]  (Q) at (2*\dx, 0.3*\dy) {$\widehat{\vec{v}}_t$};
\draw[-{Triangle[scale=1.5]}, line width=1] (P) -- (O);
\draw[-{Triangle[scale=1.5]}, line width=1] (Q) -- (P);

\node[loop, fill=lightorange, line width=1.5, draw=black, minimum width=3cm, minimum height=1.5cm] (R) at (1*\dx,-2*\dy) {Label smoothing};
\draw[-{Triangle[scale=1.5]}, line width=1] (R) -- (O);

\draw[-{Triangle[scale=1.5]}, line width=1] (L) -| (M);
\draw[-{Triangle[scale=1.5]}, line width=1] (L) -| (R);

\node[loop, fill=lightorange, line width=1.5, draw=black, minimum width=3cm, minimum height=1.5cm] (S) at (2*\dx,1.3*\dy) {Encoding};

\node[loop, fill=lightblue, draw=black, line width=1, minimum width=2cm]  (T) at (2*\dx, 2.8*\dy-1) {$\vec{v}_t$};
\draw[-{Triangle[scale=1.5]}, line width=1] (S) -- (Q);
\draw[-{Triangle[scale=1.5]}, line width=1] (T) -- (S);

\node[anchor=south west] at (0.5*\dx, -5.4*\dy) {Outside of the generator};

\end{tikzpicture}

%% file: text/04_case_study.tex
\section{Case studies}
\label{sec:case_study}

In this section, we present the case study for this article. First, we introduce the datasets in Section~\ref{sec:datasets}. For each dataset, we provide a short description of the dataset, a summary of the variables, and the DAG used in the DATGAN. Then, in Section~\ref{sec:result_assessments}, we present the new methods used to assess the quality of the synthetic datasets generated by all the models presented in this article. Finally, Section~\ref{sec:training} gives a detailed overview of the training method used with all the models in order to bring as much fairness as possible in the comparison of the models.

\subsection{Datasets}
\label{sec:datasets}

The first dataset is a travel survey made by the Chicago Metropolitan Agency for Planning. We named this dataset CMAP. The original dataset is a household travel survey of the Chicago metropolitan area, conducted from January 2007 to February 2008. The trips are given as one and two-day travel diaries, provided by all the members of the households. The data is therefore hierarchical. The dataset has first been cleaned to remove incomplete entries. Then, we selected one unique trip per individual per household for the final dataset to remove data leakage. It thus contains a total of 8'929 trips with 15 columns. A complete description of this dataset is given in Table~\ref{tab:data_Chicago} in the appendix. The DAG used for the DATGAN with this dataset can also be found in the appendix; see Figure~\ref{fig:DAG_Chicago}.

The second dataset is the London Passenger Mode Choice (LPMC) dataset~\citep{hillel_recreating_2018}. It combines the London Travel Demand Survey (LTDS) records with matched trip trajectories and corresponding mode alternatives. The LTDS has been conducted between April 2012 and March 2015 and records trips made by individuals residing within Greater London. The trip trajectories are extrapolated from Google Maps API. The final dataset has been processed to not lead to data leakage. Similar to the CMAP dataset, we selected only one trip per household. The final dataset contains a total of 17'616 trips with 27 columns. A complete description of this dataset is given in Table~\ref{tab:data_LPMC} in the appendix. The DAG used for the DATGAN with this dataset can also be found in the appendix; see Figure~\ref{fig:DAG_LPMC}. In addition, we created a smaller version of the LPMC dataset by randomly selecting 50\% of the rows for testing the DATGAN versions. This dataset is conveniently named LPMC\_half. We mainly use it to understand the effect of the number of rows on the performance of the models. 

The third and final dataset is the ADULT dataset~\citep{kohavi_scaling_1996}, also known as the Census-Income dataset. This dataset contains socio-economic variables on multiple individuals to predict if their income is below or above \$50k/yr. From the original dataset, we removed all the rows with unknown values. A complete description of this dataset is given in Table~\ref{tab:data_adult} in the appendix. The DAG used for the DATGAN with this dataset can also be found in the appendix; see Figure~\ref{fig:DAG_adult}. Due to its larger size compared to the other datasets, the ADULT dataset is only used when comparing the DATGAN with state-of-the-art models. 

\begin{table}[H]
    \centering
    \caption{Summary of the datasets used in the case studies. Full description of the datasets can be found in the Appendix.}
    \label{tab:summary_dataset}
    \begin{tabularx}{0.7\textwidth}{l||C|C|C||C}
        \textbf{Name} & \boldmath$\#$\textbf{columns} & \boldmath$\#$\textbf{continuous} & \boldmath$\#$\textbf{categorical} & \boldmath$\#$\textbf{rows} \\ \midrule[1.5pt]
        CMAP & 15 & 3 & 12 & 8'929 \\
        LPMC & 27 & 13 & 14 & 17'616 \\
        LPMC\_half & 27 & 13 & 14 & 8'808 \\
        ADULT & 14 & 4 & 10 & 45'222
    \end{tabularx}
\end{table}

\subsection{Training process}
\label{sec:training}

This article aims to propose a new way to generate synthetic data. However, for fairness towards state-of-the-art methods in the literature, we need to test all these generative models on a similar playground. We, thus, decide to train every model on 1'000 epochs with a batch size $N_b$ of 500, even if the optimization process could be stopped earlier. In addition, we decided to keep the original hyperparameters provided in the articles. While optimizing these parameters would most likely lead to better results, most users would use the models as the authors provide them. In addition, each model is trained five times on each dataset, and each of the five models generates five synthetic datasets with the same number of rows as the original dataset. This means that each test is performed on a total of 25 synthetic datasets. Thus, the results provided in Section~\ref{sec:results} correspond to the average value of the performed tests. 

The training process is slightly different for the DATGAN versions (summary given in the supplementary materials). Indeed, the sampling can be done independently after the training process. Therefore, we only have to train a total of nine different models (combinations of loss functions and label smoothing). Each of these models is trained five times. Then, for each of these 45 models, we use the four different sampling methods to produce five synthetic datasets for each method. Therefore, we get a total of 900 synthetic datasets to be compared, \emph{i.e.} 25 synthetic datasets for each model presented in the summary table in the supplementary materials for each case study presented in Section~\ref{sec:datasets}.

%% file: text/05_results.tex
\section{Results}
\label{sec:results}

In this section, we present the results obtained using the assessment methods presented in Section~\ref{sec:result_assessments} on the different case studies presented in Section~\ref{sec:case_study}. The first section, see Section~\ref{sec:comp_lit}, compares the DATGAN against state-of-the-art models found in the literature. The second section, see Section~\ref{sec:analysis_DAG}, performs a sensitivity analysis on the DAG used for the DATGAN in order to understand its effect on the performance of the model. 

In the supplementary materials, we analyze the results of the possible DATGAN versions. We provide here the conclusions of this analysis. The reader may refer to this document for more information. Across all case studies, the DATGAN performs better when using the two-sided label smoothing and the simulation sampling strategy for both continuous and categorical variables. However, we have found that using the \texttt{WGAN} loss function leads to better results on datasets with fewer variables, such as the CMAP dataset. On the other hand, the \texttt{WGGP} loss performs better on the LPMC dataset. Since the comparison of DATGAN version was not performed on the ADULT dataset (due to its larger size compared to the other case studies), we will test both the \texttt{WGAN} and the \texttt{WGGP} loss functions for this case study. Results on the other case studies indicate that the \texttt{WGAN} loss should outperform the \texttt{WGGP} loss.

\subsection{Comparison with state-of-the-art models}
\label{sec:comp_lit}

At this point of the article, we have presented our new model for generating synthetic datasets and have selected the best version depending on the type of case study. In addition, we have studied the effect of the DAG on the generated synthetic data. However, we now want to compare our model against state-of-the-art models presented in the literature. We, thus, test our DATGAN model against the four models presented in the literature (see Section~\ref{sec:sota_models}) on the four different case studies. We use the same assessment methods as previously to compare all the models. We have compiled all the results in Table~\ref{tab:summary_all_models}. It shows the average rankings on both assessment methods for the four different case studies. We see that the DATGAN model outperforms all the other models in the first three case studies. For example, it is consistently the best model when using the Machine Learning efficacy method. For the ADULT case study, we decided to test the DATGAN with the \texttt{WGAN} and the \texttt{WGGP} loss functions. Since the ADULT dataset contains more categorical variables than continuous, we expect the \texttt{WGAN} loss to perform better, as it is shown in Table~\ref{tab:summary_all_models}. It performs the best on the statistical assessments. However, it seems to struggle with the Machine Learning efficacy method. While the results are quite close on the continuous variables, the two DATGAN models are the only models that do not fail the test on the categorical variables. Indeed, it seems that the other models tend not to produce enough of the low probability categories. Thus, these models tend to oversimplify the generated synthetic data compared to the DATGAN models. One of the reasons why such a thing happens might come from the sampling process. Indeed, using simulation to get the final categories allows for more representation of the low probability values than using the maximum probability estimator. 

\begin{table}[h!]
	\centering
	\caption{Average rankings of the state-of-the-art models against the DATGAN on the four case studies.}
	\label{tab:summary_all_models}
	\input{tables/summary_all}
\end{table}

Since we compared models across multiple articles, it is interesting to look at their conclusions. For example, \cite{xu_modeling_2019} show that their CTGAN model is consistently outperforming the TVAE model. We see the same ranking between the two models except for the ADULT case study. Therefore, we can draw the same conclusion on these two models as the authors. However, \cite{zhao_ctab-gan_2021} claim that the CTAB-GAN outperforms the CTGAN across all their assessments. In our case, we see that the CTAB-GAN outperforms the CTGAN only when the case studies contain a small number of data points. While we have used both models as intended to be used by their authors, we only changed the final number of epochs. In their article, \cite{zhao_ctab-gan_2021} have trained both models on 150 epochs. Therefore, the CTAB-GAN may be providing a better early optimization process than the CTGAN. However, further work would be required to analyze this result, and it is out of the scope of this article. 

In addition, we have tested both models presented by \cite{garrido_prediction_2019}: a WGAN and a VAE. Unfortunately, results are not shown in the article because both models failed to produce adequate continuous variables. Indeed, the encoding of continuous variables is done such that they are binned based on their original distributions, thus treating them as categorical variables. This, therefore, lead to especially poor results when comparing continuous distributions. 

\subsection{Sensitivity analysis of the DAG}
\label{sec:analysis_DAG}

In this section, we want to analyze how the DAG can affect the performances of the DATGAN and how it can be used to modify the generation of synthetic datasets. Section~\ref{sec:structure_DAG} performs the analysis on the DAG using different versions of the latter, and Section~\ref{sec:effect_DAG} explains how we can alter the DAG to generate hypothetical synthetic datasets.

\subsubsection{Structure of the DAG}
\label{sec:structure_DAG}

In the previous section, we have analyzed all the different versions of the DATGAN to select the best one. However, in this section, we want to investigate how the DAG will influence the results. Thus, we only use the best possible model for each case study. The idea is to start with the DAG presented in the appendix for each case study and make variations of it to study the generated datasets. Therefore, we created five different DAGs for each case study:
\begin{itemize}
    \item \texttt{full}: Complete DAG presented in the appendix for each case study.
    \item \texttt{trans. red.}: Transitive reduction of the \texttt{full} DAG. The transitive reduction consists in removing as many edges as possible in a DAG such that there exists only one path between two vertices in the graph.
    \item \texttt{linear}: This DAG consists in taking the variables in the order provided by the dataset and linking them to each other linearly. Thus, there are no multi-inputs within the DAG. This is similar to the technique used in the TGAN~\citep{xu_synthesizing_2018}.
    \item \texttt{prediction}: This DAG consists of only one sink node. All the other nodes are considered source nodes linked to the sink node. The source nodes are not linked to each other. The sink nodes are the \texttt{choice} for the CMAP case study and the \texttt{travel\_mode} for the LPMC case studies.
    \item \texttt{no links}: This DAG consists of only nodes without any edges. This, thus, cuts all the links between the variables.
\end{itemize}
At first glance, we can already predict that the last two DAGs should all perform badly since the connections between them are either badly implemented or absent. However, we are mostly interested in the first three DAGs. Indeed, if the DAG can help the model to generate more representative synthetic data, the \texttt{full} DAG or the \texttt{trans. red.} DAG should outperform the \texttt{linear} DAG. As for the previous section, the details of the results are given in the supplementary materials. Table~\ref{tab:summary_Chicago_DAG} shows the rankings of the DAGs on the CMAP case study. As expected, the best two DAGs are the two complete ones. It is interesting to note that the \texttt{full} DAG provides better results on the Machine Learning efficacy assessment while the \texttt{trans. red.} provides better results on the statistical assessments. Since the model with the \texttt{full} DAG contains more edges than the other DAGs, it is, therefore, larger and more complex to train. Thus, this can hurt the LSTM cells' performance when creating the synthetic variables. However, the correlations between the variables are better with the complete DAG since it always performs best compared to the other DAGs. 

\begin{table}[H]
	\centering
	\caption{Average rankings of the different DAGs on the CMAP dataset}
	\label{tab:summary_Chicago_DAG}
	\input{tables/summary_Chicago_DAG}
\end{table}

The CMAP dataset is relatively small. Therefore, learning the correlations between the variables can be pretty difficult. The LPMC dataset, on the other hand, is twice as big. It is thus interesting if the DAG can still provide the same kind of help on a larger dataset. Table~\ref{tab:summary_LPMC_DAG} shows the rankings of the DAG on the LPMC case study. We see that this time, the \texttt{linear} DAG is the best one, closely followed by the two complete DAGs. While the \texttt{full} DAG has some issues with the statistical assessments, it remains the best on the Machine Learning efficacy method. Therefore, it seems that if one wants to generate a synthetic dataset with column data as close as possible to the original dataset, a smaller dataset leads to better results. However, if one wants to keep as much correlation as possible, one should opt for a complete DAG.

\begin{table}[H]
	\centering
	\caption{Average rankings of the different DAGs on the LPMC dataset}
	\label{tab:summary_LPMC_DAG}
	\input{tables/summary_LPMC_DAG}
\end{table}

Finally, we want to confirm our hypothesis that the completeness of the DAG is less important with more data points. We, thus, make the same tests on the smaller LPMC case study. Table~\ref{tab:summary_LPMC_half_DAG} shows the rankings of the DAG on the LPMC\_half case study. Results show that the \texttt{linear} DAG performs better than the \texttt{full} DAG on both metrics. At first glance, this result can be quite surprising. However, when we look at the DAG for both the CMAP and the LPMC case study, we see that the LPMC DAG is much more complex than the CMAP DAG. Indeed, the CMAP DAG contains a total of 25 edges for 15 nodes for an average of $1.\bar{6}$ edges per node. On the other hand, the LPMC DAG contains a total of 63 edges for 27 nodes for an average of $2.\bar{3}$ edges per node. It is, thus, possible that the model struggles to train correctly with a more complex DAG. However, we see that the \texttt{trans. red.} version of the LPMC DAG leads to even worse results than the \texttt{full} DAG. Therefore, it might also be possible that this DAG is not well constructed for this particular case study. It, thus, requires further investigation to fully understand how the DAG affects the results of this case study.

\begin{table}[H]
	\centering
	\caption{Average rankings of the different DAGs on the LPMC\_half dataset}
	\label{tab:summary_LPMC_half_DAG}
	\input{tables/summary_LPMC_half_DAG}
\end{table}

\subsubsection{Effect of the DAG on the synthetic dataset}
\label{sec:effect_DAG}

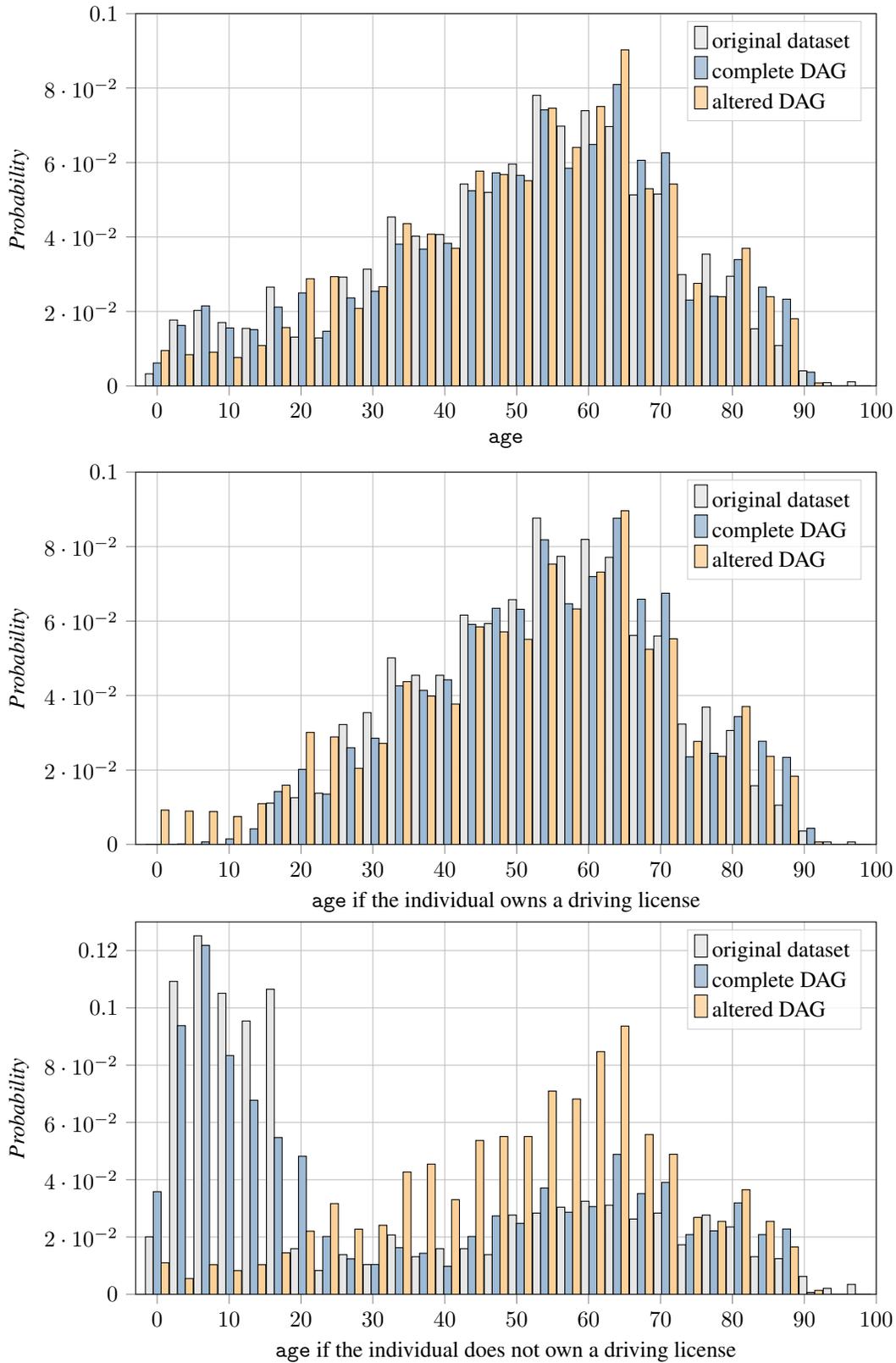
\begin{figure}[tp!]
\begin{subfigure}{\textwidth}
    \centering
    \input{tikz/age}
    %\caption{Age distribution}
    %\label{fig:age}
\end{subfigure}\\
\begin{subfigure}{\textwidth}
    \centering
    \input{tikz/age_with_license}
    %\caption{Age distribution only if the individual owns a driving license}
    %\label{fig:age_with_license}
\end{subfigure}\\
\begin{subfigure}{\textwidth}
    \centering
    \input{tikz/age_without_license}
    %\caption{Age distribution only if the individual does not own a driving license}
    %\label{fig:age_without_license}
\end{subfigure}\\
\caption{Age distributions for the original CMAP dataset, for a synthetic CMAP dataset with a complete DAG, and for a synthetic CMAP dataset with an altered DAG. }
\label{fig:age_distributions}
\end{figure}

The final section of the results shows what can be achieved with the DAG depending on the desire of the modeler. Indeed, we have shown that using a complete DAG allows generating the best possible synthetic datasets compared to state-of-the-art models. However, the DAG can also be used to create hypothetical synthetic datasets. Since the DAG controls the causal links between the variables, removing any relationships between two or more variables is simple. For example, in the CMAP case study, we could imagine a hypothetical population with no minimum age requirement to get a driving license. In order to achieve this, we can simply remove the link between the variables \texttt{age} and \texttt{license} in the DAG presented in Figure~\ref{fig:DAG_Chicago}. If the modeler wants to ensure that two variables do not interact, these variables should be defined as source nodes in the DAG. Figure~\ref{fig:age_distributions} shows the age distributions for the original CMAP dataset, for a synthetic dataset generated with a complete DAG, and for a synthetic dataset generated with the altered DAG. The top figure distribution shows the distribution of the variable \texttt{age} in all of the datasets. As we can see, the synthetic probability distributions are quite similar to the original probability distributions. However, if we look at the age distribution when the individual owns a driving license (middle figure), we see that the synthetic dataset with the altered DAG still produces individuals of age lower than 18 owning a driving license. It is also the case with the synthetic dataset generated from a complete DAG. However, the number of minors with a driving license is marginally less than the data generated with the altered DAG. The effect of the altered DAG can be seen even better when looking at the age distribution of individuals not owning a driving license (bottom figure). Indeed, both the original and synthetic datasets with a complete DAG show most young individuals who do not own a driving license. However, the altered DAG produces the same type of distributions compared to the previous two. This, thus, shows that we eliminated the correlation between age and owning a driving license with the altered DAG. 

Since the DAG can remove the correlations between variables altogether, the modeler must ensure that the source nodes used in the DAG do not have any correlations in the original dataset. This can be done by analyzing the data before creating the DAG. However, this is a small price to pay in order to be able to have more control on the dependencies between the variables.

%% file: tables/summary_all.tex
\begin{tabularx}{0.7\textwidth}{l||C|C||C}
\multicolumn{1}{c||}{\textbf{Name}} & \multicolumn{1}{c|}{\textbf{Avg. rank stats}} & \multicolumn{1}{c||}{\textbf{Avg. rank ML}} & \multicolumn{1}{c}{\textbf{rank}}  \\ \midrule[1.5pt]
	\hline \multicolumn{4}{c}{\cellcolor{Gray!25}\textbf{CMAP case study}} \\ \hline
	DATGAN (\texttt{WGAN}) & \textbf{1.00} & \textbf{1.0} & \textbf{1.00} \\
	CTAB-GAN & 2.92 & 2.5 & 2.71 \\
	TGAN & 2.60 & 3.0 & 2.80 \\
	CTGAN & 4.24 & 4.0 & 4.12 \\
	TVAE & 4.24 & 4.5 & 4.37 \\
	\hline \multicolumn{4}{c}{\cellcolor{Gray!25}\textbf{LPMC case study}} \\ \hline
	DATGAN (\texttt{WGGP}) & \textbf{1.00} & \textbf{1.0} & \textbf{1.00} \\
	TGAN & 2.36 & 2.5 & 2.43 \\
	CTGAN & 3.04 & 3.0 & 3.02 \\
	CTAB-GAN & 4.08 & 3.5 & 3.79 \\
	TVAE & 4.52 & 5.0 & 4.76 \\
	\hline \multicolumn{4}{c}{\cellcolor{Gray!25}\textbf{LPMC\_half case study}} \\ \hline
	DATGAN (\texttt{WGGP}) & \textbf{1.08} & \textbf{1.0} & \textbf{1.04} \\
	TGAN & 2.60 & 2.5 & 2.55 \\
	CTAB-GAN & 3.04 & 4.0 & 3.52 \\
	CTGAN & 3.64 & 3.5 & 3.57 \\
	TVAE & 4.64 & 4.0 & 4.32 \\
	\hline \multicolumn{4}{c}{\cellcolor{Gray!25}\textbf{ADULT case study}} \\ \hline
	DATGAN (\texttt{WGAN}) & \textbf{1.08} & \textbf{3.0} & \textbf{2.04} \\
	TGAN & 1.92 & 3.5 & 2.71 \\
	DATGAN (\texttt{WGGP}) & 3.32 & 3.5 & 3.41 \\
	TVAE & 3.80 & 3.5 & 3.65 \\
	CTGAN & 4.88 & \textbf{3.0} & 3.94 \\
	CTAB-GAN & 6.00 & 4.5 & 5.25 \\
\end{tabularx}

%% file: tables/summary_Chicago_DAG.tex
\begin{tabularx}{0.7\textwidth}{l||C|C||C}
\multicolumn{1}{c||}{\textbf{Name}} & \multicolumn{1}{c|}{\textbf{Avg. rank stats}} & \multicolumn{1}{c||}{\textbf{Avg. rank ML}} & \multicolumn{1}{c}{\textbf{rank}}  \\ \midrule[1.5pt]
	\texttt{trans. red.} & \textbf{1.96} & 2.0 & \textbf{1.98} \\
	\texttt{full} & 3.04 & \textbf{1.0} & 2.02 \\
	\texttt{linear} & 2.80 & 3.0 & 2.90 \\
	\texttt{prediction} & 3.80 & 4.0 & 3.90 \\
	\texttt{no links} & 3.40 & 5.0 & 4.20 \\
\end{tabularx}

%% file: tables/summary_LPMC_DAG.tex
\begin{tabularx}{0.7\textwidth}{l||C|C||C}
\multicolumn{1}{c||}{\textbf{Name}} & \multicolumn{1}{c|}{\textbf{Avg. rank stats}} & \multicolumn{1}{c||}{\textbf{Avg. rank ML}} & \multicolumn{1}{c}{\textbf{rank}}  \\ \midrule[1.5pt]
	\texttt{linear} & \textbf{2.08} & 2.0 & \textbf{2.04} \\
	\texttt{full} & 2.80 & \textbf{1.5} & 2.15 \\
	\texttt{trans. red.} & 2.16 & 2.5 & 2.33 \\
	\texttt{prediction} & 4.00 & 4.0 & 4.00 \\
	\texttt{no links} & 3.96 & 5.0 & 4.48 \\
\end{tabularx}

%% file: tables/summary_LPMC_half_DAG.tex
\begin{tabularx}{0.7\textwidth}{l||C|C||C}
\multicolumn{1}{c||}{\textbf{Name}} & \multicolumn{1}{c|}{\textbf{Avg. rank stats}} & \multicolumn{1}{c||}{\textbf{Avg. rank ML}} & \multicolumn{1}{c}{\textbf{rank}}  \\ \midrule[1.5pt]
	\texttt{linear} & \textbf{1.84} & \textbf{1.0} & \textbf{1.42} \\
	\texttt{full} & 2.04 & 2.0 & 2.02 \\
	\texttt{trans. red.} & 3.76 & 3.0 & 3.38 \\
	\texttt{prediction} & 4.08 & 4.0 & 4.04 \\
	\texttt{no links} & 3.28 & 5.0 & 4.14 \\
\end{tabularx}

%% file: tikz/age.tex
% This file was created with tikzplotlib v0.9.17.
\begin{tikzpicture}

\definecolor{lightgray}{rgb}{.9,.9,.9}
\definecolor{lightblue}{rgb}{.62,.73,.83}
\definecolor{lightorange}{rgb}{.99,.82,.60}

\pgfplotstableread{
0.0 0.0032478441034830344 0.0061597043341919625 0.00951954306193302 
3.3666666666666667 0.017695150632769538 0.01623922051741506 0.008399596819352612 
6.733333333333333 0.020271026990704613 0.02150296785754308 0.009071564564900821 
10.1 0.01702318288722182 0.015567252771867324 0.0076156344495464726 
13.466666666666667 0.015455258147609278 0.015119274274835164 0.010863478553029715 
16.833333333333332 0.026542725949155077 0.02116698398476924 0.015679247396125362 
20.2 0.013103371038190481 0.02497480120954254 0.028782618434315836 
23.566666666666666 0.012879381789674405 0.014671295777803017 0.029342591555606035 
26.933333333333334 0.029230596931347996 0.02363086571844608 0.020831000111995124 
30.3 0.031358494792250724 0.025422779706574694 0.026654720573413115 
33.666666666666664 0.04535782282450551 0.03807817224773302 0.0435659088363769 
37.03333333333333 0.04020607010863572 0.036734236756636535 0.04076604322992591 
40.4 0.0406540486056679 0.0383021614962491 0.03695822600515264 
43.766666666666666 0.05420539814089054 0.052413484152761924 0.057677231492889725 
47.13333333333333 0.05196550565572977 0.05722925299585757 0.05678127449882542 
50.5 0.059581140105276376 0.05655728525030934 0.055101355134954844 
53.86666666666667 0.07806025310785264 0.0741404412588213 0.07458841975585345 
57.233333333333334 0.06977265091275786 0.05846119386269599 0.06406092507559791 
60.6 0.07391645201030528 0.06484488744540418 0.07503639825288566 
63.96666666666667 0.06966065628849982 0.08097211333856169 0.09026766715197887 
67.33333333333333 0.05129353791018154 0.06058909172359872 0.052973457274052116 
70.7 0.05151752715869762 0.06260499496024341 0.05420539814089054 
74.06666666666666 0.029902564676896226 0.02307089259715589 0.027550677567477422 
77.43333333333334 0.0353903012655401 0.024078844215478235 0.023966849591220196 
80.8 0.029454586179864073 0.033934371150185605 0.03695822600515264 
84.16666666666667 0.015343263523351247 0.026542725949155077 0.023966849591220196 
87.53333333333333 0.010863478553029715 0.023294881845671966 0.018031134505544166 
90.9 0.004031806473289379 0.003695822600515153 0.0007839623698061571 
94.26666666666667 0.0008959569940643064 0.0 0.0 
97.63333333333334 0.001119946242580272 0.0 0.0 
}\dataset

\begin{axis}[
width=.8\textwidth,
height=.45\textwidth,
x tick label style={/pgf/number format/1000 sep=},
ylabel=\emph{Probability},
xlabel=\texttt{age},
ybar=0,
bar width=3.5pt,
legend cell align={left},
legend style={fill opacity=0.8, draw opacity=1, text opacity=1, draw=white!80!black},
tick align=outside,
tick pos=left,
xmajorgrids,
ymajorgrids,
legend image code/.code={
        \draw [#1] (0cm,-0.1cm) rectangle (0.2cm,0.25cm); },
xmin=-3, xmax=100,
ymin=0, ymax=0.1
]

\addplot [black!100!white,fill=lightgray] table[x index=0,y index=1] \dataset;

\addplot [black!100!white,fill=lightblue] table[x index=0,y index=2] \dataset;

\addplot [black!100!white,fill=lightorange] table[x index=0,y index=3] \dataset;

\legend{original dataset,complete DAG,altered DAG}

\end{axis}

\end{tikzpicture}

%% file: tikz/age_with_license.tex
% This file was created with tikzplotlib v0.9.17.
\begin{tikzpicture}

\definecolor{lightgray}{rgb}{.9,.9,.9}
\definecolor{lightblue}{rgb}{.62,.73,.83}
\definecolor{lightorange}{rgb}{.99,.82,.60}

\pgfplotstableread{
0.0 0.0 0.0 0.009228300120369145 
3.3666666666666667 0.0 0.0001352447930754666 0.008960813160358455 
6.733333333333333 0.0 0.000676223965377333 0.008827069680353107 
10.1 0.0 0.0014876927238301316 0.007489634880299523 
13.466666666666667 0.0 0.00419258858533947 0.010966965360438424 
16.833333333333332 0.011091808098356293 0.014200703272923942 0.01591547412063625 
20.2 0.012561806761993827 0.020151474168244445 0.030092283001203 
23.566666666666666 0.013764532941333597 0.013524479307546605 0.028888591681154874 
26.933333333333334 0.03220633435787801 0.025967000270490398 0.020462752440819687 
30.3 0.035413604169452356 0.0285366513389248 0.027149926441088185 
33.666666666666664 0.05011359080582879 0.04260210981877399 0.04373411796175289 
37.03333333333333 0.04543632233061812 0.041384906681094735 0.03985555704159713 
40.4 0.04543632233061812 0.04422504733567967 0.03771566136150384 
43.766666666666666 0.06160630763062022 0.0591019745739817 0.058445900762330416 
47.13333333333333 0.05933449151408898 0.06342980795239683 0.05710846596227709 
50.5 0.06574903113723374 0.06315931836624589 0.0551023137621971 
53.86666666666667 0.08766537484965348 0.0818230998106611 0.075297579243017 
57.233333333333334 0.07737538420421575 0.06464701109007609 0.06326066604254865 
60.6 0.0819190164372785 0.07195022991615163 0.07315768356294738 
63.96666666666667 0.07710811171991794 0.0876386259129065 0.08960813160361014 
67.33333333333333 0.05612722170253992 0.06586421422775535 0.0524274441621122 
70.7 0.055993585460391015 0.06748715174466102 0.055236057242225356 
74.06666666666666 0.032339970600034906 0.0235325939951323 0.027684900361115372 
77.43333333333334 0.03688360283309766 0.02447930754666061 0.023672595960953724 
80.8 0.030602699452099147 0.03435217744117014 0.03704694396149255 
84.16666666666667 0.01576907657357074 0.027725182580471963 0.023672595960953724 
87.53333333333333 0.010557263129763461 0.023397349202056827 0.018322856760738193 
90.9 0.0036081785380204234 0.004327833378415025 0.0006687174000269414 
94.26666666666667 0.0006681812107445229 0.0 0.0 
97.63333333333334 0.0006681812107444118 0.0 0.0 
}\dataset

\begin{axis}[
width=.8\textwidth,
height=.45\textwidth,
x tick label style={/pgf/number format/1000 sep=},
ylabel=\emph{Probability},
xlabel=\texttt{age} if the individual owns a driving license,
ybar=0,
bar width=3.5pt,
legend cell align={left},
legend style={fill opacity=0.8, draw opacity=1, text opacity=1, draw=white!80!black},
tick align=outside,
tick pos=left,
xmajorgrids,
ymajorgrids,
legend image code/.code={
        \draw [#1] (0cm,-0.1cm) rectangle (0.2cm,0.25cm); },
xmin=-3, xmax=100,
ymin=0, ymax=0.10
]

\addplot [black!100!white,fill=lightgray] table[x index=0,y index=1] \dataset;

\addplot [black!100!white,fill=lightblue] table[x index=0,y index=2] \dataset;

\addplot [black!100!white,fill=lightorange] table[x index=0,y index=3] \dataset;

\legend{original dataset,complete DAG,altered DAG}

\end{axis}

\end{tikzpicture}

%% file: tikz/age_without_license.tex
% This file was created with tikzplotlib v0.9.17.
\begin{tikzpicture}

\definecolor{lightgray}{rgb}{.9,.9,.9}
\definecolor{lightblue}{rgb}{.62,.73,.83}
\definecolor{lightorange}{rgb}{.99,.82,.60}

\pgfplotstableread{
0.0 0.020055325034578145 0.03583061889250815 0.011019283746556474 
3.3666666666666667 0.10926694329183924 0.0938110749185671 0.005509641873278237 
6.733333333333333 0.1251728907330563 0.12182410423452567 0.010330578512396695 
10.1 0.10511756569847819 0.08338762214983575 0.008264462809917356 
13.466666666666667 0.09543568464730257 0.06775244299674155 0.010330578512396695 
16.833333333333332 0.1065006915629319 0.05472312703582971 0.014462809917355372 
20.2 0.015905947441217094 0.04820846905537429 0.022038567493113136 
23.566666666666666 0.008298755186721962 0.020195439739415066 0.03168044077135018 
26.933333333333334 0.013831258644536604 0.012377850162867299 0.022727272727272957 
30.3 0.010373443983402453 0.010423452768730357 0.02410468319559253 
33.666666666666664 0.020746887966804906 0.016286644951141183 0.04269972451790677 
37.03333333333333 0.013139695712309774 0.01433224755700424 0.045454545454545914 
40.4 0.015905947441217094 0.00977198697068471 0.03305785123966848 
43.766666666666666 0.015905947441217094 0.020195439739415066 0.05371900826446119 
47.13333333333333 0.013831258644536604 0.027361563517917187 0.055096418732780705 
50.5 0.027662517289073207 0.024755700325734598 0.055096418732780705 
53.86666666666667 0.028354080221300038 0.037133550488601896 0.07093663911845516 
57.233333333333334 0.030428769017980528 0.02866449511400848 0.06818181818181612 
60.6 0.03250345781466102 0.030618892508145423 0.08471074380165033 
63.96666666666667 0.03112033195020736 0.04885993485342355 0.0936639118457272 
67.33333333333333 0.026279391424619547 0.035179153094464954 0.055785123966940464 
70.7 0.028354080221300038 0.03908794788273884 0.048898071625342876 
74.06666666666666 0.017289073305670755 0.020846905537460714 0.026859504132230594 
77.43333333333334 0.027662517289073207 0.02214983713355201 0.025482093663911076 
80.8 0.023513139695712226 0.03192182410423672 0.03650137741046722 
84.16666666666667 0.013139695712309774 0.020846905537460714 0.025482093663911076 
87.53333333333333 0.012448132780082943 0.022801302931597656 0.01652892561983421 
90.9 0.006224066390041472 0.0006514657980455363 0.0013774104683195176 
94.26666666666667 0.0020746887966804906 0.0 0.0 
97.63333333333334 0.003457814661134151 0.0 0.0
}\dataset

\begin{axis}[
width=.8\textwidth,
height=.45\textwidth,
x tick label style={/pgf/number format/1000 sep=},
ylabel=\emph{Probability},
xlabel=\texttt{age} if the individual does not own a driving license,
ybar=0,
bar width=3.5pt,
legend cell align={left},
legend style={fill opacity=0.8, draw opacity=1, text opacity=1, draw=white!80!black},
tick align=outside,
tick pos=left,
xmajorgrids,
ymajorgrids,
legend image code/.code={
        \draw [#1] (0cm,-0.1cm) rectangle (0.2cm,0.25cm); },
xmin=-3, xmax=100,
ymin=0, ymax=0.13
]

\addplot [black!100!white,fill=lightgray] table[x index=0,y index=1] \dataset;

\addplot [black!100!white,fill=lightblue] table[x index=0,y index=2] \dataset;

\addplot [black!100!white,fill=lightorange] table[x index=0,y index=3] \dataset;

\legend{original dataset,complete DAG,altered DAG}

\end{axis}

\end{tikzpicture}

%% file: text/06_conclusion.tex
\section{Conclusion}
\label{sec:conclusion}

This article presents a novel GAN architecture, the DATGAN, that integrates expert knowledge to control the causal links between the variables. In the methodology, we show how a Directed Acyclic Graph (DAG) can model the generator's structure and how synthetic variables are generated using Long-Short Term Memory (LSTM) cells. In addition, we provide an efficient way to encode categorical and continuous variables. While the core mechanics of the DATGAN remain the same, we explore different loss functions for training the DATGAN, the use of label smoothing on categorical variables, and multiple sampling methods. In order to compare the results as fairly as possible, we provide two new systematic assessment methods for comparing synthetic datasets: a statistical method and a Machine Learning efficacy method. Using these two methods, we show that two-sided label smoothing and using simulation to sample the final synthetic variables lead to the best performances. The most optimal loss function, on the other hand, depends on the ratio of continuous and categorical variables. Indeed, if a dataset contains primarily categorical variables, we recommend using a \texttt{WGAN} loss function. On the contrary, if the dataset contains more continuous variables than categorical variables, we recommend using a \texttt{WGGP} loss function. We then show that using a complete DAG leads to better correlations in the final synthetic dataset than a stripped one. The optimal DATGAN models are then compared against state-of-the-art models. We show that the DATGAN models outperform all the other models on all the case studies using both assessment methods. Finally, we show how the DATGAN can create hypothetical synthetic populations.

The DATGAN architecture has been developed to improve the representativity of its generated synthetic data. Such datasets can then be used in simulations and might improve the latter's results since it has shown better results than other synthetic datasets generated by state-of-the-art models found in the literature. However, it might be possible to improve the DATGAN even further by upgrading the encoding of the tabular data. In the DATGAN, we have only considered two types of variables: continuous and categorical. However, \cite{zhao_ctab-gan_2021} consider four different types of variables: continuous, categorical, mixed data, and integers. Therefore, a straightforward improvement to the DATGAN would be to add more data types in the encoding process. Indeed, if the encoding of the data is improved, the generated synthetic data will most likely improve as well. In addition, the DATGAN, as every other GAN, achieves differential privacy~\citep{jordon_pate-gan_2018} since the generator never sees the original data. Therefore, privacy preservation is generally not a concern for GANs. For the Machine Learning efficacy research axis, the DATGAN is already showing improvements thanks to the Machine Learning evaluation metric results. However, the DATGAN does not consider bias in the original data. Therefore, it also generates biased data. Thus, improving the bias correction of the DATGAN will also improve the Machine Learning efficacy. A simple fix that already exists in the literature is the use of conditionality on GANs. We could, thus, update the DATGAN such that it can conditionally generate synthetic data to reduce the bias in the data. Furthermore, one of the difficult tasks with the DATGAN is to create a good DAG to hinder the models' performances. While the relationship between some variables might be easy to find, it is more subtle for others. Therefore, it would be interesting to add a feature to combine multiple variables in a cluster such that they all influence each other without having to decide in which specific order. However, such an improvement might not work with the current design of the DATGAN using LSTM cells since each cell is assigned to a single variable. Therefore, a complete redesign of the DATGAN might be needed to implement such an improvement. Finally, we have discussed four of the five research axes presented in the literature review. The final axis is transfer learning. Researchers have already been working on this topic with synthetic image generators. Therefore, one of the future steps for synthetic tabular data generators is to follow this trend and start implementing models that can transfer knowledge between multiple datasets.

%% file: text/07_appendix.tex
\section{Appendix}
\label{sec:app}

\subsection{Case studies - dataset description}

\begin{xltabular}{\textwidth}{l||c|X|X}
    
    \caption{CMAP dataset}
	\label{tab:data_Chicago} \\

    \textbf{Variables} & \textbf{Type} & \textbf{Details} & \textbf{Description}  \\ \midrule[1.5pt] 
	\endfirsthead

	\multicolumn{4}{c}{\tablename\ \thetable{} -- continued from previous page} \\
	\textbf{Variables} & \textbf{Type} & \textbf{Details} & \textbf{Description}   \\ \midrule[1.5pt]
	\endhead

	\multicolumn{4}{r}{{Continues on next page...}}
	\endfoot

	\endlastfoot
    \texttt{choice} & Categorical & 5 string categories & Chosen mode \\ \hline
    \texttt{travel\_dow} & Categorical & 1-7 (Monday-Sunday) & Day of the week travel \\ \hline
    \texttt{trip\_purpose} & Categorical & 7 string categories & Primary purpose for making trip \\ \hline
    \texttt{distance} & Continuous & 8'743 unique values between 0 and 69.71 & Straight-line trip distance in miles \\ \hline
    \texttt{hh\_vehicles} & Categorical & 0-8 & Number of vehicles in household \\ \hline
    \texttt{hh\_size} & Categorical & 1-8 & Number of people in household \\ \hline
    \texttt{hh\_bikes} & Categorical & 0-7 & Number of bikes in household \\ \hline
    \texttt{hh\_descr} & Categorical & 3 string categories & Household type \\ \hline
    \texttt{hh\_income} & Categorical & 7 categories & Household income level \\ \hline
    \texttt{gender} & Categorical & 0 (female), 1 (male) & Gender of individual \\ \hline
    \texttt{age} & Continuous & 97 unique integers between 0 and 98 & Age of individual \\ \hline
    \texttt{license} & Categorical & 0 (none), 1 (has driving license) & Driving license ownership \\ \hline
    \texttt{education\_level} & Categorical & 6 categories & Highest level of education achieved \\ \hline
    \texttt{work\_status} & Categorical & 8 string categories & Working status \\ \hline
    \texttt{departure\_time} & Continuous & 949 unique values between 0 and 23.86 & Departure time of trip (in decimal hours)
\end{xltabular}

\vspace{0.5cm}

\begin{xltabular}{\textwidth}{l||c|X|X}
    
    \caption{LPMC dataset}
    \label{tab:data_LPMC} \\

    \textbf{Variables} & \textbf{Type} & \textbf{Details} & \textbf{Description}  \\ \midrule[1.5pt] 
    \endfirsthead

    \multicolumn{4}{c}%
    {\tablename\ \thetable{} -- continued from previous page} \\
    \textbf{Variables} & \textbf{Type} & \textbf{Details} & \textbf{Description}  \\ \midrule[1.5pt] 
    \endhead
    
    \multicolumn{4}{r}{{Continues on next page...}}
    \endfoot
    
    \endlastfoot

    \texttt{travel\_mode} & Categorical & 4 string categories & Mode of travel chosen by LTDS trip \\ \hline
    \texttt{purpose} & Categorical & 5 string categories & Journey purpose for trip \\ \hline
    \texttt{fueltype} & Categorical & 6 string categories & Fuel type of passenger’s vehicle\\ \hline
    \texttt{faretype} & Categorical & 5 string categories & Public transport fare type of passenger \\ \hline
    \texttt{bus\_scale} & Categorical & 3 values (0, 0.5, 1) & Percentage of the full bus fare paid by the passenger \\ \hline
    \texttt{travel\_year} & Categorical & 4 values between 2012 and 2015& Year of travel \\ \hline
    \texttt{travel\_month} & Categorical & 12 values between 1 and 12 & Month of year of travel \\ \hline
    \texttt{travel\_date} & Categorical & 31 values between 1 and 31& Date of month of travel\\ \hline
    \texttt{day\_of\_week} & Categorical & 7 values between 1 and 7 & Day of the week of travel\\ \hline
    \texttt{start\_time\_linear} & Continuous & 609 unique values between 0 and 23.92 & Start time of trip (in decimal hours)\\ \hline
    \texttt{age} & Continuous & 90 unique values between 5 and 94 & Age of passenger in years \\ \hline
    \texttt{female} & Categorical & 0 (male), 1 (female) & Gender of passenger \\\hline
    \texttt{driving\_license} & Categorical & 0 (none), 1 (has driving license) & Whether the traveller has a driving licence\\ \hline
    \texttt{car\_ownership} & Categorical & 3 values between 0 and 2 & Car ownership of household \\ \hline
    \texttt{distance} & Continuous & 8'972 unique values between 77 and 40'941 & Straight line trip distance \\ \hline
    \texttt{dur\_walking} & Continuous & 8'416 unique values between 0.028 and 9.28 & Duration of walking route\\ \hline
    \texttt{dur\_cycling} & Continuous & 4'350 unique values between 0.0075 and 3.05 & Duration of cycling route\\ \hline
    \texttt{dur\_pt\_access} & Continuous & 1'656 unique values between 0 and 1.06 & Duration walking to/from first/last stop on public transport route\\ \hline
    \texttt{dur\_pt\_rail} & Continuous & 76 unique values between 0 and 1.37 & Duration spent on rail services on public transport route\\ \hline
    \texttt{dur\_pt\_bus} & Continuous & 2'469 unique values between 0 and 2.15 & Duration spent on bus services on public transport route\\ \hline
    \texttt{dur\_pt\_int} & Continuous & 725 unique values between 0 and 0.57 & Total duration of public transport interchanges\\ \hline
    \texttt{pt\_n\_interchanges} & Categorical & 5 values between 0 and 4 & Total number of public transport interchanges\\ \hline
    \texttt{dur\_driving} & Continuous & 3'544 unique values between 0.0042 and 1.79 & Duration of driving route\\ \hline
    \texttt{cost\_transit} & Continuous & 146 unique values between 0 and 11.7 & Cost of public transport route\\ \hline
    \texttt{cost\_driving\_fuel} & Continuous & 507 unique values between 0.02 and 10.09 & Vehicle operation costs of driving route\\ \hline
    \texttt{cost\_driving\_con\_charge} & Categorical & 2 values (0 or 10.5) & Congestion charge for driving route\\ \hline
    \texttt{driving\_traffic\_percent} & Continuous & 15'787 unique values between 0 and 1.04 & Traffic variability \\ \hline
\end{xltabular}

\vspace{0.5cm}

\begin{xltabular}{\textwidth}{l||c|X|X}

	\caption{ADULT dataset}
	\label{tab:data_adult} \\
	
    \textbf{Variables} & \textbf{Type} & \textbf{Details} & \textbf{Description}  \\ \midrule[1.5pt] 
	\endfirsthead

	\multicolumn{4}{c}%
{\tablename\ \thetable{} -- continued from previous page} \\
    \textbf{Variables} & \textbf{Type} & \textbf{Details} & \textbf{Description}  \\ \midrule[1.5pt] 
	\endhead

	\multicolumn{4}{r}{{Continues on next page...}}
	\endfoot

	\endlastfoot
    \texttt{age} & Continuous & 74 unique integers between 17 and 90 & Age of individual \\ \hline
    \texttt{workclass} & Categorical & 9 string categories & Work status of the individual \\ \hline
    \texttt{education} & Categorical & 16 string categories & Education of the individual \\ \hline
    \texttt{education-num} & Categorical & 16 integer categories & Number of years of education \\ \hline
    \texttt{marital-status} & Categorical & 7 string categories & Marital status of the individual \\ \hline
    \texttt{occupation} & Categorical & 14 string categories & Occupation of the individual \\ \hline
    \texttt{relationship} & Categorical & 6 string categories & Relationship with a partner \\ \hline
    \texttt{race} & Categorical & 5 string categories & Race of the individual \\ \hline
    \texttt{gender} & Categorical & 2 string categories & Gender of the individual \\ \hline
    \texttt{capital-gain} & Continuous & 121 unique integers between 0 and 99'999 & Capital gains \\ \hline
    \texttt{capital-loss} & Continuous & 97 unique integers between 0 and 4'356 & Capital losses \\ \hline
    \texttt{hours-per-week} & Categorical & 96 unique integers between 1 and 99 & Number of hours of work per week \\ \hline
    \texttt{native-country} & Categorical & 41 string categories &  Native country of the individual\\ \hline
    \texttt{income} & Categorical & 2 string categories & Income greater or equal to 50k per year \\ \hline
\end{xltabular}

\subsection{Case studies - Directed Acyclic Graphs}

\begin{figure}[H]
    \centering
    \input{tikz/DAG_CMAP}
    \caption{DAG used for the CMAP case study. We identified three categories of variables: purple corresponds to individuals, blue to households, and red to trips.}
    \label{fig:DAG_Chicago}
\end{figure}
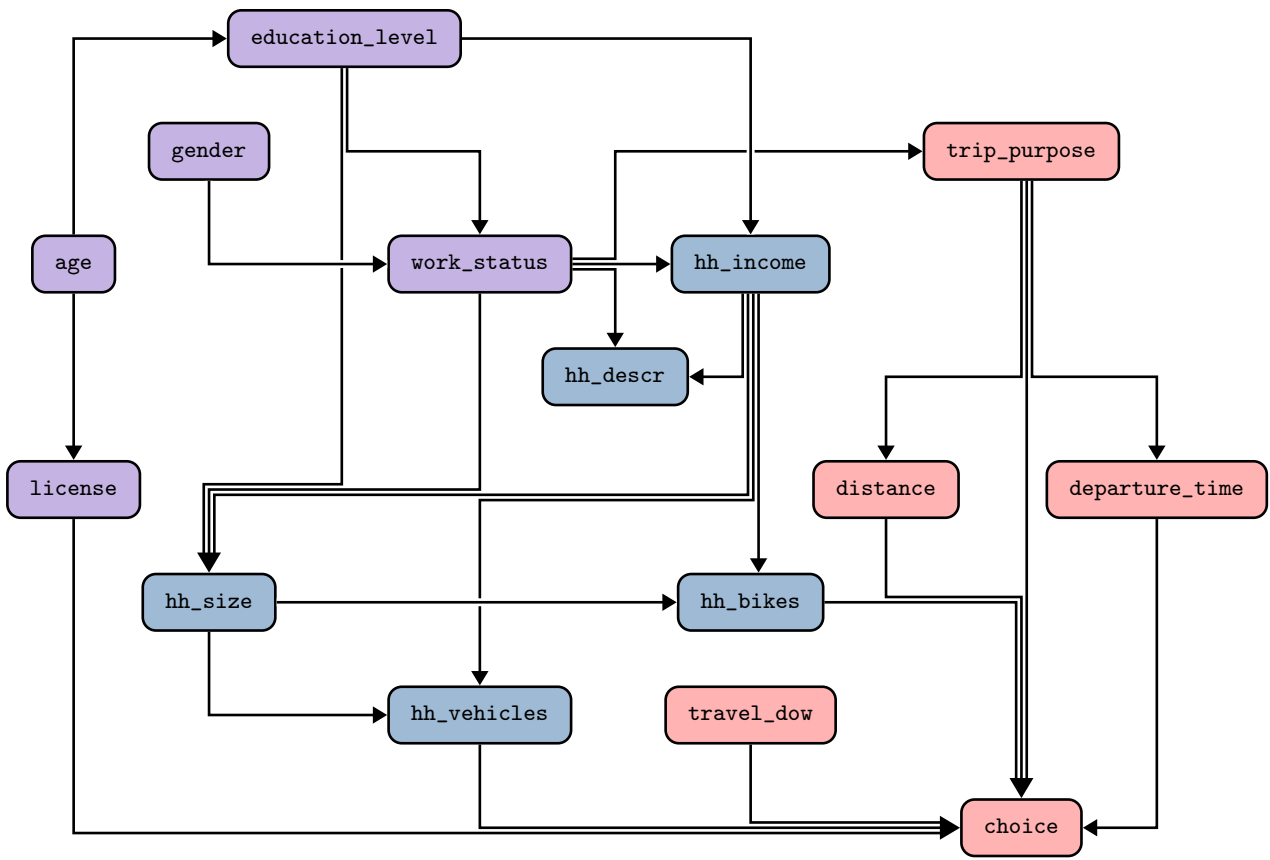

\begin{landscape}
\mbox{}\vfill
\begin{figure}[h!]
    \centering
    \input{tikz/DAG_LPMC}
    \caption{DAG used for the LPMC case study. We identified five categories of variables: purple corresponds to individuals, blue to households, red to trips, orange to survey, and yellow to alternatives. }
    \label{fig:DAG_LPMC}
\end{figure}
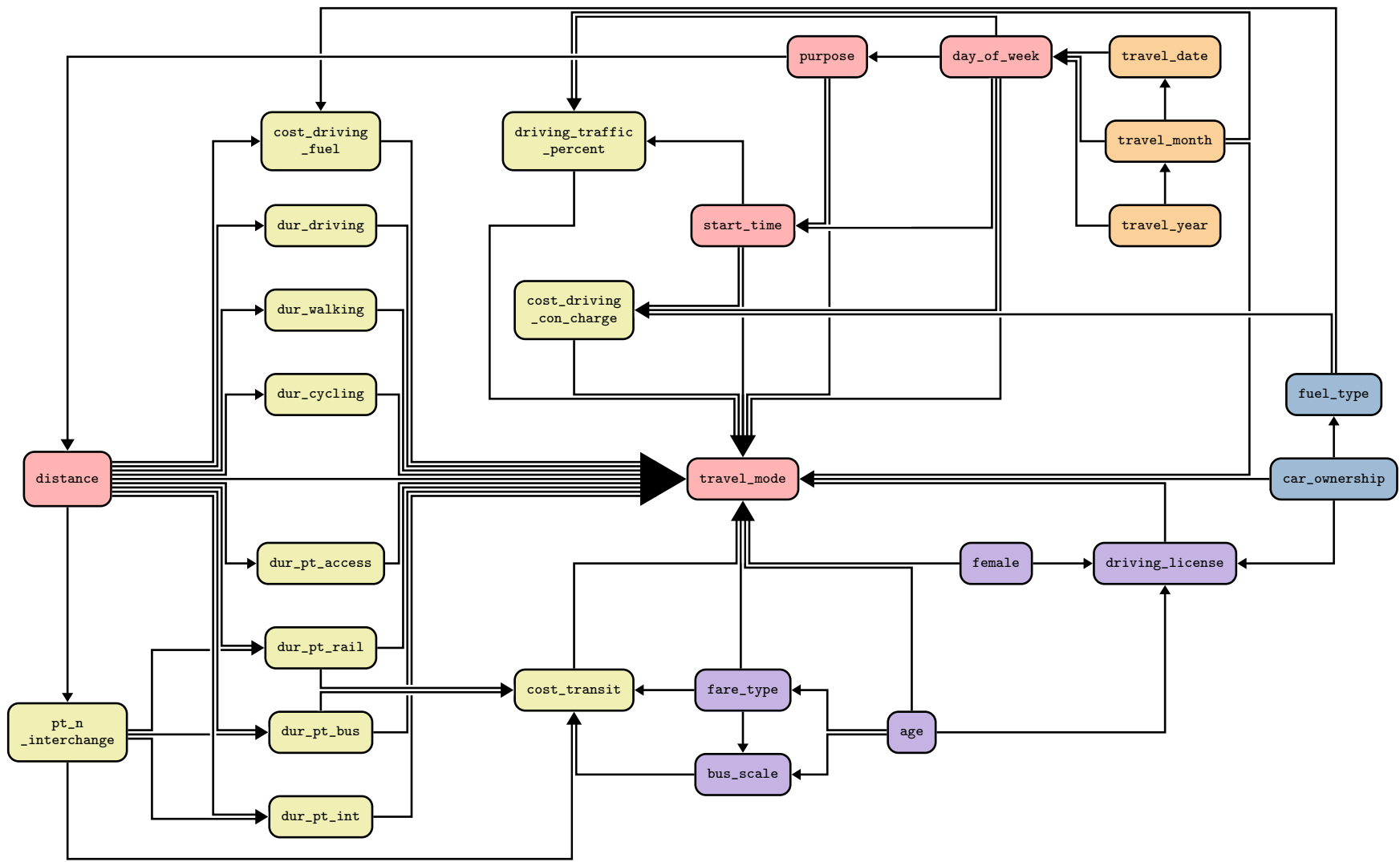
\vfill
\end{landscape}

\begin{figure}[H]
    \centering
    \input{tikz/DAG_ADULT}
    \caption{DAG used for the ADULT case study. We identified two categories of variables: orange corresponds to individuals, blue to occupation.}
    \label{fig:DAG_adult}
\end{figure}
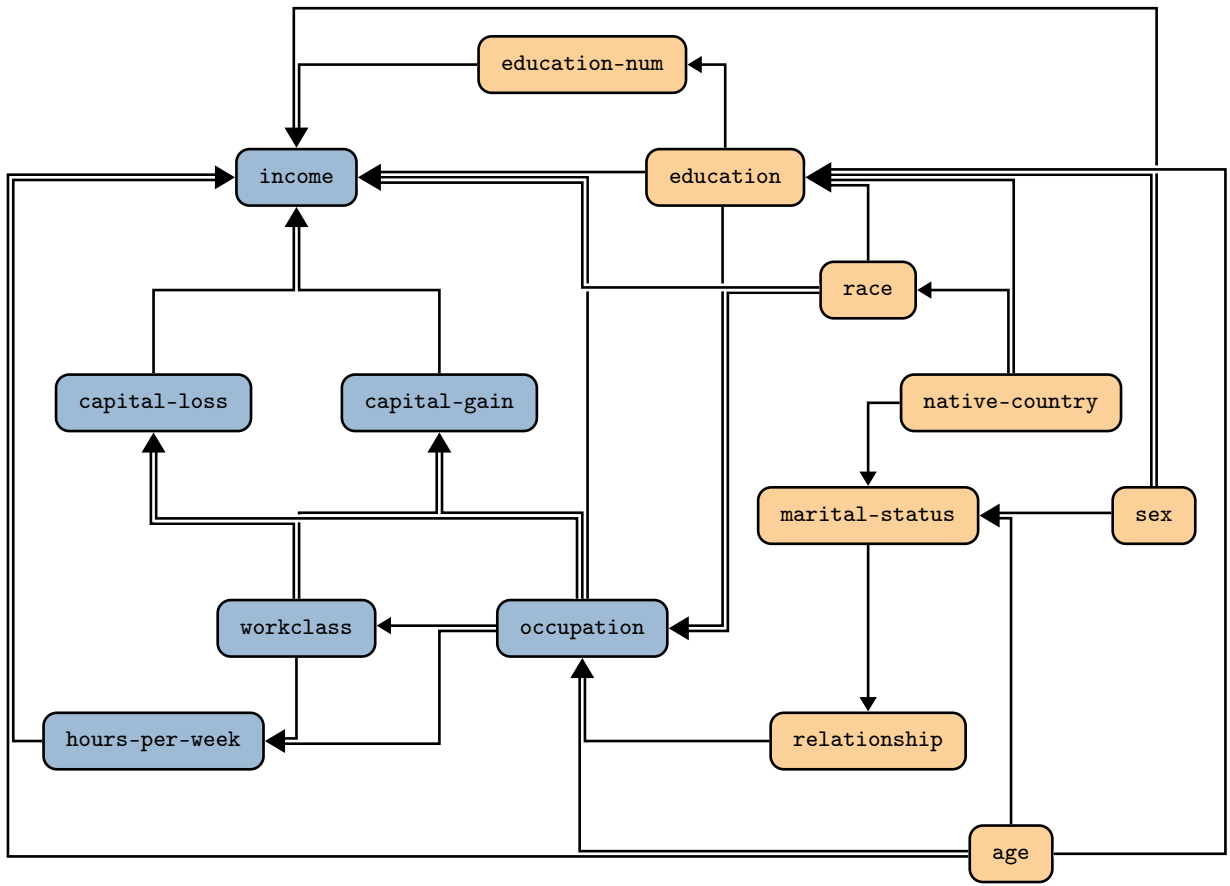

%% file: tikz/DAG_CMAP.tex
\footnotesize
\begin{tikzpicture}[every text node part/.style={align=center}]

\definecolor{lightgray}{rgb}{.9,.9,.9}
\definecolor{lightblue}{rgb}{.62,.73,.83}
\definecolor{lightorange}{rgb}{.99,.82,.60}
\definecolor{lightpurple}{RGB}{197,180,227}

% Helpers
\def\ny{11}
\def\nx{8}
%\draw[help lines,xstep=1,ystep=1, color=black!20] (-\nx,-\ny) grid (\nx,\ny);
%\draw[help lines,xstep=.5,ystep=.5, color=black!10, dotted] (-\nx,-\ny) grid (\nx,\ny);
%\foreach \x in {-\nx,...,\nx} { \node [anchor=north, color=black!30] at (\x,0) {\x}; }
%\foreach \y in {-\ny,...,\ny} { \node [anchor=east, color=black!30] at (0,\y) {\y}; }

\def\dy{1.5}
\def\dx{1.8}
\def\shift{2pt}

\tikzset{
block/.style={
	rounded corners=5pt,
    rectangle,
	fill=lightgray,
    draw=black,
	line width=1,
    text depth=.3\baselineskip, 
    text height=.7\baselineskip,
	inner ysep=0.2cm,
	inner xsep=0.3cm
}}

\node[block, fill=lightpurple] (A) at (0,2*\dy) {\ttfamily age};
\node[block, fill=lightpurple] (B) at (0,0*\dy) {\ttfamily license};
\node[block, fill=lightpurple] (C) at (2*\dx,4*\dy) {\ttfamily education\_level};
\node[block, fill=lightpurple] (D) at (\dx,3*\dy) {\ttfamily gender};
\node[block, fill=lightpurple] (E) at (3*\dx,2*\dy) {\ttfamily work\_status};
\node[block, fill=lightblue] (F) at (\dx,-1*\dy) {\ttfamily hh\_size};
\node[block, fill=lightblue] (G) at (3*\dx,-2*\dy) {\ttfamily hh\_vehicles};
\node[block, fill=red!30!white] (H) at (5*\dx,-2*\dy) {\ttfamily travel\_dow};
\node[block, fill=lightblue] (I) at (4*\dx,1*\dy) {\ttfamily hh\_descr};
\node[block, fill=lightblue] (J) at (5*\dx,2*\dy) {\ttfamily hh\_income};
\node[block, fill=red!30!white] (K) at (7*\dx,3*\dy) {\ttfamily trip\_purpose};
\node[block, fill=red!30!white] (L) at (8*\dx,0*\dy) {\ttfamily departure\_time};
\node[block, fill=red!30!white] (M) at (6*\dx,0*\dy) {\ttfamily distance};
\node[block, fill=lightblue] (N) at (5*\dx,-1*\dy) {\ttfamily hh\_bikes};
\node[block, fill=red!30!white] (O) at (7*\dx,-3*\dy) {\ttfamily choice};

\draw[-{Triangle[scale=1]}, line width=1] (A) -- (B);
\draw[-{Triangle[scale=1]}, line width=1] (A) |- (C);

\coordinate (AA) at (2*\dx,0);

\draw[-{Triangle[scale=1, color=white]}, line width=1] ([xshift=-0.5*\shift]C.south) -- ([xshift=-0.5*\shift, yshift=\shift]AA) -| ([xshift=-\shift]F.north);

\draw[line width=3, draw=white] (D) |- (E);
\draw[-{Triangle[scale=1]}, line width=1] (D) |- (E);

\coordinate (AB) at (2*\dx,3*\dy);
\draw[-{Triangle[scale=1]}, line width=1] ([xshift=0.5*\shift]C.south) -- ([xshift=0.5*\shift]AB) -| (E);

\draw[-{Triangle[scale=1]}, line width=1] (E) -- (J);
\coordinate (AC) at (4*\dx,2*\dy);
\draw[-{Triangle[scale=1]}, line width=1] ([yshift=\shift]E.east) -- ([yshift=\shift]AC) |- (K);
\draw[-{Triangle[scale=1]}, line width=1] ([yshift=-\shift]E.east) -- ([yshift=-\shift]AC) -- (I);

\draw[line width=3, draw=white] (C) -| (J);
\draw[-{Triangle[scale=1]}, line width=1] (C) -| (J);

\draw[-{Triangle[scale=1]}, line width=1] ([xshift=-1.5*\shift]J.south) |- (I);
\draw[-{Triangle[scale=1, color=white]},line width=1] ([xshift=-0.5*\shift]J.south) |- ([yshift=-\shift]AA) -| ([xshift=\shift]F.north);
\draw[-{Triangle[scale=1.4]}, line width=1] (E) |- (AA) -| (F.north);

\coordinate (AH) at (3*\dx, 0);
\draw[-{Triangle[scale=1]},line width=1] ([xshift=0.5*\shift]J.south) |- ([yshift=-2*\shift]AH) -| (G);
\draw[-{Triangle[scale=1]},line width=1] ([xshift=1.5*\shift]J.south) -- ([xshift=1.5*\shift]N.north);

\draw[line width=3, draw=white] (F) -- (N);
\draw[-{Triangle[scale=1]}, line width=1] (F) -- (N);
\draw[-{Triangle[scale=1]}, line width=1] (F) |- (G);

\coordinate (AD) at (4*\dx, -3*\dy);
\draw[-{Triangle[scale=1, color=white]}, line width=1] (B) |- ([yshift=-\shift]O.west);
\draw[-{Triangle[scale=1, color=white]},line width=1] (H.south) |- ([yshift=\shift]O.west);
\draw[-{Triangle[scale=1.4]},line width=1] (G.south) |- (O.west);

\coordinate (AE) at (7*\dx, -1*\dy);
\draw[-{Triangle[scale=1, color=white]},line width=1] (N.east) -| ([xshift=-\shift]AE) -- ([xshift=-\shift]O.north);
\draw[-{Triangle[scale=1, color=white]}, line width=1] ([xshift=\shift]K.south) -- ([xshift=\shift]O.north);
\draw[-{Triangle[scale=1.4]},line width=1] (M.south) |- ([yshift=\shift]AE) -- (O.north);

\coordinate (AF) at (7*\dx, 1*\dy);
\draw[-{Triangle[scale=1]}, line width=1] ([xshift=0]K.south) -- ([xshift=0]AF) -| (M);
\draw[-{Triangle[scale=1]}, line width=1] ([xshift=2*\shift]K.south) -- ([xshift=2*\shift]AF) -| (L);
\draw[-{Triangle[scale=1]}, line width=1] (L) |- (O);

%\draw[fill=lightgray, rounded corners=10pt] (1*\dx, -3.5*\dy) -- (7*\dx, -3.5*\dy) -- (7*\dx, -4.9*\dy) -- (1*\dx, -4.9*\dy) -- cycle;
%\node[block, fill=lightpurple] at (2*\dx,-4.4*\dy) {Individual level};
%\node[block, fill=lightblue] at (4*\dx,-4.4*\dy) {Household level};
%\node[block, fill=red!30!white] at (6*\dx,-4.4*\dy) {Trip level};
%\node at (4*\dx, -3.8*\dy) {\bf Legend};

\end{tikzpicture}

%% file: tikz/DAG_LPMC.tex
\scriptsize
\begin{tikzpicture}[every text node part/.style={align=center}]

% continue here: https://docs.google.com/presentation/d/1mj_6AYo9uRjRb74uAyt_aPRQOVGmPL7gJU9ghFnUTEU/edit#slide=id.g10da6551be6_0_46

\definecolor{lightgray}{rgb}{.9,.9,.9}
\definecolor{lightblue}{rgb}{.62,.73,.83}
\definecolor{lightorange}{rgb}{.99,.82,.60}
\definecolor{lightpurple}{RGB}{197,180,227}
\definecolor{lightyellow}{RGB}{240, 240, 180}

% Helpers
\def\ny{11}
\def\nx{8}
%\draw[help lines,xstep=1,ystep=1, color=black!20] (-\nx,-\ny) grid (\nx,\ny);
%\draw[help lines,xstep=.5,ystep=.5, color=black!10, dotted] (-\nx,-\ny) grid (\nx,\ny);
%\foreach \x in {-\nx,...,\nx} { \node [anchor=north, color=black!30] at (\x,0) {\x}; }
%\foreach \y in {-\ny,...,\ny} { \node [anchor=east, color=black!30] at (0,\y) {\y}; }

\def\dy{1.4}
\def\dx{2.8}
\def\shift{2pt}
\def\scale{0.8}

\tikzset{
block/.style={
	rounded corners=5pt,
    rectangle,
	fill=lightgray,
    draw=black,
	line width=1,
    text depth=.3\baselineskip, 
    text height=.7\baselineskip,
	inner ysep=0.2cm,
	inner xsep=0.2cm
},
doubleblock/.style={
	rounded corners=5pt,
    rectangle,
	fill=lightgray,
    draw=black,
	line width=1,
    text depth=.3\baselineskip, 
    text height=1.7\baselineskip,
	inner ysep=0.2cm,
	inner xsep=0.2cm
},
largeblock/.style={
	rounded corners=5pt,
    rectangle,
	fill=lightgray,
    draw=black,
	line width=1,
    text depth=.3\baselineskip, 
    text height=.7\baselineskip,
	minimum height=0.9cm,
	inner ysep=0.2cm,
	inner xsep=0.2cm
}}

\node[largeblock, fill=red!30!white] (A) at (-\dx,3*\dy) {\ttfamily distance};
\node[doubleblock, fill=lightyellow] (B) at (-\dx,0*\dy) {\ttfamily pt\_n\\\ttfamily \_interchange};

\node[doubleblock, fill=lightyellow] (C) at (0.5*\dx,7*\dy) {\ttfamily cost\_driving\\\ttfamily \_fuel};
\node[block, fill=lightyellow] (D) at (0.5*\dx,6*\dy) {\ttfamily dur\_driving};
\node[block, fill=lightyellow] (E) at (0.5*\dx,5*\dy) {\ttfamily dur\_walking};
\node[block, fill=lightyellow] (F) at (0.5*\dx,4*\dy) {\ttfamily dur\_cycling};
\node[block, fill=lightyellow] (G) at (0.5*\dx,2*\dy) {\ttfamily dur\_pt\_access};
\node[block, fill=lightyellow] (H) at (0.5*\dx,1*\dy) {\ttfamily dur\_pt\_rail};
\node[block, fill=lightyellow] (I) at (0.5*\dx,0*\dy) {\ttfamily dur\_pt\_bus};
\node[block, fill=lightyellow] (J) at (0.5*\dx,-1*\dy) {\ttfamily dur\_pt\_int};

\node[doubleblock, fill=lightyellow] (L) at (2*\dx,7*\dy) {\ttfamily driving\_traffic\\\ttfamily \_percent};
\node[doubleblock, fill=lightyellow] (M) at (2*\dx,5*\dy) {\ttfamily cost\_driving\\\ttfamily \_con\_charge};
\node[block, fill=lightyellow] (N) at (2*\dx,0.5*\dy) {\ttfamily cost\_transit};

\node[block, fill=red!30!white] (O) at (3*\dx,6*\dy) {\ttfamily start\_time};
\node[block, fill=red!30!white] (P) at (3*\dx,3*\dy) {\ttfamily travel\_mode};
\node[block, fill=lightpurple] (Q) at (3*\dx,0.5*\dy) {\ttfamily fare\_type};
\node[block, fill=lightpurple] (R) at (3*\dx,-0.5*\dy) {\ttfamily bus\_scale};

\node[block, fill=red!30!white] (S) at (3.5*\dx,8*\dy) {\ttfamily purpose};
\node[block, fill=lightpurple] (V) at (4*\dx,0*\dy) {\ttfamily age};

\node[block, fill=red!30!white] (U) at (4.5*\dx,8*\dy) {\ttfamily day\_of\_week};
\node[block, fill=lightpurple] (T) at (4.5*\dx,2*\dy) {\ttfamily female};

\node[block, fill=lightorange] (W) at (5.5*\dx,8*\dy) {\ttfamily travel\_date};
\node[block, fill=lightorange] (X) at (5.5*\dx,7*\dy) {\ttfamily travel\_month};
\node[block, fill=lightorange] (Y) at (5.5*\dx,6*\dy) {\ttfamily travel\_year};
\node[block, fill=lightpurple] (Z) at (5.5*\dx,2*\dy) {\ttfamily driving\_license};

\node[block, fill=lightblue] (A2) at (6.5*\dx,4*\dy) {\ttfamily fuel\_type};
\node[block, fill=lightblue] (B2) at (6.5*\dx,3*\dy) {\ttfamily car\_ownership};

% Links
\draw[-{Triangle[scale=\scale]}, line width=1] (A) -- (B);
\draw[-{Triangle[scale=\scale]}, line width=1] (T) -- (Z);
\draw[-{Triangle[scale=\scale]}, line width=1] (V) -| (Z);
\draw[-{Triangle[scale=\scale]}, line width=1] (B2) |- (Z);
\draw[-{Triangle[scale=\scale]}, line width=1] (B2) -- (A2);
\draw[-{Triangle[scale=\scale]}, line width=1] (Y) -- (X);
\draw[-{Triangle[scale=\scale]}, line width=1] (X) -- (W);
\draw[-{Triangle[scale=\scale]}, line width=1] (Q) -- (N);
\draw[-{Triangle[scale=\scale]}, line width=1] (Q) -- (R);
\draw[-{Triangle[scale=\scale]}, line width=1] (O) |- (L);
\draw[-{Triangle[scale=\scale]}, line width=1] (U) -- (S);

\coordinate (AA) at (3.5*\dx, 0*\dy);
\draw[-{Triangle[scale=\scale]}, line width=1] ([yshift=0.5*\shift]V.west) -- ([yshift=0.5*\shift]AA) |- (Q);
\draw[-{Triangle[scale=\scale]}, line width=1] ([yshift=-0.5*\shift]V.west) -- ([yshift=-0.5*\shift]AA) |- (R);

\coordinate (AB) at (2*\dx, -1.5*\dy);
\draw[-{Triangle[scale=\scale, color=white]}, line width=1] (R) -| ([xshift=0.5*\shift]N.south);
\draw[-{Triangle[scale=\scale, color=white]}, line width=1] (B) |- ([xshift=-0.5*\shift]AB) -- ([xshift=-0.5*\shift]N.south);
\draw[-{Triangle[scale=1.4*\scale, color=black]}, line width=1, draw=white] (AB) -- (N.south);

\coordinate (AC) at (3*\dx, 2*\dy);
\draw[-{Triangle[scale=\scale, color=white]}, line width=1] (T) -| ([xshift=1.5*\shift, yshift=-2*\shift]P.south);
\draw[-{Triangle[scale=\scale, color=white]}, line width=1] (N) |- ([xshift=-1.5*\shift]AC) -- ([xshift=-1.5*\shift, yshift=-2*\shift]P.south);
\draw[-{Triangle[scale=\scale, color=white]}, line width=1] (V) |- ([xshift=0.5*\shift, yshift=-\shift]AC) -- ([xshift=0.5*\shift]P.south);
\draw[-{Triangle[scale=\scale, color=white]}, line width=1] ([xshift=-0.5*\shift]Q.north) -| ([xshift=-0.5*\shift]P.south);
\draw[-{Triangle[scale=2.2*\scale, color=black]}, line width=1, draw=white] (AC) -- (P.south);

\coordinate (AD) at (6*\dx, 3*\dy);
\draw[-{Triangle[scale=\scale, color=white]}, line width=1] (Z) |- ([yshift=-\shift]P.east);
\draw[-{Triangle[scale=\scale, color=white]}, line width=1] ([yshift=-0.5*\shift]X.east) -| ([yshift=\shift]AD) -- ([yshift=\shift]P.east);
\draw[-{Triangle[scale=1.6*\scale]}, line width=1] (B2) -- (P.east);

\coordinate (AE) at (5*\dx, 8*\dy);
\draw[-{Triangle[scale=\scale, color=white]}, line width=1] ([yshift=\shift]W.west) -- ([yshift=\shift]U.east);
\draw[-{Triangle[scale=\scale, color=white]}, line width=1] (Y.west) -| ([yshift=-\shift, xshift=-\shift]AE) -- ([yshift=-\shift]U.east);
\draw[-{Triangle[scale=1.6*\scale]}, line width=1] (X.west) -| (AE) --(U.east);

\coordinate (AF) at (6*\dx, 8.5*\dy);
\draw[-{Triangle[scale=\scale, color=white]}, line width=1] ([yshift=0.5*\shift]X.east) -| ([yshift=0.5*\shift]AF) -| ([xshift=-0.5*\shift]L.north);

\coordinate (AG) at (4.5*\dx, 8.5*\dy);
\draw[-{Triangle[scale=\scale, color=white]}, line width=1] (U.north) -- ([yshift=-0.5*\shift]AG) -| ([xshift=0.5*\shift]L.north);
\draw[-{Triangle[scale=1.4*\scale, color=black]}, line width=1, draw=white]  (AG) -| (L.north);

\coordinate (AK) at (3*\dx, 4*\dy);

\coordinate (AH) at (3*\dx, 5*\dy);
\coordinate (AN) at (1.5*\dx, 6*\dy);
\draw[-{Triangle[scale=\scale, color=white]}, line width=1] ([xshift=0*\shift]O.south) |- ([yshift=\shift]AK) -| ([xshift=0*\shift]P.north);
\draw[-{Triangle[scale=\scale, color=white]}, line width=1] ([xshift=0.5*\shift]S.south) |- ([xshift=\shift]AK) -| ([xshift=\shift]P.north);
\draw[line width=3, draw=white] ([yshift=-\shift]U.south) |- (M.east);
\draw[-{Triangle[scale=\scale, color=white]}, line width=1] ([xshift=\shift]U.south) |- ([yshift=-\shift, xshift=2*\shift]AK) -| ([xshift=2*\shift, yshift=2*\shift]P.north);
\draw[line width=3, draw=white] ([xshift=-0.5*\shift]A2.north) |- ([yshift=-\shift]M.east);
\draw[-{Triangle[scale=\scale, color=white]}, line width=1] ([xshift=-0.5*\shift]A2.north) |- ([yshift=-\shift]M.east);
\draw[-{Triangle[scale=\scale, color=white]}, line width=1] ([xshift=-\shift]O.south) |- ([yshift=\shift]M.east);
\draw[-{Triangle[scale=\scale, color=white]}, line width=1] (M.south) |- ([xshift=-\shift]AK) -| ([xshift=-\shift, yshift=2*\shift]P.north);
\draw[-{Triangle[scale=\scale, color=white]}, line width=1] (L.south) |- (AN) |- ([yshift=-\shift, xshift=-2*\shift]AK) -- ([xshift=-2*\shift, yshift=2*\shift]P.north);

\draw[-{Triangle[scale=2.4*\scale]}, line width=1] (AK) -| (P.north);

\coordinate (AI) at (6.5*\dx, 8.5*\dy);
\draw[-{Triangle[scale=\scale]}, line width=1] ([xshift=0.5*\shift]A2.north) -- ([xshift=0.5*\shift, yshift=1.5*\shift]AI) -| (C);

\coordinate (AJ) at (4*\dx, 6*\dy);
\draw[line width=3, draw=white] ([xshift=-\shift]U.south) |- ([yshift=-0.5*\shift]O.east);
\draw[-{Triangle[scale=1.6*\scale]}, line width=1] (U.south) |- (M.east);
\draw[-{Triangle[scale=\scale, color=white]}, line width=1] ([xshift=-0.5*\shift]S.south) |- ([yshift=0.5*\shift]O.east);
\draw[-{Triangle[scale=\scale, color=white]}, line width=1] ([xshift=-\shift]U.south) |- ([yshift=-0.5*\shift]O.east);
\draw[-{Triangle[scale=1.4*\scale, color=black]}, line width=1, draw=white]  (AJ) -- (O.east);

\coordinate (AM) at (-0.5*\dx,0*\dy);
 \draw[-{Triangle[scale=\scale, color=white]}, line width=1] ([yshift=0.5*\shift]B.east) -- ([yshift=0.5*\shift]AM) |- ([yshift=-0.5*\shift]H.west);
 \draw[-{Triangle[scale=\scale, color=white]}, line width=1] ([yshift=-0.5*\shift]B.east) --([yshift=-0.5*\shift]I.west);
 \draw[-{Triangle[scale=\scale, color=white]}, line width=1] ([yshift=-1.5*\shift]B.east) -- ([yshift=-1.5*\shift]AM) |- ([yshift=-0.5*\shift]J.west);

\coordinate (AL) at (-0.1*\dx, 3*\dy);
\draw[-{Triangle[scale=\scale]}, line width=1] ([yshift=4*\shift]A.east) -- ([yshift=4*\shift, xshift=-1.5*\shift]AL) |- (C);
\draw[-{Triangle[scale=\scale]}, line width=1] ([yshift=3*\shift]A.east) -- ([yshift=3*\shift, xshift=-0.5*\shift]AL) |- (D);
\draw[-{Triangle[scale=\scale]}, line width=1] ([yshift=2*\shift]A.east) -- ([yshift=2*\shift, xshift=0.5*\shift]AL) |- (E);
\draw[-{Triangle[scale=\scale]}, line width=1] ([yshift=\shift]A.east) -- ([yshift=\shift, xshift=1.5*\shift]AL) |- (F);

\draw[line width=3, draw=white] ([yshift=-3*\shift]A.east) -- ([yshift=-3*\shift, xshift=-0.5*\shift]AL) |- ([yshift=0.5*\shift]I.west);
\draw[line width=3, draw=white] ([yshift=-4*\shift]A.east) -- ([yshift=-4*\shift, xshift=-1.5*\shift]AL) |- ([yshift=0.5*\shift]J.west);

\draw[-{Triangle[scale=\scale]}, line width=1] ([yshift=-\shift]A.east) -- ([yshift=-\shift, xshift=1.5*\shift]AL) |- (G);
\draw[-{Triangle[scale=\scale, color=white]}, line width=1] ([yshift=-2*\shift]A.east) -- ([yshift=-2*\shift, xshift=0.5*\shift]AL) |- ([yshift=0.5*\shift]H.west);
\draw[-{Triangle[scale=\scale, color=white]}, line width=1] ([yshift=-3*\shift]A.east) -- ([yshift=-3*\shift, xshift=-0.5*\shift]AL) |- ([yshift=0.5*\shift]I.west);
\draw[-{Triangle[scale=\scale, color=white]}, line width=1] ([yshift=-4*\shift]A.east) -- ([yshift=-4*\shift, xshift=-1.5*\shift]AL) |- ([yshift=0.5*\shift]J.west);

\draw[-{Triangle[scale=1.4*\scale, color=black]}, line width=1, draw=white]  ([xshift=-2*\shift]H.west) -- (H.west);
\draw[-{Triangle[scale=1.4*\scale, color=black]}, line width=1, draw=white]  ([xshift=-2*\shift]I.west) -- (I.west);
\draw[-{Triangle[scale=1.4*\scale, color=black]}, line width=1, draw=white]  ([xshift=-2*\shift]J.west) -- (J.west);

\coordinate (AO) at (1*\dx, 3*\dy);
\draw[-{Triangle[scale=\scale, color=white]}, line width=1] (C) -| ([yshift=4*\shift, xshift=1.5*\shift]AO) -- ([yshift=4*\shift, xshift=-5*\shift]P.west);
\draw[-{Triangle[scale=\scale, color=white]}, line width=1] (D) -| ([yshift=3*\shift, xshift=0.5*\shift]AO) -- ([yshift=3*\shift, xshift=-5*\shift]P.west);
\draw[-{Triangle[scale=\scale, color=white]}, line width=1] (E) -| ([yshift=2*\shift, xshift=-0.5*\shift]AO) -- ([yshift=2*\shift, xshift=-5*\shift]P.west);
\draw[-{Triangle[scale=\scale, color=white]}, line width=1] (F) -| ([yshift=1*\shift, xshift=-1.5*\shift]AO) -- ([yshift=1*\shift, xshift=-5*\shift]P.west);
\draw[-{Triangle[scale=\scale, color=white]}, line width=1] (G) -| ([yshift=-1*\shift, xshift=-1.5*\shift]AO) -- ([yshift=-1*\shift, xshift=-5*\shift]P.west);
\draw[-{Triangle[scale=\scale, color=white]}, line width=1] (H) -| ([yshift=-2*\shift, xshift=-0.5*\shift]AO) -- ([yshift=-2*\shift, xshift=-5*\shift]P.west);
\draw[-{Triangle[scale=\scale, color=white]}, line width=1] (I) -| ([yshift=-3*\shift, xshift=0.5*\shift]AO) -- ([yshift=-3*\shift, xshift=-5*\shift]P.west);
\draw[-{Triangle[scale=\scale, color=white]}, line width=1] (J) -| ([yshift=-4*\shift, xshift=1.5*\shift]AO) -- ([yshift=-4*\shift, xshift=-5*\shift]P.west);
\draw[-{Triangle[scale=5*\scale]}, line width=1] (A.east) -- (P.west);

\draw[line width=3, draw=white] (H.south) |- ([yshift=0.5*\shift]N.west); 
\draw[line width=3, draw=white] (I.north) |- ([yshift=-0.5*\shift]N.west); 

\draw[-{Triangle[scale=\scale, color=white]}, line width=1] (H.south) |- ([yshift=0.5*\shift]N.west); 
\draw[-{Triangle[scale=\scale, color=white]}, line width=1] (I.north) |- ([yshift=-0.5*\shift]N.west); 
\draw[-{Triangle[scale=1.4*\scale]}, line width=1] ([xshift=-2*\shift]N.west) -- (N.west);

\draw[line width=3, draw=white] (S) -| (A);
\draw[-{Triangle[scale=1]}, line width=1] (S) -| (A);

\end{tikzpicture}

%% file: tikz/DAG_ADULT.tex
\footnotesize
\begin{tikzpicture}[every text node part/.style={align=center}]

\definecolor{lightgray}{rgb}{.9,.9,.9}
\definecolor{lightblue}{rgb}{.62,.73,.83}
\definecolor{lightorange}{rgb}{.99,.82,.60}
\definecolor{lightpurple}{RGB}{197,180,227}

% Helpers
\def\ny{11}
\def\nx{8}
%\draw[help lines,xstep=1,ystep=1, color=black!20] (-\nx,-\ny) grid (\nx,\ny);
%\draw[help lines,xstep=.5,ystep=.5, color=black!10, dotted] (-\nx,-\ny) grid (\nx,\ny);
%\foreach \x in {-\nx,...,\nx} { \node [anchor=north, color=black!30] at (\x,0) {\x}; }
%\foreach \y in {-\ny,...,\ny} { \node [anchor=east, color=black!30] at (0,\y) {\y}; }

\def\dy{1.5}
\def\dx{1.9}
\def\shift{2pt}

\tikzset{
block/.style={
	rounded corners=5pt,
    rectangle,
	fill=lightgray,
    draw=black,
	line width=1,
    text depth=.3\baselineskip, 
    text height=.7\baselineskip,
	inner ysep=0.2cm,
	inner xsep=0.3cm
}}

\node[block, fill=lightblue] (A) at (0,4*\dy) {\ttfamily capital-loss};
\node[block, fill=lightblue] (B) at (0, 1*\dy) {\ttfamily hours-per-week};

\node[block, fill=lightblue] (C) at (1*\dx, 6*\dy) {\ttfamily income};

\node[block, fill=lightblue] (D) at (2*\dx, 4*\dy) {\ttfamily capital-gain};
\node[block, fill=lightblue] (E) at (1*\dx, 2*\dy) {\ttfamily workclass};

\node[block, fill=lightorange] (F) at (3*\dx, 7*\dy) {\ttfamily education-num};

\node[block, fill=lightorange] (G) at (4*\dx, 6*\dy) {\ttfamily education};
\node[block, fill=lightblue] (H) at (3*\dx, 2*\dy) {\ttfamily occupation};

\node[block, fill=lightorange] (I) at (5*\dx, 5*\dy) {\ttfamily race};
\node[block, fill=lightorange] (J) at (5*\dx, 3*\dy) {\ttfamily marital-status};
\node[block, fill=lightorange] (K) at (5*\dx, 1*\dy) {\ttfamily relationship};

\node[block, fill=lightorange] (L) at (6*\dx, 4*\dy) {\ttfamily native-country};
\node[block, fill=lightorange] (M) at (6*\dx, 0*\dy) {\ttfamily age};

\node[block, fill=lightorange] (N) at (7*\dx, 3*\dy) {\ttfamily sex};

\draw[-{Triangle[scale=1]}, line width=1] (G) |- (F);
\draw[-{Triangle[scale=1]}, line width=1] (L) -| (J);
\draw[-{Triangle[scale=1]}, line width=1] (J) -- (K);

\draw[-{Triangle[scale=1, color=white]}, line width=1] (M) |- ([yshift=-0.5*\shift]J.east);
\draw[-{Triangle[scale=1, color=white]}, line width=1] ([yshift=0.5*\shift]N.west) -- ([yshift=0.5*\shift]J.east);
\draw[-{Triangle[scale=1.4, color=black]}, line width=1, draw=white] ([xshift=2*\shift]J.east) -- (J.east);

\draw[-{Triangle[scale=1]}, line width=1] ([xshift=-0.5*\shift]L.north) |- (I);
\draw[-{Triangle[scale=1, color=white]}, line width=1] ([xshift=0.5*\shift]L.north) |- ([yshift=-0.5*\shift, xshift=\shift]G.east);
\draw[-{Triangle[scale=1, color=white]}, line width=1] (I.north) |- ([yshift=-1.5*\shift, xshift=\shift]G.east);
\draw[-{Triangle[scale=1, color=white]}, line width=1] ([xshift=-0.5*\shift]N.north) |- ([yshift=0.5*\shift]G.east);

\coordinate (AB) at (7*\dx, 7.5*\dy);
\draw[-{Triangle[scale=1, color=white]}, line width=1] ([xshift=0.5*\shift]N.north)-- ([xshift=0.5*\shift]AB) -| ([xshift=-0.5*\shift]C.north);
\draw[-{Triangle[scale=1, color=white]}, line width=1] (F.west) -| ([xshift=0.5*\shift]C.north);
\draw[-{Triangle[scale=1.4, color=black]}, line width=1, draw=white] ([yshift=2*\shift]C.north) -- (C.north);

\coordinate (AA) at (7.5*\dx, 0*\dy);
\draw[line width=3, draw=white] (M.east) -- (AA) |- ([yshift=1.5*\shift]G.east);
\draw[-{Triangle[scale=1, color=white]}, line width=1] (M.east) -- (AA) |- ([yshift=1.5*\shift, xshift=\shift]G.east);
\draw[-{Triangle[scale=1.8, color=black]}, line width=1, draw=white] ([xshift=2*\shift]G.east) -- (G.east);

\draw[-{Triangle[scale=1, color=white]}, line width=1] (K) -| ([xshift=0.5*\shift]H.south);
\draw[-{Triangle[scale=1, color=white]}, line width=1] ([yshift=0.5*\shift]M.west) -| ([xshift=-0.5*\shift]H.south);
\draw[-{Triangle[scale=1.4, color=black]}, line width=1, draw=white] ([yshift=-2*\shift]H.south) -- (H.south);

\coordinate (AC) at (-1*\dx, 6*\dy);
\draw[-{Triangle[scale=1, color=white]}, line width=1] ([yshift=-0.5*\shift]M.west) -| ([yshift=0.5*\shift, xshift=-0.5*\shift]AC) -- ([yshift=0.5*\shift]C.west);
\draw[-{Triangle[scale=1, color=white]}, line width=1] (B.west) -| ([yshift=-0.5*\shift, xshift=0.5*\shift]AC) -- ([yshift=-0.5*\shift]C.west);
\draw[-{Triangle[scale=1.4, color=black]}, line width=1, draw=white] ([xshift=-2*\shift]C.west) -- (C.west);

\coordinate (AD) at (1*\dx, 5*\dy);
\draw[-{Triangle[scale=1, color=white]}, line width=1] (A.north) |- ([xshift=-0.5*\shift]AD) -- ([xshift=-0.5*\shift]C.south);
\draw[-{Triangle[scale=1, color=white]}, line width=1] (D.north) |- ([xshift=0.5*\shift]AD) -- ([xshift=0.5*\shift]C.south);
\draw[-{Triangle[scale=1.4, color=black]}, line width=1, draw=white] ([yshift=-2*\shift]C.south) -- (C.south);

\coordinate (AE) at (1*\dx, 3.025*\dy);
\coordinate (AH) at (1*\dx, 2.975*\dy);

\draw[-{Triangle[scale=1, color=white]}, line width=1] ([xshift=-0.5*\shift]E.north)  -- ([xshift=-0.5*\shift, yshift=-\shift]AH) -| ([xshift=-0.5*\shift]A.south);
\draw[-{Triangle[scale=1, color=white]}, line width=1] ([xshift=0.5*\shift]E.north)  -- ([xshift=0.5*\shift]AE) -| ([xshift=-0.5*\shift]D.south);

\coordinate (AF) at (3*\dx, 3.025*\dy);
\draw[-{Triangle[scale=1, color=white]}, line width=1] ([xshift=0*\shift]H.north)  -- ([xshift=0*\shift]AF) -| ([xshift=0.5*\shift]D.south);
\coordinate (AG) at (3*\dx, 2.975*\dy);
\draw[line width=3, draw=white] ([xshift=-\shift]H.north)  -- ([xshift=-\shift]AG) -| ([xshift=1*\shift, yshift=-\shift]A.south);
\draw[-{Triangle[scale=1, color=white]}, line width=1] ([xshift=-\shift]H.north)  -- ([xshift=-\shift]AG) -| ([xshift=0.5*\shift]A.south);
\draw[-{Triangle[scale=1.4, color=black]}, line width=1, draw=white] ([yshift=-2*\shift]A.south) -- (A.south);
\draw[-{Triangle[scale=1.4, color=black]}, line width=1, draw=white] ([yshift=-2*\shift]D.south) -- (D.south);

\coordinate (AI) at (2*\dx, 1*\dy);
\draw[-{Triangle[scale=1]}, line width=1] ([yshift=0.5*\shift]H.west) -- ([yshift=0.5*\shift]E.east);
\draw[-{Triangle[scale=1, color=white]}, line width=1] ([yshift=-0.5*\shift]H.west) -| ([yshift=-0.5*\shift]AI) -- ([yshift=-0.5*\shift]B.east);
\draw[-{Triangle[scale=1, color=white]}, line width=1] (E) |- ([yshift=0.5*\shift]B.east);
\draw[-{Triangle[scale=1.4, color=black]}, line width=1, draw=white] ([xshift=2*\shift]B.east) -- (B.east);

\coordinate (AJ) at (4*\dx, 2*\dy);

\draw[-{Triangle[scale=1, color=white]}, line width=1] ([yshift=-0.5*\shift]I.west) -| ([yshift=-0.5*\shift, xshift=0.5*\shift]AJ) -- ([yshift=-0.5*\shift]H.east);
\draw[-{Triangle[scale=1, color=white]}, line width=1] ([xshift=-0.5*\shift]G.south) -| ([yshift=0.5*\shift, xshift=-0.5*\shift]AJ) -- ([yshift=0.5*\shift]H.east);
\draw[-{Triangle[scale=1.4, color=black]}, line width=1, draw=white] ([xshift=2*\shift]H.east) -- (H.east);

\coordinate (AK) at (3*\dx, 6*\dy);

\draw[-{Triangle[scale=1, color=white]}, line width=1] ([yshift=\shift]G.west) -- ([yshift=\shift]C.east);
\draw[-{Triangle[scale=1, color=white]}, line width=1] ([yshift=0.5*\shift]I.west) -| ([yshift=-\shift]AK) -- ([yshift=-\shift]C.east);
\draw[-{Triangle[scale=1, color=white]}, line width=1] ([xshift=\shift]H.north) |- ([xshift=-\shift]AK) |- (C.east);
\draw[line width=3, draw=white] ([yshift=0.5*\shift]I.west) -| ([yshift=-\shift]AK) -- ([yshift=-\shift]C.east);
\draw[-{Triangle[scale=1, color=white]}, line width=1] ([yshift=0.5*\shift]I.west) -| ([yshift=-\shift]AK) -- ([yshift=-\shift]C.east);
\draw[-{Triangle[scale=1.6, color=black]}, line width=1, draw=white] ([xshift=2*\shift]C.east) -- (C.east);

\end{tikzpicture}